\documentclass[
11pt, 
oneside, 
english, 
singlespacing, 
headsepline, 
]{MastersDoctoralThesis} 

\usepackage[utf8]{inputenc} 
\usepackage[T1]{fontenc} 

\usepackage{mathpazo} 

\usepackage{pgfgantt}
\usepackage{amsmath}
\usepackage{algorithm}
\usepackage[noend]{algpseudocode}
\usepackage{amsthm}
\usepackage{amssymb}
\usepackage{subcaption}
\usepackage{multirow}
\usepackage{siunitx}
\usepackage{lipsum}
\usepackage{bbm}
\usepackage{tikz}
\usepackage[skins]{tcolorbox}
\usetikzlibrary{shapes,arrows}
\usetikzlibrary{shadows}
\usepackage{epigraph} 

\usepackage[hyphens]{url}  
\usepackage{graphicx} 
\usepackage{caption} 
\usepackage{amsmath}
\usepackage{amssymb}
\usepackage{amsthm}
\usepackage{amsfonts}
\usepackage{multirow}
\usepackage{bbold}
\usepackage{mathtools}
\usepackage{tikz}
\usepackage{adjustbox}
\usepackage[switch]{lineno}

\usepackage[utf8]{inputenc} 
\usepackage{hyperref}       
\usepackage{booktabs}       
\usepackage{amsfonts}       
\usepackage{nicefrac}       
\usepackage{microtype}      
\usepackage{cleveref}
\usepackage{enumitem}

\usepackage[square,sort,comma,numbers]{natbib}
\usepackage{bibentry}
\nobibliography*

\usepackage[autostyle=true]{csquotes} 

\allowdisplaybreaks

\newtheorem{theorem}{Theorem}[chapter]

\theoremstyle{definition}
\newtheorem{definition}{Definition}

\newcommand{\ignore}[1]{}

\DeclareMathOperator*{\argmax}{arg\,max}
\DeclareMathOperator*{\argmin}{arg\,min}

\newcommand{\bbE}{\mathbb{E}}
\newcommand{\bbP}{\mathbb{P}}

\newcommand{\calS}{\mathcal{S}}

\newcommand{\modelAccro}{SB-CB}
\newcommand{\modelAccroLong}{State-based Contextual Bandits}

\newcommand{\xhdr}[1]{\vspace{1.7mm}\noindent{{\bf #1.}}}
\newcommand{\RB}{\mathit{RB}}
\newcommand{\RBmeasure}{\RB}

\newcommand{\yhat}{\hat{y}}

\newcommand{\etc}{\emph{etc.}}
\newcommand{\ie}{\emph{i.e.}}
\newcommand{\eg}{\emph{e.g.}}

 \DeclareMathOperator{\sign}{sign}

\newcommand{\FRT}{FRT}
\newcommand{\rCE}{\textsf{$rCE$}}
\newcommand{\mrCE}{\textsf{$mrCE$}}

\newtoggle{review}

\newcommand{\dataname}{InterRace}
\newcommand{\nsurvey}{545}

\newcommand{\dataset}{\url{https://dataverse.harvard.edu/privateurl.xhtml?token=730a3a39-8caf-47b7-8475-c3dd00f5c993}}

\usepackage{color}
\usepackage{comment}
\definecolor{Orange}{rgb}{1,0.5,0}
\definecolor{Purple}{rgb}{0.75,0,1}
\definecolor{Green}{rgb}{0,0.8,0.5}
\definecolor{dodgerblue}{rgb}{0.12, 0.56, 1.0}

\newcommand{\samuel}[1]{\textsf{\textbf{\textcolor{Green}{[[SD: #1]]}}}}

\thesistitle{How Technology Impacts and Compares to Humans in Socially Consequential Arenas} 
\supervisor{John P. Dickerson} 
\examiner{} 
\degree{Doctor of Philosophy} 
\author{Samuel Dooley} 
\addresses{} 

\subject{Computer Science} 
\university{University of Maryland} 
\department{Department of Computer Science} 

\AtBeginDocument{
\hypersetup{pdftitle=\ttitle} 
\hypersetup{pdfauthor=\authorname} 
\hypersetup{pdfkeywords=\keywordnames} 
}

\begin{document}

\frontmatter 

\pagestyle{plain} 


\begin{titlepage}
\begin{center}

\vspace*{.06\textheight}
{\scshape\LARGE \univname\par}\vspace{1.5cm} 
\textsc{\Large Doctoral Thesis Proposal}\\[0.5cm] 

\HRule \\[0.4cm] 
{\huge \bfseries \ttitle\par}\vspace{0.4cm} 
\HRule \\[1.5cm] 

\begin{minipage}[t]{0.4\textwidth}
\begin{flushleft} \large
\emph{Author:}\\
\authorname 
\end{flushleft}
\end{minipage}
\begin{minipage}[t]{0.4\textwidth}
\begin{flushright} \large
\emph{Supervisor:} \\
\supname 
\end{flushright}
\end{minipage}\\[3cm]

\vfill
\deptname\\[2cm] 

\vfill

{\large 10 January, 2022}\\[4cm] 
\includegraphics{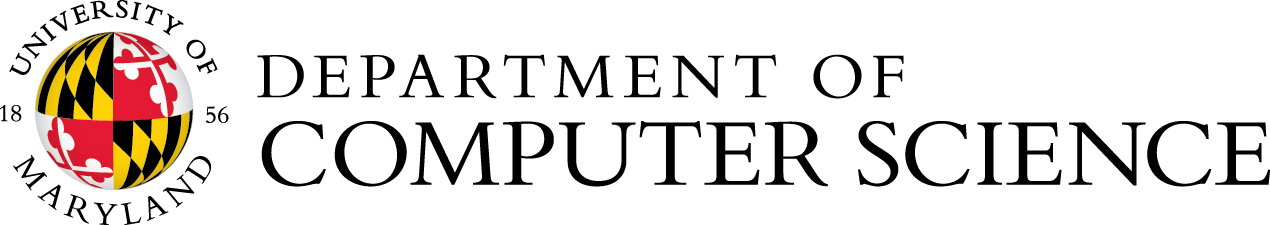} 

\vfill
\end{center}
\end{titlepage}

\begin{abstract}
One of the main promises of technology development is for it to be adopted by people, organizations, societies, and governments --- incorporated into their life, work stream, or processes. Often, this is socially beneficial as it automates mundane tasks, frees up more time for other more important things, or otherwise improves the lives of those who use the technology. However, these beneficial results do not apply in every scenario and may not impact everyone in a system the same way. Sometimes a technology is developed which produces both benefits and inflicts some harm. These harms may come at a higher cost to some people than others, raising the question: {\it how are benefits and harms weighed when deciding if and how a socially consequential technology gets developed?} The most natural way to answer this question, and in fact how people first approach it, is to compare the new technology to what used to exist. As such, in this work, I make comparative analyses between humans and machines in three scenarios and seek to understand how sentiment about a technology, performance of that technology, and the impacts of that technology combine to influence how one decides to answer my main research question.

In this work, I look at three such scenarios: (1) decision support tools, (2) facial analysis technology, and (3) Covid-19 technology. In the first setting, I explore a setting where human evaluators are tasked with finding the best individuals from a population (of people or things) and can pull on a variety of data sources to help them. An example of this is in mental health screening applications where a clinician with a variety of information sources (in-person sessions, audio recordings, social media posts) wants to find the most at-risk individuals from a population. In this area, I
develop novel algorithms for this problem and evaluate the efficacy and comparative improvements on my algorithms when compared to human evaluators alone.

In the second setting, I compare how humans and machines are vulnerable to making errors in facial analysis technology. I explore errors in facial verification, identification, and detection. For facial verification and identification, I compare the biases exhibited by humans to those of machines and conclude similar biases exist for both. For facial detection, I examine the robustness of commercial systems to perturbations under synthetic, naturally-simulated noise corruptions, finding biases along age, gender, skin type, and lighting conditions.

Finally, with Covid-19, I show people's perceptions about privacy and security have altered and been altered by the Covid-19 pandemic with data from field studies and survey collections. In all three settings, my findings from these three scenarios contribute to our understanding of the expansiveness of and the limits to technological interventions. 
\end{abstract}

\setcounter{chapter}{-1}




\tableofcontents 

\mainmatter 

\pagestyle{thesis} 


\chapter{Proposal Outline} 

\label{chpt:prelim} 

\epigraph{He talks, he talks, how he talks, and waves his arms.\\
He fills up ornate vases.\\
Twenty-seven an hour. And keeps the words in with cork stoppers\\
(If you hold the vases to your ears you can hear the muted syllables colliding into each other).\\
I want vases, some of them ornate,\\
But simple ones too.\\
And most of them\\
Will have flowers}{\textit{On Verbosity} \\ Annette Ryan}

We live in a world of gray --- there are few things that are truly black or white.
Of course, some things are purely good and some are purely bad, but most things have elements of both. In those instances, we then get to choose whether the good is more important than the bad.

How do we do this when we are designing technology which might genuinely help some people while also negatively impacting others? Some technology developments fall into an ethnically neutral category --- posing no real harms to anyone, at least none that are immediately obvious. Others, of course, are more hotly contested and each individual may weigh the good and bad parts of a technology in their own way.

The most common way that people make these analyses is by a priori comparison: a new technology development is evaluated in comparison to what existed beforehand. Perceptions about technological change often start by considering how the change would impact the way we were doing something before and how good that technology is. Of course, sometimes, the technology change happens without our full consent (governments adopting facial recognition technology) or deceitfully achieved consent (endless user agreements). In those instances, while we have little say over its adoption, we certainly may bear the harms (and benefit from the goods). 

In this proposal, I primarily study the ways in which technology interventions impact people: both users of the technology and those who are impacted or monitored by it. I have three main application areas where I study this: (1) clinical decision support tools, (2) facial analysis technology, and (3) Covid-19 technology.

In this preliminary section, I will outline the work that my collaborators and I have done so far and then what I plan to do for the rest of my dissertation in each of these areas

\section{Clinical Decision Support}

\subsection{What I've done done so far}
As part of the NSF-funded Smart and Connected Health program, I have been working to build a novel technology intervention which could be used in clinical decision support settings. The goal of this technology is to aid clinicians with screening tasks where they are monitoring a population of patients but have limited resources (time and money), many data sources about each person in the population, and need to prioritize their time on the most at risk individuals. 

I have already proven this idea through simulation-based work on the UMD Reddit Suicidality Dataset where we show that the algorithm outperforms human equivalents under reasonable resource constraint assumptions. This work is discussed in Chapter~\ref{chpt:mh}. 

The above model, while outperforming humans, has some assumptions that could be adjusted and would make the algorithms more general and applicable to other areas. In Chapter~\ref{chpt:sbcb}, I develop a novel multi-armed bandit approach which assumes there is underlying structure to how the different data streams are related to one another. I prove certain desirable properties of this algorithm and place these advancements in the larger multi-armed bandit literature. 

\subsection{Proposed directions forward}

In the next phase of my dissertation, I will take the work in Chapter~\ref{chpt:sbcb} and expand the analysis to include further theories and algorithms which assume that you can choose which type of information you receive about an individual at your will. This analysis will make the theoretical developments more applicable to the overall clinical decision support program. 

Additionally, as the NSF-funded Smart and Connected Health program evolves and collects more data, I will apply the theoretical advancements discussed above to these data and prove out my technology intervention in this scenario. 

Finally, I may conduct survey and/or interview work with interested parties who would be impacted by these technology interventions to (1) understand their perceptions of the technology change, and (2) drive more responsible and applicable development in this space.

\section{Facial Analysis Technology}
\subsection{What I've done done so far}

In Chapter~\ref{chpt:AdvRobustness}, I detail how machine learning datasets, some of which use pictures of human faces, may systematically have classes which are more or less robust than others. Put another way, some classes, like individuals who are Black, may across many different models be mores susceptible to adversarial attack than their counterparts in other races.  

In Chapter~\ref{chpt:ProductionRobustness}, I show how various commercial facial detection systems are not very robust to natural corruptions. This is particularly the case for older, masculine-presenting, darker skinned, and dimly lit people. 

In Chapter~\ref{chpt:fr_hum}, I compare the performance of humans and machines in facial recognition tasks. Using a hand-curated dataset with biases minimized, I asked the same questions of non-expert crowdsourcers, academically-trained machine learning models, and commercial systems. I found that both humans and machines have biases towards males and lighter skinned people, and the biases were similar. Further, humans had a bias towards people who looked like them. 

\subsection{Proposed directions forward}

In the future, I will expand on the above work by testing the robustness disparities found in academically-trained machine learning models for facial detection systems. We may also ask humans to perform a similar task in order to compare human and machine performance in this domain. This is a question that has not be explored in previous work.

\section{Covid-19}
\subsection{What I've done done so far}

As part of Elissa M. Redmile's Max Planck Institute's group studying Covid-19, I have investigated how people's interest in Covid contact tracing apps are moderated by the language used to describe these apps. We analyzed field data from over 7 million impressions of 14 different Google ads which changed the description of privacy and security features of the apps. This work, described in Chapter~\ref{chpt:CovidAds}, provides us with data on how individuals actually behaved in the wild when presented with a decision about engagement with a Covid contact tracing app. 

Additionally, Covid-19 has caused changes to how we perceive technology as well. In ~\cite{goetzen2021ctrl} I detail how privacy attitudes have changed over time from 2019 to 2021. Using survey data from Pew in 2019 and replicated surveys in 2020 and 2021, I explore how pandemic, and to some lesser extent the 2020 Presidential election, has had lasting impacts on what technology interventions Americans view as acceptable and unacceptable. 

\subsection{Proposed directions forward}

Chapter~\ref{chpt:CovidAds} describes how people actually behaved in the wild when presented with text about contact tracing apps. However, how do these behaviors compare to how humans think they would behave? In other words, does the way that we see ourselves using a technology intervention actually comport with reality? In the future, Elissa M. Redmiles and I will study this question by examining prospective-looking survey results which asked participants to think about what would make them comfortable downloading a contact tracing app. We will compare these results to those from the Ad study.

\section{Reading List}
As part of the proposal, I have curated this reading list of 30 relevant papers from three areas. 

\subsection{Multi-Armed Bandits and Applications to Mental Health}
\begin{enumerate}
    \item \bibentry{Auer02:Finite-time}
    \item \bibentry{audibert2010best}
    \item \bibentry{bubeck2011pure}
    \item \bibentry{bubeck2012regret}
    \item \bibentry{Chen14:Combinatorial}
    \item \bibentry{DeChoudhury2016}
    \item \bibentry{Coppersmith2018}
    \item \bibentry{Slivkins19:Introduction}
    \item \bibentry{Schumann2019}
    \item \bibentry{Zirikly2019}
\end{enumerate}

\subsection{Facial Recognition Robustness and Performance}
\begin{enumerate}
    \item \bibentry{o2007face}
    \item \bibentry{klare2012face}
    \item \bibentry{o2012demographic}
    \item \bibentry{phillips2018face}
    \item \bibentry{hu2017person}
    \item \bibentry{hosseini2017google}
    \item \bibentry{buolamwini2018gendershades}
    \item \bibentry{grother2019face}
    \item \bibentry{hendrycks2019benchmarking}
    \item \bibentry{singh2020robustness}
\end{enumerate}

\subsection{Technology Privacy Beliefs and Behaviors}
\begin{enumerate}
    \item \bibentry{tucker2014social}
    \item \bibentry{acquisti2015privacy}
    \item \bibentry{mikal2016ethical}
    \item \bibentry{ford2019public}
    \item \bibentry{nicholas2020ethics}
    \item \bibentry{andalibi2020human}
    \item \bibentry{redmiles_user_2020}
    \item \bibentry{simko2020covid19}
    \item \bibentry{seberger2021us}
    \item \bibentry{munzert2021tracking}
\end{enumerate}

Other work I have completed include~\cite{lam2018xview,kuo2020proportionnet,peri2021preferencenet,mace2018overhead,cherepanova2022deep,knittel2022dichotomous,dooley2020affiliate}.
\chapter{Mental Health Simulations} 
This work has been completed with collaborators Candice Schumann, Han-Chin Shing, John P. Dickerson, and Philip Resnik
\label{chpt:mh} 

\section{Background}\label{sec:intro}

\noindent Machine learning is beginning to have a large impact on the ways that people think about addressing problems in healthcare~\citep{ma2018risk,zhang2019metapred} and mental health \cite[\emph{inter alia}]{Alonso2018,Linthicum2019}, just as it is having large impacts everywhere else. The ability to obtain data about people’s day to day thoughts and experiences via social media---unobtrusive windows into what \cite{Coppersmith2018} call the ``clinical whitespace'' between clinician encounters, in the form of social media posts, wearables data, etc.---is looking to be thoroughly disruptive, and the ability to engage with people via natural spoken interactions on all manner of electronic devices creates potential for even more windows into people’s everyday thoughts and experiences, enhancing the ability to detect new problems earlier and monitor patients under treatment more effectively and at lower cost.

This is no small matter, because mental illness is one of the most significant problems in healthcare. Considering both direct and indirect costs, mental illness exceeds cardiovascular diseases in the projected 2011-2030 economic toll of noncommunicable diseases (\$16.3T worldwide) and that total is more than the cost of cancer, chronic respiratory diseases, and diabetes \emph{combined} \citep{Bloom}. Schizophrenia ranks higher in costs than congestive heart failure and stroke \citep{Insel2008}\ignore{[CITATION NEEDED for Soni 2009].}. The personal and societal toll is also enormous.
In~2016 suicide became the second leading cause of death in the U.S. among those aged 10-34~\citep{hedegaard2018suicide}
and is a major contributor to mortality among those with schizophrenia and depression.

\ignore{
Suicide and Suicidal Attempts in the United States: Costs and Policy Implications. Donald S. Shepard, Deborah Gurewich et al. Suicide and Life-Threatening Behavior, 46, 3, 6 2016.  From summary at sprc.org/about-suicide/costs:
The average cost of one suicide was \$1,329,553. More than 97 percent of this cost was due to lost productivity. The remaining 3 percent were costs associated with medical treatment. The total cost of suicides and suicide attempts was \$93.5 billion. Every \$1.00 spent on psychotherapeutic interventions and interventions that strengthened linkages among different care providers saved \$2.50 in the cost of suicides.}

It is becoming clear that traditional approaches to these problems do not suffice. \cite{Franklin2017}, for example, conclude from a large meta-analysis that there has been no improvement in predictive ability for suicidal thoughts and behaviors over the last 50 years, and argue their findings ``suggest the need for a shift in focus from \emph{risk factors} to machine learning-based risk \emph{algorithms}'' (their emphasis). The technological community is increasingly aware of this problem space and enthusiastic about contributing (e.g.~\cite{Milne2016a,Losada2018,Zirikly2019}), with significant progress in ethical data collection~\citep{Coppersmith2018,Padrez2016} and effective use of those data in predictive models~\citep{Coppersmith2018,Milne2019,Jaroszewski2019,Corcoran2019,Iter2018}.

\ignore{
  Researchers have already demonstrated that sensitive personal data can be elicited ethically\ignore{while still paying careful attention to ethical considerations} \cite{Coppersmith2018,Padrez2016}; that using social media as a window into people’s lives can yield prediction of suicide attempts well beyond the typical accuracy of face to face clinical evaluations \cite{Coppersmith2018}; that automated prediction of crisis can improve speed of response \cite{Milne2019} and help substantially increase the likelihood that a person in acute distress will seek crisis services \cite{Jaroszewski2019}; and that text analysis can capture subtle mental state changes to accurately predict emergent psychosis \cite{Corcoran2019,Iter2018}.  The problem of accurate prediction is far from solved, but these recent developments suggest it is time to begin serious consideration of how machine learning methods can make a wider transition from research into practice.\ignore{, (Resnik in \cite{Lyons2019}).}
  }

Moving machine learning out of the lab will raise new challenges, however, because the mental health ecosystem is highly resource-limited.\ignore{ when it comes to evaluation and intervention.} Even setting aside the unavoidable problem of false positives, an increased ability to identify true positives with the help of machine learning is going to add an influx of new cases that require clinical interaction and potentially action, significantly increasing stress on an ecosystem that cannot easily scale up.\ignore{ to much larger numbers of people for whom risk is identified.} \ignore{An extreme example is Facebook’s suicide risk detection program,\ignore{for detecting and responding to suicide risk} where interventions are limited to providing information (e.g. crisis line numbers), on the one hand, or triggering resource-intensive (and potentially harmful) measures like hospitalization or law enforcement involvement, on the other \cite{GomesdeAndrade2018}.}  As detection of potential problems gets easier and more widespread, effective and scalable methods will be needed so that cases can be prioritized in terms of the attention needed, and so appropriate interventions can be offered across the entire range of severity.
\ignore{However, in considering this explosion of activity, we have identified a clear barrier to wide applicability: the fact that machine learning can scale up much, much more effectively than the human ecosystems in which it is embedded.}

In this paper we introduce a concrete technological proposal for addressing this problem, involving a basic shift in the way we think about machine learning in mental health: the dominant paradigm of individual-level classification is not an end in itself; rather it provides components in a population-based framework involving both machines and humans, where limited resources give rise to a critical need for effective and appropriate ways to set priorities.

\begin{figure}
  \centering
  \includegraphics[width=.9\textwidth]{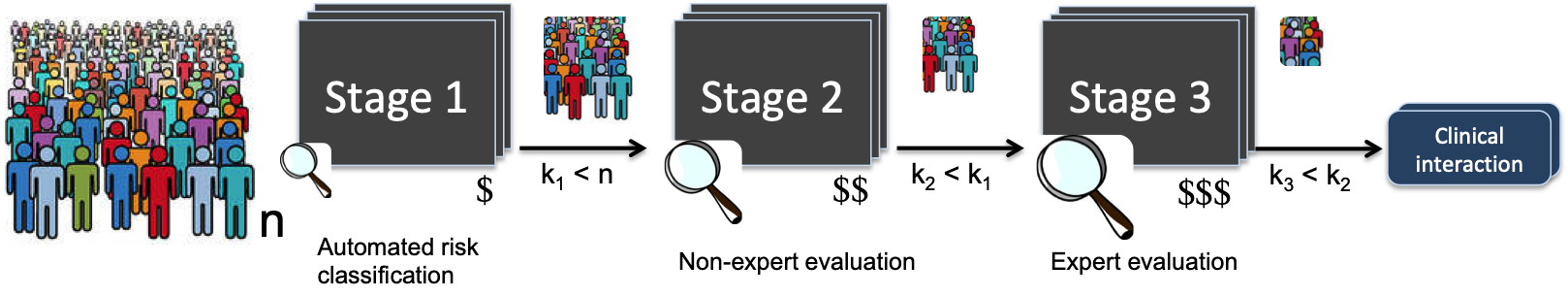}
  \caption{We apply a multi-armed bandit framework in mental health to identify at-risk individuals using a multi-stage approach. The goal is to optimize the number of people at high risk who go on to receive detailed clinical attention, given limited resources.
In our experimental scenario, we instantiate a pipeline progressing from automated analysis of social media posts, to risk evaluation by people without expert training, to expert evaluation.}
  \label{fig:teaser}
\end{figure}

At the core of our technical approach is the recognition that the multi-armed bandit problem in machine learning is a good fit for the real-world scenario created by scaling up the application of technology for detection and monitoring in mental health: what is the best way to allocate limited resources among competing choices, given only limited information?  \ignore{The name of the problem comes from the idea of being in a casino with a fixed amount of money and having a bank of slot machines, “one-armed bandits”, to choose from. If you are able to get a little bit of information at a time from each machine --- pull the arm and see how well it pays off --- what is the best way to decide which subset of machines to spend your money on?\footnote{The abstract formulation of the problem ignores the fact that in the real world slot machines are designed to make you lose and playing them is never a good idea in the first place.}  Here each slot machine is an individual in the population, and playing the machine involves investing some of the limited resources to help form a better assessment of its “reward”, in this case corresponding to severity of risk.}
We adopt a \emph{tiered} multi-armed bandit formulation originally introduced with application to hiring or admissions decisions \citep{Schumann2019}, where a succession of stages is applied to a population of applicants, each stage successively more expensive but also more informative, in order to optimize the value of the set of applicants who are chosen. Our key insight is that, by replacing a population of potential hires with a population of people with potential mental health problems, and by replacing ``value'' with ``risk'', this tiered framework maps directly to a population-level formulation of the assessment problem. Using real data and human annotation, our simulations demonstrate the value of using this framework to combine (cheap, less accurate) automation with (more expensive, more accurate) human evaluation of social media in order to identify individuals within a population who are at high risk for a suicide attempt.

\section{Related Work}

{\it Multi-Armed Bandits.} The main model on which our approach relies is derived from \cite{Schumann2019}. They introduce the concept of tiers to the extant literature on multi-armed bandits. \cite{bubeck2012regret} provide an excellent overview on the history of the field. Historically, MAB work has been focused on selecting the best arm from a population, but works recently have moved to selecting the best cohort \citep{bubeck2013multiple,Chen14:Combinatorial}. There has been extensive research into the objective functions that get used in these models.
\cite{lin2011class} introduced a monotone submodular function as a method for balancing individual utility and diversity of a set of items; this has been adapted to MAB models~\citep{Schumann19:Diverse}.
Additional work has been done on optimization algorithms for these types of functions \citep{krause2014submodular,ashkan2015optimal}. \cite{ding2013multi} and \cite{xia2016budgeted} looked at a regret minimization MAB
problem in which, when an arm is pulled, a random reward is received and a random cost is taken from the budget.
\cite{Schumann19:Diverse} introduced a concept of
``weak'' and ``strong'' pulls in the Strong Weak Arm Pull (SWAP) algorithm. Taken together, this body of literature provides the theoretical backbone for the appropriateness and functionality of our approach.

{\it Mental Health Datasets.} The data we used lacks ground truth on whether or not the individual attempted suicide. Such information is extremely difficult to obtain, and it is even rarer to see datasets linking clinical and social media data (though cf. \cite{Padrez2016}).
As a result, most work analyzing social media for mental health relies on non-ground-truth evidence such as 
online self-report \citep{Coppersmith2014,Coppersmith2015g,MacAvaney2018}
or group membership participation and changes \citep{DeChoudhury2015,DeChoudhury2016}, though see \cite{Ernala2019b} for important limitations of such proxy diagnostic signals.
As one notable exception, \cite{Coppersmith2018} report strong predictive results using a dataset that contains outcome data on suicide attempts, collected using \url{ourdatahelps.org}, an innovative platform for consented data donation. \ignore{ that permits people who have lost a loved one to suicide, suicide attempt survivors, or just people who want to help, to donate their social media data for research.  A similar dataset is being collected using the OurDataHelps infrastructure, at \url{umd.ourdatahelps.org}, focused on schizophrenia and depression.}
\ignore{As another, \cite{Padrez2016} pursued an innovative strategy for obtaining linked social media and clinical data, approaching  more than 5,000 people in an urban emergency department to obtain consent. The promising news is that, of the subset who had Facebook or Twitter accounts (about half), nearly 40\% were willing to share their social media and EMR data for research purposes.}

{\it Prediction of Risk using Machine Learning.}
Recently there has been a significant uptick in research activity in NLP and machine learning for mental health. A 2019 suicide risk prediction exercise using (an earlier version of) the UMD Reddit Suicidality Dataset took place in which an international set of 15 teams participated~\citep{Zirikly2019}; a number of other related shared tasks have also taken place \citep{Milne2016a,Milne2016,Losada2019}. In real-world settings,  automated prediction of mental health crisis has improved speed of response~\citep{Milne2019} and has been used to trigger interventions that substantially increase the likelihood that a person in acute distress will seek crisis services~\citep{Jaroszewski2019}.

\section{Problem Formulation}

Consider a population of individuals where each individual has some potential risk in a given mental health scenario, e.g. veterans at risk for suicide, or college students at risk for onset of schizophrenia. We assume a characterization of risk on a four-point scale (low, no, moderate, or severe). These labels are inherently context based and will depend upon the particular condition, but we assume they are derived by clinical experts and agreed upon for the given population (e.g. see \citet{Corbitt-Hall2019,Milne2019,shing2018}).

Given such a population, we take as our goal the identification of as many severe-risk individuals as possible, so they can receive more thorough assessment and appropriate intervention or treatment; however, we need to do this with extremely low resources. The mental health ecosystem is \emph{dramatically} under-resourced; for example, fully a third of the U.S. population live in federally designated mental healthcare provider shortage areas \citep{kff_2019}. This makes it essential to improve our ability to \emph{prioritize} clinicians' time and caseload, but in a way that minimizes the chance of missing at-risk individuals. 

One promising direction is in the increasing ability to tap into what may be happening with individuals in an ongoing way via their social media, using machine learning. \ignore{Research into the efficacy of these inferences is ongoing, e.g. \citep{Zirikly2019,Braithwaite2016,Ernala2019b,Guntuku2017}, and see our Broader Impact section for an important discussion of ethical considerations, but such approaches show significant promise.} For example, \citet{Coppersmith2018} demonstrate an ability to predict suicide attempts based on social media that is much better than typical performance of clinicans based on traditional in-person evaluation, and \citet{Milne2019} show that machine risk classification can greatly improve response latency by moderators on a peer-support forum.

At the same time, human review of individuals' social media content is also increasingly taking place, including, for example, by non-clinicians within Facebook's operations \citep{GomesdeAndrade2018} and moderators in peer support forums \citep{Milne2019}, and there is also initial work looking at the evaluation of social media content by personnel with varying levels of specialization or training \citep{shing2018,Kelly}.  This raises the possibility of exploring intermediate points between inexpensive fully automated methods and expensive clinical interactions---and, in particular, the idea that by combining different forms of evaluation, it may be possible to optimize the combination of machine and human effort in a way that produces the best outcome possible given the resources available.

\ignore{
Our approach to doing so is a multi-stage model that in some respects parallels the dominant multi-stage progression of universal, selective, and indicated interventions in public health \citep{Goldsmith2002,Olson2018}, but which approaches the problem using the tools of machine learning and optimization.
}

\ignore{The traditional mental health community has historically identified and supported at risk individuals through relatively infrequent interactions, due to the stressors on the system. At risk individuals who do seek medical help may interact with a healthcare professional rarely. However, they may be posting on social media with much more significant frequency. There is a growing body of work that suggests facets of a patients mental health can be inferred from the text which they generate for social media posts \citep{Coppersmith2018}. Research into the efficacy of these inferences is ongoing and is examining both experts and crowdsourced evaluations of these social media posts \citep{Zirikly2019,Braithwaite2016,Ernala2019be,Guntuku2017}.}

\section{Approach}\label{sec:approach}

We propose that mental health risk assessment should be viewed as a population-oriented, multi-stage problem, where subsets of individuals (who have opted in appropriately with informed consent) progress from less costly stages (that are also less informative\ignore{, e.g. automated predictive models}), to intermediate stages that require more resources but also provide potentially better information, and finally to more costly forms of assessment, such as evaluation by a trained expert or a qualified clinician. Ultimately the goal is, within given resource limitations, to have as many people as possible who are actually at high risk progress through the entire pipeline to the highly limited and resource-intensive process of traditional, interactive clinical assessment; see Figure~\ref{fig:teaser} for an example of such a pipeline.

We operationalize this approach using the recent budgeted multi-armed bandit (MAB) framework named BRUTaS~\citep{Schumann2019}.  To briefly summarize the model, we cast tiered decision making as a combinatorial pure exploration (CPE) problem in the stochastic multi-armed bandit setting~\citep{Chen14:Combinatorial}.  Here, arms represent individuals with latent true risk profiles, where $S$ is the population of arms with $|S|=n$. \ignore{(e.g., the cohort of all $n$ individuals or the first group of people in Figure~\ref{fig:teaser}).}  The end goal is to recommend a subset of $k \leq n$ \ignore{(the final, and smallest, group of people in Figure~\ref{fig:teaser})} for clinical interaction, after narrowing the pool over successive stages or tiers. Each arm (or individual) $a\in S$ has an associated unknown true risk $u(a)$, and an empirical risk $\hat{u}(a)$ that the algorithm estimates and uses to make decisions. Each analysis stage $i$ has an associated strength of arm pull defined as information gain $s_i$---a further generalization of earlier work~\citep{Schumann19:Diverse}. \ignore{The strength correlates with the confidence of the signal generated as well as the cost of performing an arm pull.} For example, if we compare the signal generated from an expert reviewer (Stage 3 in Figure~\ref{fig:teaser}) and a non-expert (Stage 2 in Figure~\ref{fig:teaser}), one would be much more confident in the signal from the expert compared to the non-expert. Additionally, each analysis stage $i$ has a cost $j_i$. Successive stages increase in both cost $j_i$ and information gain $s_i$.

In our current model we have three stages of assessment: (1) automated risk classification using an NLP model, (2) non-expert risk assessment, and (3) expert risk assessment.\footnote{The first stage is representative of automated systems currently deployed to flag risk in real-world online environments (e.g.~\cite{GomesdeAndrade2018,Milne2019}); being automated, these are less expensive and more scalable than human assessment. The second and third stages represent successively higher value but also successively more expensive and specialized resources; for example, the former might include social work trainees and the latter might include a trained crisis-line staffer or a specialist clinical psychologist. }
\ignore{\footnote{Although in this study's simulations we approximate an intermediate stage of non-experts using crowdsourced judgments, the idea of true crowdsourcing, in the sense of Mechanical Turk and similar platforms, need not, and definitely should not, be considered a part of the proposal. Rather, we use crowdsourcing to approximate some intermediate level of cost and expertise.  Such intermediate levels exist in the real world, e.g. a social work trainee or general practitioner would have less expertise in suicidality assessment than than a trained crisis-line staffer or a specialist clinical psychologist. \label{fn:nonexperts}}}
In that 3-stage setting, the goal is to select a final subset of size $k$ out of the full cohort $S$. After each stage, the pool is narrowed (that is, for some subset of the remaining cohort, intervention decisions are fixed permanently). During stage $i$, $k_i$ individuals move on to the next stage (i.e., we decide not to pursue a deeper intervention with $k_{i-1}-k_i$ individuals), where $n=k_0>k_1>k_2>k_3=k$.
\ignore{\footnote{Note that although in this paper we focus on the importance of getting as many of the right people as possible through to the end of the pipeline, this multi-stage architecture introduces new possibilities for intermediate outcomes, rather than a choice between a clinical interaction or nothing at all. For example, a low-cost intervention for people who reach Stage~2 or Stage~3 might be to send a caring contact~\citep{Comtois2019} or information about help lines or peer support, and encouragement to reach out. Such interventions and their evaluation are a topic for future work.}}
\ignore{Therefore, each stage $i$ could be considered a selection problem where $k_i$ individuals need to be selected in order to maximize the total empirical risk of the chosen individuals.} More concretely, at each stage $i$ a cohort $M_i$ is chosen where $|M_i|=k_i$ where $M_i$ is chosen as follows: $M_i=\arg\max_{M}\sum_{a\in M}\hat{u}(a)$.
Finally, at each stage $i$, there is a budget $T_i$ associated with how much information gathering can be performed at that stage, leading to a total budget of $T=\sum_{i=1}^3 T_i$. Thus, there are a few hyperparameters to tune before running the algorithm: the number individuals to move on to each next stage $k_i$, budgets for each stage $T_i$, information gain for each stage $s_i$, and the cost for each stage $j$.

\ignore{A key technical insight in our proposal is that mental health assessment can be formulated in a directly analogous way\ignore{, as a tiered, multi-stage problem involving resource limitations} (Figure \ref{fig:stages}). In Schumann et al.~\citep{Schumann2019}, an admissions or hiring process begins with a population of applicants, and progresses through a process of ``paring down'' the applicant pool such that predicted lower-quality workers are filtered at each stage, so that greater evaluation resources are directed most effectively, particularly at the more costly later stages.
\ignore{\footnote{We note that this framework is concerned with more effectively allocating \emph{human} review and interview resources, and is explicitly not proposing pure ``AI screening''. We will use \emph{tier} and \emph{stage} interchangeably.}}
Similarly, we propose that mental health assessment should be viewed as a population-oriented, multi-stage problem, where subsets of individuals progress from less costly stages (that are also less informative, e.g. automated predictive models of the kind emphasized in current mental health machine learning literature), to intermediate stages that require more resources but also provide better information (including traditional methods like requesting self-report scales, as well as new concepts such as automated interviews or clinician review of automated predictions), and ultimately to the most costly forms of assessment, such as in-person evaluation by a qualified clinician. Crucially, this does not obviate the need for individual-level predictive modeling, where significant advances have been achieved over the past several years by us and others \citep{Zirikly2019,Coppersmith2018,Corcoran2018a,Linthicum2019,Alonso2018,Calvo2017}.\footnote{See \citet{Alonso2018,Calvo2017} for broader discussion of computational language analysis for mental health more generally, and \citet{Linthicum2019} for a broader review of machine learning in suicide science).} Rather, the individual level predictive models are re-cast as crucial components within the multi-stage framework.

We used the recent multi-armed bandit (MAB) framework~\citep{Schumann2019} due to PI Dickerson and team.  To briefly summarize the model, we cast tiered decision making as a combinatorial pure exploration (CPE) problem in the stochastic multi-armed bandit setting~\citep{Chen14:Combinatorial}.  Here, arms represents individuals with latent true risk profiles.  The goal is to select a subset of $k \leq n$ arms $S$, with $|S|=n$ (e.g., the cohort of all $n$ individuals), after narrowing the pool over successive stages or tiers. Each analysis stage has an associated strength of arm pull---a further generalization of PI Dickerson's earlier work~\citep{Schumann19:Diverse}. The strength determines the confidence of the signal generated (e.g., by the expert reviewer or clinician) as well as the cost of performing an arm pull.

Assume $m$ stages of assessment.  Then, in that $m$-stage setting, the goal is to select a final subset of size $k_m$ of the full cohort $S$, with $|S|=n$.  After each stage, the pool is narrowed (that is, for some subset of the remaining cohort, intervention decisions are fixed permanently). In other words, during each stage $k_i$ individuals move on to the next stage (i.e., we decide not to pursue a deeper intervention with $k_{i-1}-k_i$ individuals), where $n=k_0>k_1>\cdots>k_{m-1}>k_m=k)$. Therefore, each stage $i$ could be considered a selection problem where $k_i$ individuals need to be selected in order to maximize some objective function.  Finally, at each stage, there is a budget associated with how much information gathering can be performed at that stage  In our pilot setting, we follow the intuition that individuals can and often are evaluated by an NLP-based system, by a crowd, or by an expert. These different evaluations provide signals about the potential, e.g., suicide risk of an individual, or more generally risk of a negative mental health event. \ignore{In our model, we place these different evaluations in a staged system where Stage 1 is an NLP evaluation, Stage 2 is a crowd evaluation, and Stage 3 is an expert evaluation.}}

\section{Experiments}

\ignore{In this section, we provide an overview of the data for our experiments, our comparative baselines, and the experiments we conduct.}

\subsection{Data}
\label{sec:research:data}
 
\ignore{We propose that our approach can be applied to a variety of mental health screening applications. However, testing this proposal in practice can be extremely difficult owing to data availability and privacy concerns. All data used in this project has already been collected and falls under existing IRB approvals which extend to the researchers.}

\ignore{The intent of the framework is extremely general, and its potential will ultimately need to be evaluated across a wide range of mental health conditions and scenarios.}

\ignore{Since our tiered approach to suicidality-risk assessment is a novel technical approach, we work with the closest dataset available:}
Data connected with suicide risk assessment is incredibly difficult to obtain, especially in quantity.  As an accessible approximation, we work with the UMD Reddit Suicidality Dataset \cite{shing2018}, derived from Reddit, a collection of online communities discussing an enormous range of topics in which participants post anonymously.
The dataset comprises more than 1.5M posts across Reddit subcommunities, from 11,129 users who posted to the SuicideWatch community and a corresponding set of control users who never posted to SuicideWatch. It includes human assessments of suicide risk on a four-point scale (no, low, moderate, and severe risk) based on SuicideWatch posts for a randomly selected subset of 242 of the users who posted to SuicideWatch. Four experts provided ratings, with good inter-rater reliability (Krippendorff's $\alpha = 0.81$). Crowdsource worker judgments based on SuicideWatch posts for the same~242 individuals, plus an additional~621 individuals, were also obtained, achieving moderate inter-rater reliability (Krippendorff's $\alpha = 0.55$); note that we use these judgments here only as an approximation of non-expert evaluators, and would not propose using crowdsourcing in a realization of this system. \ignore{(See Shing et al. \cite{Shing2018} for discussion of a rubric for assessment of social media posts developed in consultation with suicide prevention experts, which was used in this process.)} Taken together, these data capture people’s outreach for help (posts on SuicideWatch), along with high quality expert assessments of risk, moderate quality, non-expert assessments, and large-volume weak positive evidence for more than 10K people (by virtue of their having posted to SuicideWatch).

\ignore{In order to facilitate the comparison of our multi-armed bandit approach to existing baselines, we compute average cost for expert and the non-expert reviews.} From the UMD Reddit Suicidality Dataset metadata, we computed that the average non-expert cost \$0.09 per evaluation of an individual.  For Stage~3, discussion with experts suggests that an estimated cost of \$5.35 per individual is a reasonable first approximation. (All figures are in USD.)   In the absence of a well-founded way to measure information gain at this point, we assume that the information gain of each stage is ten times that of the previous, which is within the range of parameters explored in~\cite{Schumann2019}; further exploration of this parameter is an important subject the future.

\subsection{Baselines}\label{sec:baselines}

Recall, our goal is to identify the at-risk individuals from a population. In our setup, we have a population of 242 individuals where 42 of them are at risk (as defined by having an expert consensus risk label of \emph{severe}). An individual is determined to be at risk by a consensus of four experts. We now outline several baseline approaches, reporting the cost of each approach, the number of individuals it evaluates, and the performance statistics. 

Each of these baselines was evaluated on the UMD Reddit Suicidality Dataset with results reported in Table \ref{tab:main_table}. For those baselines with an element of randomness, for instance, selecting only 100 individuals to evaluate, the simulation of the baseline was performed 10,000 times. The mean and two standard deviations are reported.

\textbf{Expert Baselines} \quad
The first set of baselines involve only \emph{experts}. The most na\"ive approach to evaluate the population would be to have every expert evaluate every individual ({\it 4Experts}). This would be the most expensive with $242\cdot 4 = 968$ evaluations at a total cost of $968\cdot$\$5.35 = \$5,178.8. However, this will yield the best results. It would have perfect predictive power, by the definition of how we have defined the at-risk individuals. 

Another, less expensive option is to have each individual only be evaluated by one Expert ({\it 1Expert}). For instance, for each individual, randomly sample an expert to perform an evaluation, and use that evaluation as the prediction. This would only take 242 evaluations at a total cost of \$1,294.7 and has slightly lower performance than {\it 4Experts}; the population sensitivity of the former is 0.91 compared to 1.0 of the latter. The performance loss captures the noise in the evaluations of the experts. This baseline emulates likely real-world scenarios in which evaluations are distributed across a team of reviewers; it is similar, for example, to what happens to calls when they come in to a crisis line.

Yet a different approach would be to sample a cohort of the population and have experts perform evaluations only on that subset. Say we sample a cohort of 100 individuals and then have either all four experts evaluate each person in the cohort ({\it 4Experts-Sub}), or, for each individual in the cohort, randomly assign an expert to evaluate them ({\it 1Expert-Sub}). The former baseline does 400 evaluations at a cost of \$2,140, and the latter does 100 evaluations at a cost of \$535. 

\textbf{NLP Baselines} \quad
Another set of baseline approaches involve using a classifier based on natural language processing (NLP).We assume that the cost of an evaluation by an algorithm is negligible. For a mental health provider, there is likely a cost to integrate and run the technology, which we do not estimate or factor into our analysis, but each individual machine evaluation is certainly very cheap, with costs amortized over time, and so performing an evaluation on the entire population is very feasible. To do this, we have the NLP system evaluate each individual in the population and consider the predicted class (the argmax of the output probability vector) for each individual ({\it NLP-Full}). For comparison to the last two expert baselines, we also establish {\it NLP-Sub} which also first randomly selects a cohort and then runs the algorithm only on that cohort. The final pure NLP baseline would be to run the algorithm across all individuals in the population, and then only take the top $k$ most confident severe individuals ({\it NLP-Top-$k$}). This particular baseline will always perform worse than {\it NLP-Full}, but we include it for comparisons.

For classification we adopt the state of the art approach introduced by \citet{shing2020}. The classifier is a three layer hierarchical attention network~\citep[3HAN]{yang2016hierarchical}, where each layer is composed of a GRU~\citep{bahdanau2014neural} followed by an attention mechanism that learns to pay attention to different parts of the input sequence to derive the output. 3HAN aggregates a sequence of word vectors to a sentence vector, a sequence of sentence vectors to a document vector, and finally a sequence of document vectors to a individual's vector for making the prediction. Particularly useful for problems like this one, where the relevant classification is of individuals, not documents, the 3HAN approach is able to train a model for document-level ranking even when risk labels are available only at the level of the individual author, not the documents themselves (also see \cite{eldan2019}). In addition, Shing et al. discuss advantages of 3HAN in supporting a nested ranking for human review (jointly ranking individuals by highest risk, and ranking within-individual document evidence for faster review); in future work we plan to explore this as a general framework for resource-limited human review within individual stages of our pipeline.

\textbf{Combination Baseline} \quad
Finally, we can combine baselines, e.g. NLP with an expert. This combination will have the algorithm evaluate every individual in a population, then take the top $k$ individuals with highest confidence of being most severe, and then give that cohort to experts to evaluate. This aligns with a naive two-tiered system, though not using the multi-armed bandit approach that we propose. The most meaningful combination of these baselines is {\it NLP-Top-100 + 1Expert-Sub}. 

\subsection{MAB Experiments}

For our main experiments, we use the MAB framework discussed in Section~\ref{sec:approach} with the UMD Reddit Suicidality Dataset.
We translate this data (with subsets of individuals rated by non-experts and clinical experts), and the state of the art NLP classifier, into a three-stage evaluation process, where Stage 1 is an NLP evaluation, Stage 2 is a non-expert evaluation, and Stage 3 is an expert evaluation.  

\textbf{Overall Experiment}\quad
The overarching experiment aims to investigate if a three-tiered MAB approach outperforms the most realistic baselines above for given fixed budgets. The most realistic scenarios for clinician screenings are those with a limited budget, such as {\it 1Expert-Sub} and {\it 1Expert}. Therefore, through these experiments, we report overall performance for the best models for budgets of \$553, \$1,300, or \$2,200. The first offers a comparison to {\it 1Expert-Sub} baseline, the middle to {\it 1Expert}, and the last to {\it 4Experts-Sub}. Results are reported in Table \ref{tab:main_table}.

\textbf{Hyperparameter Experiments} \quad
We conduct other experiments that support our overall experiment, like hyperparameter tuning. Recall from Section \ref{sec:approach} that there are many hyperparameters to this model, such as: budget ($T$) and budget allocation at each stage ($T_i$), cohort size transferred to each stage ($k_1,k_2,k_3)$, output cohort size ($k$), and information gain and cost at each stage $(s_i,j_i)$. We set the information gain and costs associated with each successive stage in our model using the calculations described in Section \ref{sec:research:data}. For Stage 1, we assume the cost of an NLP system is negligible.

To start, we fixed total budget, $T$, at \$553, \$1,300, or \$2,200. We then can divide that total budget among the different stages,  $T_1$, $T_2$, and $T_3$, in two main ways: (1) adjusting the cohort sizes $\{k_1,k_2,k_3\}$, or (2) directly changing the number of evaluations at each stage.  For (1), we performed a simple grid search over combinations of $k_1$ and $k_2$, and $k_3$ (results visualized in Figure~\ref{fig:mab-grid}).      

For (2), we studied how budget division across the different stages impacts performance. With a fixed $T$, we could vary the division of that budget to each stage. Recall that we are assuming that the cost for the first stage (NLP) is negligible. Therefore, we can allocate $T$ to the non-expert and expert stages, $T_2$ and $T_3$ respectively. Intuitively, we could (1) allocate most of the money to the expert reviews in Stage 3 ({\it More 3}), (2) allocate most of the money to the non-expert reviews in Stage 2 ({\it More 2}), or (3) equally split it between Stages 2 and 3 ({\it Equal Split}); we detail overall budget values used in Table \ref{tbl:allocation}, in real USD. Note that at $T = \$ \num{553}$, there is only enough budget for one pull for every 100 individuals in the final cohort and a few pulls for each non-expert. Therefore, there we have no degrees of freedom to allocate the budget to the stages in these settings. We carry out an experiment with $k_1=200$, $k_2=100$ and $k_3\in\{1,2,\dots,100\}$ with results reported in Figure \ref{fig:allocation_strat}.

\begin{table}
\centering
\small{
\begin{tabular}{l|c|c|c|c|}
\cline{2-5}
                             & \multicolumn{2}{c|}{\$1,300}   &\multicolumn{2}{c|}{\$2,200}  \\ \cline{2-5} 
                             & Stage 2       & Stage 3 & Stage 2       & Stage 3            \\ \hline \hline
\multicolumn{1}{|l||}{More 3} & \$200         & \$1,100 & \$300         & \$1,900            \\ \hline
\multicolumn{1}{|l||}{More 2} & \$765       & \$535  & \$1,500       & \$700                \\ \hline
\multicolumn{1}{|l||}{Equal}  & \$620       & \$680 & \$1,100       & \$1,100             \\ \hline
\end{tabular}}
\caption{Budgets for allocation schemes distributing between Stages 2 and 3 for two budgets: \$1,300 and \$2,200. Stage 1 has no cost.}
\label{tbl:allocation}
\end{table}

\ignore{\textbf{Risk Encoding Experiment}\quad
This dataset has four ordinal rating levels: no, low, moderate, and severe risk.  Our framework maximizes a numeric objective. Thus, we tried several different encoding schemes for these discrete classes, including: Binary method ({\it Bin}) where [No, Low, Moderate, Severe] maps to [0,0,0,1]; Linear method ({\it Lin}) where [No, Low, Moderate, Severe] maps to $[0,\frac13,\frac23,1]$; and Exponential method ({\it Exp}) where [No, Low, Moderate, Severe] maps to $[0,\frac17,\frac37,1]$.}

\subsection{Evaluation}\label{sec:eval}

\ignore{Our ultimate metric is population sensitivity of a system. Any mental health evaluation tool will invariably recommend some number of individuals for further clinical attention, which we will assume is some form of clinical interaction. Depending on the form that interaction takes, we should be more tolerant of providing it (if not prohibitively dangerous, intrusive, or expensive, e.g. an in-office assessment) with someone who is not at risk (false positives) than for not providing an intervention with someone who is at risk (false negatives).  We consider a positive example to be an individual rated as at severe risk, i.e., the highest risk classification in the UMD Suicidality Dataset.}

\ignore{We report the sensitivity on the population and the cohort level. Since some of these baselines only evaluate a cohort of individuals, all those individuals in the population that were not in the cohort are treated as negatives. Therefore, we report numbers at both the cohort and population level. To illustrate, in Table \ref{tab:grand_table}, {\it 1Expert-Sub} only evaluates 100 individuals. While the sensitivity on those 100 individuals is high, it also misses many at-risk individuals in the population, which increases the number of false negatives at the population level and decreases the population sensitivity. }

For each model, we calculate its sensitivity (both for the entire population and for the cohort it evaluated), precision, and specificity. We report these metrics by calculating the true/false positives and true/false negatives. For the MAB model, we can count a positive in two ways: (1) any individual that is included in the final cohort ({\bf MAB}), or (2) any individual that is included in the final cohort where an expert evaluated them as at-risk ({\bf MAB*}). The second evaluation metric acknowledges that the system exists in an ecosystem where the final cohort individuals will also include an expert review which could be used to guide decisions made about that individual. The number of true positives will decrease from {\bf MAB} to {\bf MAB*} (because it includes the noise of the expert review process), but the number of true negatives will also increase significantly. Both metrics capture important information and neither is more appropriate than the other. 

\section{Results}

\begin{table*}[]
 \centering
  \begin{adjustbox}{max width=\textwidth}
\begin{tabular}{l|cc|cccc|cccc}
 \toprule
 Approaches & Budget & \begin{tabular}[c]{@{}c@{}}Number of\\ Individuals\\ Evaluated\end{tabular} & \begin{tabular}[c]{@{}c@{}}Population \\ Sensitivity\end{tabular} & \begin{tabular}[c]{@{}c@{}}Cohort \\ Sensitivity\end{tabular} & Precision & Specificity & \begin{tabular}[c]{@{}c@{}}Cohort \\ TP \end{tabular} & \begin{tabular}[c]{@{}c@{}}Cohort \\ FP\end{tabular} & \begin{tabular}[c]{@{}c@{}}Cohort \\ FN\end{tabular} & \begin{tabular}[c]{@{}c@{}}Cohort \\ TN\end{tabular} \\ \midrule
{\bf NLP-Full} & - & 242 & 0.64 & 0.64 & 0.24 & 0.59 & 27 & 82 & 15 & 118 \\
\midrule
\begin{tabular}[c]{@{}l@{}}\bf{NLP-Top-100} \\ {\bf + 1Expert-Sub}\end{tabular} & \$535 & 242 & 0.49 $\pm$ 0.06 & 0.90 $\pm$ 0.11 & 0.71 $\pm$ 0.10 & 0.89 $\pm$ 0.05 & 21 $\pm$ 2.7 & 9 $\pm$ 3.9 & 2 $\pm$ 2.7 & 68 $\pm$ 3.9 \\
{\bf 1Expert-Sub} & \$535 & 100 & 0.34 $\pm$ 0.03 & 0.91 $\pm$ 0.05 & 0.66 $\pm$ 0.14 & 0.92 $\pm$ 0.05 & 14 $\pm$ 2.3 & 7 $\pm$ 4.3 & 2 $\pm$ 2.3 & 77 $\pm$ 4.3 \\
\ignore{{\bf MAB} & \$553 & 242 & 0.77 $\pm$ 0.12 & 0.77 $\pm$ 0.12 & 0.33 $\pm$ 0.05 & 0.66 $\pm$ 0.01 & 33 $\pm$ 5.3 & 67 $\pm$ 5.3 & 9 $\pm$ 5.3 & 132 $\pm$ 5.3 \\}
{\bf MAB} & \$553 & 242 & 0.76 $\pm$ 0.13 & 0.76 $\pm$ 0.14 & 0.33 $\pm$ 0.11 & 0.95 $\pm$ 0.03 & 30 $\pm$ 6.0 & 10 $\pm$ 6.2 & 12 $\pm$ 6.0 & 190 $\pm$ 6.2 \\
\midrule
{\bf 1Expert} & \$1,295 & 242 & 0.91 $\pm$ 0.08 & 0.91 $\pm$ 0.08 & 0.67 $\pm$ 0.08 & 0.91 $\pm$ 0.03 & 38 $\pm$ 3.4 & 18 $\pm$ 6.6 & 4 $\pm$ 3.4 & 182 $\pm$ 6.6 \\
\ignore{{\bf MAB} & \$1,300 & 242 & 0.85 $\pm$ 0.08 & 0.85 $\pm$ 0.08 & 0.36 $\pm$ 0.04 & 0.67 $\pm$ 0.04 & 36 $\pm$ 2.0 & 64 $\pm$ 2.0 & 6 $\pm$ 2.0 & 136 $\pm$ 2.0 \\}
{\bf MAB} & \$1,300 & 242 & 0.74 $\pm$ 0.12 & 0.73 $\pm$ 0.09 & 0.73 $\pm$ 0.08 & 0.95 $\pm$ 0.03 & 31 $\pm$ 4.0 & 11 $\pm$ 6.0 & 11 $\pm$ 4.0 & 189 $\pm$ 6.0 \\ 
\bottomrule
\end{tabular}
\end{adjustbox}
 \caption{Main experimental results for the \$535 and \$1,300 budgets. Further results for higher budgets are in the Appendix \ignore{Baseline approaches are described in \ref{sec:baselines}. Evaluation protocols are reported in Section \ref{sec:eval}. The three MAB experiments all use the {\bf Linear} encoding mechanism described in Section \ref{sec:encoding}.} For approaches with an element of randomness, means and two standard deviations are reported.}
 \label{tab:main_table}
\end{table*}

\textbf{Overall Experiment} \quad

We report average statistics for each model at budgets of \$535 and \$1,300 in Table \ref{tab:main_table}. Our MAB approach outperforms the existing baselines at the lowest budget, in the severely budget-constrained setting. At the lowest \$500 budget, the 1-Expert solution has the lowest performance, and both our MAB and the pure-NLP approaches double the 1-Expert Sensitivity. However, our MAB approach does even better than the pure-NLP approaches, producing a 20\% increase over the next best NLP approach.  At a  higher budget of \$1,300, our approach achieves similar sensitivity to the expert approach (within the error bounds). The MAB does trade off increased sensitivity against lower precision and specificity, but the cost of the increased false positives is already factored into the budget as expert review of those individuals in the final stage. We also highlight that our aim is not the result in any specific condition but rather the general validation of the approach by showing it tends to work better across a range of cases, particularly when the budget is most constrained. 

\ignore{Simulating a real-world scenario in which resources are very limited, our MAB approach outperforms all comparable baselines. The most resource constrained approach ({\it 1Expert-Sub}) for the Experts, evaluates only 100 individuals and achieves a population sensitivity of 0.34. For the NLP baselines, the approach that also only evaluates 100 individuals ({\it NLP-Sub}) achieves 0.27, while the NLP baseline which evaluates every individual ({\it NLP-Full}) achieves 0.64. (Note, {\it NLP-Full} outperforms {\it 1Expert-Sub} because the former sees the entire population whereas the latter only sees the cohort subpopulation. The former's cohort sensitivity is lower than the latter's.) Finally, the combination baseline ({\it NLP-Top-100 + 1Expert-Sub}) performs at 0.49. The MAB approach with the same resources of \$553 achieves an average population sensitivity of 0.77.

Our MAB approach more than doubles the population sensitivity of the expert baseline for the same resource amount and increases the population sensitivity for a no cost NLP baseline by 20\%. At \$553, our approach averaged 9 false negatives in the population and 33 true positives. In comparison to the expert baseline with \$535, it achieved an average of 26 false negatives in the population and 14 true positives.}

\begin{figure*}
\centering
\begin{minipage}{.46\textwidth}
  \centering
    \centering
    \includegraphics[width=\linewidth]{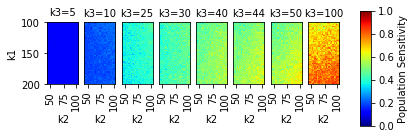}
    \caption{Grid test over $\{k_1,k_2,k_3\}$ for constant budget of \$1,300. Y-axis: first-stage cohort $k_1 \in [100,200]$; x-axis: second-stage cohort $k_2 \in [50,100]$; left-to-right: final cohort size $k_3\in\{5,10,25,30,40,44,50,100\}$. Population sensitivity is reported.}
    \label{fig:mab-grid}
\end{minipage}%
\hfill
\begin{minipage}{.46\textwidth}
    \centering
    \includegraphics[width=.7\linewidth]{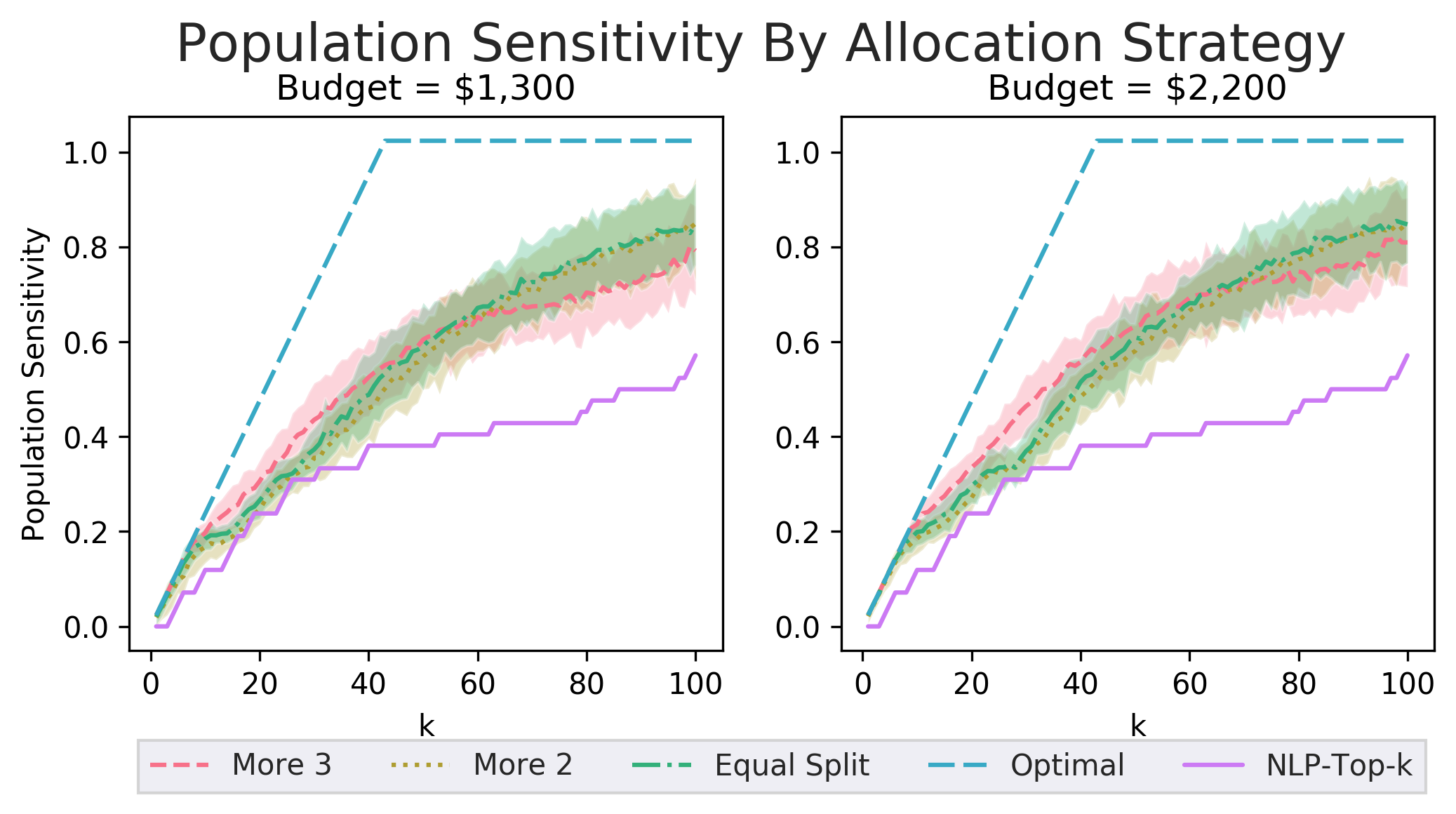}
    \caption{Budget allocation plots for {\it More 3}, {\it More 2}, and {\it Equal}.} 
    \label{fig:allocation_strat}
\end{minipage}
\end{figure*}

\textbf{Hyperparameter Experiments} \quad
\ignore{We now present results from our hyperparameter experiments. }Recall our first line of inquiry focuses on hyperparameters $k_1,k_2$, and $k_3$. These values indicate the size of the cohort that moves to each successive stage in the MAB framework. We present these results in Figure \ref{fig:mab-grid} for a budget of \$2,200, selecting the higher budget to draw out the nuances in the grid search over the $k_i$. In the figure, we report several slices of cube for eight values of $k_3$, where we plot $k_1$ on the x-axis and $k_2$ on the y-axis.

We observe two main points here: (1) as $k_3$ increases, population sensitivity increases, and (2) higher values of $k_1$ correlate to poorer population sensitivity. This first result is intuitive: since there are only 42 severe risk individuals in the population, the sensitivity will be low with low $k_3$. More interestingly, this positive correlation between $k_3$ and population sensitivity holds for all combinations of $k_1$ and $k_2$. This reveals that for any fixed combination of cohort sizes $k_1$ and $k_2$, any increase in $k_3$ will lead to an increase in population sensitivity. Put in a more policy prescriptive way, we suggest that it is always advantageous to include more individuals in the final output cohort, if budget permits. 

Our second claim from this hyperparameter result in Figure \ref{fig:mab-grid} is that higher values of $k_1$ correlate to higher population sensitivity. This is qualitatively evident by the figure, and also supported as statistically significant with a simple linear regression between $k_1$ and population sensitivity with $t$-value 135.75 and $p$-value $p \ll 0.001$. What this indicates is that moving a smaller cohort to the non-experts leads to worse population sensitivity. Put another way, the model performs worse when the NLP makes more discriminative decisions about individuals. Therefore, we conclude that while the MAB system benefits from the inclusion of the NLP system, the NLP provides a useful signal of risk, but over-reliance on the NLP system to remove individuals from the pipeline is not advisable. When adjusting hyperparameters, we must balance the power each stage has to remove individuals from the pipeline with the overall predictive power of that stage. 

We also conducted a similar analysis for $k_2$; a simple linear regression between $k_2$ and population sensitivity with $t$-value 136.02 and $p$-value $p \ll 0.001$. We find similar results from this regression analysis which indicate that there is a positive correlation between $k_2$ and population sensitivity. Again, this implies that with $k_1$ and $k_3$ fixed, system performance improves with lower $k_2$. We can deduce from this analysis that the non-experts provide useful signal to the MAB framework, but are not helpful in removing individuals from the pipeline. We will add that this conclusion is consistent with what clinical practitioners have conveyed to the researchers about non-expert evaluations.

\ignore{\begin{figure}
\begin{subfigure}[b]{0.45\linewidth}
    \centering
    \includegraphics[width=\linewidth]{allocation.png}
    \caption{Budget allocation plots for {\it More 3}, {\it More 2}, and {\it Equal} allocation strategies.} 
    \label{fig:allocation_strat}
\end{subfigure}
\hspace{.1\linewidth}
\begin{subfigure}[b]{0.45\textwidth}
    \centering
    \includegraphics[width=\linewidth]{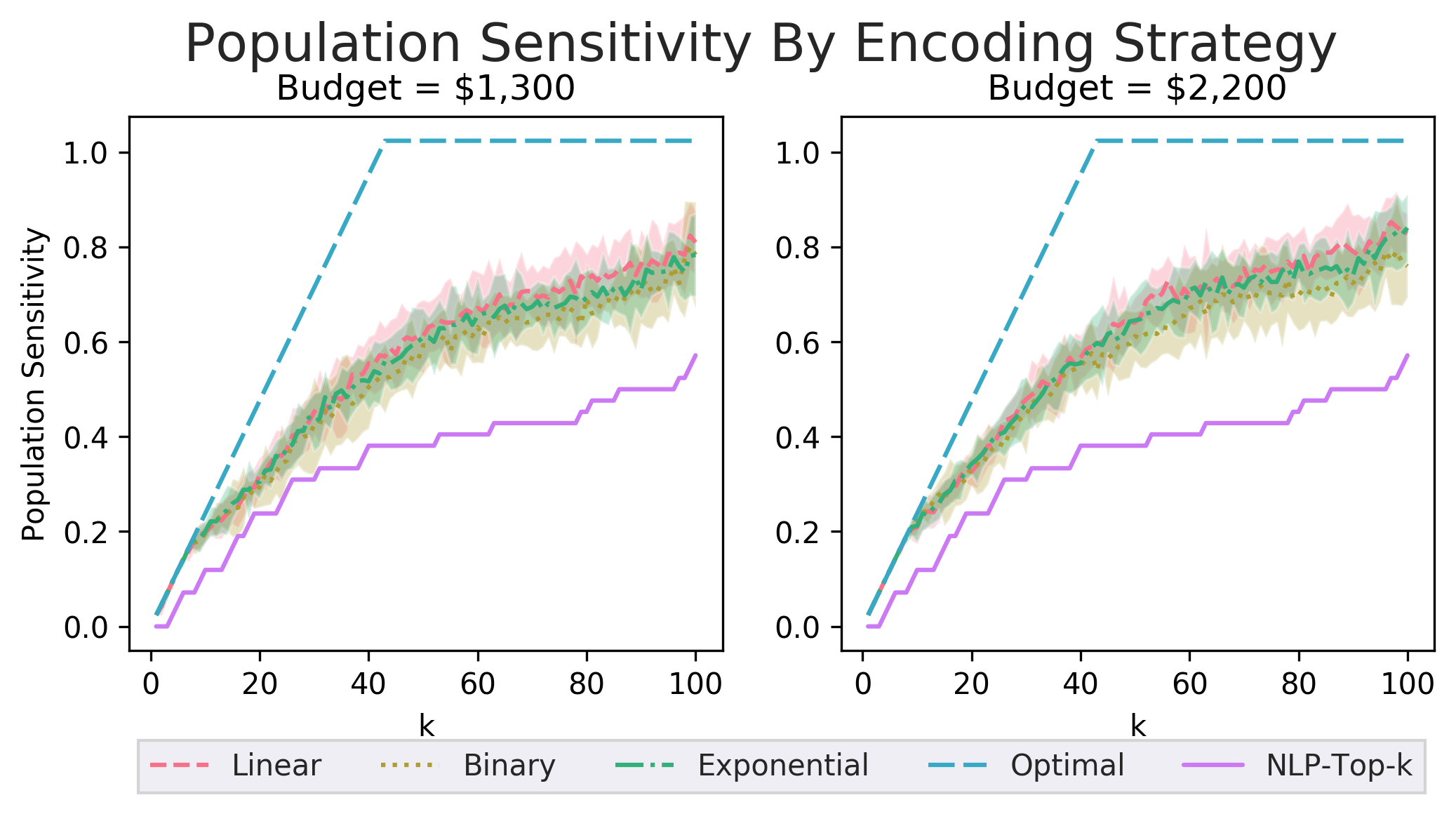}
    \caption{Risk encoding plots for {\it Linear}, {\it Binary}, {\it Exponential} encoding methods.}
    \label{fig:encoding_strat}
\end{subfigure}
\caption{Plots for a MAB model with $k_1=200$, $k_2=100$, and $k_3\in[1,100]$.}
\end{figure}}

Additionally, we analyzed the allocation strategy for a fixed budget among the different stages. For this experiment, we vary $k_3$ from 1 to 100 and keep $k_1$ and $k_2$ fixed at 200 and 100 respectively, as suggested by the results from the grid search. Our initial baseline against which to compare is the ``omniscient'' method that always returns a size-capped cohort with as many at-risk individuals as possible.  Given our evaluation metric, the optimal baseline ({\it Opt}) is one which achieves the highest possible sensitivity for the dataset.  Since there are 42 ground truth severe risk individuals in the dataset, if $k\leq42$, then the best a model could do would be to choose $k$ severe risk individuals and achieve a sensitivity of $k/42$. 
For $k>44$, the best possible would choose 42 severe risk individuals and $k-42$ others, which would result in a sensitivity of $1$.  This optimal baseline can also be thought of as only having experts evaluate the entire population, without any cap on budget.  

In Figure \ref{fig:allocation_strat}, we see that there are no significant differences between {\it More 2} and {\it Equal} which indicates that allocating more budget to the non-expert level does not improve the population sensitivity for various final cohort sizes $k$. This, again aligns with intuition provided by the clinical practitioners about the non-expert evaluations. However, we note that the analysis of strategy {\it More 3} is  more nuanced. For low final cohort values, {\it More 3} outperforms the other two allocation strategies. This flips for higher $k$. We also note that when comparing the magnitude of this difference between budgets of \$1,300 and \$2,200, the magnitude is slightly more pronounced in the former. This suggests that in resource constrained settings, the allocation strategy matters more. Further, the allocation strategy that one would choose for a given scenario would depend upon the final cohort size. For example, if the final cohort is constrained to be 30 individuals, then More 3 outperforms the other two methods. However, this does not hold for larger $k_3$.

\ignore{\textbf{Risk Encoding Experiments}\quad 
 Using the same settings of $k_1,k_2,$ and $k_3$ as in the budget allocation experiment: we varied the three encoding schemes of the ordinal variables into numeric values: {\it Linear}, {\it Binary}, and {\it Exponential}. Results from this experiment are reported in Figure \ref{fig:encoding_strat}. We found no significant difference between the encoding schemes. There is a rough ordering over the different encodings: Linear performs better than Exponential, which performs better than Binary.}




\section{Ethical Considerations}
\label{sec:ethical}

This research underwent appropriate IRB review and its conduct has been informed by the ethical guidelines in~\cite{Benton2017ethical}.

\ignore{This work is fundamentally motivated by pressing needs in mental health. It is important to note that a form of multi-stage prioritization already exists in the real world: individuals going to a primary care appointment in the U.S. often receive a 2-question depression screening, which can lead to a longer 9-question screening, which might lead to discussion with a doctor who generally has not received specialized mental health assessment training, which might then lead to a specialist referral \cite{maurer2012screening}.  But all that only happens when people are seeing a doctor in the first place, severe risk is often missed by clinicians even a week before a suicide death \cite{smith2013suicide}, and in general the determinants of risk are poorly understood~\cite{Franklin2017}.}

Adding social media classification has the potential to significantly improve our ability to detect people at risk~\citep{Coppersmith2018}. The fundamental insight we add is the idea of moving from \emph{automated classification} to a \emph{general framework for prioritization}: we provide a novel way forward to intelligently integrate classification with human-in-the-loop processes, with the bandit framework providing a means of optimization. Our simulations support the claim that integrating these separate kinds of evaluation in a process of population-based prioritization can dramatically increase the likelihood of an at-risk individual successfully being identified as requiring attention, while keeping resource levels the same.  Concretely, we showed that---to the extent our assumptions and abstraction of the problem are reasonable---we can more than double the number of at-risk individuals identified, for the population in our dataset.  This represents an initial validation of the approach and a significant step toward bridging the gap between idealized machine learning experimentation and deployment of technology in a resource-limited world.

However, the idea of actually deploying a system of the kind envisioned here raises questions that require careful consideration, since even in trying to help a population, one can actually hurt individuals in that population~(e.g. see~\cite{eubanks2018automating}). One set of ethical questions involves the broader socio-technical problem of social media data use in mental healthcare \citep{de2014opportunities,Conway2016,Mikal2016,Benton2017ethical,Linthicum2019}.  Privacy is of course a central consideration, and taking the wrong approach can undermine the larger goals; for example, well intentioned but insufficiently thought out applications of technology have in some cases already caused backlash~\citep{lee2014:samaritan,Horvitz2015}.

In addition, the integration of social media analysis into the mental health ecosystem could have impacts on the labor and economy of both mental health professionals and non-experts. Even if the ultimate goal is to improve the efficacy and efficiency of the system, the most well-implemented changes can have negative impacts. Along with the health and well-being of the potentially at-risk population, the well-being of the humans in the loop needs to be considered~\citep{chen2014laborers,Irani2016,Maitra2020}. Related considerations include issues for clinicians assessing social media of people they are not themselves treating, and how that relates to professional codes of ethics (e.g. \cite{AmericanPsychologicalAssociation}), particularly the duties to warn and to inform \citep{Rothstein2012}; how far would those codes extend in the context of this framework, where would these responsibilities lie, and how would this affect professional risks and protections?

\ignore{, making discussion of new technologies that much more difficult. In addition, it is important to carefully evaluate the validity of signals that are extracted from social media in support of clinical decision-making~\cite{Ernala2019b}.}\ignore{and could trigger paranoia about surveillance in vulnerable populations, which ultimately hurts the patient and may provide no net benefit to them or their group \cite{color}. Further, the mere possibility of social media monitoring can harm those in some populations that researchers or clinicians are trying to help. This concept can reinforce power dynamics between patients and healthcare providers which may lead to symptomatic onsets or regressions.}

Other questions are more specific to our multi-stage framework. Bias, a general issue in machine learning, may manifest in our scenario when some populations present differently than the majority and could be filtered out too early \citep{Schumann2019a,Monahan2008}. And ``filtering out'' is clearly not an acceptable end state: at each stage in the pipeline, suitable forms of attention and potential intervention need to be defined for those whose evaluated risk is not severe enough to require progression to the next later stage. Our work also surfaces questions about resource allocation, introducing new degrees of freedom in budget allocation (e.g. Table~\ref{tbl:allocation}). Simulations can help evaluate alternatives, but ultimately decisions about technological deployment, staffing, and then ensuing adjustments in clinical assessment and intervention, will involve considerations well outside the scope of any optimization strategy.

We conclude with the observation that our results are only a first step on the way to practical deployment. To get the rest of the way there, further theoretical research and experimentation are required in order to expand the evidence base for this approach. We also must carefully consider the ethical issues with conversations that integrate the voices of (at least) technologists, in-the-trenches clinicians, policy makers, and those with lived experience of the conditions.

\ignore{How these individuals make their decisions can be fraught. In the future, we would like to be able to include whatever heuristics they may have established through their own ethical review board processes. We must wrestle with introducing a system like ours which perhaps improves population-level assistance but may hide individual accountability.}

\ignore{Where do these lines get drawn in the context of multinational institutions which may have these data or implement these approaches?  Responsibility must be attributed somewhere in the pipeline, and that responsibility becomes easier to avoid and harder to grasp with a technical solution.}


\section{Conclusions}

Our framework for identification of mental health risk introduces a multi-stage assessment, using tiered multi-armed bandits to navigate tradeoffs between the quality of information and the cost of obtaining it.  Our simulations suggest that such a pipeline can dramatically improve the ability to detect at-risk individuals with severe resource constraints. The competitiveness of our approach diminishes as more resources are available, though we believe the resource-constrained version has wide applicability in the field.

This is a starting point. Our simulations currently use expert ratings, not outcomes or clinically obtained data. In addition, in lieu of access to real-world intermediate levels of expertise, e.g. a social work trainee or general practitioner who may have less expertise in suicidality assessment than than a trained specialist, we currently approximate our intermediate stage of non-experts and its cost using crowdsourced judgments. \ignore{crisis-line staffer or a specialist clinical psychologist.} Finally, for any specific condition, appropriate review and interventions need to be defined at all stages, not just at the end. Common to all these limitations is the observation that the design of appropriate solutions is not just a machine learning research problem, it is a challenge that requires significant engagement between technologists and practitioners.

\ignore{
\section{Discussion}

Current technological research in mental health tends to treat machine-derived and human evaluations very differently. Our simulations support the claim that integrating these separate kinds of evaluation in a process of population based prioritization can dramatically increase the likelihood of an at-risk individual successfully being identified as requiring attention, while keeping resource levels the same.  Concretely, we showed that---to the extent our assumptions and abstraction of the problem are reasonable---we can more than double the number of at-risk individuals identified, for the population in our dataset.

Section~\ref{sec:ethical} noted ethical challenges that need to be considered; let us also consider here the technical challenge of interpretability for the  algorithmic elements of our approach.
\ignore{
. We expect that NLP evaluations will always be superseded by human evaluations for those individuals who progress past the first round.\footnote{Though this is not a given; see \citet{Coppersmith2018} for very strong predictive results and analysis relative to typical clinician performance.}  Further, the}

The MAB framework is interpretable insofar as a decision to omit an individual from a successive stage can be interrogated by examining the individual's human or computer ratings. In that sense, it does not inject any \emph{more} obfuscation of decision making than is already present at each of the tiers. Say, for example, we want to understand what happened for those, on average, 9 individuals in the \$553 setting of the MAB framework who were not identified as being at risk. We observe that those individuals that were excluded after the NLP round were more likely to have been rated by the NLP as being of `No' risk; the average probability of `No' Risk for those excluded after the first round was 71\% versus 8\% for those that progressed. This included some individuals that were truly at high risk. This tells us that the MAB framework is (1) behaving like we would expect, and (2) it is only as good as the individual evaluations. In this case, we see that false negatives in the MAB model are a direct result of the NLP false negatives.

\ignore{
Put more directly, when we ask the MAB system why it made a particular decision, it can point us to the individuals who made particular evaluations, whether algorithmic or human, and for purposes of understanding and improvement, one can interrogate them as to what led them to those decisions. 
}
This ability to track from the MAB system's decisions to the component evaluations is one of the reasons we selected the hierarchical attention network approach \cite{yang2016hierarchical}, for our classifier: its hierarchical attention mechanism has greater potential for interpretation than many other models.\footnote{Although there is controversy over the relationship between attention and interpretability \cite{Jain2019AttentionIN}, that has generally been in the context of non-hierarchical networks. Our experience, though at this point only anecdotal, is that network attention in the hierarchical setting does tend to highlight evidence that is subjectively relevant. We plan to explore this further in future work.}  In general, we find it imperative that any evaluator (both human and machine) used in our proposed ecosystem is well trained and able to explain why evaluations were made. We are hopeful that, as we progress in developing and validating the model, the properties of the MAB setting with regard to interpretability will increase the likelihood that policy makers will want to engage with our proposed solution \cite{Zhang}.

These results are only a first step on the way to practical deployment. To get the rest of the way there, further theoretical research and experimentation are required in order to expand the evidence base for this approach. Equally important, for this and any other proposal, careful consideration of the balance between privacy and prevention must continue and, crucially, that conversation needs to integrate the voices of (at least) technologists, in-the-trenches clinicians, policy makers, and those with lived experience of the conditions we are trying to help address.
}

\chapter{State-based Contextual Bandits} 

\label{chpt:sbcb} 
This work is co-authored with John P. Dickerson.

\section{Introduction}\label{sec:intro}
Internet retailers, social networking services, and traditional businesses frequently test new products and services on sub-populations before deploying globally.  Firms may temporarily tweak services in different ways---e.g., changing the color of a purchase button or the text of a hyperlink in an email---and sub-populations may respond to those tweaks in different ways~\cite{Siroker13:AB}.  Firms may then choose to globally change a service (e.g., the UI of a messaging client or a brand's language) or product (e.g., the size of a package or ingredients of a food) based on the heterogeneous feedback they received during a trial.  How should a firm act in the face of noisy, heterogeneous feedback while needing to make one global decision?

Similarly, in a healthcare triage setting, different recommendations and interventions, potentially at different costs and with different levels of efficacy, may be offered based on information gathered by different mechanisms (e.g., self-reporting, non-expert family members, or doctors)~\cite{Daniels16:Resource}.  How should a healthcare provider allocate time, resources, or interventions in the face of uncertainty?

Indeed, the \emph{multi-armed bandit} (MAB) paradigm has been used to great success throughout industry~\cite{bubeck2012regret,Slivkins19:Introduction} to address this style of problem and others, including recommendation systems~\cite{Bresler14:Latent}, revenue management~\cite{Ferreira18:Online}, and adaptive medical trials~\cite{Chow08:Adaptive}.  This mature literature has developed a variety of specific MAB settings based on the type and nature of feedback received, level of noise in the system, goal of the central decisionmaker, access to information, and other concerns.  These include: stochastic and nonstochastic (e.g., adversarial); (non-)combinatorial with various structural assumptions~\cite{Kveton14:Matroid}, exogeneous constraints~\cite{Balakrishnan19:Incorporating}, and different objectives~\cite{Chen14:Combinatorial}; contextual~\cite{Woodroofe79:One-armed,Auer02:Finite-time}; and, when feasible, combinations of the previous categories.

\noindent\textbf{Our contributions.}  To the best of our knowledge, no current MAB model addresses our earlier motivational settings, particularly the need to identify a best treatment even if it is evaluated in different sub-populations or contexts.  Thus, in this paper, we propose a novel contextual-bandit-based setting that balances particular \emph{local} and \emph{global} properties while remaining amenable to best arm identification analysis.  We see our primary contributions as follows:
\begin{itemize}
    \item We propose a new model, \modelAccroLong{} (\modelAccro{}), that adds the concept of \emph{global true utility} to contextual bandits, and we motivate this with a new application of best arm identification in a contextual bandit setting;
    \item We conduct a theoretical analysis of global best arm identification in contextual bandits within the \modelAccro{} setting, proving extensions of existing results; and
    \item We identify one scenario where intuition from stochastic-MAB best arm identification does not hold in \modelAccro{}. 
\end{itemize}

\section{Preliminaries \& Placement in the Literature}\label{sec:prelims}
In this section, we describe our model's connection to existing multi-armed bandit models, motivate the applications of the model, and set up preliminaries for our analysis in future sections.

{\bf Stochastic Bandits} \quad
We start with the classic multi-armed bandit problem.\footnote{We lightly overview a broad set of MAB-related concepts and settings.  The MAB literature is quite large and mature; toward that end, we recommend as starting points for learning more the survey due to~\cite{bubeck2012regret} and the recent introductory tutorial due to~\cite{Slivkins19:Introduction}.} Bandits have been an important tool in performing resource allocation in decision making under uncertainty. In the classic \emph{stochastic} multi-armed bandit problem with i.i.d.\ rewards~\cite{Lai85:Asymptotically}, an algorithm is given a set of $K$ arms, $A=\{a_i\}_{i=1}^K$ where each arm has a true utility $\mu_i\in[0,1]$. The algorithm is given some $T$ time steps where, at each time $t$, the algorithm makes a choice $I_t$ of an arm $a_i\in A$ and collects a reward derived i.i.d. from a $\sigma$-subgaussian distribution centered at $\mu_i$. An algorithm is usually evaluated by comparing it to the optimal action, meaning the arm with highest true utility: $\mu^\ast = \max_i \mu_i$. This quantity is called \emph{pseudo-regret} and can be written $\overline{R}_T = n\mu^\ast - \sum_{i=1}^T \bbE[\mu_{I_t}].$ For any given set of arms and corresponding true utilities, we often analyze how hard this environment is by looking at the suboptimality gap between the true utility of the optimal arm and each other arm: $\Delta_i = \Delta^\mu_i$. 

\ignore{{\bf Restless Bandits and Markov Decision Processes} \quad
In a restless bandit, we relax the assumption of stochastic rewards by letting  ...}

{\bf Contextual Bandits} \quad
In stochastic bandits, it is assumed that no information changes over time, the true utility of an arm is the same each time it is pulled.
With \emph{contextual} bandits, we assume that there is some side information available to the learner about each time step. The canonical application of contextual bandits is behavioral advertising (see, e.g.,~\cite{Langford08:Epoch-greedy} for the seminal casting of that problem into the stochastic contextual bandit setting, as well as~\cite{Beygelzimer11:Contextual} and~\cite{Agarwal14:Taming} for important follow-on works). When a user comes to a website, the learner may have contextual information about that user which can help inform which is the optimal action to take. In contextual bandits, we assume that each arm $a\in A$ has a $\sigma$-subgaussian reward distribution, $\nu_{a,x}$, for each context $x$ in a possible set of contexts $\mathcal{X}$. 
For a time horizon $T$, and for each time step $t$, the algorithm observes a context $x_t\in\mathcal{X}$, chooses an arm $a_t\in A$ and observes a reward $r_t\in[0,1]$ drawn from $\nu_{a_t,x_t}$. An analysis of an algorithm generally compares the choice $I_t$ of an arm at time $t$ with the arm that has highest expected reward in that given context, i.e., $\pi^\ast(x) = \max_a \bbE[\nu_{a,x}]$. 

The relaxation in contextual bandits is very natural for various advertising or behavioral modeling problems in online web-traffic.  A classic paper due to~\citet{Li10:Contextual} describes such a pipeline with respect to news article recommendation.  Contextual bandits, particularly as they are applied, impose two important additional assumptions: (1) the context at time $t$ is not known in advance, i.e., the system doesn't know the next individual coming to the website, and (2) the evaluation of an arm in one context has little meaning outside of that context, i.e., the end goal is for identifying the best action in a given contexts without regard for the connection between the contexts.

{\bf Toward Our Model} \quad To motivate our model, consider a relaxation of (1) and (2). We believe that the emphasis on unknown contexts in contextual bandits is limiting, insofar that there are growing scenarios where the context may be known a priori and where there is a desire for a global best arm. For example, problems in the realm of sequential decision making often perform tests with the same treatment in many different settings to determine the best outcome globally \cite{littman1996algorithms,roijers2013survey}. Hiring is one such setting where sequential decision making has a notion of global true utility and local estimates of that utility. One can view each interview round or interviewer of a potential applicant as a context or state in a multi-tiered decision making process. \citet{Schumann19:Making} even cast tiered hiring has as a multi-armed bandit problem which is a promising prospect for the application of our model (although they and others note that issues of fairness and bias would likely arise in such a setting as well~\cite{BoRi18a,schumann2020we}).

Further, the marketing literature has studied sequential product launches, for example, releasing a movie in a limited number of theatres, followed by a wide release, followed by an at-home release~\cite{Lim18:Optimal,Thorbjornsen20:Tomorrow}.  These models focus on the movement of a single item (e.g., movie) across multiple states (e.g., types of release), with a combined goal of learning the value of a particular state (e.g., expected revenue per unit of time) so as to transition between states and ultimately maximize utility (e.g., total revenue) over time.  Our model could be seen a complement to the sequential product launch problem; casting to that model, we would learn the value of arms (e.g., movies) in different contexts (e.g., geographic regions) before launch to identify globally optimal arms.

\section{Contextual Bandits with Global True Utility}\label{sec:model}
We now formalize the setting in which we operate. Let there be a set of arms $A=\{a_i\}_{i=1}^K$ where each arm  $i$ has a global true utility, $\mu_i\in [0,1]$. Let there be a set of states $\calS$ of size $S$. The states serve the same role as contexts in contextual bandits. Fix a state sequence $(s_t)$ for times $t=1,2,\dots,n$ such that $s_t\in \calS$. This sequence will govern the state of our system as time evolves. We assume that the state sequence is given to us in advance. 
 This is similar in spirit to the \emph{restless bandits} setting~\cite{Whittle88:Restless} with deterministic transitions, except for our model's integration with contexts, described below.

For each arm $i$, let $\nu_i$ be a distribution from which we will instantiate the reward means for each state. In particular, make $\bbE[\nu_i]=\mu_i$. We will sample from $\nu_i$ to derive the means of the arm pulls at different states. Given an arm $a_i$ and state $s$, let $m_{i,s}$, considered a local, state-based estimate of the global true utility, be an i.i.d.\ sample from $\nu_i$. 
\ignore{\jpd{Why is this state based? Index is only $i$ = arm?}\samuel{This is a misplaced modifier. hopefully it is more clear now}} Fix a distribution $\eta_{i,s}$ with mean $m_{i,s}$, and denote samples from $\eta_{i,s}$ as $X_{i,s}\sim \eta_{i,s}$. So at time $t$, if we choose arm $i$, we receive reward $X_{i,{s_t}}\sim\eta_{i,s_t}$. The description of the game is summarized in the following model definition.

\begin{figure}
    \centering
\noindent\fbox{%
    \parbox{\linewidth}{%
        \textbf{The \modelAccro{} Model}
        \\
        
        \textit{Known parameters}: Arms $A$, state space $S$, evolution sequence $(s_t)$, number of rounds $n\geq K$.\\
        \textit{Unknown parameters}: true rewards $\mu_i$; state-defining distributions, $\nu_i$, state-instantiated rewards, $m_{i,s}$; arm-state reward distributions, $\eta_{i,s}$.
        \\
        
        For each round $t=1,\dots, n$
        \begin{enumerate}
            \item The forecaster chooses arm $I_t\in A$,
            \item The environment chooses a reward $X_{I_t,s_t}\sim\eta_{I_t,s_t}$ iid and reveals it to the forecaster
        \end{enumerate}
    }%
}    
    \label{fig:model_desc}
\end{figure}

This setup is very similar to the setup for contextual bandits, except for two important distinctions: in this setting, (1) the state or context at a time $t$ may be known to the forecaster before time $t$ (or may even be chosen by the forecaster, which is a possibility discussed in our conclusion), and (2) we impose a concept of a true utility for a given arm. \ignore{In contextual bandits, the arm's utility is only meaningful within a given context or state. The forecaster is concerned with identifying the optimal action for a given context, not the evaluation of an arm's utility outside of the pre-determined contexts.} In our model, we assume that an arm \emph{does} have an underlying true utility, and that these contexts can be seen as approximations or samples from a distribution over possible different contexts for a broader, globally true utility for a given arm. This assumption opens the possibility to analyze best arm identification in contextual bandits---the main subject of the remainder of this paper. See \ref{BAI analysis} for our analysis.

We claim that our \modelAccro{} model behaves as regular contextual bandits do in the standard UCB setting~\cite{Auer02:Finite-time}. This comparison shows that our model is reasonably founded, aligning with existing theory on stochastic and contextual bandits. To build this theory out, we discuss regret in the \modelAccro{} first. 

We play the game described in the \modelAccro{} model against the optimal policy which has full knowledge of the $m_{i,s}$. With this knowledge, the optimal arm to pull at time $t$ is that which has largest $m_{i,s_t}$. Therefore, define
$$m^\ast_t = \max_{i=1,\dots,K} m_{i,s_t} \quad \text{and} \quad i^\ast_t = \argmax_{i=1,\dots,K} m_{i,s_t}.$$
$$m^\ast_s = \max_{i=1,\dots,K} m_{i,s} \quad \text{and} \quad i^\ast_s = \argmax_{i=1,\dots,K} m_{i,s}.$$

So if a forecaster chooses actions $I_t$ at each time step $t=1,\dots,n$, the pseudo-regret in the \modelAccro{} model is best defined as
$$\overline{R_n} = \sum_{t=1}^n m^\ast_t - \bbE\left[ \sum_{t=1}^n m_{I_t,s_t} \right].$$

As is the case with contextual bandits, observe that we can use a strategy that essentially works $S$ different UCB strategies in parallel. As in~\cite{bubeck2012regret}, assume that for all chosen $\eta_{i,s}$ the following moment conditions hold: there exists a convex function $\psi$ such that for all $\lambda\geq0$, if $X\sim \eta_{i,s}$,

{\tiny \begin{equation}\label{subgauss}
    \ln\left[ \bbE \left[ e^{\lambda(X-\bbE[X])}\right]\right] \leq \psi(\lambda) \quad \text{and} \quad \ln\left[ \bbE \left[ e^{\lambda(\bbE[X]-X)}\right]\right] \leq \psi(\lambda).
\end{equation}}

This generality is not strictly necessary, and can be reduced to distributions with compact support on $[0,1]$ by taking $\psi(\lambda) = \lambda^2/8$. Further, let $\psi^\ast$ be the Legendre–Fenchel transform of $\psi$ defined by $$\psi^\ast(\epsilon) = \sup_{\lambda\in\mathbb{R}}(\lambda\epsilon-\psi(\lambda)).$$

Consider the straightforward relaxation of the UCB strategy as follows: 
\\

\noindent\fbox{%
    \parbox{\linewidth}{%
        \textbf{The ($\alpha,\psi$) SB-UCB Strategy}
        \\
        
        For each round $t=1,\dots, n$, the forecaster chooses to pull an arm by

            \begin{align*}
                I_t\in\argmax_{i=1,\dots,K} \left[ \widehat{m}_{i,s_t,N_i^{s_t}(n)} + (\psi^\ast)^{-1}\left(  \frac{\alpha\ln t}{N_i^{s_t}(t-1)}\right)  \right]
            \end{align*}
    }%
}
\\

where $N_i^{s_t}(n)$ is the number of times the forecaster chose arm $i$ while in state $s_t$ by the $n^{\text{th}}$ round, and $\widehat{m}_{i,s_t,N_i^{s_t}(n)}$ is the sample mean for arm $i$ in stage $s_t$ after pulling it $N_i^{s_t}(n)$ times. We often will abuse notation and unambiguously write $\widehat{m}_{i,s,N_i^{s}(n)}$ as $\widehat{m}_{N_i^{s}(n)}$ when it is clear which arm and state we are considering. 

To analyze this strategy, we will use the standard concepts of hardness in stochastic multi-armed bandits, but we need to extend its definition to consider the states. Let $\Delta^m_{i,s}$ be the optimality gap for each arm at each stage defined as:
$$\Delta^m_{i,s}=m^\ast_s - m_{i,s}.$$ We observe that this optimally gap measures the difference between the highest sampled mean for a given state. In general, it need not be true that $m^\ast_s$ is also the arm with highest true utility. As we'll see in the proof below, this modified definition of an optimally gap is the proper extension of the standard UCB strategy to handle the \modelAccro{} setting. 

We now have everything needed to state and prove an upper bound on the pseudo-regret for the above strategy.

\begin{theorem}\label{double DAB UCB pseudoregret}
Assume that the reward distributions $\eta_{i,s}$ satisfy equation \ref{subgauss}, then the \textbf{$(\alpha,\psi)$ SB-UCB Strategy} with $\alpha>2$ satisfies:
\[
    \overline{R_n} \leq \sum_{i,s : \Delta^m_{i,s}>0} \Delta^m_{i,s}\left[\frac{\alpha\ln(n)}{\psi^\ast(\Delta^m_{i,s}/2)}  + \frac{\alpha}{\alpha-2}\right].
\]
\end{theorem}

\ignore{\jpd{Let's include an algorithm block (try to make it ``not too wide'', so we can wrapfigure it to 50\% of the column, with SB-UCB (name is a macro) so we can reference it in Thm 1 and elsewhere.}}

This theorem connects stochastic multi-armed bandits and contextual bandits. With $S=1$, Theorem \ref{double DAB UCB pseudoregret} is the same as Theorem 2.1 in \cite{bubeck2012regret}. However, the extension to contextual bandits with global true utility will allow us to take contextual bandits into the rich area of best arm identification. 

\section{Best Arm Identification Preliminaries}\label{sec:best-arm}

\begin{figure}
    \centering
    \resizebox{\linewidth}{!}{
        \tikzset{every picture/.style={line width=0.75pt}} 

\begin{tikzpicture}

\draw[domain=-3:3,smooth,variable=\x,blue] plot ({\x},{e^-(\x*\x/2)});

\draw    (0,.1) -- (0,-.1) ;
\draw (0,-.2) node [anchor=north][inner sep=0.75pt]   {$\mu _{1}$};

\draw    (1,.1) -- (1,-.1) ;
\draw (1,-.2) node [anchor=north][inner sep=0.75pt]   {$m _{1,1}$};

\draw    (1.75,.1) -- (1.75,-.1) ;
\draw (1.75,-.2) node [anchor=north][inner sep=0.75pt]   {$m _{1,2}$};

\draw (-3,.25) node [anchor=west][inner sep=0.75pt]   {$\nu _{1}$};

\draw [color=gray]    (-3,0) -- (3,0) ;
\draw [color=gray]    (3,.05) -- (3,-.05) ;
\draw [color=gray]    (2,.05) -- (2,-.05) ;
\draw [color=gray]    (1,.05) -- (1,-.05) ;
\draw [color=gray]    (0,.05) -- (0,-.05) ;
\draw [color=gray]    (-1,.05) -- (-1,-.05) ;
\draw [color=gray]    (-2,.05) -- (-2,-.05) ;
\draw [color=gray]    (-3,.05) -- (-3,-.05) ;

\draw[domain=4:10,smooth,variable=\x,blue] plot ({\x},{e^-((\x-8)*(\x-8)/2)});

\draw    (8,.1) -- (8,-.1) ;
\draw (8,-.2) node [anchor=north][inner sep=0.75pt]   {$\mu _{2}$};

\draw    (7.3,.1) -- (7.3,-.1) ;
\draw (7.3,-.2) node [anchor=north][inner sep=0.75pt]   {$m _{2,1}$};

\draw    (6.5,.1) -- (6.5,-.1) ;
\draw (6.5,-.2) node [anchor=north][inner sep=0.75pt]   {$m _{2,2}$};

\draw (4,.25) node [anchor=west][inner sep=0.75pt]   {$\nu _{2}$};

\draw [color=gray]    (-3+7,0) -- (3+7,0) ;
\draw [color=gray]    (7+3,.05) -- (7+3,-.05) ;
\draw [color=gray]    (7+2,.05) -- (7+2,-.05) ;
\draw [color=gray]    (7+1,.05) -- (7+1,-.05) ;
\draw [color=gray]    (7+0,.05) -- (7+0,-.05) ;
\draw [color=gray]    (7+-1,.05) -- (7+-1,-.05) ;
\draw [color=gray]    (7+-2,.05) -- (7+-2,-.05) ;
\draw [color=gray]    (7+-3,.05) -- (7+-3,-.05) ;

\end{tikzpicture}
    }
    \caption{Consider a scenario with two arms with means $\mu_1,\mu_2$, two states, and two random variables $\nu_1,\nu_2$ which select two state means for each arm as $m_{1,1},m_{1,2}\sim\nu_1$ and $m_{2,1},m_{2,2}\sim\nu_2$ depicted above. We see that while $\mu_1<\mu_2$, we have $\frac{m_{2,1}+m_{2,2}}{2} < \frac{m_{1,1}+m_{1,2}}{2}$. This illustrates the challenge when choosing a ``best'' arm. Choosing the arm with highest $\mu_i$ is the \emph{global best} reporting requirement. Choosing the arm with highest average $m_{i,s}$ across states is the \emph{empiric best} reporting requirement. In the above setting, $j^\ast = 2$ while $j^{\hat{\ast}}=1$.  }
    \label{fig:empiric vs global simple regret}
\end{figure}
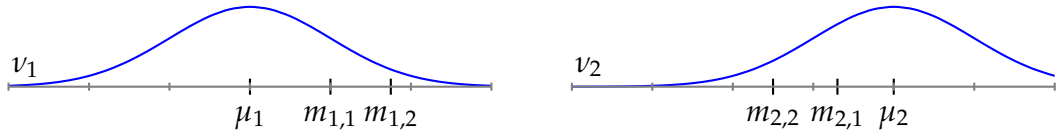

Using the global true utility setting described above, we begin to explore what it means to choose a best arm. 
In a standard stochastic bandit setting, the ``best'' arm is unambiguously defined: the best arm is that which has the highest true utility. 
But in this \modelAccro{} setting, there can be two meanings to the word ``best.''
First, we may be interested to report the arm with highest, global, true utility $\mu_i$. We think of this as an intrinsic property of an arm, which is, generally, quite unknown to an observer. We call this the {\bf global best} reporting requirement. 
However, in practice, the environment only observes some $S$ contexts or states of any arm's global true utility. This is limited by the draws of the $m_{i,s}$, as these values are picked by the environment (via $\nu_i$). Since the environment selects the number of states, $S$, as well as the $m_{i,s}$, another definition of ``best'' would be the arm with highest average $m_{i,s}$ over the states. Call this reporting requirement {\bf empiric best}. See Figure \ref{fig:empiric vs global simple regret} for a depiction of this difference. There may be other ways to describe what ``best'' means in this new setting; we will start the discussion with these two, as they are the most straightforward.

In notation, denote the global best arm as $j^\ast$ with true mean $\mu^\ast$ and empiric best arm as $i^{\hat{\ast}}$ with empiric mean $m^{\hat{\ast}}$ defined as
\begin{alignat*}{3}
    \text{global best:} \quad\quad j^\ast &= \argmax_{i=1,\dots,K} \mu_i  \\
    \mu^\ast &= \max_{i=1,\dots,K} \mu_i\\
    \text{empiric best:} \quad\quad j^{\hat{\ast}} &= \argmax_{i=1,\dots,K} \frac{1}{S}\sum_{s=1}^S m_{i,s} \\
    m^{\hat{\ast}} &= \max_{i=1,\dots,K} \frac{1}{S}\sum_{s=1}^S m_{i,s}.
\end{alignat*}

A recommendation strategy, $J_t$, is a choice for an arm that the forecaster makes at time $t$ to satisfy a particular reporting requirement. Recall that $I_t$ chooses the arm to pull at time $t$, while $J_t$ chooses the arm which it believes to be best arm. We then are interested in the quality of the recommendation strategy for each reporting requirement. We measure the quality of a recommendation strategy through two notions of simple regret. First, we have {\bf global simple regret}
\[
r_n = \mu^\ast - \mu_{J_n}
\]
which compares the best true mean to the chosen true mean. We also have {\bf empiric simple regret}
\[
\hat{r}_n = m^{\hat{\ast}} - \frac{1}{S}\sum_{s=1}^S m_{J_n,s}.
\]
which compares the highest sample estimate of true utility to the estimate of true utility for the chosen arm.  

\ignore{This section will be concerned with an analysis of the probability of choosing $\mu^\ast$ or $m^{\hat{\ast}}$ based off of different reporting and sampling strategies. The reporting strategy to satisfy both reporting requirements will often be the same, but the analysis of their performance may differ.}

\ignore{Observe that an analysis of global simple regret will require an expectation over $\nu_i$. Since the number of states, often, will be chosen by the operating environment, it is often not the case that the forecaster will be able to control $S$. As one of our main motivating examples of sequential decision making involves a small number of states, we will leave a detailed analysis of this concept for a later manuscript. However, there are instances when the number of states is large enough for meaningful analysis of global simple regret. Note that as $S$ increases, we can intuitively see that we will sample $\nu_i$ more frequently which will, in the limit, make the two concepts of global best and empiric best arms converge.}

Both types of simple regret are intricately related to the probability of using a recommendation strategy that chooses the \emph{wrong}  best arm. Accordingly, let ${e}_n$ be the probability that the choice from the recommendation strategy is not $j^{{\ast}}$, i.e., 
$${e}_n = \bbP(J_n \neq j^{{\ast}}),$$ 
and let $\hat{e}_n$ be the probability that the choice from the recommendation strategy is not $j^{\hat{\ast}}$, i.e., 
$$\hat{e}_n = \bbP(J_n \neq j^{\hat{\ast}}).$$ 

Observe that, as in the stochastic bandit case in  \cite{audibert2010best}, the behavior of $\hat{e}_n$ and $\hat{r}_n$ is similar up to a second order term, so we will analyze both below in Section \ref{BAI analysis}.  

\ignore{\jpd{Something I've found helpful when reading/writing MAB papers -- create a big tabular, two columns, with the symbol and then a brief (few wordS) explanation of that symbol, $\forall$ symbols in the paper.  We can throw that in the supplemental material---so just make a shell tabular now, reference it in the main paper, and then flesh it out between June 3rd and June 9th or whenever supp.\ material is due.}}

Recall from above, that the cumulative (pseudo-) regret problem hinges on identifying, for each stage $s$, the arm with highest $m_{i,s}$. For this problem, we defined the hardness as $\Delta^m_{i,s}$ where for all $s$ and for all $i\neq i_s^\ast$,  we put $\Delta^m_{i,s} = m_s^\ast - m_{i,s}$.

Alternatively, the empiric simple regret problem tries to choose the single arm with highest $\sum_{s} m_{i,s}$. This means that we are not interested in the highest $m_{i,s}$ for any stage; thus the hardness definition for cumulative regret is inappropriate for empiric simple regret. Here we define for each arm $i\neq j^{\hat{\ast}}$, an alternate version of hardness which captures the empiric estimate of $\mu_i$,
$$\Delta^\Sigma_i = \max_{j} \left[\frac{1}{S}\sum_{s=1}^S m_{j,s}\right] - \frac{1}{S}\sum_{s=1}^S m_{i,s} = m^{\hat{\ast}} - \frac{1}{S}\sum_{s=1}^S m_{i,s}.$$
We can also put $\Delta^\Sigma_{j^{\hat{\ast}}} = \min_{i\neq j^{\hat{\ast}}} \Delta^\Sigma_i$.

Even further, the optmality gap notions described above fail to capture how the global simple regret operates. A natural notion of optimality gap for global simple regret is defined for $i\neq j^\ast$ as
$$\Delta^\mu_i = \Delta^\mu_i.$$
We put $\Delta^\mu_{j^\ast} = \min_{i\neq j^\ast} \Delta^\mu_i.$

\ignore{We note that the arm with the maximal $m_{i,s}$ for a given stage $s$ doesn't necessarily correspond to the arm with highest average $m_{i,s}$. Therefore, we cannot relate the $\Delta^m_{i,s}$ to the $\Delta^\Sigma_i$ for a given arm $i$. To accommodate this, for an arm $i$,  define $\Delta_{i} = \min_{s}\Delta^m_{i,s}$ and define
$$\delta'_{i} = \min(\Delta^\Sigma_i, \Delta_{i}).$$}
The state sequence $(s_t)$ adds additional complexity presented in this new formulation of contextual bandits with global utility. Since this sequence is provided by the environment, we should expect that our analysis of a recommendation strategy will include a measure of complexity for the state sequence. As such, $\Sigma_s(t)$ be the number of times the environment has been in state $s$ by time $t$.

\section{Analysis of Best Arm Identification}\label{BAI analysis}

To analyze best arm identification in contextual bandits with global true utility, we will need to analyze an allocation strategy together with a recommendation strategy. We analyze a uniform allocation strategy with empiric best arm recommendation strategy; we report results for the global and empiric probability of error (Theorem \ref{Uniform + EBA}) and simple regret (Theorem \ref{Uniform + EBA, ETR}). Additionally, we analyze the global and empiric probability of error in a hybrid recommendation-allocation strategy called Successive Rejects (Theorem \ref{SR}). 

We begin with the simple uniform allocation strategy which will pull each arm a uniform number of times for each stage. 
\\

\noindent\fbox{%
    \parbox{\linewidth}{%
        \textbf{\modelAccro{} Uniform Allocation Strategy}
        \\
        
        For each round $t=1,\dots, n$, the forecaster chooses to pull an arm by
            \begin{align*}
                \Sigma_s(t) &\longleftarrow | \{ t'\leq t | s_{t'} = s\}|\\
                I_t &\longleftarrow \Sigma_{s_t}(t) \mod K
            \end{align*}
    }%
}
\\

We also begin with the simplest recommendation strategy: the empiric best arm which deterministically recommends the arm with highest empiric estimate of true utility. 
\\

\noindent\fbox{%
    \parbox{\linewidth}{%
        \textbf{Empiric Best Arm (EBA) Recommendation Strategy}
        \\
        
        For each round $t=1,\dots, n$, the forecaster recommends arm by
            \begin{align*}
                R_t \in \argmax_{i=1,\dots, K} \frac{1}{S}\sum_{s=1}^S \widehat{m}_{N^{s}_i(t)}
            \end{align*}
    }%
}
\\

When using these two strategies, we can bound the probability of error based off of the number of times a state has been visited and the hardness as measured by $\Delta^\Sigma_i$. We state this theorem with the general moment condition in (\ref{subgauss}). Proofs of all theorems can be found in the supplemental material.

\begin{theorem}[Uniform + EBA]\label{Uniform + EBA}
If we use a uniform allocation strategy and empiric best arm recommendation strategy, we can bound the probability of error of both the global best and empiric best arms with
\begin{equation*}
    {e}_n \leq \sum_{i,s} 2e^{-\lfloor\frac{\Sigma_s(n)}{K}\rfloor\psi^\ast\left(\frac{\Delta^\Sigma_i}{4}\right)} + \sum_{i} 2S\Phi\left(-\frac{\Delta^\mu_i}{4\sigma^2}\right)
    \tag{2.1} \label{myeq}
\end{equation*}
\begin{equation*}
    \hat{e}_n \leq \sum_{i=1}^K\sum_{s=1}^S 2e^{-\lfloor\frac{\Sigma_s(n)}{K}\rfloor\psi^\ast\left(\frac{\Delta^\Sigma_i}{2}\right)}
    \tag{2.2} \label{myeq}
\end{equation*}
\end{theorem}

Now we provide a bound directly on empiric simple regret for the same uniform allocation and EBA recommendation strategies. To do this, we will move to considering distributions $\nu_i,\eta_{i,s}$ with compact support on [0,1]. We make this additional assumption to aide in ease of proof, though these bounds could be made for a general compact interval $[a,b]$.

Now, we prove a bound on the expected global and empiric simple regret for a recommendation strategy $J_n$. Recall that 
$$\bbE[{r}_n] = \sum_{i=1}^K \Delta^\mu_i \bbP(J_n = i).$$
$$\bbE[\hat{r}_n] = \sum_{i=1}^K \Delta^\Sigma_i \bbP(J_n = i).$$

\begin{theorem}[Uniform + EBA]\label{Uniform + EBA, ETR}
Let all $\nu_i$ and $\eta_{i,s}$ have support almost surely on [0,1]. If we use a uniform allocation strategy and empiric best arm recommendation strategy, we can bound the empiric simple regret with 
\begin{equation*}
\bbE[{r}_n] \leq \sum_{i=1}^K\sum_{s=1}^S (\Delta^\mu_i) e^{-\lfloor\Sigma_s(n)/K\rfloor(m_{j^{{\ast}},s} - m_{i,s})^2}     \tag{3.1} \label{myeq}
\end{equation*}
\begin{equation*}
\bbE[\hat{r}_n] \leq \sum_{i=1}^K\sum_{s=1}^S \Delta^\Sigma_i e^{-\lfloor\Sigma_s(n)/K\rfloor(m_{j^{\hat{\ast}},s} - m_{i,s})^2}     \tag{3.2} \label{myeq}
\end{equation*}

\end{theorem}
Note that when $S=1$, the above theorem aligns with Proposition 1 of \cite{bubeck2011pure}.
Further, note that the statement of this theorem is in terms of the quantity $m_{j^\ast,s} - m_{i,s}$ and $m_{j^{\hat{\ast}},s} - m_{i,s}$ respectively. Note that this is not the same quantity as $\Delta^m_{i,s}$ as $m^\ast_s$ need not be equal to $m_{j^{\hat{\ast}},s}$ or $m_{j^\ast,s}$.






Finally, we analyze the Successive Rejects (SR) algorithm as it is applied to \modelAccro{}. Again, we see that the upper bound on the probability of error includes terms to accommodate the state sequence $(s_t)$. Recall that the successive rejects algorithm breaks a time horizon $n$ into $K-1$ rounds where the allocation of arm pulls are performed uniformly during any given round. At the end of a round, the arm with lowest estimate is removed from the pool of active arms. We now state the reframing of the SR algorithm in this setting:
\\

\noindent\fbox{%
    \parbox{\linewidth}{%
        \textbf{Successive Rejects (SR) in \modelAccro{}}
        \\

        Let $A_1 = \{1,\dots,K\}$. Separate our time horizon $n$ into $K-1$ rounds $R_1 = [1,t_1], R_2=[t_1+1, t_2], \dots, R_{K-1} = [t_{K-2}+1,n]$. 
        
        For each phase $k=1,\dots,K-1$, 
        \begin{enumerate}
            \item Identify the number of times a state $s$ is visited during round $k$: $\Sigma_s(t_k) - \Sigma_s(t_{k-1})$.
            \item Allocate the pulls evenly across the states for each arm.
            \item Let $A_{k+1} = A_k \setminus \{\argmin_{i\in A_k} \sum_s \widehat{m}_{N_i^s(t_k)}\}.$
        \end{enumerate}
        Let $J_n$ be the unique element in the set $A_K$.
    }%
}
\\

Denote $n_{s,k}$ as the \emph{total} number of times an arm is pulled in state $s$ by the end of round $k$, i.e., $n_{s,k} = N_i^S(t_k)$ for all the arm $i$ which was rejected at round $k$. We will note that since $(s_t)$ is fixed a priori in this setting, then $n_{s,k}$ could be negligible if we so happen to choose rounds which do not include visits to all the states enough times. As such, this variable appropriately appears in the upper bound on the probability of error.

\begin{theorem}[Successive Rejects Algorithm]\label{SR}
For the Successive Rejects algorithm, with $K-1$ rounds with chosen $t_1,\dots,t_{K-2}$, the probability of error is bounded by
\begin{equation*}
    {e}_n \leq \sum_{k=1}^{K-1} \sum_{s=1}^S k \exp\left\{-n_{s,k}(m_{j^{{\ast}},s} - m_{(K+1-k),s})^2\right\}
    \tag{4.2} \label{myeq}
\end{equation*}
\begin{equation*}
    \hat{e}_n \leq \sum_{k=1}^{K-1} \sum_{s=1}^S k \exp\left\{-n_{s,k}(m_{j^{\hat{\ast}},s} - m_{(K+1-k),s})^2\right\}
    \tag{4.1} \label{myeq}
\end{equation*}
\end{theorem}

\section{Experiments}
We now present results that complement the theoretical results given above. Below, we run simulations of the \modelAccro{} in various configurations and report interesting observations derived therefrom. From Theorems \ref{Uniform + EBA} and \ref{Uniform + EBA, ETR}, we see the importance of the $\Sigma_s(n)$ term in the upper bounds. Finally, we find a surprising result for our intuition of the SR algorithm which shows a unique difference from standard stochastic SR results.

\subsection{Uniform Allocation + EBA}

\begin{figure}
\centering
\includegraphics[width=0.45\linewidth]{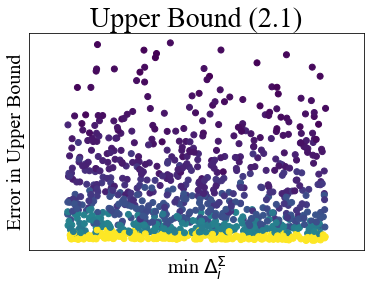}
\includegraphics[width=0.45\linewidth]{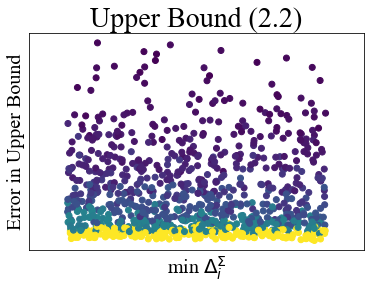}\\
\includegraphics[width=0.45\linewidth]{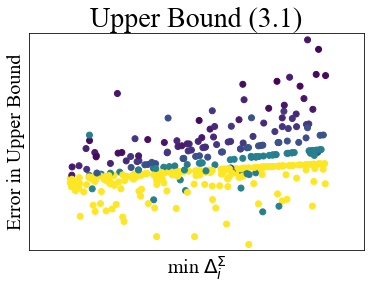}
\includegraphics[width=0.45\linewidth]{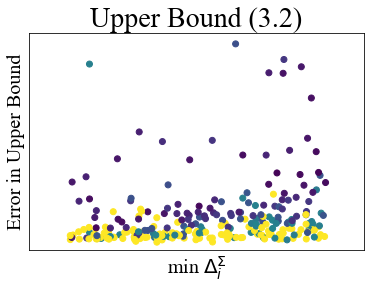}\\
\caption{Each plot shows how tight the bound given in statements 2.1, 2.2, 3.1, and 3.2. Each simulation is colored according to the number of times the model visited its least visited state, i.e., $\min_s \Sigma_s(n)$. Yellow represents a higher number of minimum visits. Dark purple represents a fewer number of minimum visits.}
\label{fig:uni+EBA simulation}
\end{figure}

In Theorems \ref{Uniform + EBA} and \ref{Uniform + EBA, ETR} we prove upper bounds on the global and empiric simple regret and probability for error. To analyze the complexities of these bounds, we simulated a \modelAccro{} model across a random set hyperparameters. We randomly chose hyperparemeters of models with between 3 and 10 arms, 1 and 10 states, and $\sigma^2$ between 0 and 0.3. Each $\mu_i$ was chosen uniformly across the unit interval and arm pulls were Bernoulli. We initialized 1,000 instantiations of the \modelAccro{} model and ran the Uniform Allocation and EBA algorithm 100 times for each. This allowed us to estimate $e_n,\hat{e}_n, r_n$, and $\hat{r}_n$ empirically and compare them against the bounds defined in the statements of Theorems \ref{Uniform + EBA} and \ref{Uniform + EBA, ETR}.

We plot the tightness of these four bounds compared to their empirically estimated values and plot these in Figure \ref{fig:uni+EBA simulation} against the minimum $\Delta^\Sigma_i$. Each simulation is colored according to the number of times the model visited its least visited state, i.e., $\min_s \Sigma_s(n)$.

Recall that in Theorem \ref{Uniform + EBA}, there are two terms, $\Sigma_s(n)$ and $\Delta^\Sigma_i$ in statement (2.2) with the addition of the term $\Delta^\mu_i$ in statement (2.1). The plots in Figure \ref{fig:uni+EBA simulation} clearly demonstrate that the dominating term is $\Sigma_s(n)$. This term uniformly impacts the error of the upper bound across all $\Delta^\Sigma_i$. This makes sense as the smaller that number is, the larger $e^{-\Sigma_s(n)}$ is, leading to a blow up of our uncertainty, i.e., the error in the upper bound. This same effect can be seen for the statements of Theorem \ref{Uniform + EBA, ETR} where the $\Sigma_s(n)$ still dominates uniformly across the $x$-axis of the lower two figures.

\subsection{Successive Rejects}
The above Theorem \ref{SR}, unlike the corresponding Theorem 2 of~\cite{audibert2010best}, does not tell us how to choose the $t_k$. The setting in \cite{audibert2010best} presents a method to choose the $t_k$ for the stochastic bandit because the state space does not alternate. In the \modelAccro{} model, we have less control over the states in which the arms are pulled, and correspondingly, less expressive power in the associated analysis. Nevertheless, we can still apply their approach to choosing the $t_k$. Alternatively, we could allocate the $t_k$ uniformly across the round $[0,n]$ by putting $t_k = \lceil k \frac{n}{K-1}\rceil$. We simulated these two strategies for choosing $t_k$ by randomly selecting all the model's hyperparameters over 10,000 simulations. 
\begin{figure}
\centering
\includegraphics[width=0.45\linewidth]{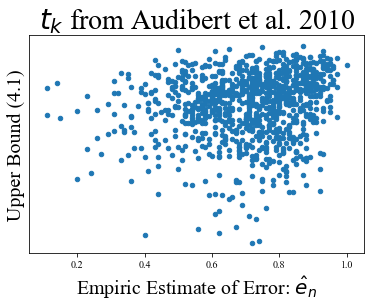}
\includegraphics[width=0.45\linewidth]{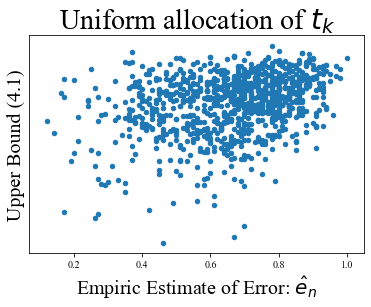}
\caption{Comparing simulations of Successive Rejects where the $t_k$ are chosen by the algorithm in \protect{\cite{audibert2010best}} (left) and where the $t_k$ are chosen uniformly across the interval $[0,n]$ (right).}
\label{fig:SR simulation}
\end{figure}

As can be seen in Figure~\ref{fig:SR simulation}, both data exhibit that the upper bound on the empiric estimates of $\hat{e}_n$ do indeed appear to be non-vacuous. Further, as $\hat{e}_n$ gets closer to 1, the upper bound gets tighter. Finally, we can see that the figures imply that the \cite{audibert2010best} approach has a higher probability of error than the uniform strategy. This result shows that the intuition about the SR algorithm in the purely stochastic setting does not hold in this model. This gives us further motivation for the necessity of better understanding this model and how it differs from existing results in stochastic MAB best arm identification.

\ignore{is possible in the standard stochastic bandit problem because the state space does not alternate. However, in our new model, we do not know the state sequence until it is given to us for a specific problem. The above Theorem provides an upper bound on this pessimistic scenario where the algorithm has no control over the state sequence. We believe that an important area of future work on this model will include relaxing this assumption and allowing the forecaster to allocate not just the arm to be pulled, but what state the arm should be in. We believe this will remove the dependence on the complexity of the state sequence from the above analysis.}

\section{Conclusions \& Future Research}\label{sec:conclusions}

In the above work, we have framed the question of best arm identification in a contextual bandit setting where each arm has a global true utility by way of a new model, \modelAccroLong{} (\modelAccro{}). Our results provide a first analysis in this new area, and we extend existing literature to generalize well-known results in best arm identification.  Of particular interest, we also uncover at least one setting where intuition from the traditional stochastic MAB literature does not hold in our new setting.

We see two main avenues for future research to extend our work. First, we assumed that the number of states was small, and this restricted our analysis of global simple regret, i.e., the ability to choose the arm which actually has the highest global true utility. However, this assumption can be relaxed to accommodate larger state sequences where global simple regret is practical. Finally, the above setting assumed that a state sequence was handed to the learner. This assumption was necessary to begin to build the theory on this problem. However, there are many practical settings where the learner has the ability to alter the state of an arm. Therefore, future work should be conducted to analyze algorithms which jointly decide where to allocate arm pulls while simultaneously deciding what the state of the arm should be.






\chapter{Adversarial Robustness} 

\label{chpt:AdvRobustness} 
This work was done in collaboration with my co-first author Vedant Nanda, as well as Sahil Singla, John P. Dickerson, and Soheil Feizi. See~\cite{nanda2021fairness}.

\section{Introduction}
Automated decision-making systems that are driven by data are being used in a variety of different real-world applications. In many cases, these systems make decisions on data points that represent humans (\eg, targeted ads~\cite{speicher2018potential,ribeiro2019microtagging}, personalized recommendations~\cite{singh2018fairness, beiga2018equity}, hiring~\cite{schumann2019diverse, schumann2020hiring}, credit scoring~\cite{khandani2010consumer}, or recidivism prediction~\cite{Chouldechova17:Fair}). In such scenarios, there is often concern regarding the fairness of outcomes of the systems~\cite{barocas2016big,sainyam2017fairness}. This has resulted in a growing body of work from the nascent Fairness, Accountability, Transparency, and Ethics (FATE) community that---drawing on prior legal and philosophical doctrine---aims to define, measure, and (attempt to) mitigate manifestations of unfairness in automated systems~\cite{Chouldechova17:Fair,Feldman15:Certifying,Leben20:Normative,Binns17:Fairness}.

Most of the initial work on fairness in machine learning considered notions that were one-shot and considered the model and data distribution to be static~\cite{zafar2019constraints, zafar2017parity, Chouldechova17:Fair, barocas2016big, dwork2012fairness,  zemel2013learning}. Recently, there has been more work exploring notions of fairness that are dynamic and consider the possibility that the world (\ie, the model as well as data points) might change over time~\cite{heidari2019longterm, heidari2018preventing, hashimoto2018fairness, liu2018delayed}. Our proposed notion of robustness bias has subtle difference from existing one-shot and dynamic notions of fairness in that it requires each partition of the population be equally robust to imperceptible changes in the input (\eg, noise, adversarial perturbations, etc).

We propose a simple and intuitive notion of \textit{robustness bias} which requires subgroups of populations to be equally ``robust.'' Robustness can be defined in multiple different ways~\cite{szegedy2013intriguing, goodfellow2014explaining, papernot2015limitations}. We take a general definition which assigns points that are farther away from the decision boundary higher robustness. Our key contributions are as follows:
\begin{itemize}
    \item We define a simple, intuitive notion of \textbf{\textit{robustness bias}} that requires all partitions of the dataset to be equally robust. We argue that such a notion is especially important when the decision-making system is a deep neural network (DNN) since these have been shown to be susceptible to various attacks~\cite{carlini2017towards,moosavi2016deepfool}. Importantly, our notion depends not only on the outcomes of the system, but also on the distribution of distances of data-points from the decision boundary, which in turn is a characteristic of \emph{both} the data distribution and the learning process.
    \item We propose different methods to \textbf{\textit{measure this form of bias}}. Measuring the exact distance of a point from the decision boundary is a challenging task for deep neural networks which have a highly non-convex decision boundary. This makes the measurement of robustness bias a non-trivial task. In this paper we leverage the literature on adversarial machine learning and show that we can efficiently approximate robustness bias by using adversarial attacks and randomized smoothing to get estimates of a point's distance from the decision boundary.
    \item We do an in-depth analysis of \textit{robustness bias} on popularly used datasets and models. Through \textit{\textbf{extensive empirical evaluation}} we show that unfairness can exist due to different partitions of a dataset being at different levels of robustness for many state-of-the art models that are trained on common classification datasets. We argue that this form of unfairness can happen due to both the data distribution and the learning process and is an important criterion to consider when auditing models for fairness.
\end{itemize}

\subsection{Related Work}

\xhdr{Fairness in ML} Models that learn from historic data have been shown to exhibit unfairness, \ie, they disproportionately benefit or harm certain subgroups (often a sub-population that shares a common sensitive attribute such as race, gender \etc) of the population~\cite{barocas2016big, Chouldechova17:Fair, khandani2010consumer}. This has resulted in a lot of work on quantifying, measuring and to some extent also mitigating unfairness~\cite{dwork2012fairness, dwork2018fairness, zemel2013learning, zafar2019constraints, zafar2017parity, hardt2016equality, nina2018beyond, tameem2019one, wadsworth2018achieving,Saha20:Measuring, donini2018empirical, calmon2017optimized, kusner2017counterfactual, Kilbertus17:Avoiding, Pleiss17:Fairness, Wang2020Towards}. Most of these works consider notions of fairness that are one-shot---that is, they do not consider how these systems would behave over time as the world (\ie, the model and data distribution) evolves. Recently more works have taken into account the dynamic nature of these decision-making systems and consider fairness definitions and learning algorithms that fare well across multiple time steps~\cite{heidari2019longterm, heidari2018preventing, hashimoto2018fairness, liu2018delayed}. We take inspiration from both the one-shot and dynamic notions, but take a slightly different approach by requiring all subgroups of the population to be equally robust to minute changes in their features. These changes could either be random (\eg natural noise in measurements)  or carefully crafted adversarial noise. This is closely related to~\citet{heidari2019longterm}'s effort-based notion of fairness; however, their notion has a very specific use case of societal scale models whereas our approach is more general and applicable to all kinds of models. Our work is also closely related to and inspired by Zafar et al.'s use of a regularized loss function which captures fairness notions and reduces disparity in outcomes~\cite{zafar2019constraints}. There are major differences in both the {\it approach} and {\it application} between our work and that of Zafar et al's. Their disparate impact formulation aims to equalize the average distance of points to the decision boundary, $\mathbb{E}[d(x)]$; our approach, instead, aims to equalize the number of points that are ``safe'', i.e.,  $\mathbb{E}[\mathbbm{1}\{d(x)>\tau\}]$ (see section~\ref{sec:robustness_bias} for a detailed description). Our proposed metric is preferable for applications of adversarial attack or noisy data, the focus of our paper; whereas the metric of Zafar et al is more applicable for an analysis of the consequence of a decision in a classification setting.

\xhdr{Robustness} Deep Neural Networks (DNNs) have been shown to be susceptible to carefully crafted adversarial perturbations which---imperceptible to a human---result in a misclassification by the model~\cite{szegedy2013intriguing, goodfellow2014explaining, papernot2015limitations}. In the context of our paper, we use adversarial attacks to approximate the distance of a data point to the decision boundary. For this we use state-of-the-art white-box attacks proposed by~\citet{moosavi2016deepfool} and~\citet{carlini2017towards}. Due to the many works on adversarial attacks, there have been many recent works on provable robustness to such attacks. The high-level goal of these works is to estimate a (tight) lower bound on the distance of a point from the decision boundary~\cite{cohen2019smoothing, salman2019provable, singla2020curvature}. We leverage these methods to estimate distances from the decision boundary which helps assess robustness bias (defined formally in Section~\ref{sec:robustness_bias}).

\xhdr{Fairness and Robustness} Recent works have proposed  poisoning attacks on fairness~\cite{solans2020poisoning, mehrabi2020exacerbating}.~\citet{khani2019noise} analyze why noise in features can cause disparity in error rates when learning a regression. We believe that our work is the very first to show that different subgroups of the population can have different levels of robustness which can lead to unfairness. We hope that this will lead to more work at the intersection of these two important sub fields of ML.

\begin{figure}[b!]
    \centering
        \includegraphics[width=\linewidth]{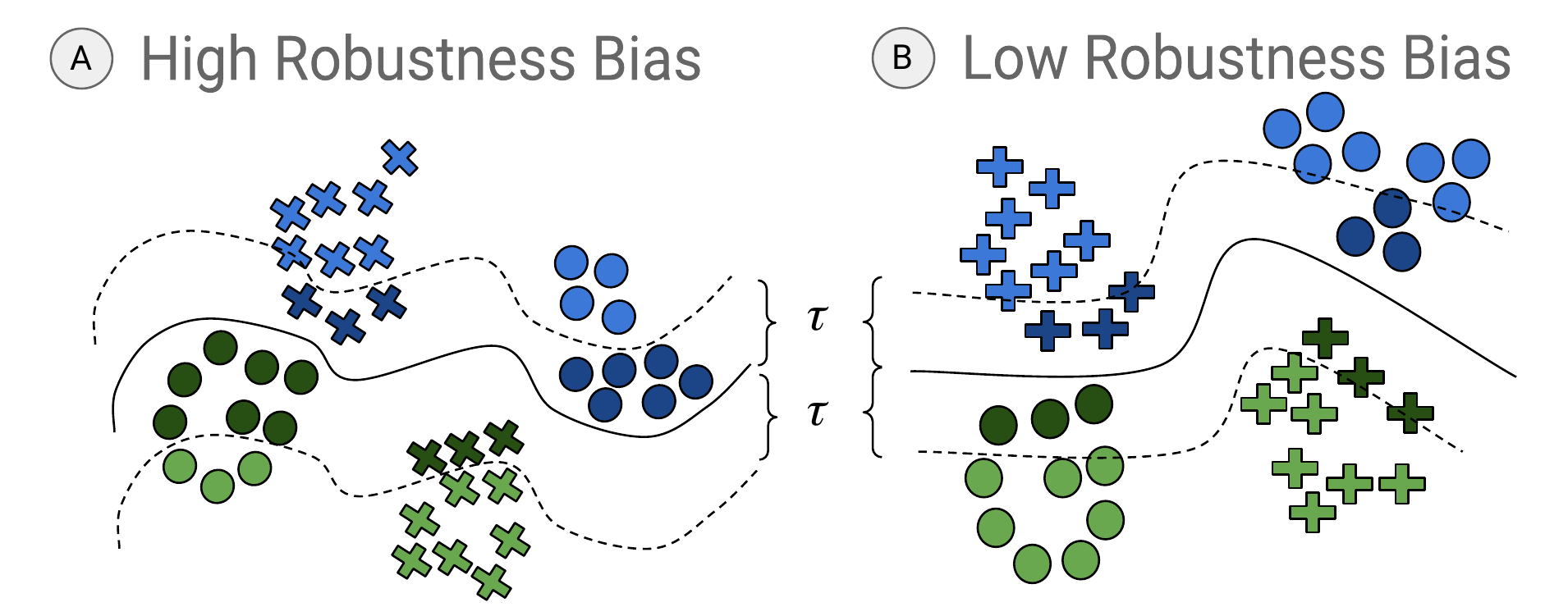}
        \caption{A toy example showing robustness bias. A.) the classifier (solid line) has $100\%$ accuracy for blue and green points. However for a budget $\tau$ (dotted lines), $70\%$ of points belonging to the ``round'' subclass (showed by dark blue and dark green) will get attacked while only $30\%$ of points in the ``cross'' subclass will be attacked. This shows a clear bias against the ``round'' subclass which is less robust in this case. B.) shows a different classifier for the same data points also with $100\%$ accuracy. However, in this case, with the same budget $\tau$, $30\%$ of both ``round'' and ``cross'' subclass will be attacked, thus being less biased.}
        \label{fig:fig1}
\vspace{-3mm}
\end{figure}

\begin{figure*}[t!]
    \centering
    \begin{subfigure}[t]{0.475\textwidth}
        \centering
        \includegraphics[width=.7\linewidth]{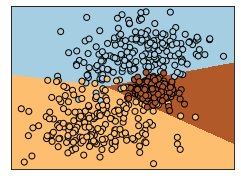}
        \caption{Three-class classification problem for randomly generated data.}
        \label{MLR-example}
    \end{subfigure}%
    ~ 
    \begin{subfigure}[t]{0.5\textwidth}
        \centering
        \includegraphics[width=.7\linewidth]{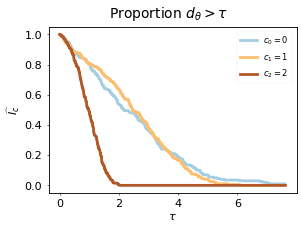}
        \caption{Proportion samples which are greater than $\tau$ away from a decision boundary.}
        \label{MLR-prop}
    \end{subfigure}
    \caption{An example of multinomial logistic regression.}
    \label{fig:MLR}
\end{figure*}

\section{Heterogeneous Robustness}

In a classification setting, a learner is given data $\mathcal{D} = \{(x_i,y_i)\}_{i=1}^N$ consisting of inputs $x_i\in\mathbb{R}^d$ and outputs $y_i\in\mathcal{C}$ which are labels in some set of classes $\mathcal{C}=\{c_1,\dots,c_k\}$. These classes form a partition on the dataset such that $\mathcal{D} = \bigsqcup_{c\in\mathcal{C}} \{(x_i,y_i) \mid y_i=c_j\}$. The goal of learning in decision boundary-based optimization is to draw delineations between points in feature space which sort the data into groups according to their class label. The learning generally tries to maximize the classification accuracy of the decision boundary choice. A learner chooses some loss function $\mathcal{L}$ to minimize on a training dataset, parameterized by parameters $\theta$, while maximizing the classification accuracy on a test dataset.

Of course there are other aspects to classification problems that have recently become more salient in the machine learning community. Considerations about the fairness of classification decisions, for example, are one such way in which additional constraints are brought into a learner's optimization strategy. In these settings, the data $\mathcal{D}=\{(x_i,y_i,s_i)\}_{i=1}^N$ is imbued with some metadata which have a sensitive attribute $\mathcal{S}=\{s_1,\dots,s_t\}$ associated with each point. Like the classes above, these sensitive attributes form a partition on the data such that $\mathcal{D} = \bigsqcup_{s\in\mathcal{S}} \{(x_i,y_i,s_i) \mid s_i = s\}$. Without loss of generality, we assume a single sensitive attribute. Generally speaking, learning with fairness in mind considers the output of a classifier based off of the partition of data by the sensitive attribute, where some objective behavior, like minimizing disparate impact or treatment~\cite{zafar2019constraints}, is integrated into the loss function or learning procedure to find the optimal parameters $\theta$.

There is not a one-to-one correspondence between decision boundaries and classifier performance. For any given performance level on a test dataset, there are infinitely many decision boundaries which produce the same performance, see Figure \ref{fig:fig1}. This raises the question: \emph{if we consider all decision boundaries or model parameters which achieve a certain performance, how do we choose among them? What are the properties of a desirable, high-performing decision boundary?} As the community has discovered, one \emph{undesirable} characteristic of a decision boundary is its proximity to data which might be susceptible to adversarial attack~\cite{goodfellow2014explaining, szegedy2013intriguing, papernot2015limitations}.  This provides intuition that we should prefer boundaries that are as far away as possible from example data ~\cite{suykens1999least, boser1992svm}.

Let us look at how this plays out in a simple example. In multinomial logistic regression, the decision boundaries are well understood and can be written in closed form. This makes it easy for us to compute how close each point is to a decision boundary. Consider for example a dataset and learned classifier as in Figure \ref{MLR-example}. For this dataset, we observe that the brown class, as a whole, is closer to a decision boundary than the yellow or blue classes. We can quantify this by plotting the proportion of data that are greater than a distance $\tau$ away from a decision boundary, and then varying $\tau$. Let $d_\theta(x)$ be the minimal distance between a point $x$ and a decision boundary corresponding to parameters $\theta$. For a given partition $\mathcal{P}$ of a dataset, $\mathcal{D}$, such that $\mathcal{D} = \bigsqcup_{P\in\mathcal{P}} P$, we define the function:  
\[\widehat{I_P}(\tau)=\frac{|\{(x,y)\in P \mid d_\theta(x) > \tau, y = \yhat\} |}{|P|}\]
If each element of the partition is uniquely defined by an element, say a class label, $c$, or a sensitive attribute label, $s$, we equivalently will write $\widehat{I_{c}}(\tau)$ or $\widehat{I_s}(\tau)$ respectively. We plot this over a range of $\tau$ in Figure \ref{MLR-prop} for the toy classification problem in Figure \ref{MLR-example}. Observe that the function for the brown class decreases significantly faster than the other two classes, quantifying how much closer the brown class is to the decision boundary.

\begin{figure*}[t!]
    \centering
        \includegraphics[width=0.75\linewidth]{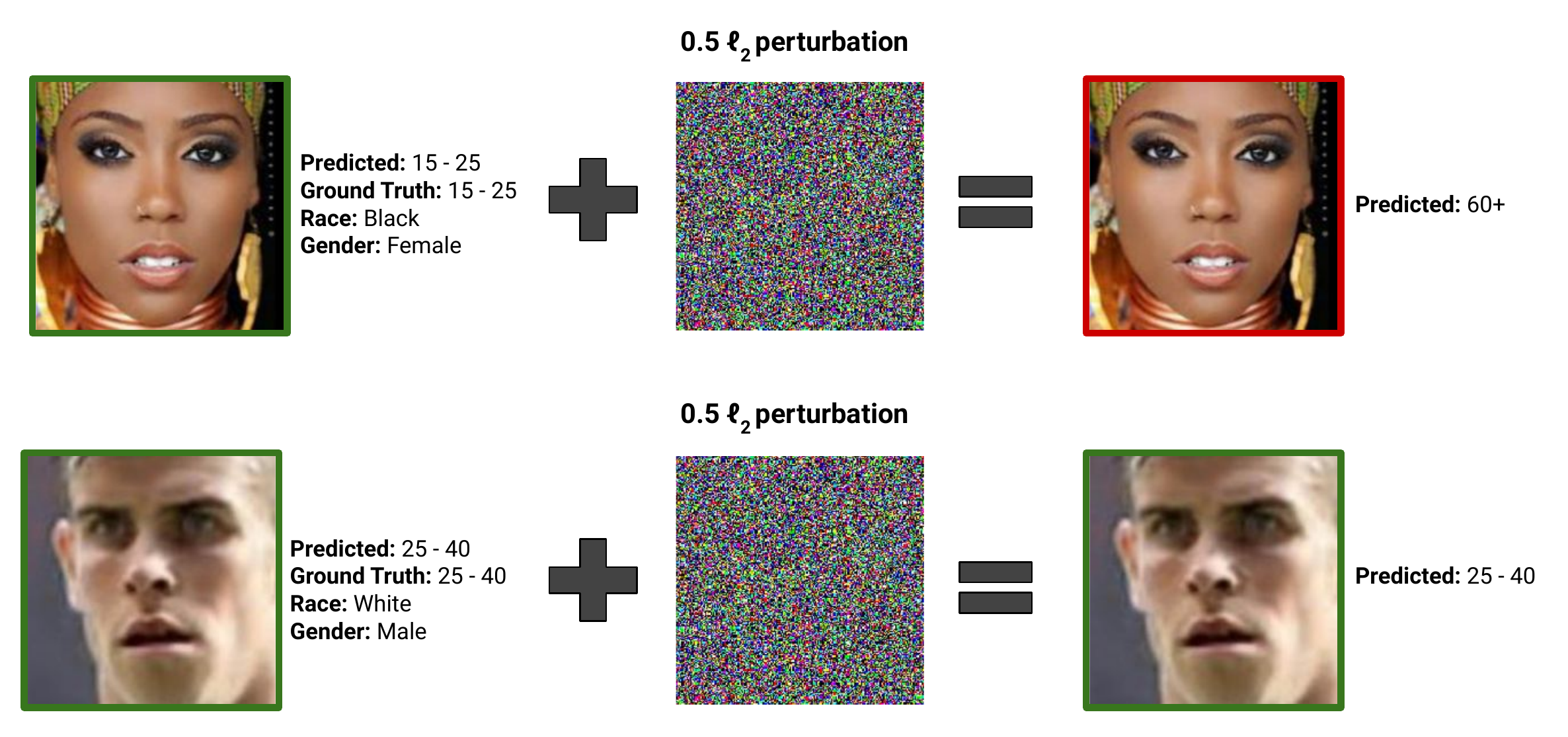}
        \caption{An example of robustness bias in the UTKFace dataset. A model trained to predict age group from faces is fooled for an inputs belonging to certain subgroups (black and female in this example) for a given perturbation, but is robust for inputs belonging to other subgroups (white and male in this example) for the \textit{same magnitude} of perturbation. We use the UTKFace dataset to make a broader point that robustness bias can cause harms. In the specific case of UTKFace (and similar datasets), the task definition of predicting age from faces itself is flawed, as has been noted in many previous studies~\cite{cramer2019challenges,crawford2019excavating,buolamwini2018gender}.}
        \label{fig:real_world_robustness_bias}
\end{figure*}

From a strictly classification accuracy point of view, the brown class being significantly closer to the decision boundary is not of concern; all three classes achieve similar classification accuracy. However, when we move away from this toy problem and into neural networks on real data, this difference between the classes could become a potential vulnerability to exploit, particularly when we consider adversarial examples.

\section{Robustness Bias}\label{sec:robustness_bias}


Our goal is to understand how susceptible different classes are to perturbations (e.g., natural noise, adversarial perturbations). Ideally, no one class would be more susceptible than any other, but this may not be possible.  We have observed that for the same dataset, there may be some classifiers which have differences between the distance of that partition to a decision boundary; and some which do not. There may also be one partition $\mathcal{P}$ which exhibits this discrepancy, and another partition $\mathcal{P}'$ which does not. Therefore, we make the following statement about robustness bias:

\begin{definition}\label{def:robustness-bias}
A dataset $\mathcal{D}$ with a partition $\mathcal{P}$ and a classifier parameterized by $\theta$ exhibits {\bf robustness bias} if there exists an element $P\in\mathcal{P}$ for which the elements of $P$ are either significantly closer to (or significantly farther from) a decision boundary than elements not in $P$.
\end{definition}

A partition $\mathcal{P}$ may be based on sensitive attributes such as race, gender, or ethnicity---or other class labels. For example, given a classifier and dataset with sensitive attribute ``race'', we might say that classifier exhibits robustness bias if, partitioning on that sensitive attribute, for some value of ``race'' the average distance of members of that particular racial value are substantially closer to the decision boundary than other members.

We might say that a dataset, partition, and classifier do not exhibit robustness bias 
if for all $P,P'\in\mathcal{P}$ and all $\tau>0$

\begin{equation}
\begin{split}
    \mathbb{P}_{(x,y)\in\mathcal{D}}\{d_\theta (x) > \tau \mid x\in P , y = \yhat \} \approx \\
    \mathbb{P}_{(x,y)\in\mathcal{D}}\{d_\theta (x) > \tau \mid x\in P', y = \yhat \}. 
\end{split}
\end{equation}

Intuitively, this definition requires that for a given perturbation budget $\tau$ and a given partition $P$, one should not have any incentive to perturb data points from $P$ over points that do not belong to $P$.
Even when examining this criteria, we can see that this might be particularly hard to satisfy. Thus, we want to quantify the disparate susceptibility of each element of a partition to adversarial attack, i.e., how much farther or closer it is to a decision boundary when compared to all other points. We can do this with the following function for a dataset $\mathcal{D}$ with partition element $P\in\mathcal{P}$ and classifier parameterized by $\theta$:

\begin{equation}
\begin{split}
    \RBmeasure{(P,\tau)} = \, | \, \mathbb{P}_{x\in\mathcal{D}}\{ d_\theta (x) > \tau \mid x\in P, y = \yhat\} - \\ \mathbb{P}_{x\in\mathcal{D}}\{d_\theta (x) > \tau \mid x\notin P, y = \yhat\} \, | 
\end{split}
\end{equation}

Observe that $\RBmeasure{(P,\tau)}$ is a large value if and only if the elements of $P$ are much more (or less) adversarially robust than elements not in $P$. 
We can then quantify this for each element $P\in\mathcal{P}$---but a more pernicious variable to handle is $\tau$. We propose to look at the area under the curve $\widehat{I_P}$ for all $\tau$:

\begin{equation}\label{eq:sigma}
    \sigma(P) = \frac{AUC(\widehat{I_P}) - AUC(\sum_{P'\neq P}\widehat{I_{P'}})}{AUC(\sum_{P'\neq P}\widehat{I_{P'}})}
\end{equation}

Note that these notions take into account the distances of data points from the decision boundary and hence are orthogonal and complementary to other traditional notions of bias or fairness (\eg, disparate impact/disparate mistreatment~\cite{zafar2019constraints}, etc). This means that having lower robustness bias does not necessarily come at the cost of fairness as measured by these notions. Consider the motivating example shown in Figure~\ref{fig:fig1}: the decision boundary on the right has lower robustness bias but preserves all other common notions (\eg~\cite{hardt2016equality, dwork2012fairness, zafar2017parity}) as both classifiers maintain $100\%$ accuracy. 

\begin{figure*}[t!]
    \centering

        \includegraphics[width=.9\linewidth]{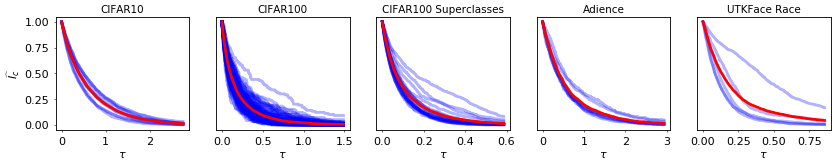}
        \caption{For each dataset, we plot $\widehat{I_c}(\tau)$ for each class $c$ in each dataset. Each blue line  represents one class. The red line represents the mean of the blue lines, i.e., $\sum_{c\in\mathcal{C}} \widehat{I_c}(\tau)$ for each $\tau$. }
        \label{fig:MLR}
\end{figure*}%
\begin{figure*}[t!]
        \centering
        \includegraphics[width=.9\linewidth]{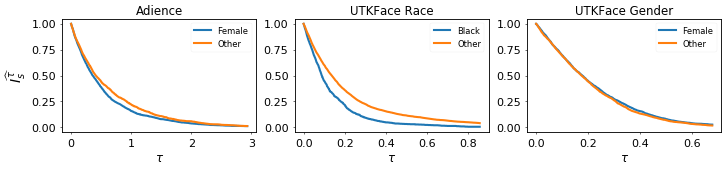}
        \caption{For each dataset, we plot $\widehat{I_s^\tau}$ for each sensitive attribute $s$ in each dataset.}
        \label{fig:MLR-sens}
\end{figure*}

\subsection{Real-world Implications: Degradation of Quality of Service}\label{subsec:robustness_bias_implications}

Deep neural networks are the core of many real world applications, for example, facial recognition, object detection, etc. In such cases, perturbations in the input can occur due to multiple factors such as noise due to the environment or malicious intent by an adversary. Previous works have highlighted how harms can be caused due to the degradation in quality of service for certain sub-populations~\cite{cramer2019challenges, holstein2019improving}. Figure~\ref{fig:real_world_robustness_bias} shows an example of inputs from the UTKFace dataset where an $\ell_2$ perturbation of $0.5$ could change the predicted label for an input with race ``black'' and gender ``female'' but an input with race ``white'' and gender ``male'' was robust to the same magnitude of perturbation. In such a case, the system worked better for a certain sub-group (white, male) thus resulting in unfairness. It is important to note that we use datasets such as Adience and UTKFace (described in detail in section~\ref{sec:experiments}) only to demonstrate the importance of having unbiased robustness. As noted in previous works, the very task of predicting age from a person's face is a flawed task definition with many ethical concerns~\cite{cramer2019challenges, buolamwini2018gender, crawford2019excavating}.

\section{Measuring Robustness Bias}

Robustness bias as defined in the previous section requires a way to measure the distance between a point and the (closest) decision boundary. For deep neural networks in use today, a direct computation of $d_\theta(x)$ is not feasible due to their highly complicated and non-convex decision boundary. However, we show that we can leverage existing techniques from the literature on adversarial attacks to efficiently approximate $d_\theta(x)$. We describe these in more detail in this section.

\subsection{Adversarial Attacks (Upper Bound)}\label{subsec:upper_bounds}


For a given input and model, one can compute an {\it upper bound} on $d_{\theta}(x)$ by performing an optimization which alters the input image slightly so as to place the altered image into a different category than the original. Assume for a given data point $x$, we are able to compute an adversarial image $\tilde{x}$, then the distance between these two images provides an upper bound on distance to a decision boundary, i.e, $\|x-\tilde{x}\| \geq d_\theta(x)$. 

We evaluate two adversarial attacks: DeepFool \cite{moosavi2016deepfool} and CarliniWagner's L2 attack \cite{carlini2017towards}. 
We extend $\widehat{I_P}$ for DeepFool and CarliniWagner as 
\begin{equation}
  \label{eq:measure_df}
  \widehat{I^{DF}_P} = \frac{| \{(x,y)\in P | \tau < \|x-\tilde{x}\|, y=\yhat\} |} { |P|}
\end{equation}
and
\begin{equation}
  \label{eq:measure_cw}
  \widehat{I^{CW}_P} = \frac{| \{(x,y)\in P | \tau < \|x-\tilde{x}\|, y=\yhat\} |} { |P|}
\end{equation}
respectively.
We use similar notation to define $\sigma^{DF}(P)$, 
and $\sigma^{CW}(P)$ ($\sigma$ as defined in Eq~\ref{eq:sigma}).  
While these methods are guaranteed to yield upper bounds on $d_\theta(x)$, they need not yield similar behavior to $\widehat{I_P}$ or $\sigma(P)$. We perform an evaluation of this in Section \ref{eval_of_measures}.

\subsection{Randomized Smoothing (Lower Bound)}\label{subsec:lower_bounds}
Alternatively one can compute a lower bound on $d_\theta(x)$ using techniques from recent works on training provably robust classifiers~\cite{salman2019provable, cohen2019smoothing}. For each input, these methods calculate a radius in which the prediction of $x$ will not change (i.e. the robustness certificate). In particular, we use the {\it randomized smoothing} method \cite{cohen2019smoothing,salman2019provable} since it is scalable to large and deep neural networks and leads to the state-of-the-art in provable defenses. Randomized smoothing transforms the base classifier $f$ to a new smooth classifier $g$ by averaging the output of $f$ over noisy versions of $x$. This new classifier $g$ is more robust to perturbations while also having accuracy on par to the original classifier. It is also possible to calculate the radius $\delta_x$ (in the $\ell_2$ distance) in which, with high probability, a given input's prediction remains the same for the smoothed classifier (i.e. $d_{\theta}(x)\geq \delta_x$). A given input $x$ is then said to be provably robust, with high probability, for a $\delta_x$ $\ell_2$-perturbation where $\delta_x$ is the robustness certificate of $x$.



For each point we use its $\delta_x$, calculated using the method proposed by~\cite{salman2019provable}, as a proxy for $d_{\theta}(x)$. The magnitude of $\delta_x$ for an input is a measure of how robust an input is. Inputs with higher $\delta_x$ are more robust than inputs with smaller $\delta_x$. Again, we extend $\widehat{I_P}$ for Randomized Smoothing as
\begin{equation}
  \label{eq:measure_rs}
  \widehat{I^{RS}_P} = \frac{| \{(x,y)\in P | \tau < \delta_x, y=\yhat\} |} { |P|}
\end{equation}

We use similar notation to define $\sigma^{RS}(P)$ (see Eq~\ref{eq:sigma}).

\section{Empirical Evidence of Robustness Bias in the Wild}\label{sec:experiments}

We hypothesize that there exist datasets and model architectures which exhibit robustness bias. To investigate this claim, we examine several image-based classification datasets and common model architectures.

\xhdr{Datasets and Model Architectures:} We perform these tests of the datasets {\bf CIFAR-10} \cite{krizhevsky2009learning}, {\bf CIFAR-100} \cite{krizhevsky2009learning} (using both 100 classes and 20 super classes), {\bf Adience} \cite{eidinger2014age}, and {\bf UTKFace} \cite{zhifei2017cvpr}. The first two are widely accepted benchmarks in image classification, while the latter two provide significant metadata about each image, permitting various partitions of the data by final classes and sensitive attributes.

Our experiments were performed using PyTorch's torchvision module~\cite{pytorch}.
We first explore a simple {\bf Multinomial Logistic Regression} model which could be fully analyzed with direct computation of the distance to the nearest decision boundary. For convolutional neural networks, we focus on {\bf Alexnet} \cite{krizhevsky2014one}, {\bf VGG19} \cite{simonyan2014very}, {\bf ResNet50} \cite{he2016deep}, {\bf DenseNet121} \cite{huang2017densely}, and \textbf{ Squeezenet1\_0} \cite{iandola2016squeezenet} which are all available through torchvision.
We use these models since these are widely used for a variety of tasks. We achieve performance that is comparable to state of the art performance on these datasets for these models. Additionally we also train some other popularly used dataset specific architectures like a deep convolutional neural network (we call this \textbf{Deep CNN})\footnote{\url{http://torch.ch/blog/2015/07/30/cifar.html}} and \textbf{PyramidNet} ($\alpha=64$, depth=110, no bottleneck)~\cite{han2017pyramidnet} for CIFAR-10. We re-implemented Deep CNN in pytorch and used the publicly available repo to train PyramidNet\footnote{\url{https://github.com/dyhan0920/PyramidNet-PyTorch}}. We use another deep convolutional neural network (which we refer to as \textbf{Deep CNN CIFAR100\footnote{https://github.com/aaron-xichen/pytorch-playground/blob/master/cifar/model.py}} and \textbf{PyramidNet} ($\alpha=48$, depth=164, with bottleneck) for CIFAR-100 and CIFAR-100Super. For Adience and UTKFace we additionally take simple deep convolutional neural networks with multiple convolutional layers each of which is followed by a ReLu activation, dropout and maxpooling. As opposed to architectures from torchvision (which are pre-trained on ImageNet) these architectures are trained from scratch on the respective datasets. We refer to them as \textbf{UTK Classifier} and \textbf{Adience Classifier} respectively. These simple models serve two purposes: they form reasonable baselines for comparison with pre-trained ImageNet models finetuned on the respective datasets, and they allow us to  analyze robustness bias when models are trained from scratch.

In sections~\ref{sec:adv_attacks} and~\ref{sec:randomized_smoothing} we audit these datasets and the listed models for robustness bias. In section~\ref{sec:exact_computation}, we train logistic regression on all the mentioned datasets and evaluate robustness bias using an exact computation. We then show in section~\ref{sec:adv_attacks} and~\ref{sec:randomized_smoothing} that robustness bias can be efficiently approximated using the techniques mentioned in \ref{subsec:upper_bounds} and \ref{subsec:lower_bounds} respectively for much more complicated models, which are often used in the real world. We also provide a thorough analysis of the types of robustness biases exhibited by some of the popularly used models on these datasets.

\section{Exact Computation in a Simple Model: Multinomial Logistic Regression}\label{sec:exact_computation}

We begin our analysis by studying the behavior of multinomial logistic regression. Admittedly, this is a simple model compared to modern deep-learning-based approaches; however, it enables is to explicitly compute the exact distance to a decision boundary, $d_\theta(x)$. We fit a regression to each of our vision datasets to their native classes and plot $\widehat{I_c}(\tau)$ for each dataset. Figure~\ref{fig:MLR} shows the distributions of $\widehat{I_c}(\tau)$, from which we observe three main phenomena: (1) the general shape of the curves are similar for each dataset, (2) there are classes which are significant outliers from the other classes, and (3) the range of support of the $\tau$ for each dataset varies significantly. We discuss each of these individually.

First, we note that the shape of the curves for each dataset is qualitatively similar. Since the form of the decision boundaries in multinomial logistic regression are linear delineations in the input space, it is fair to assume that this similarity in shape in Figure \ref{fig:MLR} can be attributed to the nature of the classifier.

Second, there are classes $c$ which indicate disparate treatment under $\widehat{I_c}(\tau)$. The treatment disparities are most notable in UTKFace, the superclass version CIFAR-100, and regular CIFAR-100. This suggests that, when considering the dataset as a whole, these outlier classes are less suceptible to adversarial attack than other classes. Further, in UTKFace, there are some classes that are considerably more susceptible to adversarial attack because a larger proportion of that class is closer to the decision boundaries.

We also observe that the median distance to decision boundary can vary based on the dataset. The median distance to a decision boundary for each dataset is: 0.40 for CIFAR-10;
0.10 for CIFAR-100; 0.06 for the superclass version of CIFAR-100; 0.38 for Adience; and 0.12 for UTKFace.
This is no surprise as $d_\theta(x)$ depends both on the location of the data points (which are fixed and immovable in a learning environment) and the choice of architectures/parameters.  \ignore{\jpd{Do I know if 0.40 is big or small at this point?  Can I compare distances across architectures and/or datasets?}}

Finally, we consider another partition of the datasets. Above, we consider the partition of the dataset which occurs by the class labels. With the Adience and UTKFace datasets, we have an additional partition by sensitive attributes. Adience admits partitions based off of gender; UTKFace admits partition by gender and ethnicity. We note that Adience and UTKFace use categorical labels for these multidimensional and socially complex concepts. We know this to be reductive and serves to minimize the contextualization within which race and gender derive their meaning \cite{hanna2020towards,buolamwini2018gender}. Further, we acknowledge the systems and notions that were used to reify such data partitions and the subsequent implications and conclusions draw therefrom. We use these socially and systemically-laden partitions to demonstrate that the functions we define, $\widehat{I_P}$ and $\sigma$ depend upon how the data are divided for analysis. To that end, the function $\widehat{I_P}$ is visualized in Figure \ref{fig:MLR-sens}. We observe that the Adience dataset, which exhibited some adversarial robustness bias in the partition on $\mathcal{C}$ only exhibits minor adversarial robustness bias in the partition on $\mathcal{S}$ for the attribute `Female'. On the other hand, UTKFace which had signifiant adversarial robustness bias does exhibit the phenomenon for the sensitive attribute `Black' but not for the sensitive attribute `Female'.

This emphasizes that adversarial robustness bias is dependant upon the dataset and the partition. We will demonstrate later that it is also dependant on the choice of classifier. First, we talk about ways to approximate $d_\theta(x)$ for more complicated models.

\begin{figure*}[h!]
    \centering
    \begin{subfigure}[b]{0.23\textwidth}
        \includegraphics[trim={0cm 0cm 0cm 0cm},clip,width=1\textwidth]{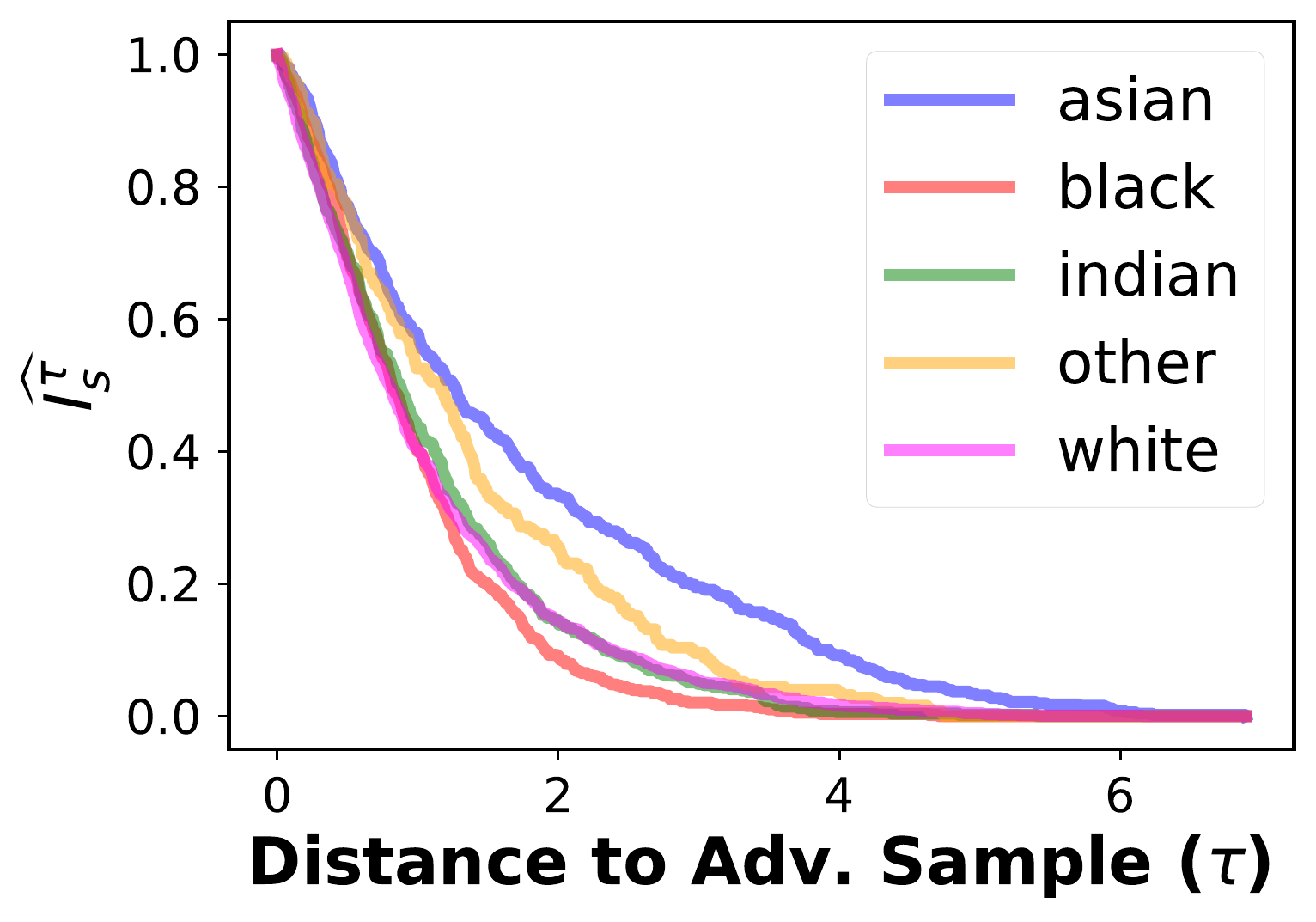}
        \caption{UTK Classifier: DeepFool}
        \label{fig:utkface_race_utk_classifier_df}
    \end{subfigure}
    \begin{subfigure}[b]{0.23\textwidth}
        \includegraphics[trim={0cm 0cm 0cm 0cm},clip,width=1\textwidth]{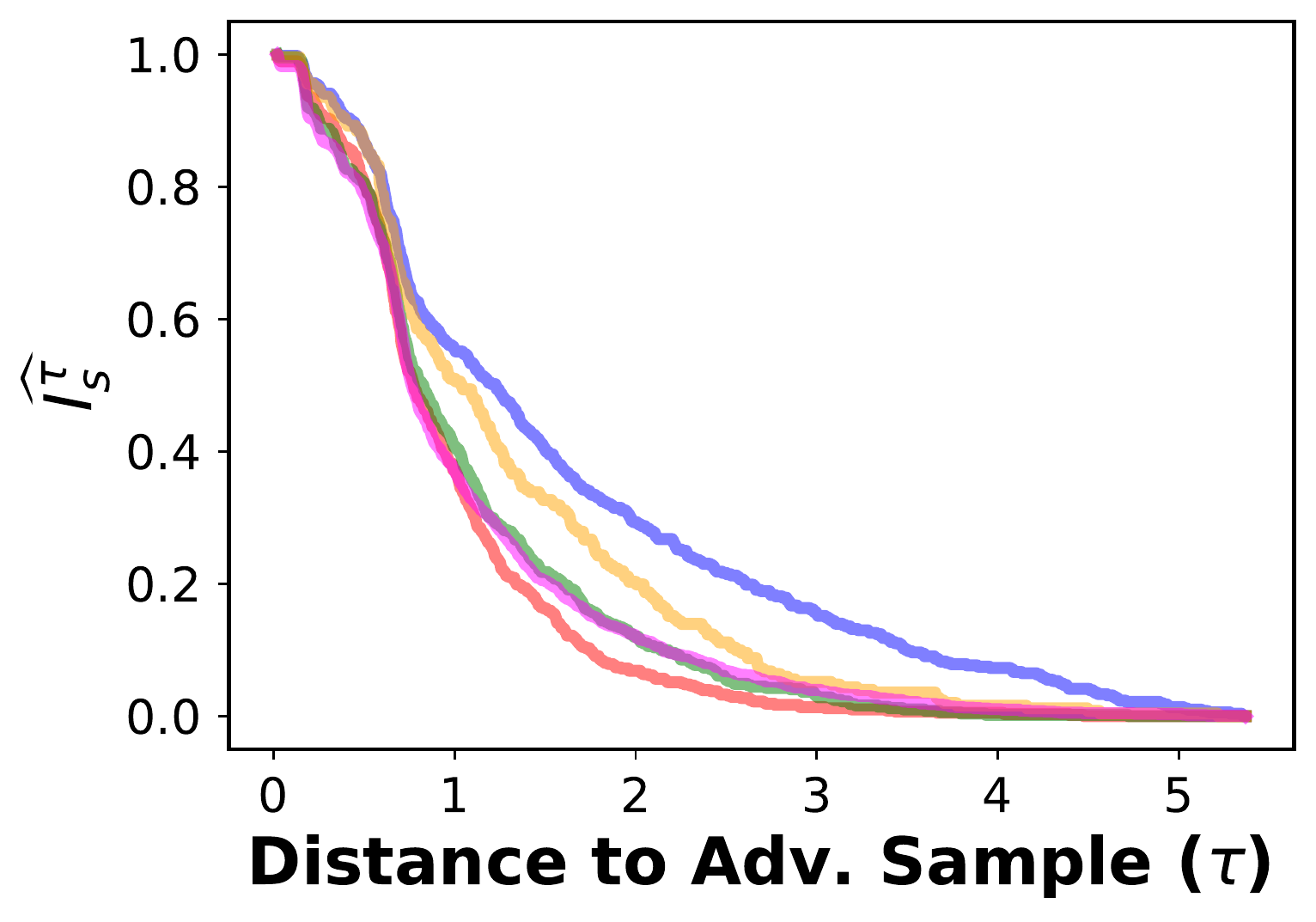}
        \caption{UTK Classifier: CarliniWagner}
        \label{fig:utkface_race_utk_classifier_cw}
    \end{subfigure}
    \begin{subfigure}[b]{0.23\textwidth}
        \includegraphics[trim={0cm 0cm 0cm 0cm},clip,width=1\textwidth]{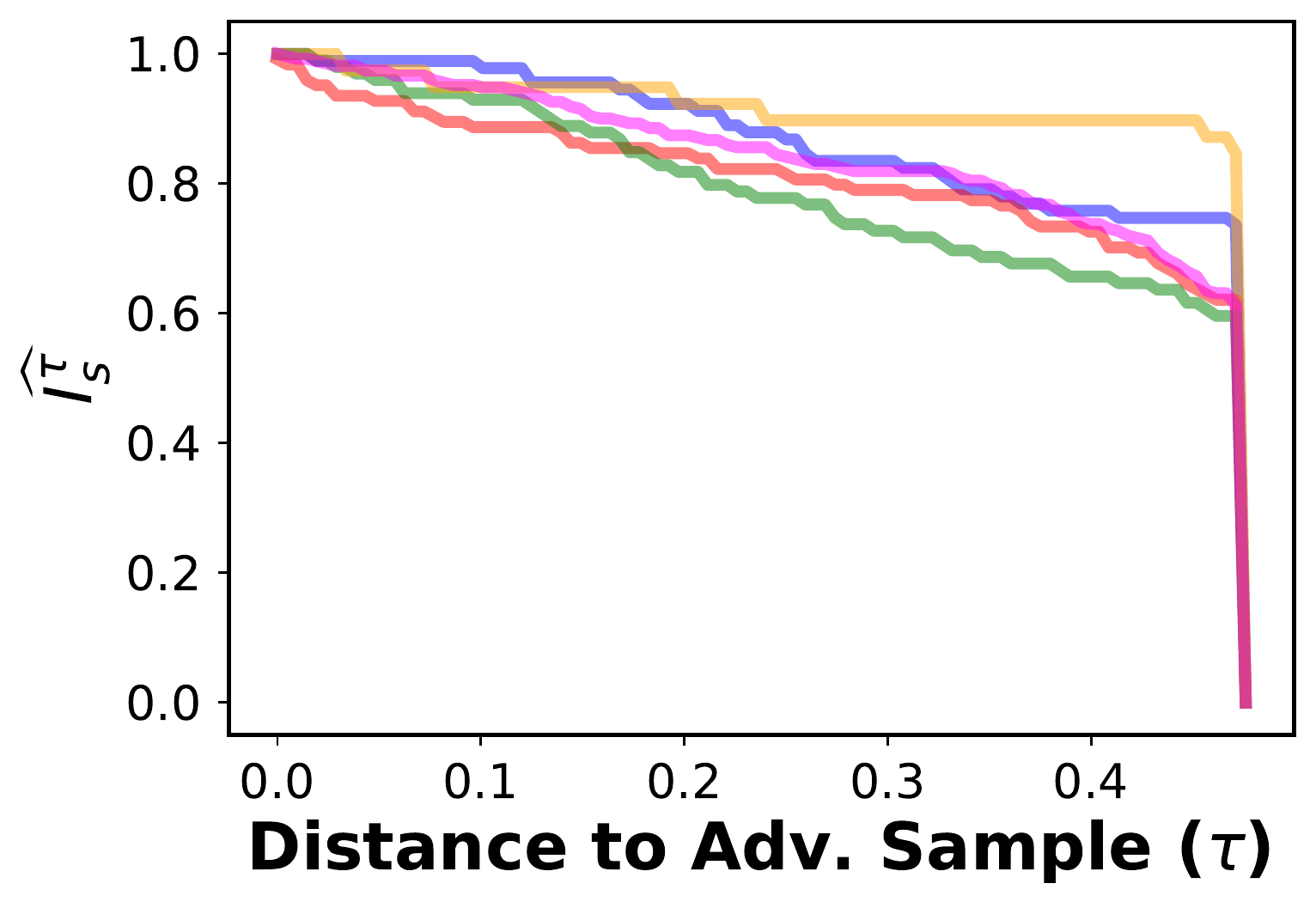}
        \caption{UTK Classifier: Rand. Smoothing}
        \label{fig:lb_utkface_race_utk_classifier}
    \end{subfigure}

    \begin{subfigure}[b]{0.23\textwidth}
        \includegraphics[trim={0cm 0cm 0cm 0cm},clip,width=1\textwidth]{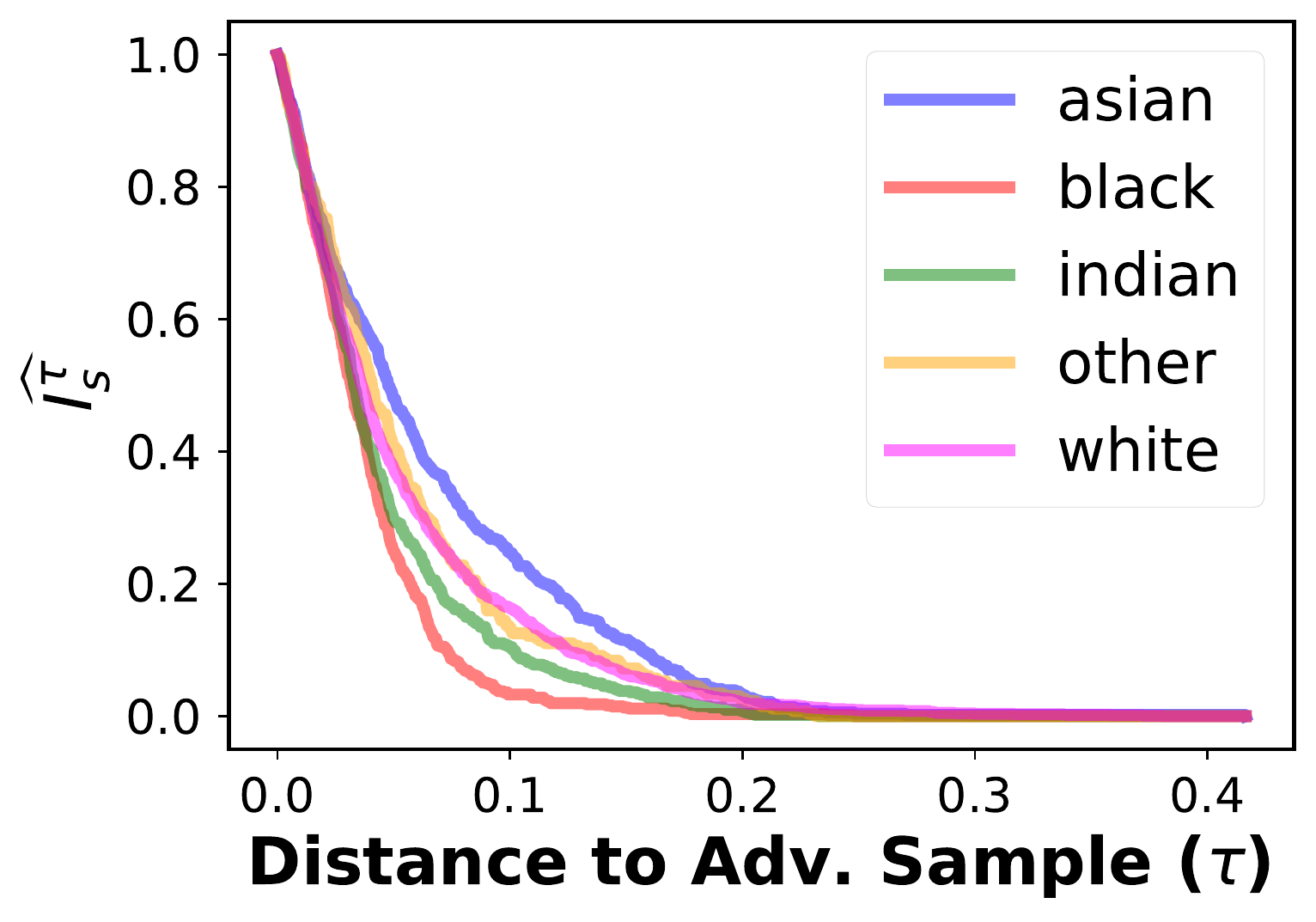}
        \caption{ResNet50: DeepFool}
        \label{fig:utkface_race_resnet_df}
    \end{subfigure}
    \begin{subfigure}[b]{0.23\textwidth}
        \includegraphics[trim={0cm 0cm 0cm 0cm},clip,width=1\textwidth]{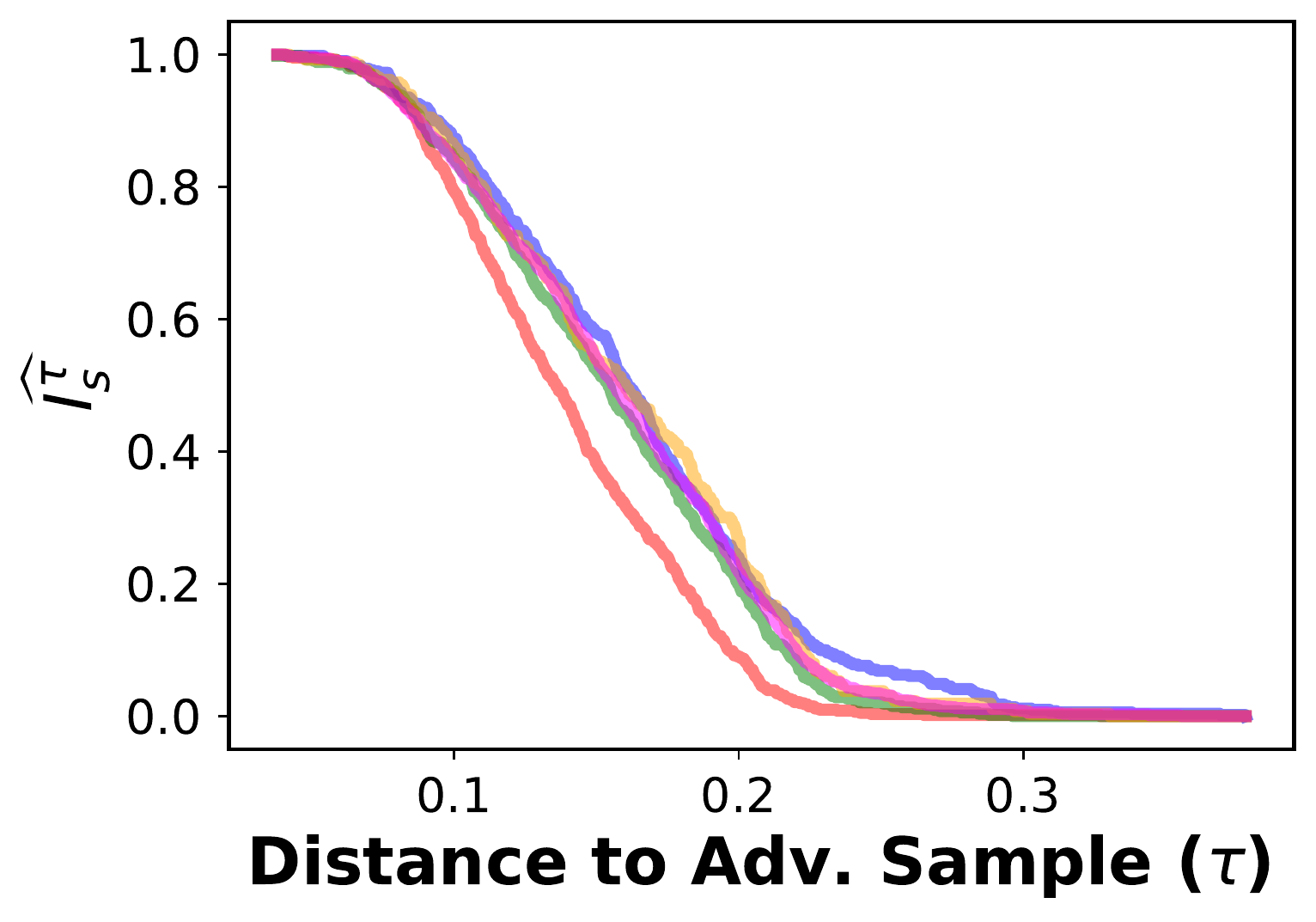}
        \caption{ResNet50: CarliniWagner}
        \label{fig:utkface_race_resnet_cw}
    \end{subfigure}
    \begin{subfigure}[b]{0.23\textwidth}
        \includegraphics[trim={0cm 0cm 0cm 0cm},clip,width=1\textwidth]{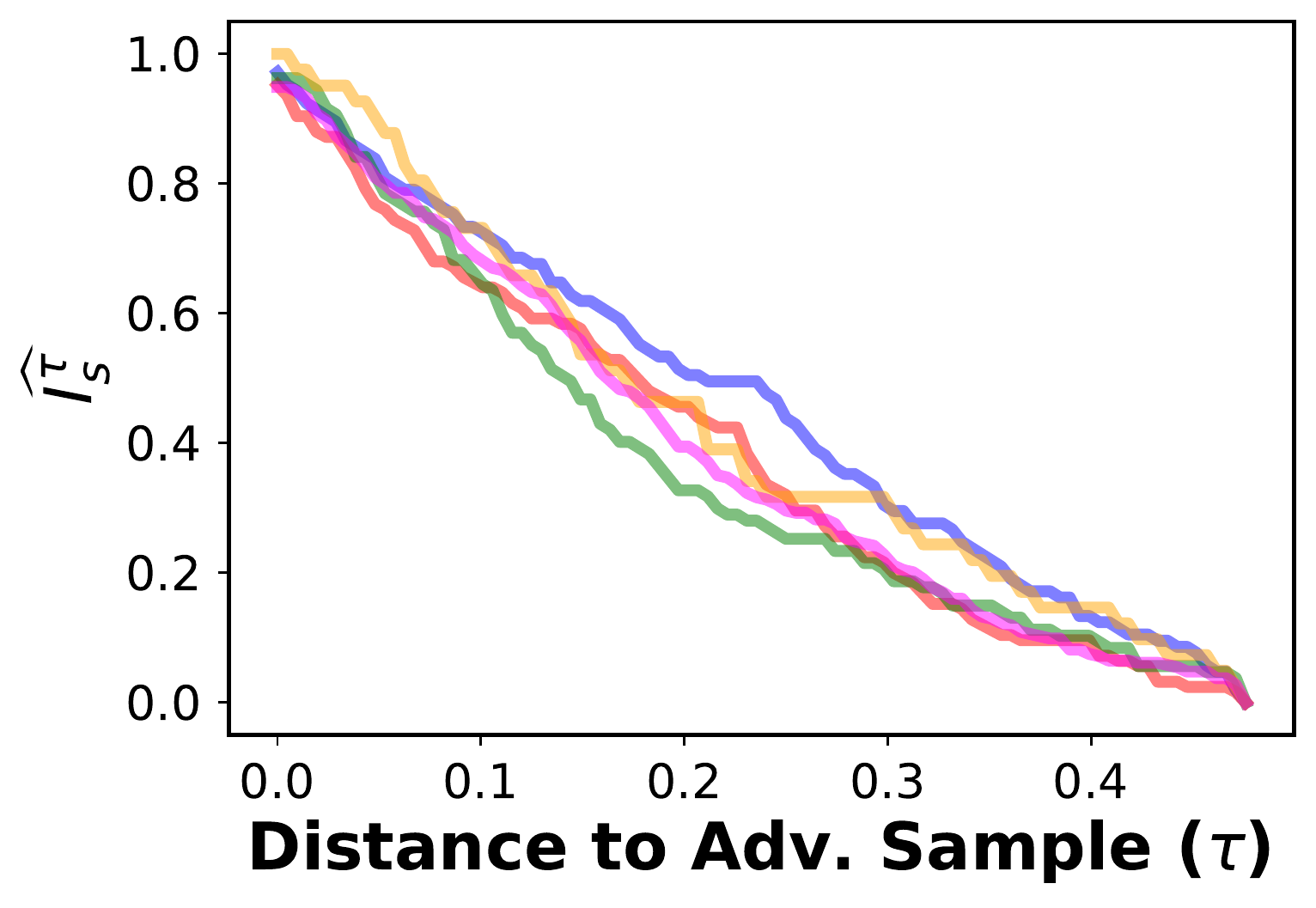}
        \caption{ResNet50: Rand. Smoothing}
        \label{fig:lb_utkface_race_resnet}
    \end{subfigure}

    \begin{subfigure}[b]{0.23\textwidth}
        \includegraphics[trim={0cm 0cm 0cm 0cm},clip,width=1\textwidth]{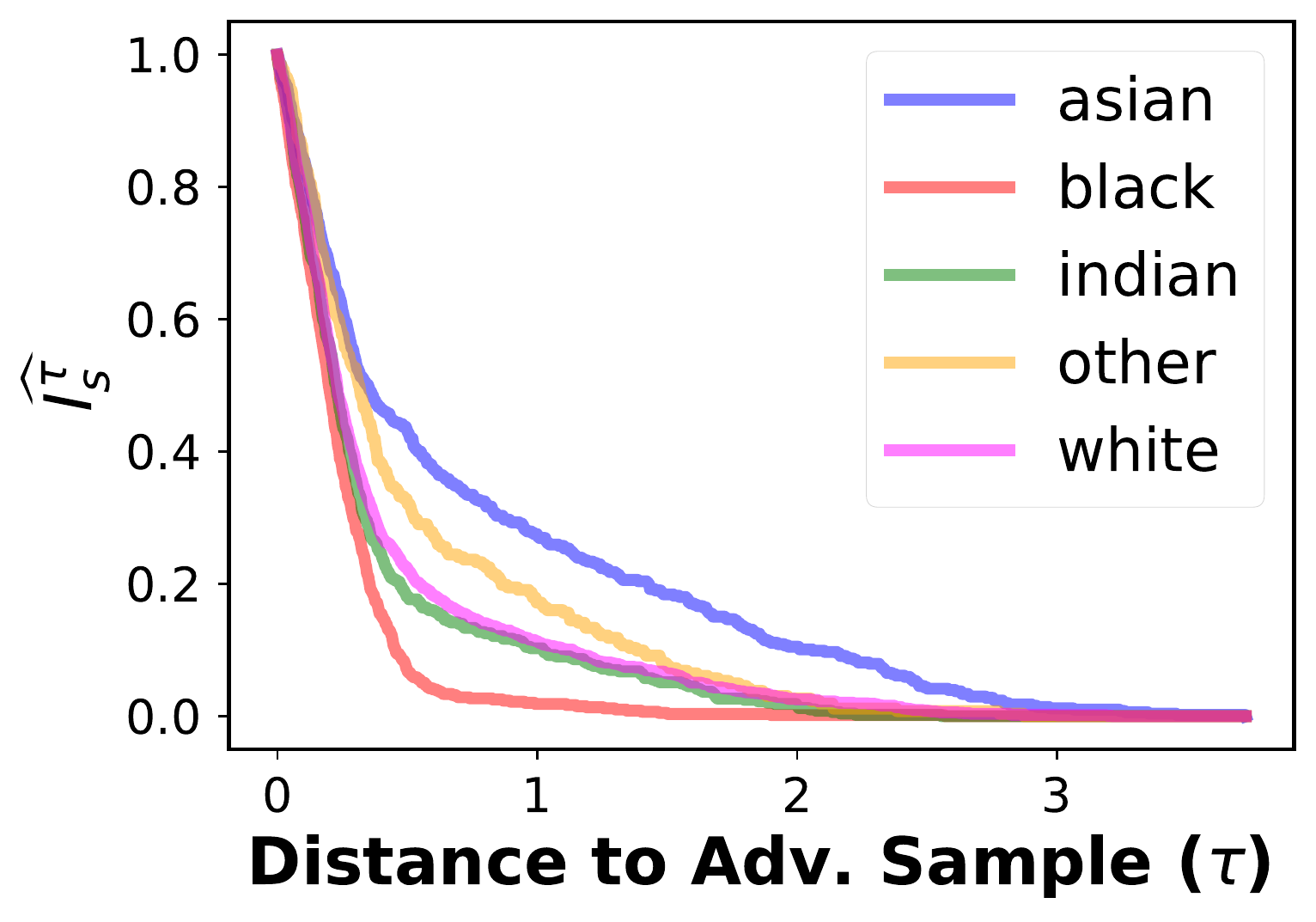}
        \caption{Alexnet: DeepFool}
        \label{fig:utkface_race_alexnet_df}
    \end{subfigure}
    \begin{subfigure}[b]{0.23\textwidth}
        \includegraphics[trim={0cm 0cm 0cm 0cm},clip,width=1\textwidth]{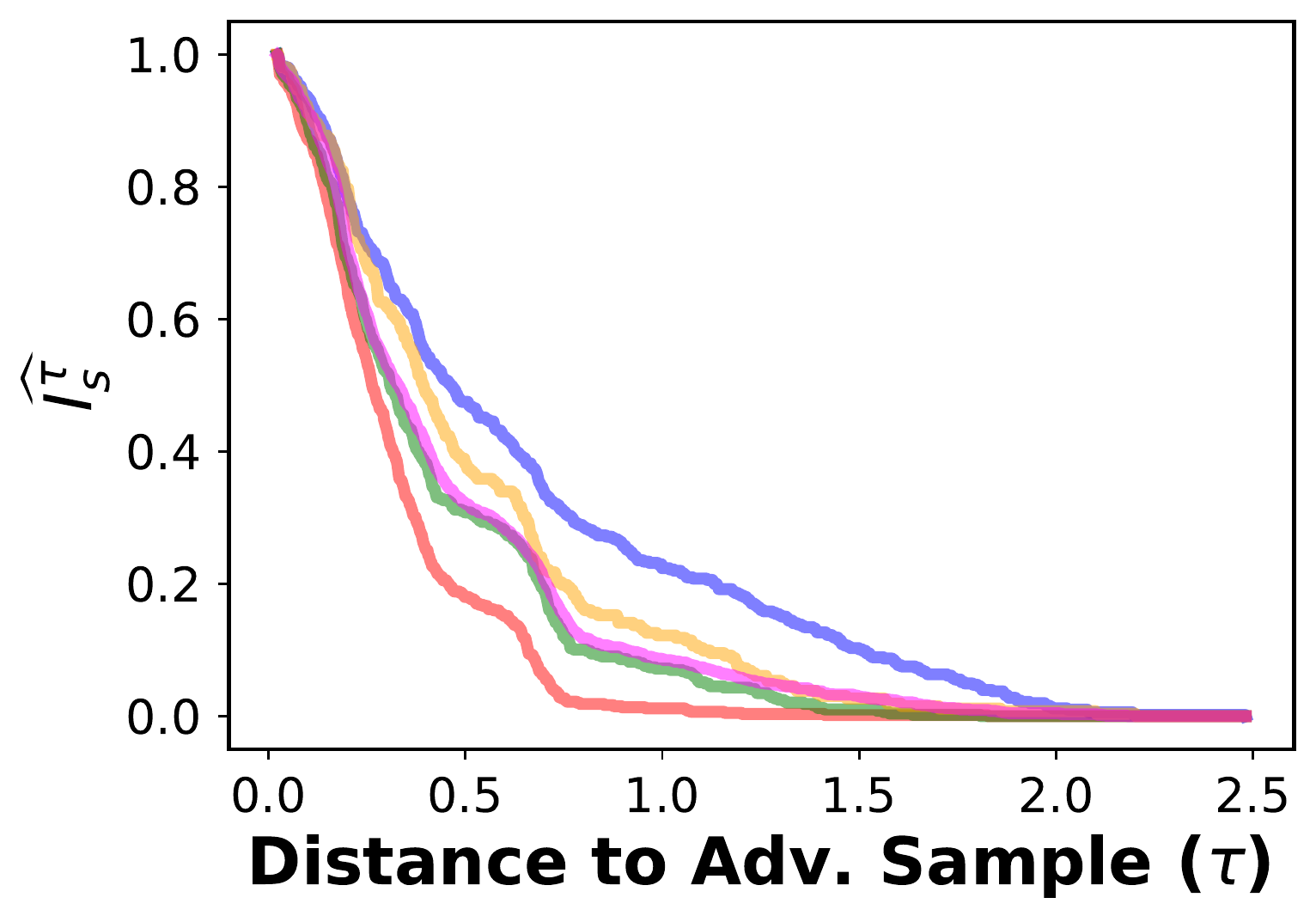}
        \caption{Alexnet: CarliniWagner}
        \label{fig:utkface_race_alexnet_cw}
    \end{subfigure}
    \begin{subfigure}[b]{0.23\textwidth}
        \includegraphics[trim={0cm 0cm 0cm 0cm},clip,width=1\textwidth]{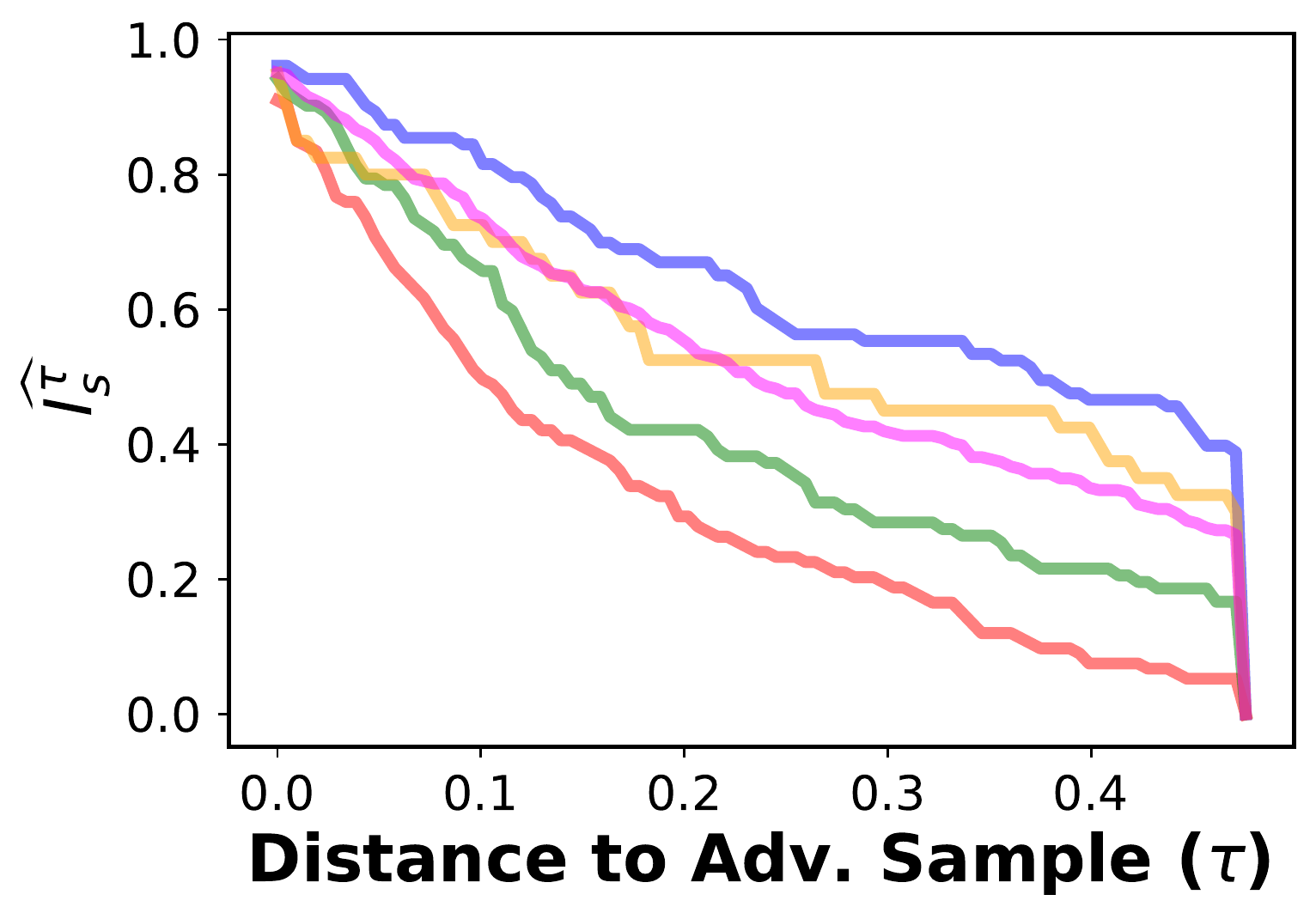}
        \caption{Alexnet: Rand. Smoothing}
        \label{fig:lb_utkface_race_alexnet}
    \end{subfigure}

    \begin{subfigure}[b]{0.23\textwidth}
        \includegraphics[trim={0cm 0cm 0cm 0cm},clip,width=1\textwidth]{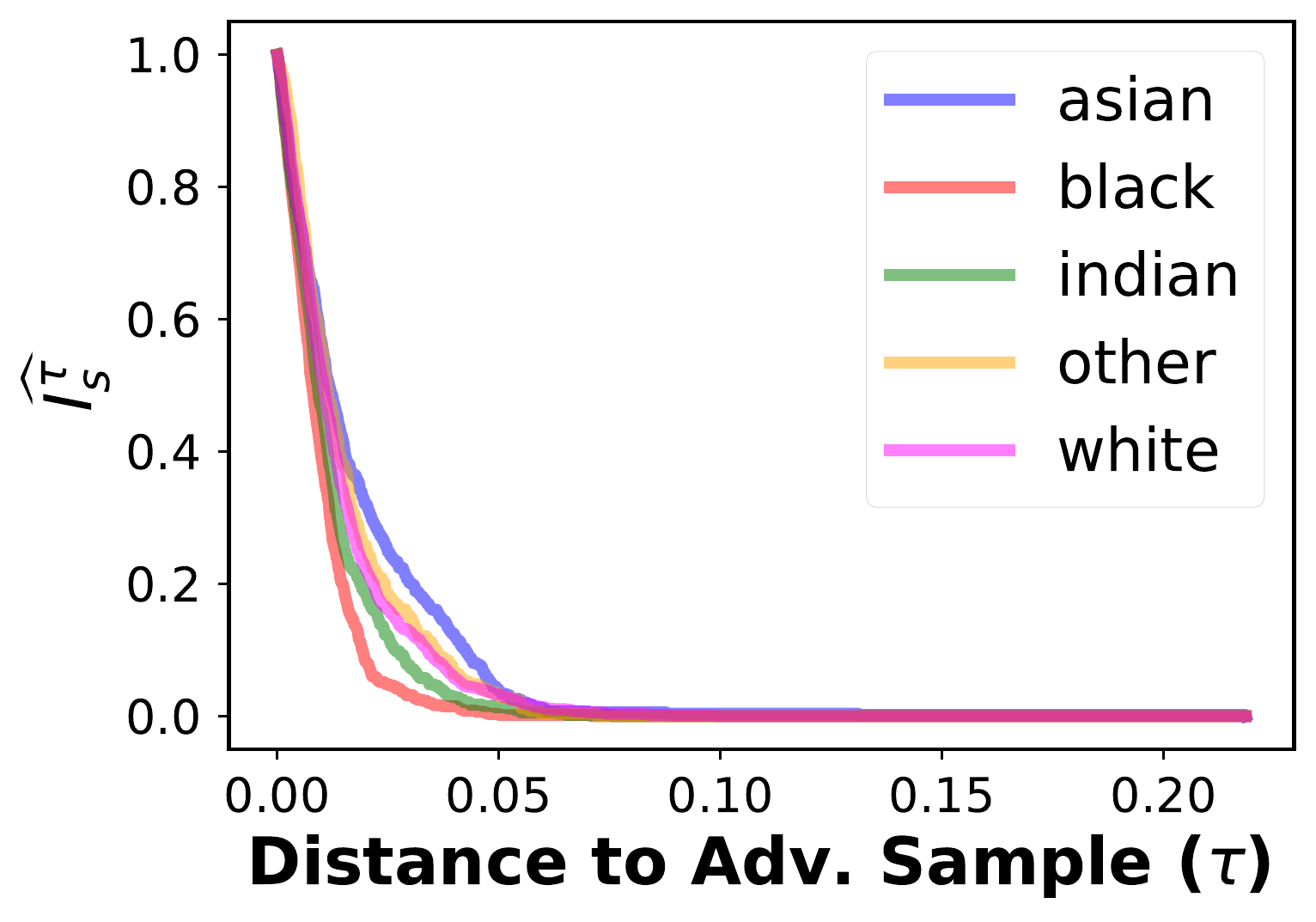}
        \caption{VGG-19: DeepFool}
        \label{fig:utkface_race_vgg_df}
    \end{subfigure}
    \begin{subfigure}[b]{0.23\textwidth}
        \includegraphics[trim={0cm 0cm 0cm 0cm},clip,width=1\textwidth]{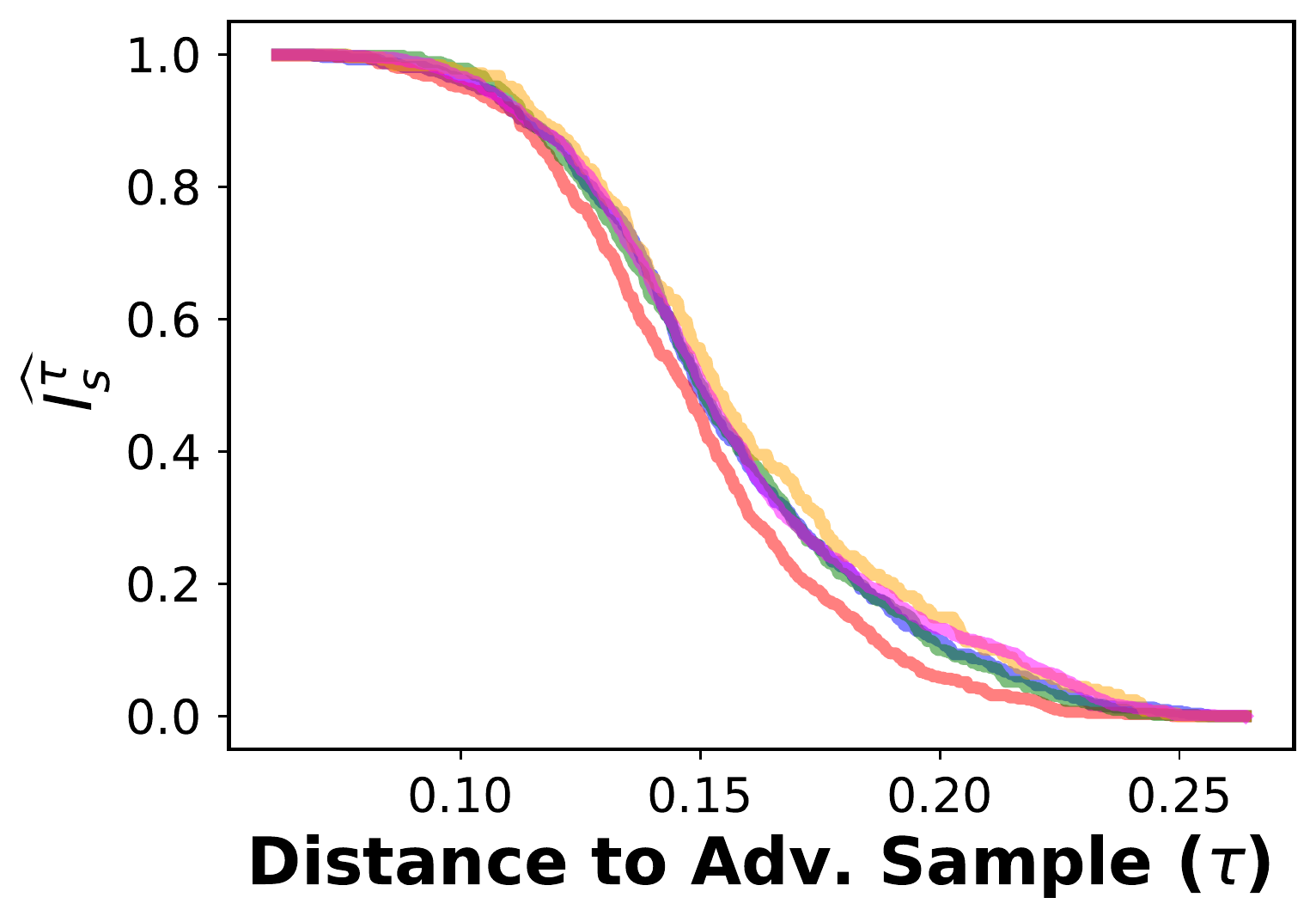}
        \caption{VGG-19: CarliniWagner}
        \label{fig:utkface_race_vgg_cw}
    \end{subfigure}
    \begin{subfigure}[b]{0.23\textwidth}
        \includegraphics[trim={0cm 0cm 0cm 0cm},clip,width=1\textwidth]{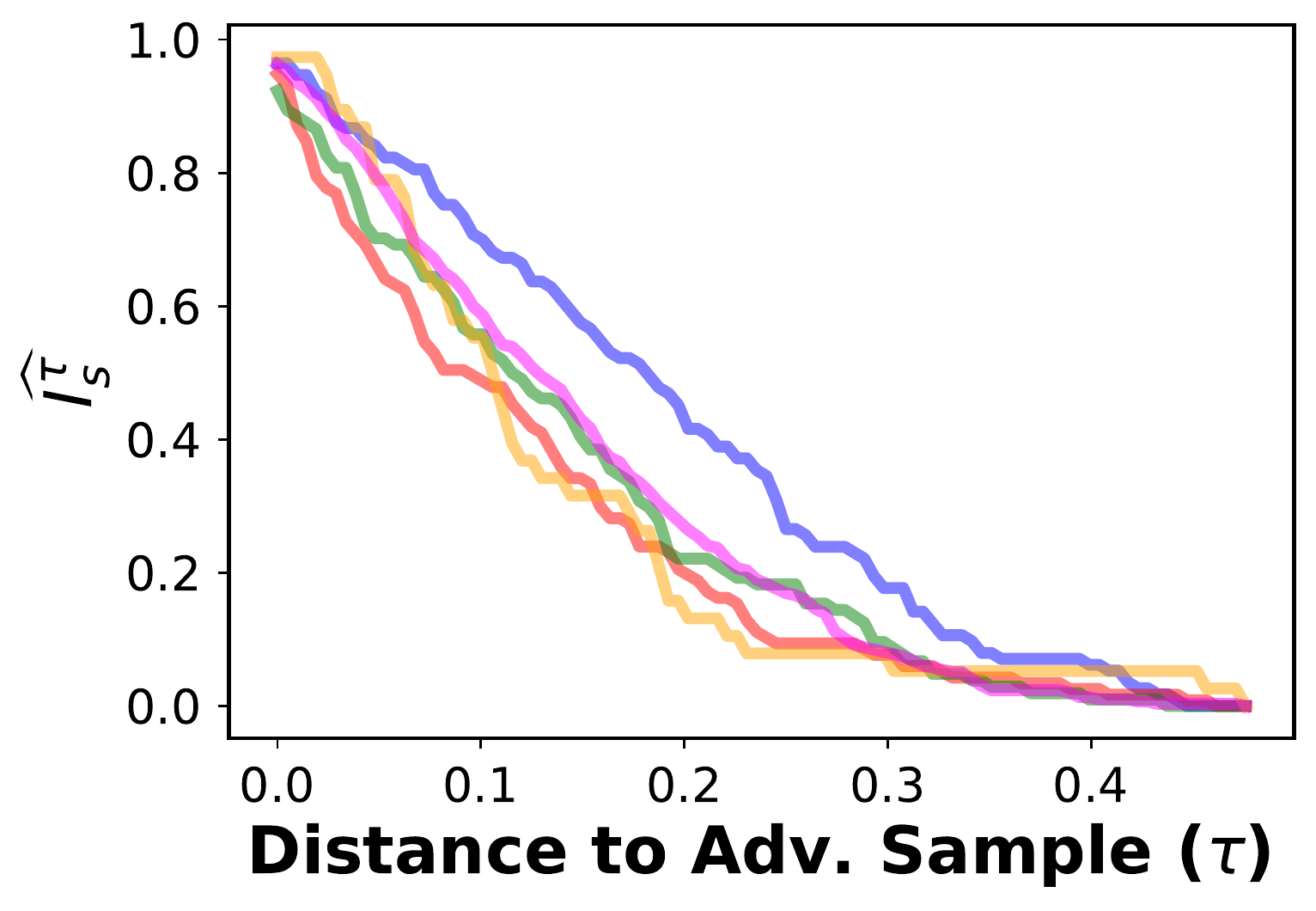}
        \caption{VGG-19: Rand. Smoothing}
        \label{fig:lb_utkface_race_vgg}
    \end{subfigure}

    \begin{subfigure}[b]{0.23\textwidth}
        \includegraphics[trim={0cm 0cm 0cm 0cm},clip,width=1\textwidth]{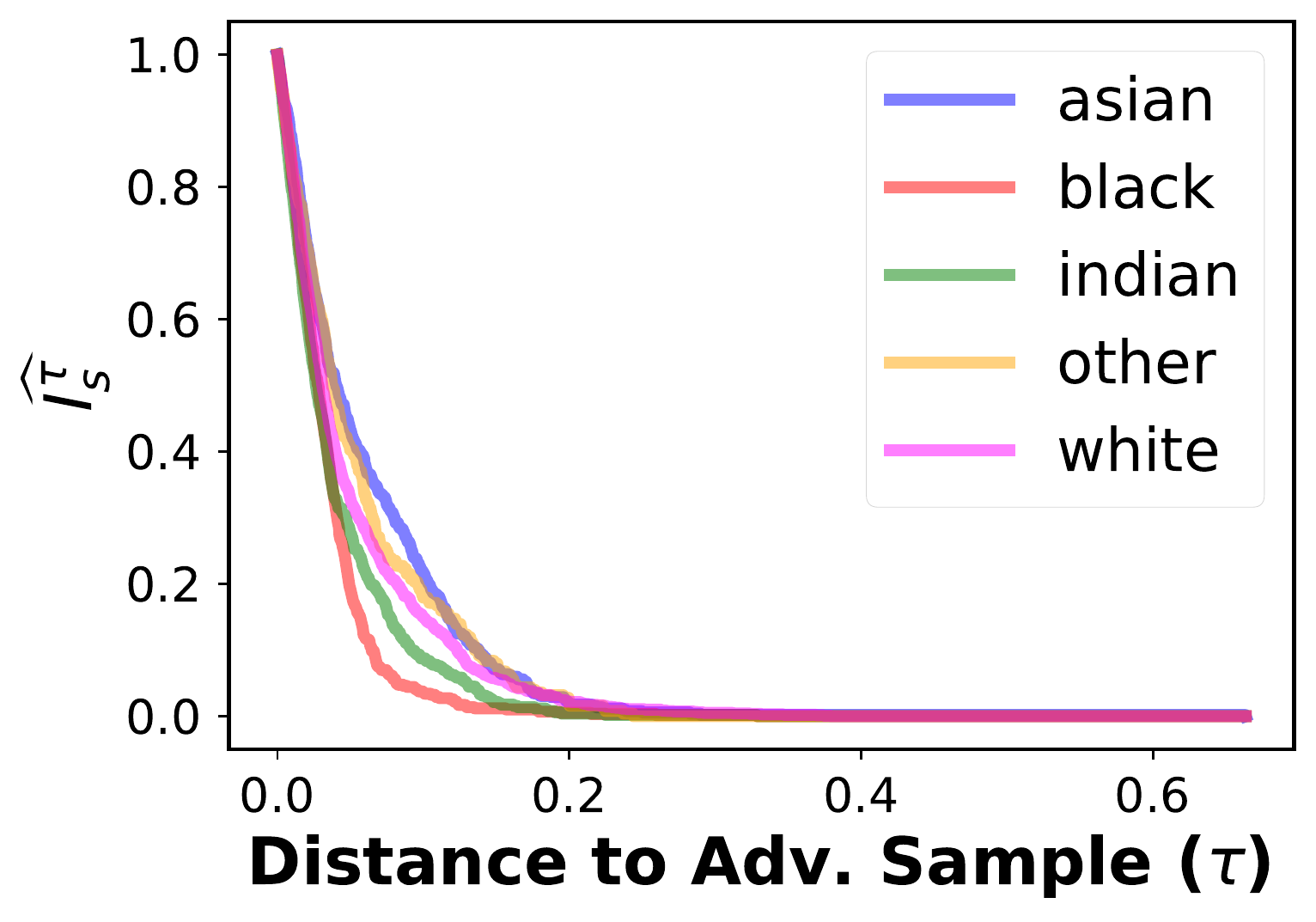}
        \caption{Densenet: DeepFool}
        \label{fig:utkface_race_densenet_df}
    \end{subfigure}
    \begin{subfigure}[b]{0.23\textwidth}
        \includegraphics[trim={0cm 0cm 0cm 0cm},clip,width=1\textwidth]{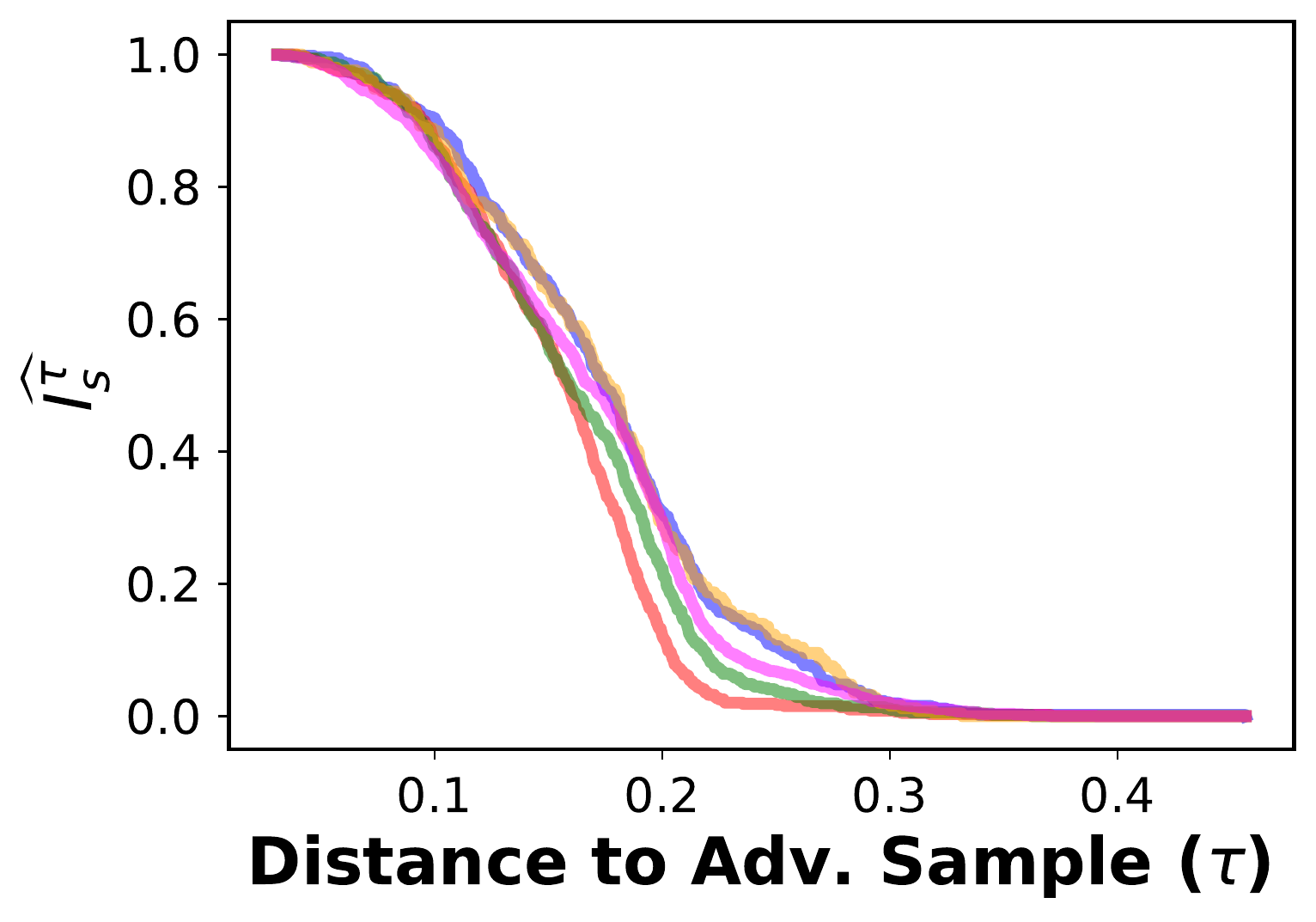}
        \caption{Densenet: CarliniWagner}
        \label{fig:utkface_race_densenet_cw}
    \end{subfigure}
    \begin{subfigure}[b]{0.23\textwidth}
        \includegraphics[trim={0cm 0cm 0cm 0cm},clip,width=1\textwidth]{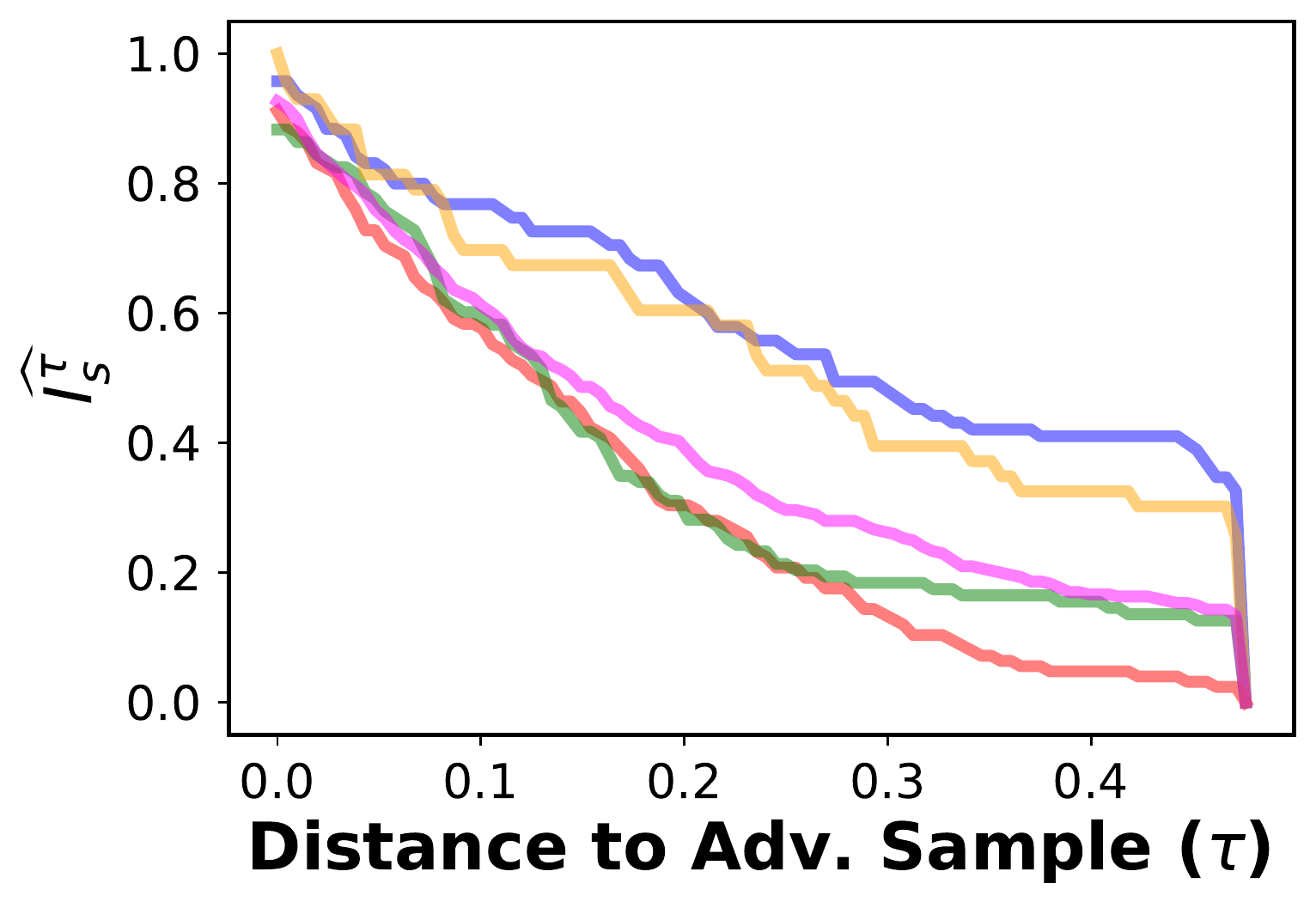}
        \caption{Densenet: Rand. Smoothing}
        \label{fig:lb_utkface_race_densenet}
    \end{subfigure}

    \begin{subfigure}[b]{0.23\textwidth}
        \includegraphics[trim={0cm 0cm 0cm 0cm},clip,width=1\textwidth]{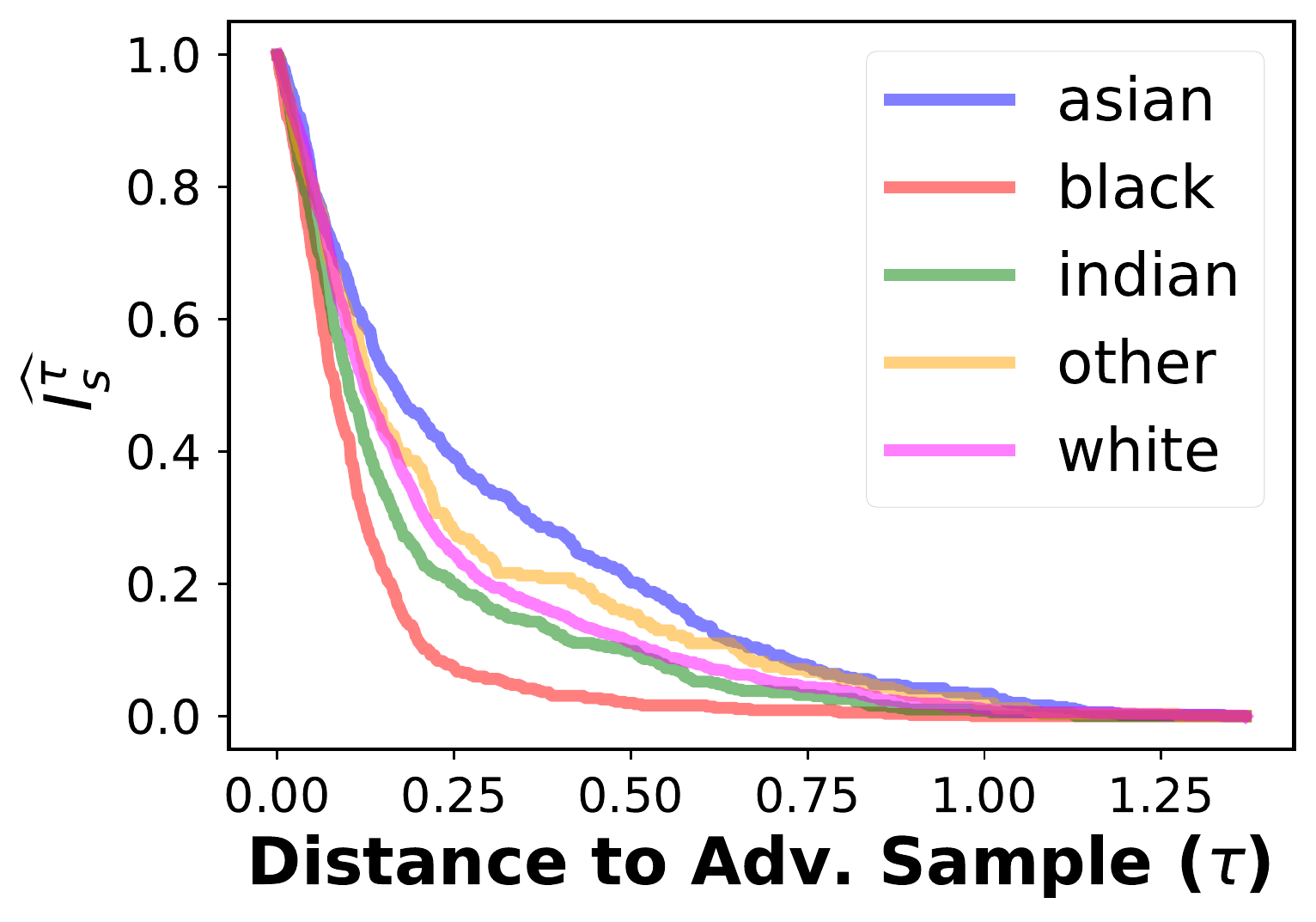}
        \caption{Squeezenet: DeepFool}
        \label{fig:utkface_race_squeezenet_df}
    \end{subfigure}
    \begin{subfigure}[b]{0.23\textwidth}
        \includegraphics[trim={0cm 0cm 0cm 0cm},clip,width=1\textwidth]{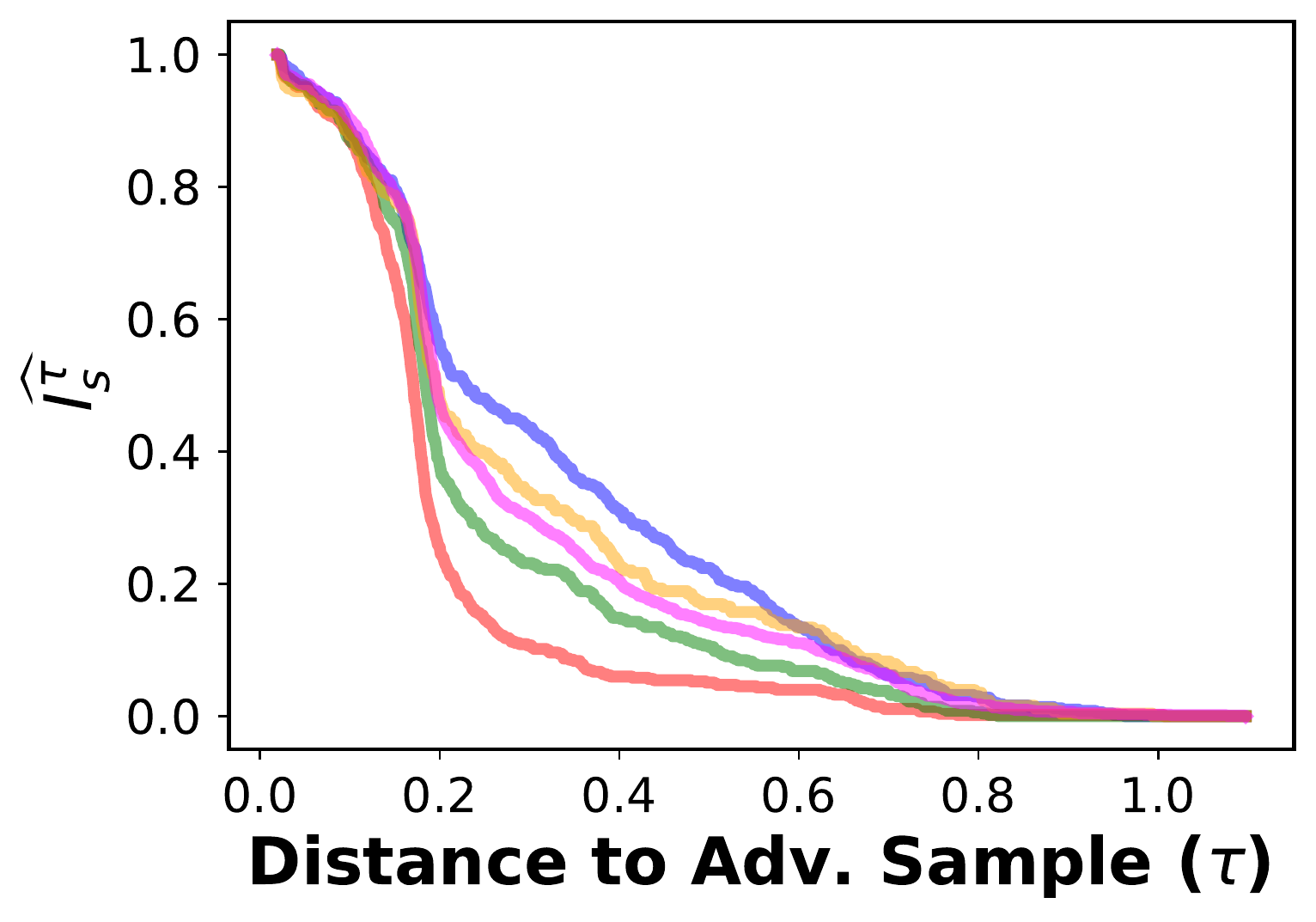}
        \caption{Squeezenet: CarliniWagner}
        \label{fig:utkface_race_squeezenet_cw}
    \end{subfigure}
    \begin{subfigure}[b]{0.23\textwidth}
        \includegraphics[trim={0cm 0cm 0cm 0cm},clip,width=1\textwidth]{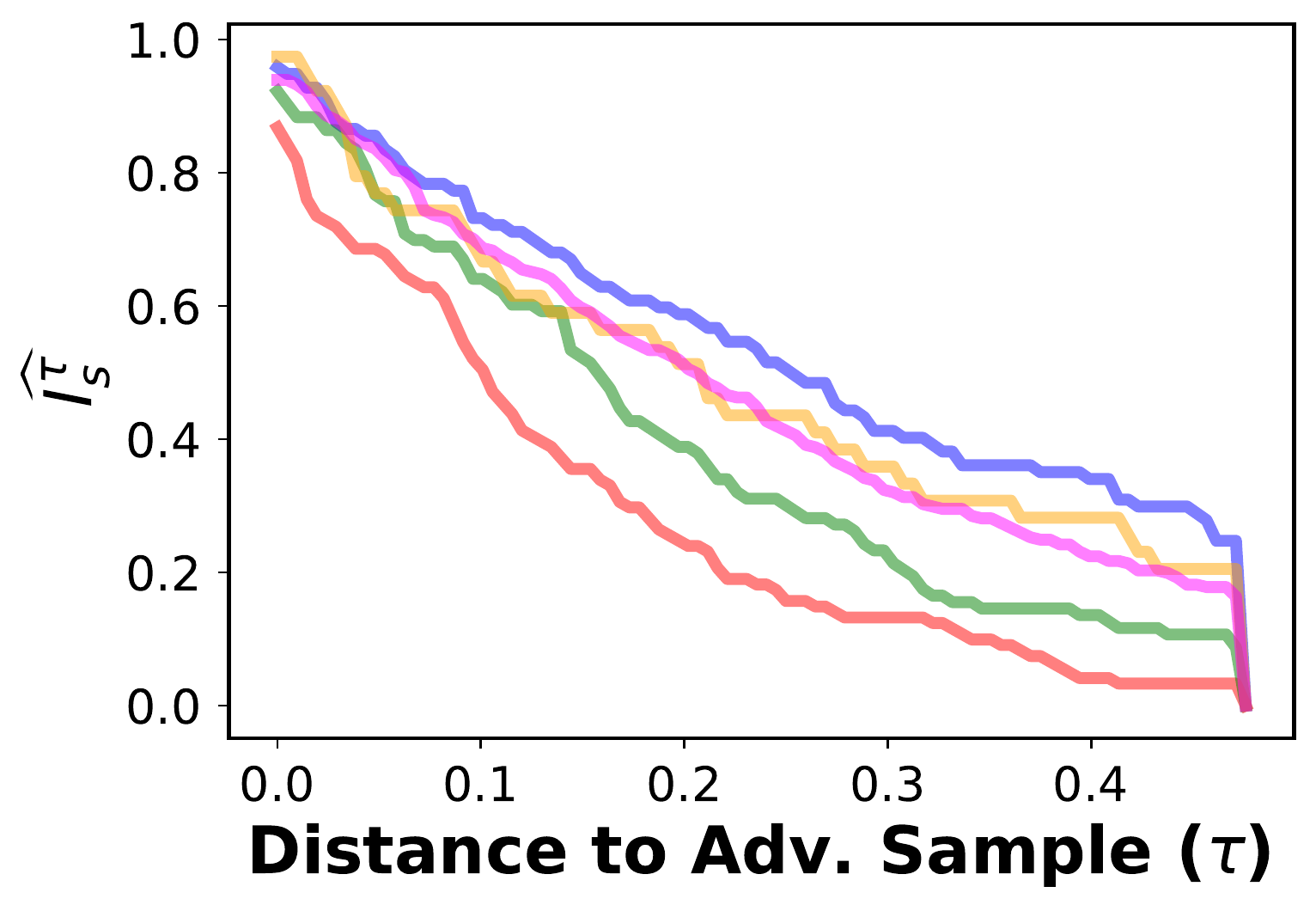}
        \caption{Squeezenet: Rand. Smoothing}
        \label{fig:lb_utkface_race_squeezenet}
    \end{subfigure}

\caption{ UTKFace partitioned by race. We can see that across models, that different populations are at different levels of robustness as calculated by different proxies (DeepFool on the left, CarliniWagner in the middle and Randomized Smoothing on the right). This suggests that robustness bias is an important criterion to consider when auditing models for fairness.
}
\label{fig:utkface_race}
\end{figure*}

\section{Evaluation of Robustness Bias using Adversarial Attacks}\label{sec:adv_attacks}

As described in Section~\ref{subsec:upper_bounds}, we argued that adversarial attacks can be used to obtain upper bounds on $d_\theta(x)$ which can then be used to measure robustness bias. In this section we audit some popularly used models on datasets mentioned in Section~\ref{sec:experiments} for robustness bias as measured using the approximation given by adversarial attacks.

\subsection{Evaluation of $\widehat{I_P^{DF}}$ and $\widehat{I_P^{CW}}$}\label{eval_of_measures}

To compare the estimate of $d_\theta(x)$ by DeepFool and CarliniWagner, we first look at the signedness of $\sigma(P)$, $\sigma^{DF}(P)$, and $\sigma^{CW}(P)$. For a given partition $P$, $\sigma(P)$ captures the disparity in robustness between points in $P$ relative to points not in $P$ (see Eq~\ref{eq:sigma}).  Considering all 151 possible partitions (based on class labels and sensitive attributes, where available) for all five datasets, both CarliniWagner and DeepFool agree with the signedness of the direct computation 125 times, i.e.,
$\mathbbm{1}_P\left[ \sign(\sigma(P)) = \sign(\sigma^{DF}(P))\right] = 125 = \mathbbm{1}_P\left[ \sign(\sigma(P)) = \sign(\sigma^{CW}(P))\right]$.
Further, the mean difference between $\sigma(P)$ and $\sigma^{CW}(P)$ or $\sigma^{DF}(P)$, i.e., $(\sigma(P) - \sigma^{DF}(P))$, is 0.17 for DeepFool and 0.19 for CarliniWagner with variances of 0.07 and 0.06 respectively.

There is 83\% agreement between the direct computation and the DeepFool and CarliniWagner estimates of $\widehat{I_P}$. This behavior provides evidence that adversarial attacks provide meaningful upper bounds on $d_\theta(x)$ in terms of the behavior of identifying instances of robustness bias.

\subsection{Audit of Commonly Used Models}\label{subsec:audit_adv_attacks}

\begin{figure*}[h!]
    \centering
    \includegraphics[width = .95\linewidth]{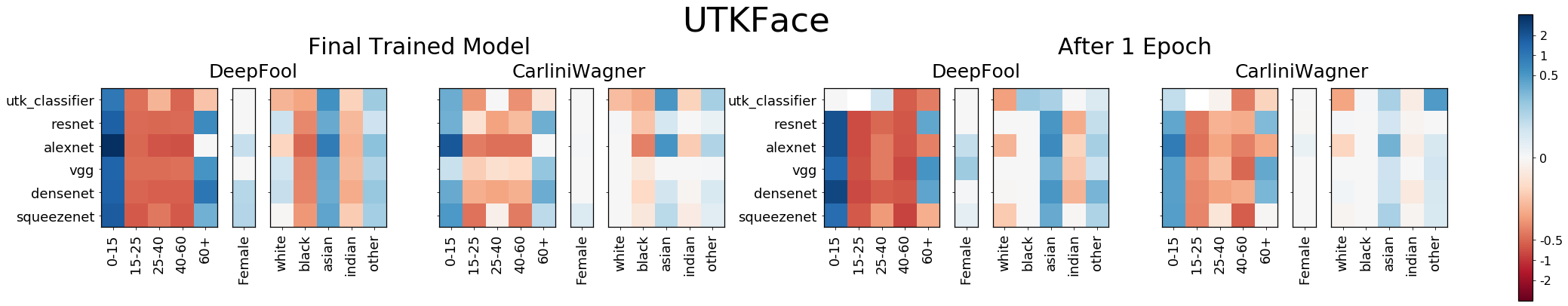}
    \caption{Depiction of $\sigma_P^{DF}$ and $\sigma_P^{CW}$ for the UTKFace dataset with partitions corresponding to the (1) class labels $\mathcal{C}$ and the, (2) gender, and (3) race/ethnicity. These values are reported for all five convolutional models both at the beginning of their training (after one epoch) and at the end. We observe that, largely, the signedness of the functions are consistent between the five models and also across the training cycle.}
    \label{fig:utkface}
\end{figure*}

We now evaluate five commonly-used convolutional neural networks (CNNs): Alexnet, VGG, ResNet, DenseNet, and Squeezenet. We trained these networks using PyTorch with standard stochastic gradient descent. We achieve comparable performance to documented state of the art for these models on these datasets. After training each model on each dataset, we generated adversarial examples using both methods and computed $\sigma(P)$ for each possible partition of the dataset. An example of the results for the UTKFace dataset can be see in Figure~\ref{fig:utkface}.

With evidence from Section~\ref{eval_of_measures} that DeepFool and CarliniWagner can approximate the robustness bias behavior of direct computations of $d_\theta$, we first ask if there are any major differences between the two methods. \emph{If DeepFool exhibits adversarial robustness bias for a dataset and a model and a class, does CarliniWagner exhibit the same? and vice versa?} Since there are 5 different convolutional models, we have $151\cdot5 = 755$ different comparisons to make. Again, we first look at the signedness of $\sigma^{DF}(P)$ and $\sigma^{CW}(P)$ and we see that $\mathbbm{1}_P\left[ \sign(\sigma^{DF}(P)) = \sign(\sigma^{CW}(P))\right] = 708$. This means there is 94\% agreement between DeepFool and CarliniWagner about the direction of the adversarial robustness bias.

To investigate if this behavior is exhibited earlier in the training cycle than at the final, fully-trained model, we compute $\sigma^{CW}(P)$ and $\sigma^{DF}(P)$ for the various models and datasets for trained models after 1 epoch and the middle epoch. For the first epoch, 637 of the 755 partitions were internally consistent, i.e., the signedness of $\sigma$ was the same in the first and last epoch, and 621  were internally consistent. We see that at the middle epoch, 671 of the 755 partitions were internally consistent for DeepFool and 665 were internally consistent for CarliniWagner. Unsurprisingly, this implies that as the training progresses, so does the behavior of the adversarial robustness bias. However, it is surprising that much more than 80\% of the final behavior is determined after the first epoch, and there is a slight increase in agreement by the middle epoch.

We note that, of course, adversarial robustness bias is not necessarily an intrinsic value of a dataset; it may be exhibited by some models and not by others. However, in our studies, we see that the UTKFace dataset partition on Race/Ethnicity does appear to be significantly prone to adversarial attacks given its comparatively low $\sigma^{DF}(P)$ and $\sigma^{CW}(P)$ values across all models.

\section{Evaluation of Robustness Bias using Randomized Smoothing}\label{sec:randomized_smoothing}

In Section~\ref{subsec:lower_bounds}, we argued that randomized smoothing can be used to obtain lower bounds on $d_\theta(x)$ which can then be used to measure robustness bias. In this section we audit popular models on a variety of datasets (described in detail in Section~\ref{sec:experiments}) for robustness bias, as measured using the approximation given by randomzied smoothing.

\begin{figure*}[h!]
    \centering
    \includegraphics[width = .95\linewidth]{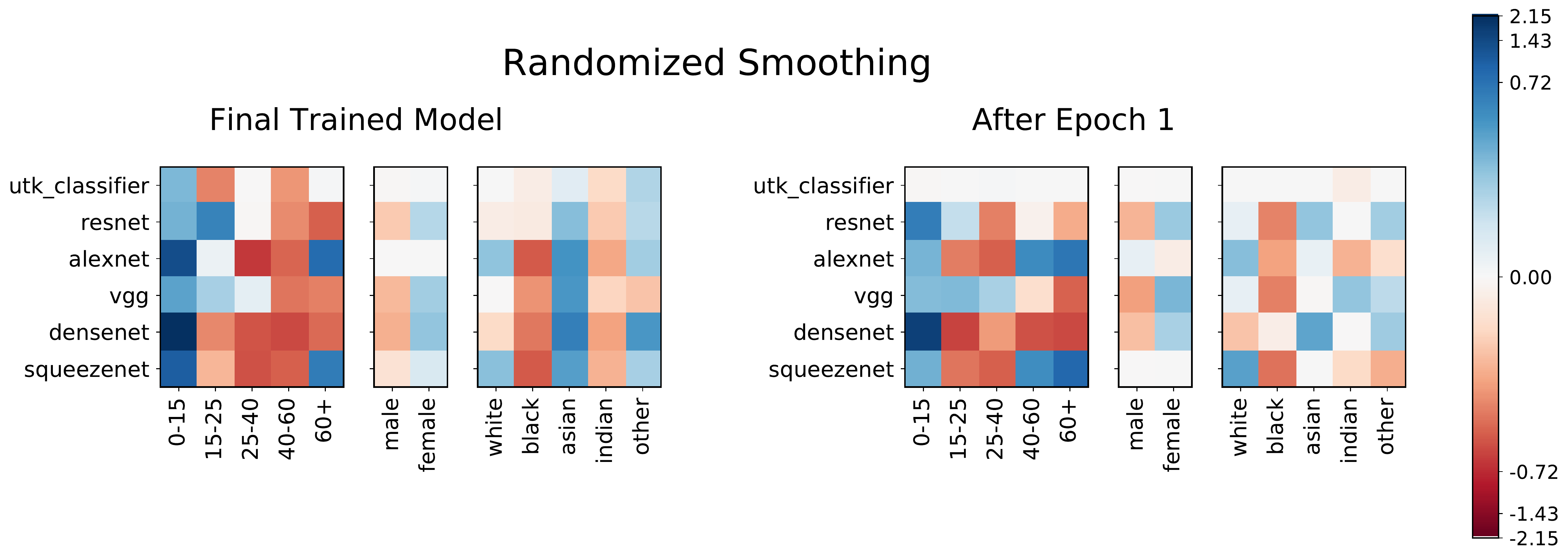}
    \caption{Depiction of $\sigma_P^{RS}$ for the UTKFace dataset with partitions corresponding to the (1) class labels $\mathcal{C}$ and the, (2) gender, and (3) race/ethnicity. A more negative value indicates less robustness bias for the partition. Darker regions indicate high robustness bias. We observe that the trend is largely consistent amongst models and also similar to the trend observed when using adversarial attacks to measure robustness bias (see Figure~\ref{fig:utkface}).}
    \label{fig:utkface_lb}
\end{figure*}

\subsection{Evaluation of $\widehat{I_P^{RS}}$}\label{eval_of_measures_rs}

To assess whether the estimate of $d_\theta(x)$ by randomized smoothing is an appropriate measure of robustness bias, we compare the signedness of $\sigma(P)$ and $\sigma^{RS}(P)$. When $\sigma(P)$ has positive sign, higher magnitude indicates a higher robustness of members of partition $P$ as compared to members not included in that partition $P$; similarly, when $\sigma(P)$ is negatively signed, higher magnitude corresponds to lesser robustness for those members of partition $P$ (see Eq~\ref{eq:sigma}). We may interpret shared signedness of both $\sigma(P)$ (where $d_\theta(x)$ is deterministic) and $\sigma^{RS}(P)$ (where $d_\theta(x)$ is measured by randomized smoothing as described in Section~\ref{subsec:lower_bounds}) as positive support for the $\widehat{I_P^{RS}}$ measure.

Similar to Section~\ref{eval_of_measures}, we consider all possible 151 partitions across CIFAR-10, CIFAR-100, CIFAR-100Super, UTKFace and Adience. For each of these partitions, we compare $\sigma^{RS}(P)$ to the corresponding $\sigma(P)$. We find that their sign agrees 101 times, \ie, $\mathbbm{1}_P\left[ \sign(\sigma(P)) = \sign(\sigma^{RS}(P))\right] = 101$, thus giving a $66.9\%$ agreement. Furthermore, the mean difference between $\sigma(P)$ and $\sigma^{RS}(P)$, \ie, $(\sigma(P) - \sigma^{RS}(P))$ is $0.08$ with a variance of $0.19$.

This provides evidence that randomized smoothing can also provide a meaningful estimate on $d_\theta(x)$ in terms of measuring robustness bias.

\subsection{Audit of Commonly Used Models}
We now evaluate the same models and all the datasets for robustness bias as measured by randomized smoothing.  Our comparison is analogous to the one performed in Section~\ref{subsec:audit_adv_attacks} using adversarial attacks. Figure~\ref{fig:utkface_lb} shows results for all models on the UTKFace dataset. Here we plot $\sigma_{P}^{RS}$ for each partition of the dataset (on x-axis) and for each model (y-axis). A darker color in the heatmap indicates high robustness bias (darker red indicates that the partition is \textit{less} robust than others, whereas a darker blue indicates that the partition is \textit{more} robust). We can see that some partitions, for example, the partition based on class label ``40-60'' and the partition based on race ``black'' tend to be less robust in the final trained model, for all models (indicated by a red color across all models). Similarly there are partitions that are more robust, for example, the partition based on class ``0-15'' and race ``asian'' end up being robust across different models (indicated by a blue color). Figure~\ref{fig:utkface_race} takes a closer look at the distribution of distances for the UTKFace dataset when partitioned by race, showing that for different models different races can be more or less robust. Figures~\ref{fig:utkface_race},~\ref{fig:utkface} and \ref{fig:utkface_lb} (we see similar trends for CIFAR-10, CIFAR-100, CIFAR-100Super and Adience) lead us to the following key conclusions:

\xhdr{Dependence on data distribution} The presence of certain partitions that show similar robustness trends as discussed above (\eg see final trained model in Figs~\ref{fig:utkface_lb} and \ref{fig:utkface}, the partitions by class ``0-15'' and race ``asian'' are more robust, whereas the class ``40-60'' and race ``black'' are less robust across \textit{all models}) point to some intrinsic property of the data distribution that results in that partition being more (or less) robust regardless of the type decision boundary. Thus we conclude that robustness bias may depend in part on the data distribution of various sub-populations.

\xhdr{Dependence on model} There are also certain partitions of the dataset (e.g., based on the classes ``15-25'' and ``60+'' as per Fig~\ref{fig:utkface_lb}) that show varying levels of robustness across different models. Moreover, even partitions that have same sign of $\sigma^{RS}(P)$ across different models have very different values of $\sigma^{RS}(P)$. This is also evident from Fig~\ref{fig:utkface_race} which shows that the distributions of $d_\theta(x)$ (as approximated by all our proposed methods) for different races can be very different for different models. Thus, we conclude that robustness bias is also dependent on the learned model.

\xhdr{Role of pre-training}
We now explore the role of pre-training on our measures of robustness bias.  Specifically, we pre-train five of the six models (Resnet, Alexnet, VGG, Densenet, and Squeezenet) on ImageNet and then fine-tune on UTKFace.  We also train a UTK classifier from scratch on UTKFace.  Figures~\ref{fig:utkface_lb} and \ref{fig:utkface} shows robustness bias scores after the first epoch and in the final, fully-trained model.  At epoch 1, we mostly see no robustness bias (indicated by close-to-zero values of $\sigma^{RS}(P)$) for UTK Classifier. This is because the model has barely trained by that first epoch and predictions are roughly equivalent to random guesses. In contrast, the other five models already have pre-trained ImageNet weights, and hence we see certain robustness biases that already exist in the model, even after the first epoch of training. Thus, we conclude that pre-trained models bring in biases due to the distributions of the data on which they were pre-trained and the resulting learned decision boundary after pre-training. We additionally see that these biases can persist even after fine-tuning.

\subsection{Comparison of Randomized Smoothing and Upper Bounds}

We have now presented two ways of measuring robustness bias: via upper bounds and via randomized smoothing. While there are important distinctions between the two methods, it is worth comparing them. To do this, we compare the sign of the randomized smoothing method and the upper bounds as
$$\mathbbm{1}_P\left[ \sign(\sigma^{RS}(P)) = \sign(\sigma^{DF}(P))\right]$$
and
$$\mathbbm{1}_P\left[ \sign(\sigma^{RS}(P)) = \sign(\sigma^{CW}(P))\right].$$
We see that there is some evidence that the two methods agree. The Adience, UTKFace, and CIFAR-10 dataset have strong agreement (at or above 75\%) between the randomized smoothing for both types of upper bounds (DeepFool and CariliniWagner), while the CIFAR-100 dataset has a much weaker agreement (above but closer to 50\%) and CIFAR-100Super has an approximately 66\% agreement.

It is important to point out that it is not entirely appropriate to perform a comparison in this way. Recall that the upper bounds provide estimates of $d_\theta$ using a trained model. However, the randomized smoothing method estimates $d_\theta$ not directly with the trained model --- instead it first modifies (smooths) the model of interest and then performs an estimation.
Since the upper bounds and randomized smoothing methods are so different in practice, there may be no truly appropriate way to compare the results therefrom.
Therefore, too much credence should not be placed on the comparison of these two methods.
Both methods indicate the existence of the robustness bias phenomenon and can be useful in distinct settings.

\section{An ``Obvious'' Mitigation Strategy}\label{sec:mitigation}

Having demonstrated the existence of this robustness bias phenomenon, it is natural to look ahead at common machine learning techniques to address it.  We have done just that by adding to the objective function a regularizer term which penalizes for large distances in the treatment of a minority and majority group. We write the empiric estimate of $\RBmeasure{(P,\tau)}$ as $\tilde{\RBmeasure}{(P,\tau)}$.  Formally,
\vspace{-2.5mm}
{\small
\[
\tilde{\RBmeasure}{(P,\tau)} = \Bigg|  \frac{1}{\displaystyle \sum_{x\notin P} \mathbbm{1}\{ y = \yhat \}} \sum_{\substack{x \notin P\\y = \yhat}} \mathbbm{1}\{ d_\theta(x) > \tau \} -   \\
    \frac{1}{\displaystyle\sum_{x\in P} \mathbbm{1}\{ y = \yhat \}} \sum_{\substack{x \in P\\y = \yhat}} \mathbbm{1}\{ d_\theta(x) > \tau \}  \Bigg|
\]
}

Experimental results based on that implementation support the idea that regularization---a typical approach taken by the fairness in machine learning community---can reduce measures of robustness bias.
However, we do believe that this type of experimentation belies the larger point of the present, largely descriptive, work: that robustness bias is a real and significant artifact of popular and commonly-used datasets and models.
Surely there are ways to mitigate some of the effects or manifestations of this bias (as we show with our fairly standard regularization-based mitigation technique).
However, we believe that any type of mitigation should be taken in concert with the contextualization of these technical systems in the social world, and thus leave ``mitigation'' research to, ideally, application-specific future work involving both machine learning practictioners and the stakeholders of particular systems.

\section{Discussion and Conclusion}\label{sec:broader}
We propose a unique definition of fairness which requires all partitions of a population to be equally robust to minute (often adversarial) perturbations, and give experimental evidence that this phenomenon can exist in some commonly-used models trained on real-world datasets. Using these observations, we argue that this can result in a potentially unfair circumstance where, in the presence of an adversary, a certain partition might be more susceptible (\ie, less secure). Susceptibility is prone to known issues with adversarial robustness such as sensitivity to hyperparameters~\cite{tramer2020adaptive}. Thus, we call for extra caution while deploying deep neural nets in the real world since this form of unfairness might go unchecked when auditing for notions that are based on just the model outputs and ground truth labels. We then show that this form of bias can be mitigated to some extent by using a regularizer that minimizes our proposed measure of robustness bias. However, we do not claim to ``solve'' unfairness; rather, we view analytical approaches to bias detection and optimization-based approaches to bias mitigation as potential pieces in a much larger, multidisciplinary approach to addressing these issues in fielded systems.

Indeed, we view our work as largely observational---we \emph{observe} that, on many commonly-used models trained on many commonly-used datasets, a particular notion of bias, \emph{robustness bias}, exists.  We show that some partitions of data  are more susceptible to two state-of-the-art and commonly-used adversarial attacks.  This knowledge could be used for \emph{attack} or to design \emph{defenses}, both of which could have potential positive or negative societal impacts depending on the parties involved and the reasons for attacking and/or defending.  We have also \emph{defined} a notion of bias as well as a corresponding notion of fairness, and by doing that we admittedly toe a morally-laden line.  Still, while we do use ``fairness'' as both a higher-level motivation and a lower-level quantitative tool, we have tried to remain ethically neutral in our presentation and have eschewed making normative judgements to the best of our ability.

\chapter{Robustness Disparities in Commercial Face Detection} 

\label{chpt:ProductionRobustness} 
This work was done in collaboration with Tom Goldstein and John P. Dickerson. See~\cite{dooley2021robustness,dooley2022commercial,dooleyrobustness}.

\section{Introduction}\label{sec:intro}
Face detection systems identify the presence and location of faces in images and video.  Automated face detection is a core component of myriad systems---including \emph{face recognition technologies} (\FRT{}), wherein a detected face is matched against a database of faces, typically for identification or verification purposes.  \FRT{}-based systems are widely deployed~\citep{hartzog2020secretive, derringer2019surveillance, weise2020amazon}.  Automated face recognition enables capabilities ranging from the relatively morally neutral (e.g., searching for photos on a personal phone~\citep{GooglePhotosFRT}) to morally laden (e.g., widespread citizen surveillance~\citep{hartzog2020secretive}, or target identification in warzones~\citep{marson_forrest_2021}).  Legal and social norms regarding the usage of \FRT{} are evolving~\citep[e.g.,][]{grother2019face}. For example, in June 2021, the first county-wide ban on its use for policing~\cite[see, e.g.,][]{garvie2016perpetual} went into effect in the US~\citep{Gutman21:King}.  Some use cases for \FRT{} will be deemed socially repugnant and thus be either legally or \emph{de facto} banned from use; yet, it is likely that pervasive use of facial analysis will remain---albeit with more guardrails than are found today~\citep{singer2018microsoft}.

One such guardrail that has spurred positive, though insufficient, improvements and widespread attention is the use of benchmarks. For example, in late 2019, the US National Institute of Standards and Technology (NIST) adapted its venerable Face Recognition Vendor Test (FRVT) to explicitly include concerns for demographic effects~\citep{grother2019face}, ensuring such concerns propagate into industry systems.   Yet, differential treatment by \FRT{} of groups has been known for at least a decade~\citep[e.g.,][]{klare2012face,el2016face}, and more recent work spearheaded by~\citet{buolamwini2018gendershades} uncovers unequal performance at the phenotypic subgroup level.  That latter work brought widespread public, and thus burgeoning regulatory, attention to bias in \FRT{}~\citep[e.g.,][]{lohr2018facial,CodedBias}.

One yet unexplored benchmark examines the bias present in a system's robustness (e.g., to noise, or to different lighting conditions), both in aggregate and with respect to different dimensions of the population on which it will be used.
Many detection and recognition systems are not built in house, instead making use of commercial cloud-based ``ML as a Service'' (MLaaS) platforms offered by tech giants such as Amazon, Microsoft, Google, Megvii, etc.  The implementation details of those systems are not exposed to the end user---and even if they were, quantifying their failure modes would be difficult.  With this in mind, our \textbf{main contribution} is a wide \emph{robustness benchmark} of three commercial-grade face detection systems (accessed via Amazon's Rekognition, Microsoft's Azure, and Google Cloud Platform's face detection APIs).  For fifteen types of realistic noise, and five levels of severity per type of noise~\citep{hendrycks2019benchmarking}, we test both APIs against images in each of four well-known datasets.  Across these more than \num{5000000} noisy images, we analyze the impact of noise on face detection performance.  Perhaps unsurprisingly, we find that noise decreases overall performance, though the result from our study confirms the previous findings of~\citet{hendrycks2019benchmarking}. Further, different types of noise impact, in an ``unfair'' way, cross sections of the population of images (e.g., based on Fitzgerald skin type, age, self-identified gender, and intersections of those dimensions).  Our method is extensible and can be used to quantify the robustness of other detection and \FRT{} systems, and adds to the burgeoning literature supporting the necessity of explicitly considering fairness in ML systems with morally-laden downstream uses.

\section{Related Work}\label{sec:rel-work}

We briefly overview additional related work in the two core areas addressed by our benchmark: robustness to noise and demographic disparity in facial detection and recognition.  That latter point overlaps heavily with the fairness in machine learning literature; for additional coverage of that broader ecosystem and discussion around fairness in machine learning writ large, we direct the reader to survey works due to~\citet{chouldechova2018frontiers} and~\citet{fairmlbook}.

\paragraph{Demographic effects in facial detection and recognition.}
The existence of differential performance of facial detection and recognition on groups and subgroups of populations has been explored in a variety of settings.  Earlier work~\citep[e.g.,][]{klare2012face,o2012demographic} focuses on single-demographic effects (specifically, race and gender) in pre-deep-learning face detection and recognition.  \citet{buolamwini2018gendershades} uncovers unequal performance at the phenotypic subgroup level in, specifically, a gender classification task powered by commercial systems.  That work, typically referred to as ``Gender Shades,'' has been and continues to be hugely impactful both within academia and at the industry level.  Indeed, \citet{raji2019actionable} provide a follow-on analysis, exploring the impact of the~\citet{buolamwini2018gendershades} paper publicly disclosing performance results, for specific systems, with respect to demographic effects; they find that their named companies (IBM, Microsoft, and Megvii) updated their APIs within a year to address some concerns that were surfaced.  Subsequently, the late 2019 update to the NIST FRVT provides evidence that commercial platforms are continuing to focus on performance at the group and subgroup level~\citep{grother2019face}. Further recent work explores these demographic questions with a focus on Indian election candidates \citep{jain2021Million}. We see our benchmark as adding to this literature by, for the first time, addressing both noise and demographic effects on commercial platforms' face detection offerings.

In this work, we focus on \emph{measuring} the impact of noise on a classification task,  like that of \citet{wilber2016can}; indeed, a core focus of our benchmark is to \emph{quantify} relative drops in performance conditioned on an input datapoint's membership in a particular group.  We view our work as a \emph{benchmark}, that is, it focuses on quantifying and measuring, decidedly not providing a new method to ``fix'' or otherwise mitigate issues of demographic inequity in a system.  Toward that latter point, existing work on ``fixing'' unfair systems can be split into three (or, arguably, four~\citep{savani2020posthoc}) focus areas: pre-, in-, and post-processing. Pre-processing work largely focuses on dataset curation and preprocessing~\citep[e.g.,][]{Feldman2015Certifying, ryu2018inclusivefacenet, quadrianto2019discovering, wang2020mitigating}. In-processing often constrains the ML training method or optimization algorithm itself~\citep[e.g.,][]{zafar2017aistats, Zafar2017www, zafar2019jmlr, donini2018empirical, goel2018non,Padala2020achieving, agarwal2018reductions, wang2020mitigating,martinez2020minimax,diana2020convergent,lahoti2020fairness}, or focuses explicitly on so-called fair representation learning~\citep[e.g.,][]{adeli2021representation,dwork2012fairness,zemel13learning,edwards2016censoring,madras2018learning,beutel2017data,wang2019balanced}. Post-processing techniques adjust decisioning at inference time to align with quantitative fairness definitions~\citep[e.g.,][]{hardt2016equality,wang2020fairness}.

\paragraph{Robustness to noise.}

Quantifying, and improving, the robustness to noise of face detection and recognition systems is a decades-old research challenge.  Indeed, mature challenges like NIST's Facial Recognition Vendor Test (FRVT) have tested for robustness since the early 2000s~\citep{phillips2007frvt}.  We direct the reader to a comprehensive introduction to an earlier robustness challenge due to NIST~\citep{phillips2011introduction}; that work describes many of the specific challenges faced by face detection and recognition systems, often grouped into Pose, Illumination, and Expression (PIE).  It is known that commercial systems still suffer from degradation due to noise~\citep[e.g.,][]{hosseini2017google}; none of this work also addresses the intersection of noise with fairness, as we do.
Recently, \emph{adversarial} attacks have been proposed that successfully break commercial face recognition systems~\citep{shan2020fawkes,cherepanova2021lowkey}; we note that our focus is on \emph{natural} noise, as motivated by~\citet{hendrycks2019benchmarking} by their ImageNet-C benchmark.  Literature at the intersection of adversarial robustness and fairness is nascent and does not address commercial platforms~\citep[e.g.,][]{singh2020robustness,nanda2021fairness}.  To our knowledge, our work is the first systematic benchmark for commercial face detection systems that addresses, comprehensively, noise and its differential impact on (sub)groups of the population.

\section{Experimental Description}\label{sec:experimental-design}

\paragraph{Datasets and Protocol.}
\begin{figure}
    \centering
    \includegraphics[width=\textwidth]{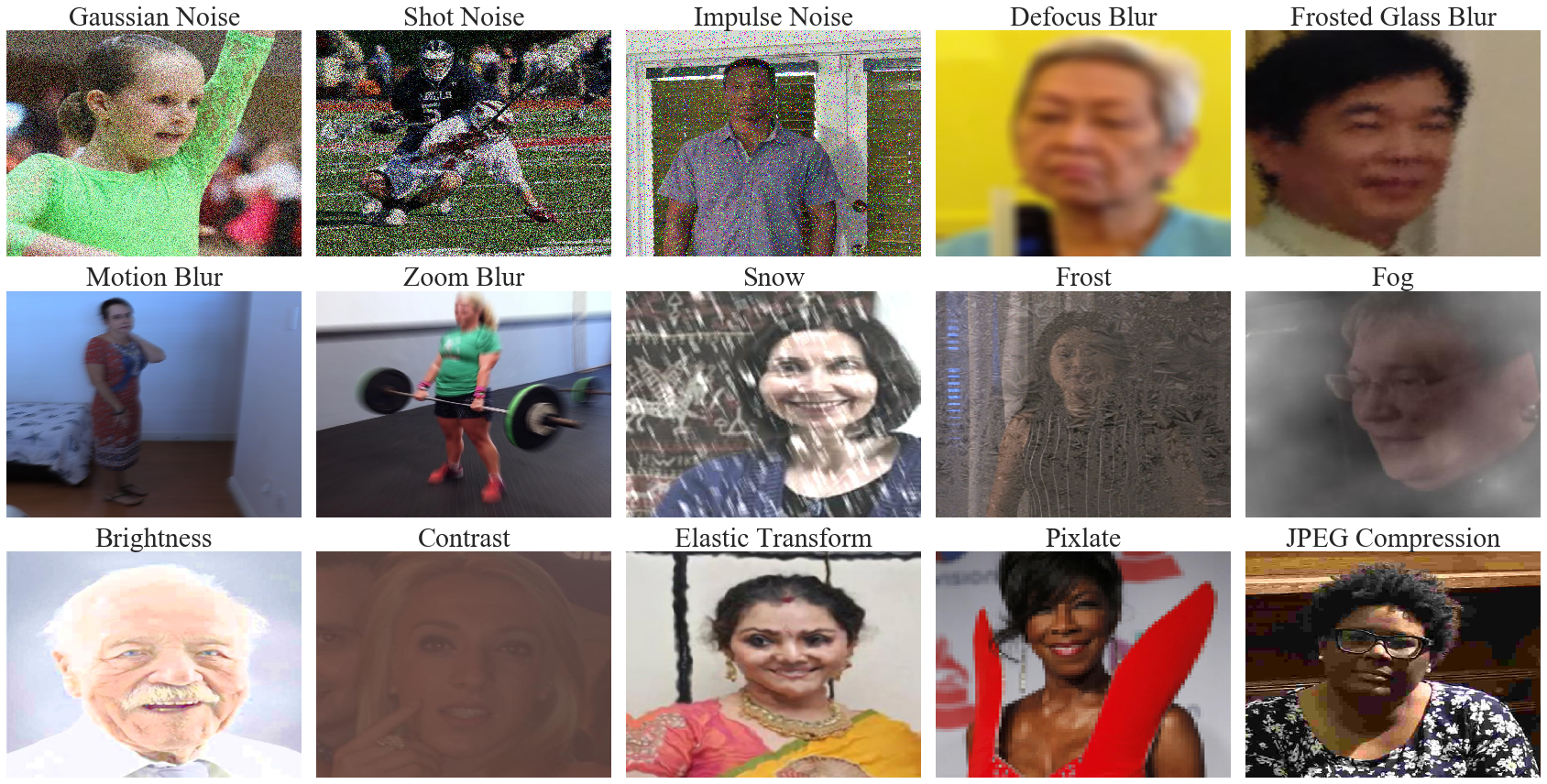}
    \caption{Our benchmark consists of 5,066,312 images of the 15 types of algorithmically generated corruptions produced by ImageNet-C. We use data from four datasets (Adience, CCD, MIAP, and UTKFace) and present examples of corruptions from each dataset here.}
    \label{fig:examples}
\end{figure}

This benchmark uses four datasets to evaluate the robustness of Amazon AWS and Microsoft Azure's face detection systems. They are described below and a repository for the experiments can be found here: \url{https://github.com/dooleys/Robustness-Disparities-in-Commercial-Face-Detection}.

The Open Images Dataset V6 -- Extended; More Inclusive Annotations for People ({\bf MIAP}) dataset \citep{miap} was released by Google in May 2021 as a  extension of the popular, permissive-licensed Open Images Dataset specifically designed to improve annotations of humans. For each image, every human is exhaustively annotated with bounding boxes for the entirety of their person visible in the image. Each annotation also has perceived gender (Feminine/Masculine/Unknown) presentation and perceived age (Young, Middle, Old, Unknown) presentation.   

The Casual Conversations Dataset ({\bf CCD}) \citep{ccd} was released by Facebook in April 2021 under limited license and includes videos of actors. Each actor consented to participate in an ML dataset and provided their self-identification of age and gender (coded as Female, Male, and Other), each actor's skin type was rated on the Fitzpatrick scale \citep{fitzpatrick1988validity}, and each video was rated for its ambient light quality. For our benchmark, we extracted one frame from each video. 

The {\bf Adience} dataset \citep{adience} under a CC license, includes cropped images of faces from images ``in the wild''. Each cropped image contains only one primary, centered face, and each face is annotated by an external evaluator for age and gender (Female/Male). The ages are reported as member of 8 age range buckets: 0-2; 3-7; 8-14; 15-24; 25-35; 36-45; 46-59; 60+.  

Finally, the {\bf UTKFace} dataset \citep{utk} under a non-commercial license, contains images with one primary subject and were annotated for age (continuous), gender (Female/Male), and ethnicity (White/Black/Asian/Indian/Others) by an algorithm, then checked by human annotators. 

For each of the datasets, we randomly selected a subset of images for our evaluation in order to cap the number of images from each intersectional identity at \num{1500} as an attempt to reduce the effect of highly imbalanced datasets. We include a total of \num{66662} images with \num{14919} images from Adience; \num{21444} images from CCD; \num{8194} images from MIAP; and \num{22105} images form UTKFace. 

Each image was corrupted a total of 75 times, per the ImageNet-C protocol with the main 15 corruptions each with 5 severity levels. Examples of these corruptions can be seen in Figure~\ref{fig:examples}. This resulted in a total of \num{5066312} images  (including the original clean ones) which were each passed through the AWS and Azure face analysis systems.  The API calls were conducted between 19 May and 29 May 2021. Images were processed and stored within AWS's cloud using S3 and EC2. 

\paragraph{Evaluation Metrics.}

We evaluate the error of the face systems. Since none of the chosen datasets have ground truth face bounding boxes, we compare the number of detected faces from the clean image to the number of faces detected in a corrupted image, using the clean image as a ground truth proxy of sorts.

Our main metric is the relative error in the number of faces a system detects after corruption; this metric has been used in other facial processing benchmarks \citep{jain2021Million}. Measuring error in this way is in some sense incongruous with the object detection nature of the APIs. However, none of the data in our datasets have bounding boxes for each face. This means that we cannot calculate precision metrics as one would usually do with other detection tasks. To overcome this, we hand-annotated bounding boxes for each face in 772 random images from the dataset. We then calculated per-image precision scores (with an IoU of 0.5) and per-image relative error in face counts and we find a Pearson's correlation of 0.91 (with $p<0.001$). This high correlation indicates that the proxy is sufficient to be used in this benchmark in the absence of fully annotated bounding boxes.

This error is calculated for each image. The way in which this works is that we first pass every clean, uncorrupted image through the commercial system's API. Then, we measure the number of detected faces, i.e., length of the system's response, and treat this number as the ground truth. Subsequently, we compare the  number of detected faces for a corrupted version of that image. If the two face counts are not the same, then we call that an error. We refer to this as the \emph{relative corruption error}. For each clean image, $i$, from dataset $d$, and each corruption $c$ which produces a corrupted image $\hat{i}_{c,s}$ with severity $s$, we compute the relative corruption error for system $r$ as

\[
    \rCE_{c,s}^{d,r}(\hat{i}_{c,s}) :=
        \begin{cases}
        1, \quad \text{if }  l_r(i) \neq l_r(\hat{i}_{c,s})\\
        0, \quad \text{if }   l_r(i) = l_r(\hat{i}_{c,s})\\
        \end{cases}
\]
 where $l_r$ computes the number of detected faces, i.e., length of the response, from face detection system $r$ when given an image. Often the super- and subscripts are omitted when they are obvious from context.

Our main metric, relative error, aligns with that of the ImageNet-C benchmark. We report mean relative corruption error ($\mrCE{}$) defined as taking the average of $\rCE{}$ across some relative set of categories. In our experiments, depending on the context, we might have any of the following categories: face systems, datasets, corruptions, severities, age presentation, gender presentation, Fitzpatrick rating, and ambient lighting. For example, we might report the relative mean corruption error when averaging across demographic groups; the mean corruption error for Azure on the UTK dataset for each age group $a$ is $\mrCE{}_a = \frac{1}{15}\frac{1}{5}\sum_{c,s} \rCE{}^{UTK,Azure}_{c,s,a}$. The subscripts on $\mrCE{}$ are omitted when it is obvious what their value is in whatever context they are presented.

Finally, we  also investigate the significance of whether the $\mrCE{}$ for two groups are equal. For example, our first question is whether the two commercial systems (AWS and Azure) have comparable $\mrCE{}$ overall. To do this, we report the raw $\mrCE{}$; these frequency or empiric probability statistics offer much insight into the likelihood of error. But we also indicate the statistical significance at $\alpha=0.05$ determined by logistic regressions for the appropriate variables and interactions. For each claim of significance, regression tables can be found in the appendix. Accordingly, we discuss the odds or odds ratio of relevant features. Finally, each claim we make for an individual dataset or service is backed up with statistical rigor through the logistic regressions. Each claim we make across datasets is done by looking at the trends in each dataset and are inherently qualitative.

\paragraph{What is not included in this study.}

There are three main things that this benchmark does not address.
First, we do not examine cause and effect. We report inferential statistics without discussion of what generates them.
Second, we only examine the types of algorithmicaly generated natural noise present in the 15 corruptions. We speak narrowly about robustness to these corruptions or perturbations. We explicitly do not study or measure robustness to other types of changes to images, for instance adversarial noise, camera dimensions, etc.
Finally, we do not investigate algorithmic training. We do not assume any knowledge of how the commercial system was developed or what training procedure or data were used.

\paragraph{Social Context.}

The central analysis of this benchmark relies on socially constructed concepts of gender presentation and the related concepts of race and age. While this benchmark analyzes phenotypal versions of these from metadata on ML datasets, it would be wrong to interpret our findings absent a social lens of what these demographic groups mean inside a society. We guide the reader to~\citet{benthall2019racial} and \citet{hanna2020towards} for a look at these concepts for race in machine learning, and~\citet{hamidi2018gender} and \citet{keyes2018misgendering} for similar looks at gender.

\section{Benchmark Results}
\begin{figure}
    \centering
    \includegraphics[width=\textwidth]{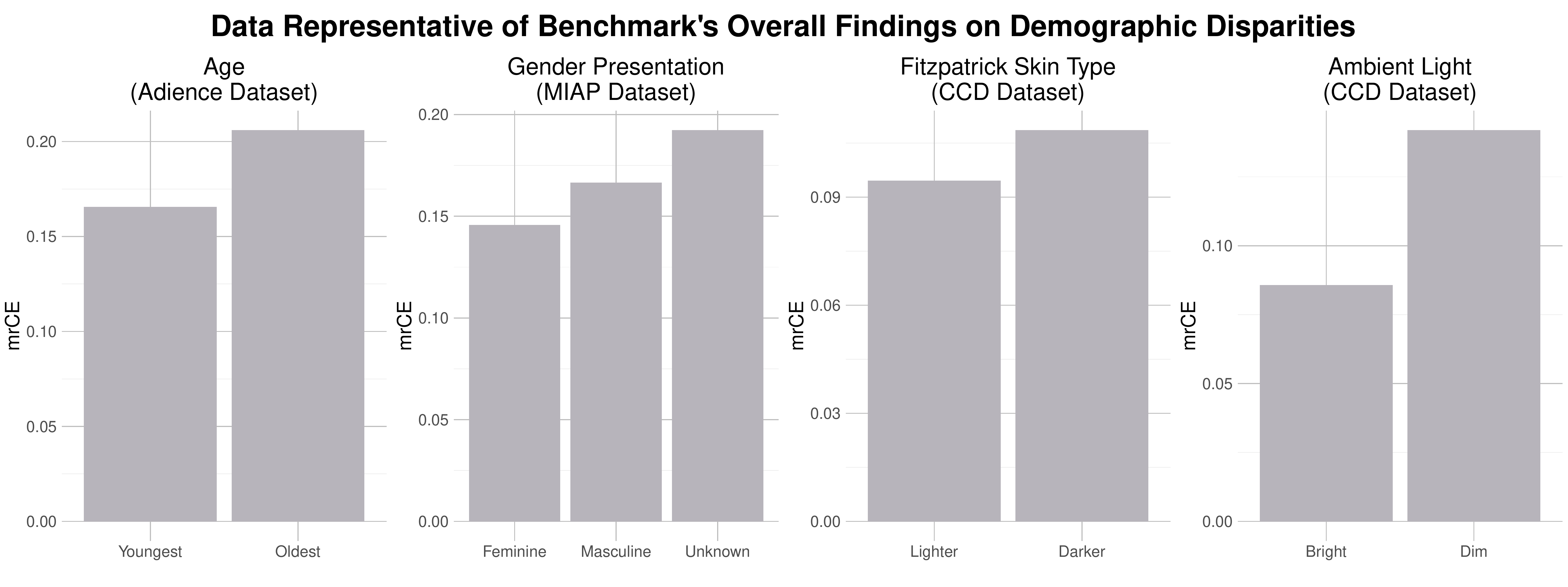}
    \caption{There are disparities in all of the demographics included in this study; we show representative evidence for each demographic on different datasets. On the left, we see (using Adience as an exemplar) that the  oldest two age groups are roughly 25\% more error prone than the  youngest two groups. Using MIAP as an exemplar, masculine presenting subjects are 20\% more error prone than feminine. On the CCD dataset, we find that individuals with Fitzpatrick scales IV-VI have a roughly 25\% higher chance of error than lighter skinned individuals. Finally, dimly lit individuals are 60\% more likely to have errors. }
    \label{fig:overall}
\end{figure}

We now report the main results of our benchmark, a synopsis of which is in Figure~\ref{fig:overall}. Our main results are derived from one regression for each dataset. Each regression includes all demographic variables and each variable is normalized for consistency across the datasets. Overall, we find that photos of individuals who are \emph{older}, \emph{masculine presenting}, \emph{darker skinned}, or are \emph{dimly lit} are more susceptible to errors than their counterparts. We see that each of these conclusions are consistent across datasets except that UTKFace has masculine presenting individuals as performing better than feminine presenting.


\subsection{ System Performance}\label{sec:comp_system}

\begin{figure}
\centering
\begin{minipage}{.36\textwidth}
  \centering
  \includegraphics[width=\linewidth]{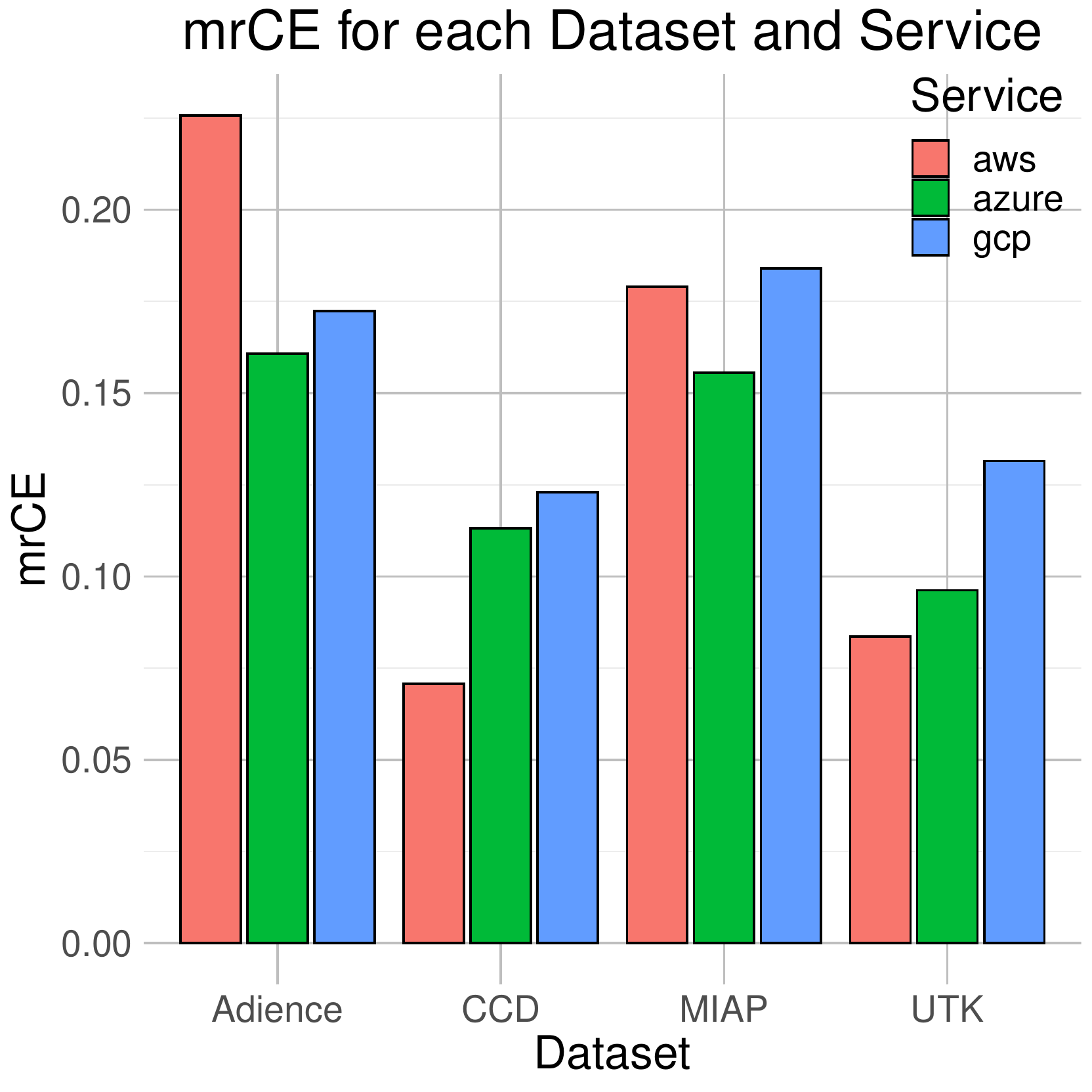}
  \captionof{figure}{Overall performance differences for each dataset and service}
  \label{fig:serice_comp}
\end{minipage}\hfill
\begin{minipage}{.6\textwidth}
  \centering
  \includegraphics[width=\linewidth]{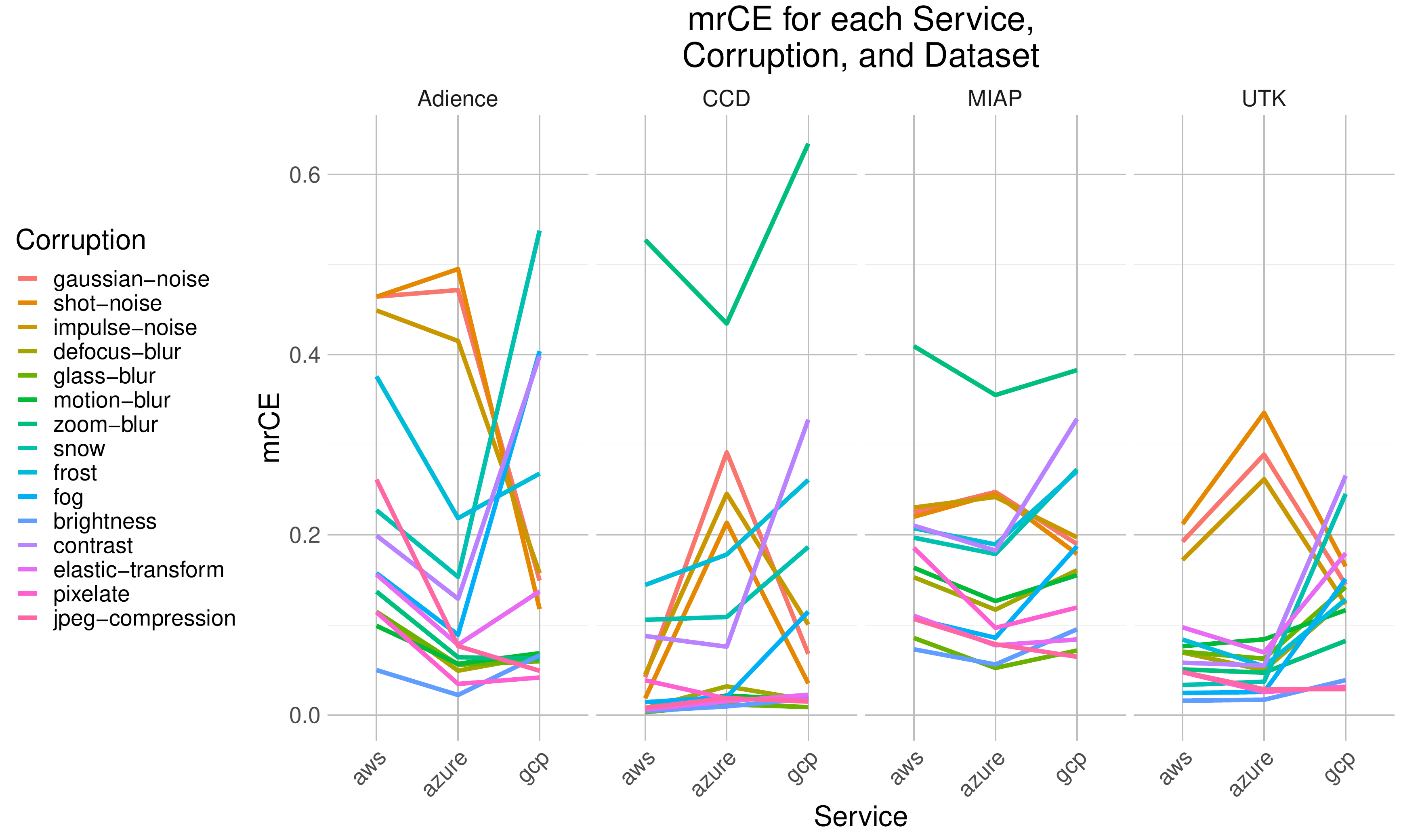}
  \captionof{figure}{A comparison of \mrCE{} for each corruption across each dataset and service.}
  \label{fig:corruption}
\end{minipage}
\end{figure}

We plot \mrCE{} for each dataset and service in Figure~\ref{fig:serice_comp}; the difference between services is statistically significant for each dataset and each service. The only consistent pattern is that GCP is always worse than Azure. AWS is sometimes higher performing than Azure and GCP and sometimes not.

\subsection{Noise corruptions are the most difficult}\label{sec:comp_corr}

Recall that there are four types of ImageNet-C corruptions: noise, blur, weather, and digital. From Figure~\ref{fig:corruption}, we observe that the noise corruptions are markedly some of the most difficult corruptions for Azure to handle across the datasets, whereas GCP has better performance on noise corruptions than Azure and AWS. Though we can only stipulate, these differences might stem from pre-processing steps that each service takes before processing their image. GCP might have a robust noise pre-processing step, which would account for their superior performance with these corruptions. 

The zoom blur corruption proves particularly difficult on the CCD and MIAP datasets, though Azure is significantly better than AWS and GCP on both datasets. We also note that all corruptions for all datasets and commercial systems are significantly differently from zero. 

\subsubsection{Comparison to ImageNet-C results}

We compare the \citet{hendrycks2019benchmarking} findings to our experiments. We recreate Figure 3 from their paper with more current results for recent models since their paper was published, as well as the addition of our findings; see Figure~\ref{fig:imagenetc_comp}. This figure reproduces their metric, mean corruption error and relative mean corruption error. From this figure, we can conclude that our results are very highly in-line with the predictions from the previous data. This indicates that, even with highly accurate models, accuracy is a strong predictor of robustness.

We also examined the corruption-specific differences between our findings (with face data) and that of the original paper (with ImageNet data). We find that while ImageNet datasets are most susceptible to blurs and digital corruptions, facial datasets are most susceptible to noise corruptions, zoom blur, and weather. These qualitative differences deserve future study.

\subsection{Errors increase on older subjects}

\begin{figure}
\centering
  \includegraphics[width=.23\textwidth]{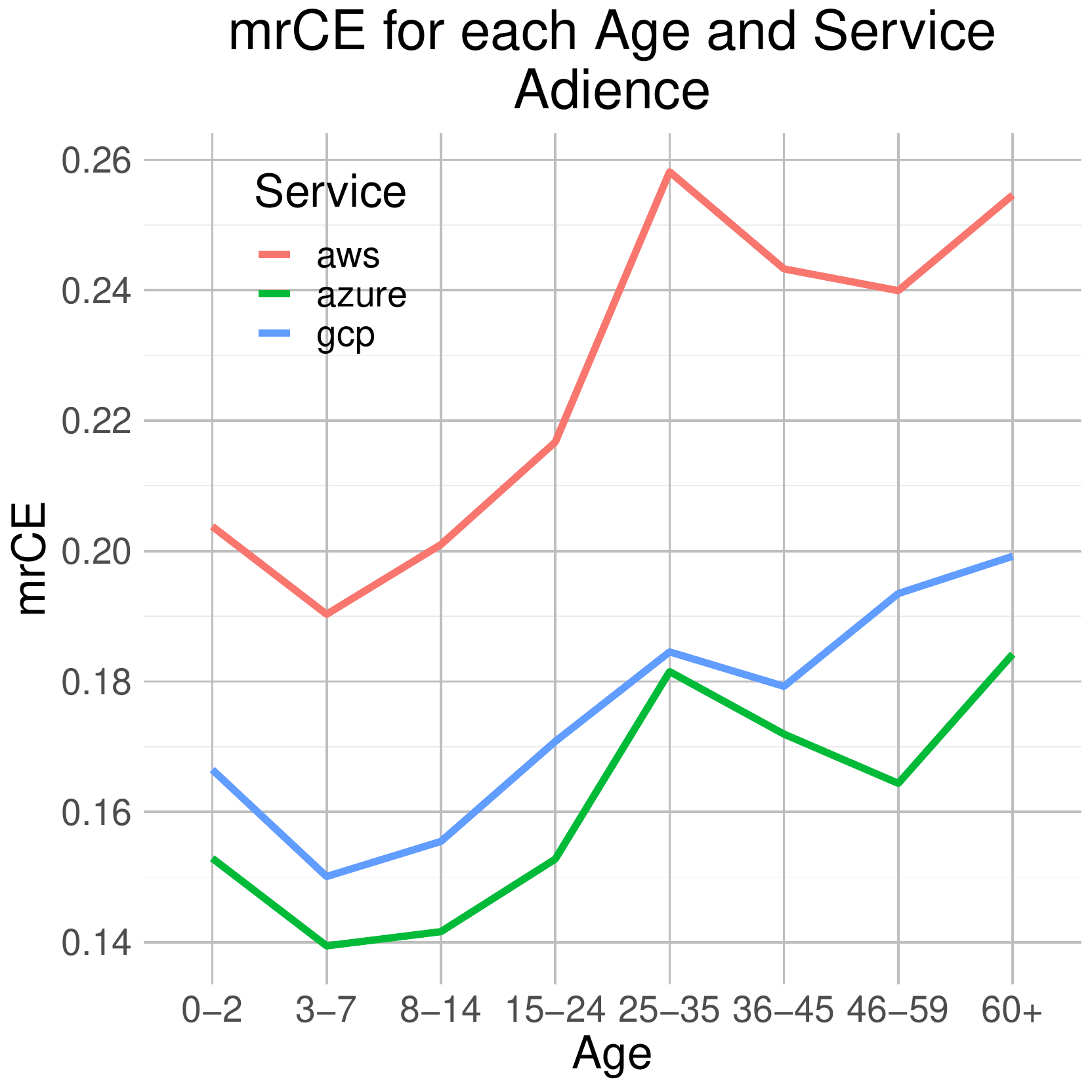}
  \includegraphics[width=.23\textwidth]{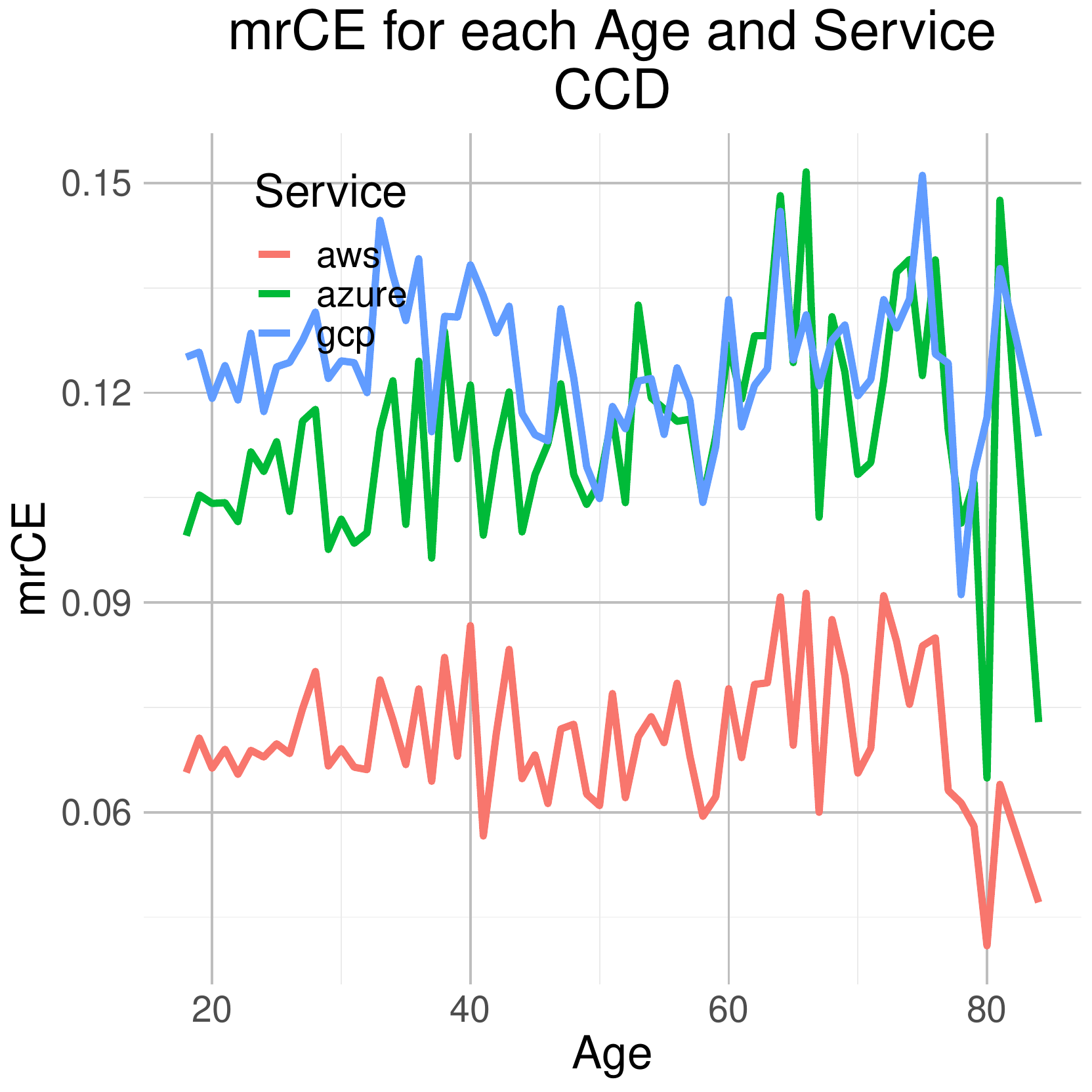}
  \includegraphics[width=.23\textwidth]{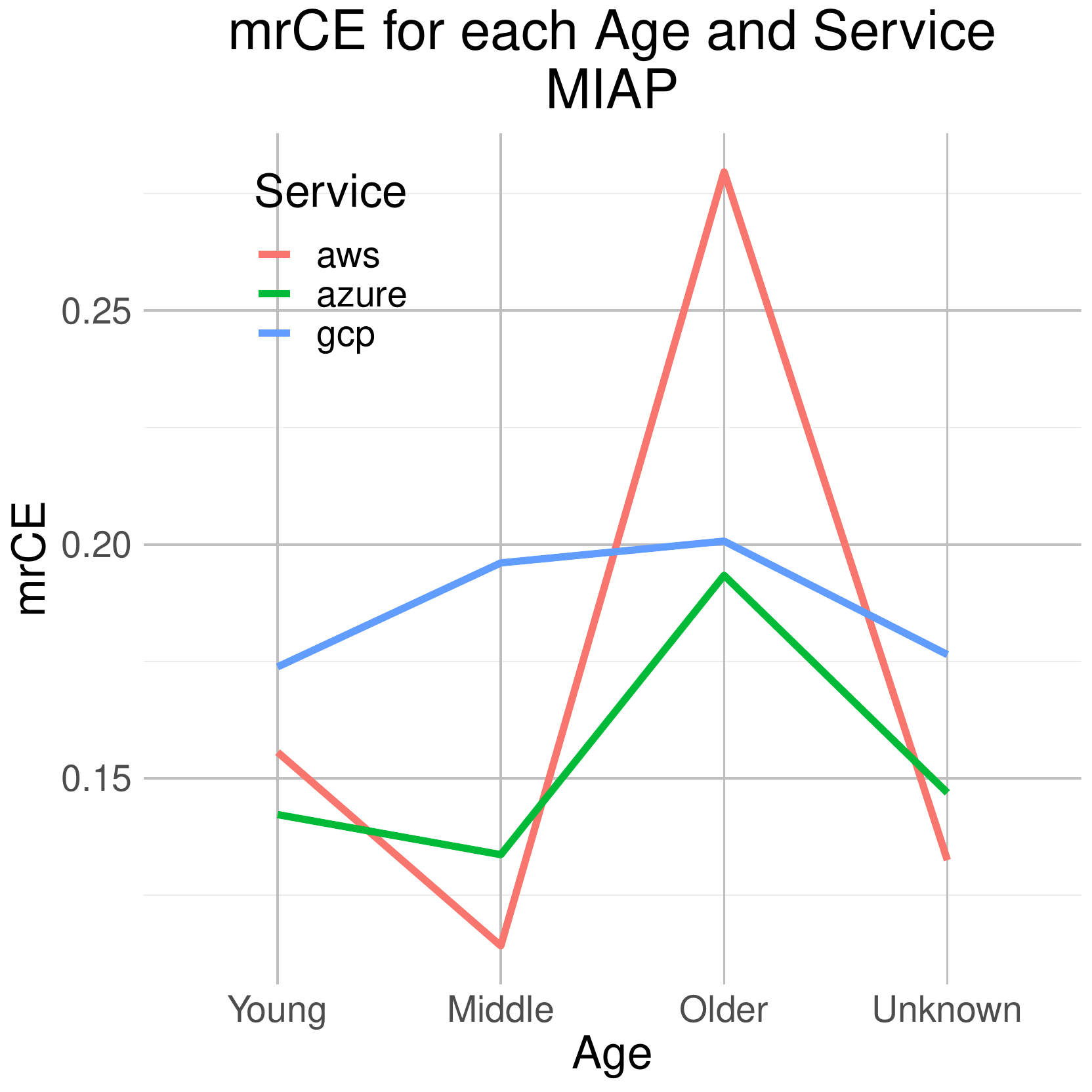}
  \includegraphics[width=.23\textwidth]{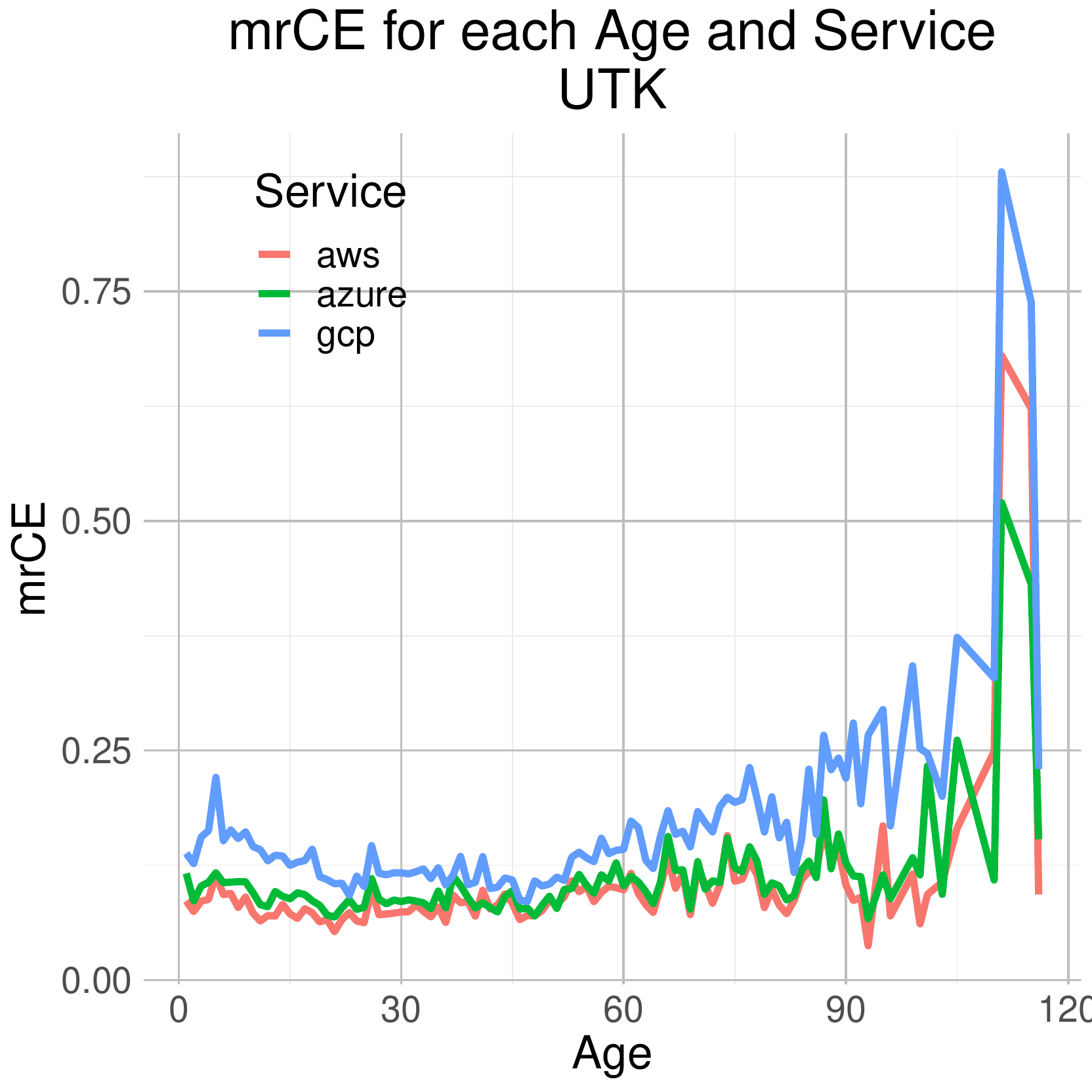}
\caption{Each figure depicts the \mrCE{} across ages. Each line depicts a commercial system. Age is a categorical variable for Adience and MIAP but a numeric for CCD and UTKFace.}
\label{fig:comp_age}
\end{figure}

We observe a significant impact of age on $\mrCE{}$; see Figure~\ref{fig:comp_age}. In every dataset and every commercial system, we see that older subjects have significantly higher error rates.

On the Adience dataset , the odds of error for the oldest group is 31\% higher than that of the youngest group. Interestingly, the shape of the \mrCE{} curves across the age groups is similar for each service. For the MIAP dataset, the age disparity is very pronounced. In AWS for instance, we see a 145\% increase in error for the oldest individuals. The overall odds ratio between the oldest and youngest is 1.383.

The CCD and UTKFace datasets have numeric age. Analyzing the  regressions indicates that for every increase of 10 years, there is a 2.3\% increase in the likelihood of error on the CCD data and 2.7\% increase for UTKFace data.

\subsection{Masculine presenting individuals have more errors than feminine presenting}

\begin{figure}
\centering
\parbox{.29\textwidth}{%
\centering
  \includegraphics[width=.29\textwidth]{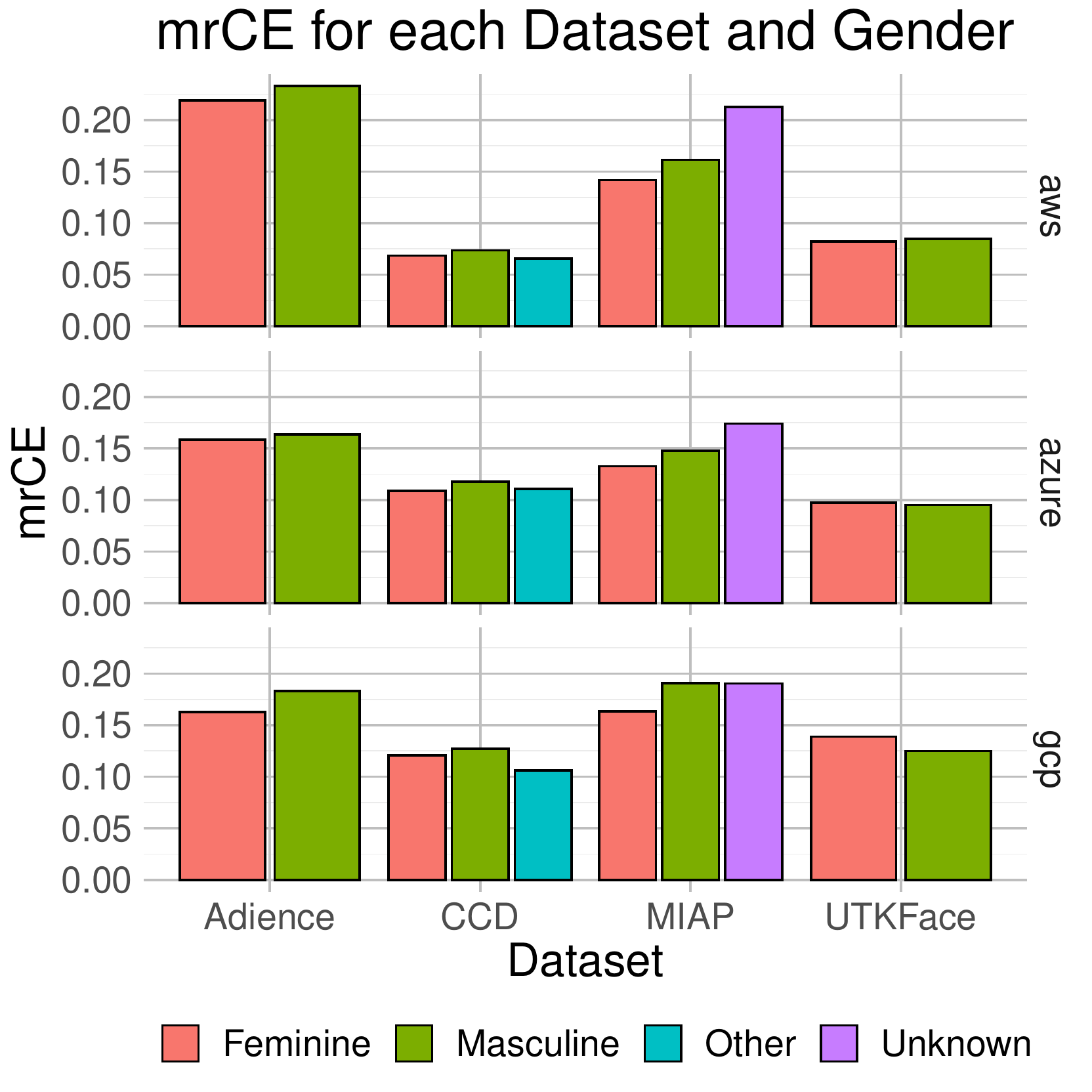}
\caption{Observe that on all datasets, except for UTKFace, feminine presenting individuals are more robust than masculine.}%
\label{fig:comp_gender}}%
\qquad
\begin{minipage}{.3\textwidth}%
  \centering
  \includegraphics[width=\textwidth]{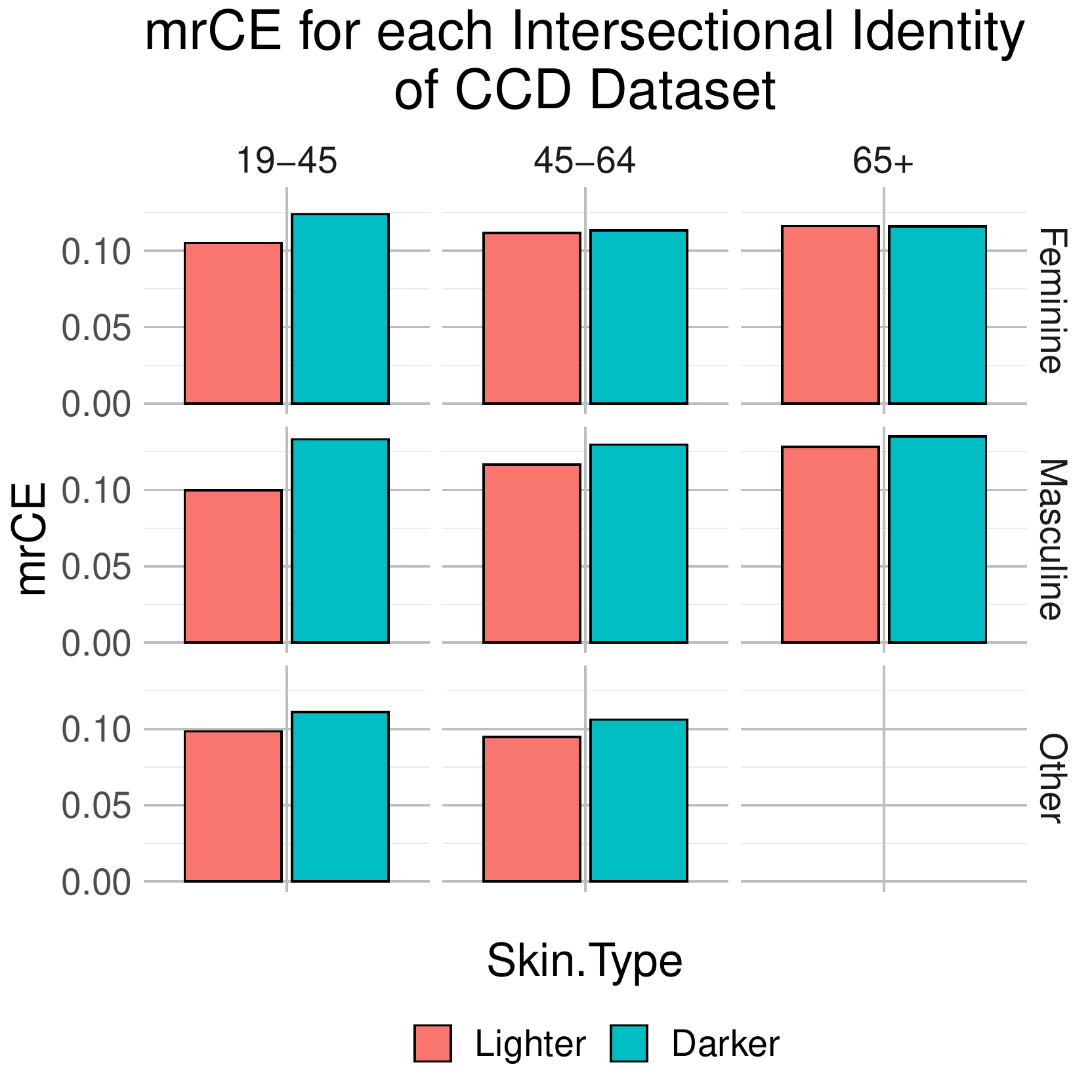}
\caption{In all intersectional identities, darker skinned individuals are less robust than those who are lighter skinned.}%
\label{fig:comp_skin}%
\end{minipage}%
\qquad
\begin{minipage}{.3\textwidth}%
  \centering
  \includegraphics[width=\textwidth]{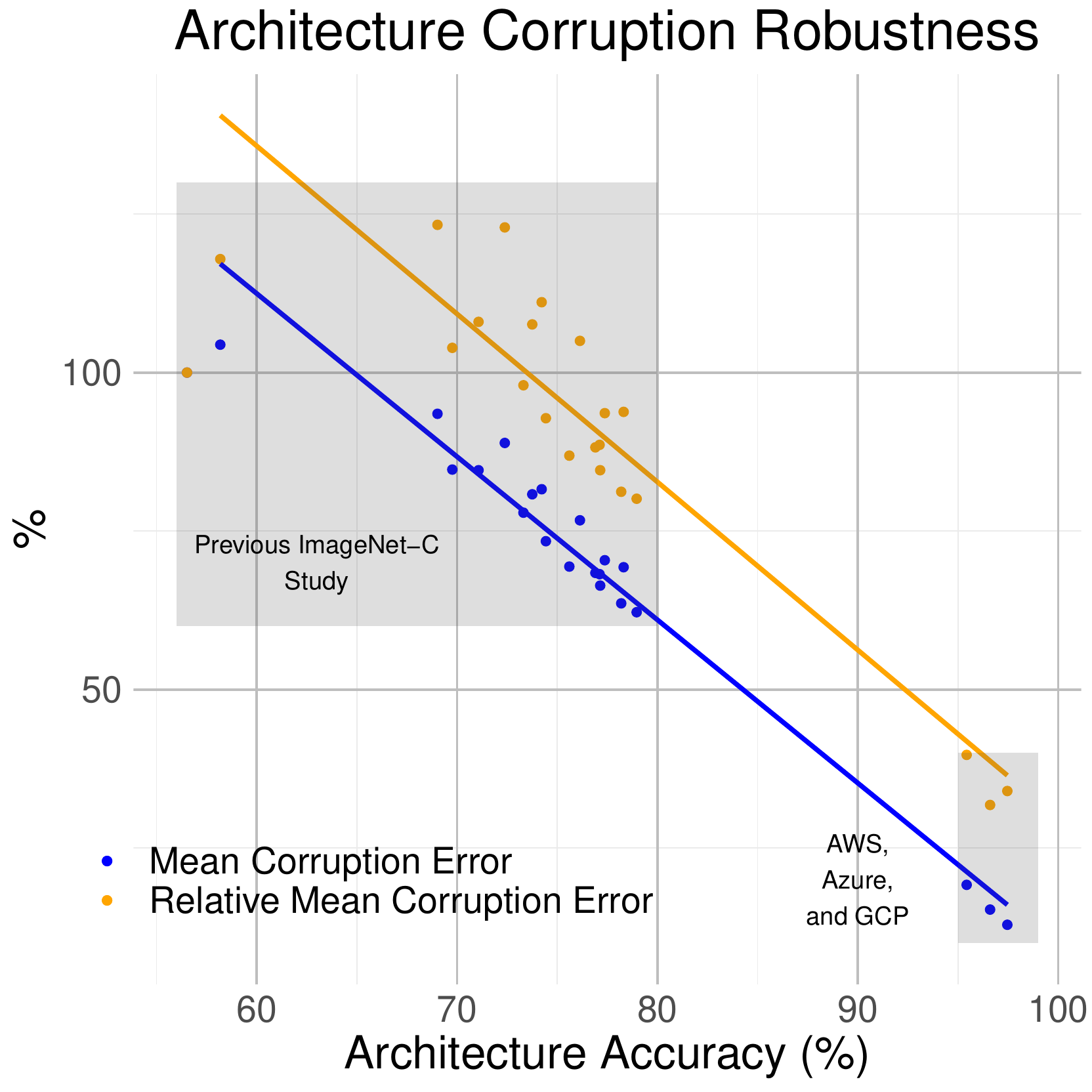}
\caption{ Recreation of Figure 3 from \citet{hendrycks2019benchmarking} with contemporary results and the addition of our findings.}%
\label{fig:imagenetc_comp}%
\end{minipage}%
\end{figure}

Across all datasets except UTKFace, we find that feminine presenting individuals have lower errors than masculine presenting individuals. See Figure~\ref{fig:comp_gender}. On Adience, feminine individuals have 18.8\% \mrCE{} whereas masculine have 19.8\%. On CCD, the $\mrCE{}$s are 8.9\% and 9.6\% respectively. On the MIAP dataset, the $\mrCE{}$ values are 13.7\% and 15.4\% respectively. On the UTKFace, both gender presentations have around 9.0\% $\mrCE{}$ (non statistically significant difference).

Stepping outside the gender binary, we have two insights into this from these data. In the CCD dataset, the subjects were asked to self-identify their gender. Two individuals selected Other and 62 others did not provide a response. Those two who chose outside the gender binary have a $\mrCE{}$ of 4.9\%. When we include those individuals without gender labels, their $\mrCE{}$ is 8.8\% and not significantly different from the feminine presenting individuals.

The other insight comes from the MIAP dataset where subjects were rated on their perceived gender presentation by crowdworkers; options were ``Predominantly Feminine", ``Predominantly Masculine", and "Unknown". For those ``Unknown", the overall $\mrCE{}$ is 19.3\%. The creators of the dataset automatically set the gender presenation of those with an age presentation of ``Young" to be ``Unknown". The $\mrCE{}$ of those annotations which are not ``Young" and have an ``Unknown" gender presentation raises to 19.9\%. One factor that might contribute to this phenomenon is that individuals with an ``Unknown'' gender presentation might have faces that are occluded or are small in the image. Further work should be done to explore the causes of his discrepancy.

\subsection{Dark skinned subjects have more errors across age and gender identities}

\begin{figure}
\centering
  \includegraphics[width=.39\textwidth, height=2.2in]{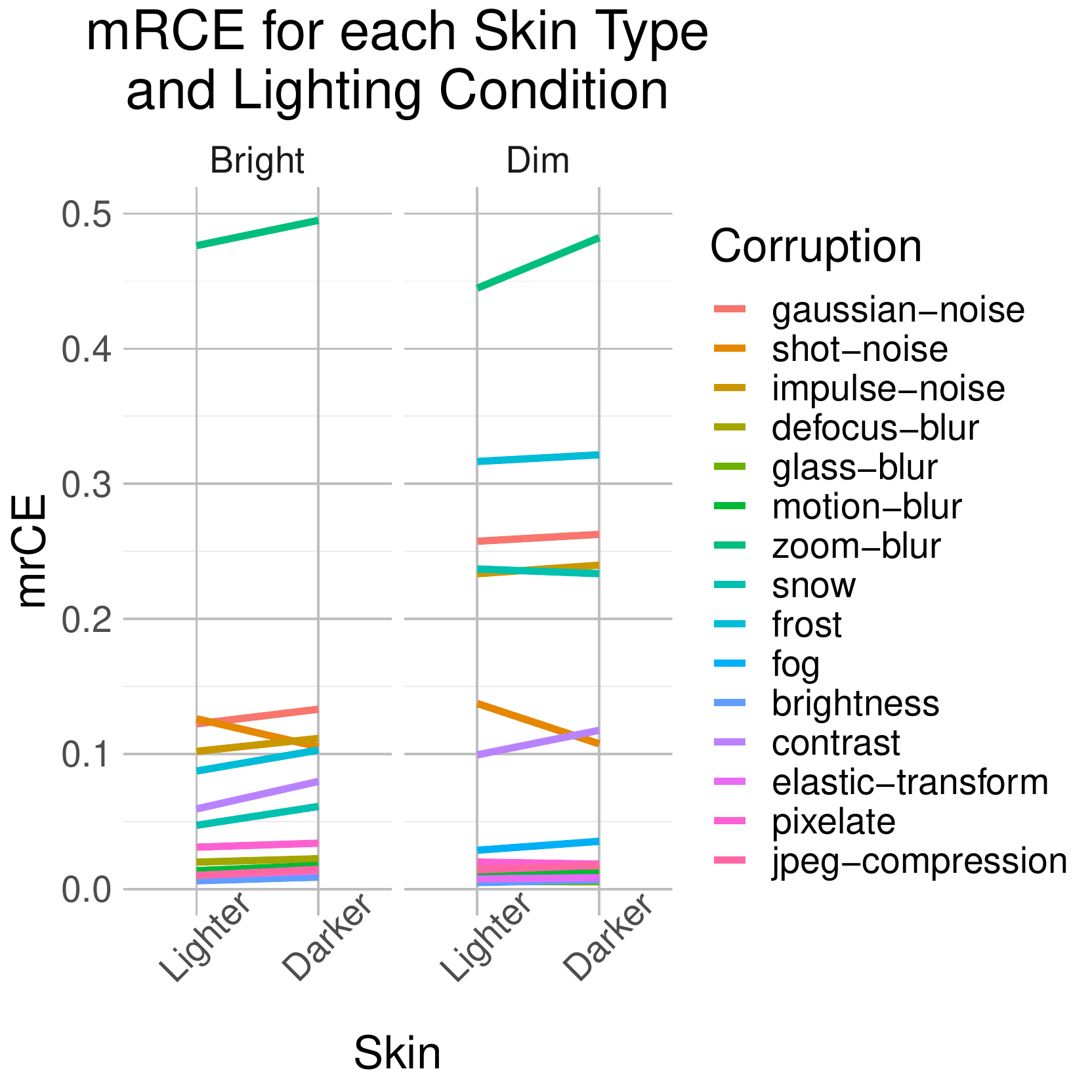}
  \includegraphics[width=.6\textwidth, height=2.2in]{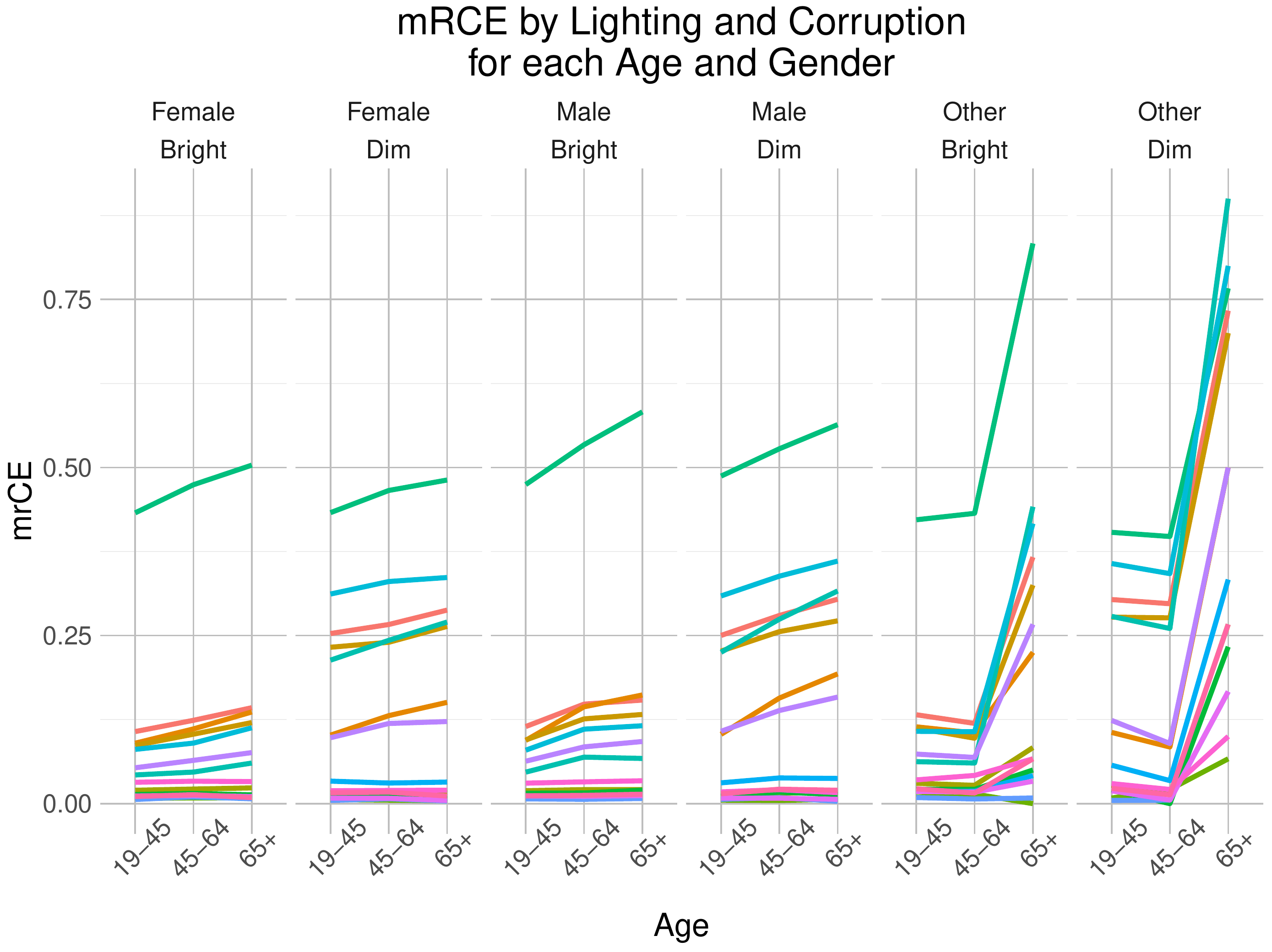}
\caption{(Left) \mrCE{} is plotted for each corruption by the intersection of lighting condition and skin type. (Right) the same is plotted by the intersection of age, gender, and lighting. Observe that for both skin types, all genders, and all ages, the dimly lit environment increases the error rates.}
\label{fig:comp_lighting}
\end{figure}

We analyze data from the CCD dataset which has ratings for each subject on the Fitzpatrick scale. As is customary in analyzing these ratings, we split the six Fitzpatrick values into two: Lighter (for ratings I-III) and Darker for ratings (IV-VI). The main intersectional results are reported in Figure~\ref{fig:comp_skin}.

The overall $\mrCE{}$ for lighter and darker skin types are 8.5\% and 9.7\% respectively, a 15\% increase for the darker skin type. We also see a similar trend in the intersectional identities available in the CCD metadata (age, gender, and skin type). We see that in every identity (except for 45-64 year old and Feminine) the darker skin type has statistically significant higher error rates. This difference is particularly stark in 19-45 year old, masculine subjects. We see a 35\% increase in errors for the darker skin type subjects in this identity compared to those with lighter skin types. For every 20 errors on a light skinned, masculine presenting individual between 18 and 45, there are 27 errors for dark skinned individuals of the same category.


\subsection{Dim lighting conditions has the most severe impact on errors}

Using lighting condition information from the CCD dataset, we observe the $\mrCE{}$ is substantially higher in dimly lit environments: 12.5\% compared to 7.8\% in bright environments. See Figure~\ref{fig:comp_lighting}.

Across the board, we generally see that the disparity in demographic groups decreases between bright and dimly lit environments. For example, the odds ratio between dark and light skinned subjects is 1.09 for bright environments, but decreases to 1.03 for dim environments. This is true for age groups (e.g., odds ratios 1.150 (bright) vs 1.078 (dim) for 45-64 compared to 19-45; 1.126 (bright) vs 1.060 (dim) for Males compared to Females). This is not true for individuals with gender identities as Other or omitted -- the disparity increases (1.053 (bright) vs 1.145 (dim) with Females as the reference).

In Figure~\ref{fig:comp_lighting} we observe the lighting differences for different intersectional identities across corruptions. We continue to see zoom blur as the most challenging corruption. Interestingly, the noise and some weather corruptions have a large increase in their errors in dimly lit environments across intersectional identities whereas many of the other corruptions do not.

\subsection{Older subjects have higher gender error disparities}

\begin{figure}
    \centering
    \includegraphics[width=\textwidth]{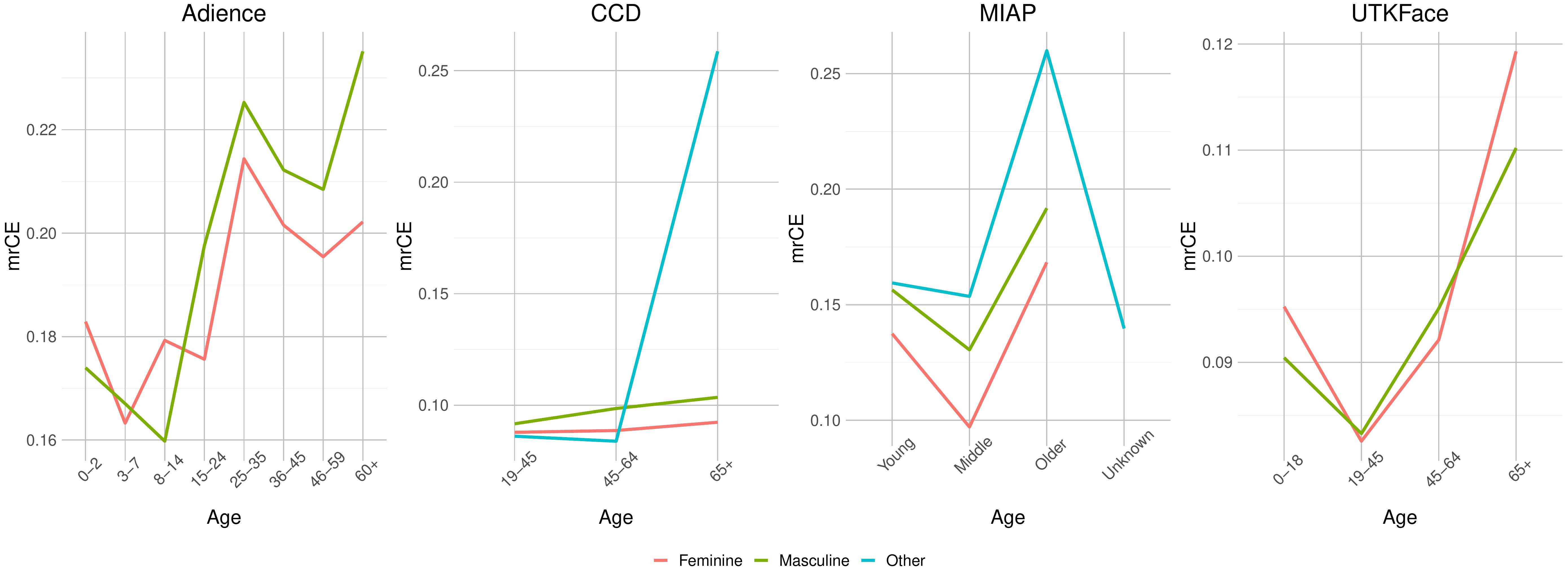}
    \caption{For each dataset, the \mrCE{} is plotted across age groups. Each gender is represented and indicates how gender disparities change across the age groups.}
    \label{fig:comp_age_gender}
\end{figure}

We plot in Figure~\ref{fig:comp_age_gender} the \mrCE{} for each dataset across age with each gender group plotted separately. From this, we can note that on the CCD and MIAP dataset, the masculine presenting group is always less robust than the feminine. On the CCD dataset, the disparity between the two groups increases as the age increases (odds ratio of 1.040 for 19-45 raises to 1.138 for 65+). On the MIAP dataset, the odds ratio is greatest between masculine and feminine for the middle age group (1.395). The disparities between the ages also increases from feminine to masculine to unknown gender identities.

On the Adience and UTKFace datasets, we see that the feminine presenting individuals sometimes have higher error rates than masculine presenting subjects. Notably, the most disparate errors in genders on these datasets occurs at the oldest categories, following the trend from the other datasets.

\section{Gender and Age Estimation Analysis}\label{sec:age_gender_prediction}

We briefly overview results from evaluating AWS's age and gender estimation commercial systems. 

\subsection{Gender estimation is at least twice as susceptible to corruptions as face detection}

The use of automated gender estimates in ML is a controversial topic. Trans and gender queer individuals are often ignored in ML research, though there is a growing body of research that aims to use these technologies in an assistive way as well \citep[e.g.,][]{ahmed2019bridging,chong2021exploring}.
To evaluate gender estimation, we only use CCD as the subjects of these photos voluntarily identified their gender. We omit from the analysis any individual who either did not choose to give their gender or falls outside the gender binary because AWS only estimates Male and Female.

AWS misgenders 9.1\% of the clean images but 21.6\% of the corrupted images. Every corruption performs worse on gender estimation than \mrCE{}. Two corruptions (elastic transform and glass blur) do not have statistically different errors from the clean images. All the others do, with the most significant being zoom blur, Gaussian noise, impulse noise, snow, frost, shot noise, and contrast. Zoom blur's probability of error is 61\% and Gaussian noise is 32\%. This compares to \mrCE{} values of 43\% and 29\% respectively. 

\subsection{Corrupted images error in their age predictions by 40\% more than clean images}

To estimate Age, AWS returns an upper and lower age estimation. Following their own guidelines on face detection~\citep{AWSGuidelines}, we use the mid-point of these numbers as a approximate estimate. On average, the estimation is 8.3 years away from the actual age of the subject for corrupted data, this compares to 5.9 years away for clean data. 

\section{Conclusion}

This benchmark has evaluated three leading commercial facial detection and analysis systems for their robustness against common natural noise corruptions. Using the 15 ImageNet-C corruptions, we measured the relative mean corruption error as measured by comparing the number of faces detected in a clean and corrupted image. We used four academic datasets which included demographic detail.

We observed through our analysis that there are significant demographic disparities in the likelihood of error on corrupted data. We found that older individuals, masculine presenting individuals, those with darker skin types, or in photos with dim ambient light all have higher errors ranging from 20-60\%. We also investigated questions of intersectional identities finding that darker males have the highest corruption errors. As for age and gender estimation, corruptions have a significant and sizeable impact on the system's performance; gender estimation is more than twice as bad on corrupted images as it is on clean images; age estimation is 40\% worse on corrupted images.

Future work could explore other metrics for evaluating face detection systems when ground truth bounding boxes are not present. While we considered the length of response on clean images to be ground truth, it could be viable to treat the clean image's bounding boxes as ground truth and measure deviations therefrom when considering questions of robustness. Of course, this would require a transition to detection-based metrics like precision, recall, and $F$-measure.

We do not explore questions of causation in this benchmark. We do not have enough different datasets or commercial systems to probe this question through regressions or mixed effects modeling. We do note that there is work that examines causation questions with such methods like that of \citet{best2017longitudinal} and \citet{cook2019demographic}. With additional data and under similar benchmarking protocols, one could start to examine this question. However, the black-box nature of commercial systems presents unique challenges to this endeavor.

\chapter{Comparing Human and Machine Bias in FaceRecognition} 

\label{chpt:fr_hum} 
This work was done in collaboration with my two co-first authors, Ryan Downing and  George Wei, as well as Nathan Shankar,  Bradon Thymes,  Gudrun Thorkelsdottir, Tiye Kurtz-Miott,  Rachel Mattson,  Olufemi Obiwumi, Valeriia Cherepanova,  Micah Goldblum,  John P. Dickerson, and Tom Goldstein. See~\cite{dooley2021comparing}.

\section{Introduction}

Facial analysis systems have been the topic of intense research for decades, and instantiations of their deployment have been criticized in recent years for their intrusive privacy concerns and differential treatment of various demographic groups. Companies and governments have deployed facial recognition systems~\citep{weise2020,derringer2019surveillance,hartzog2020secretive} which have a wide variety of applications from relatively mundane, e.g., improved search through personal photos~\citep{GooglePhotosFRT}, to rather controversial, e.g., target identification in warzones~\citep{marson2021}.
A flashpoint issue for facial analysis systems is their potential for biased results by demographics~\citep{garvie2016perpetual,lohr2018facial,buolamwini2018gendershades,grother2019face,dooley2021robustness}, which make facial recognition systems controversial for socially important applications, such as use in law enforcement or the criminal justice system.  To make things worse, many studies of machine bias in face recognition use datasets which themselves are imbalanced or riddled with errors, resulting in inaccurate measurements of machine bias.

It is now widely accepted that computers perform as well as or better than humans on a variety of facial recognition tasks~\citep{lu2015surpassing, grother2019face} in terms of {\em accuracy}, but what about {\em bias}? The algorithm's superior overall performance, as well as speed to inference, makes the use of facial recognition technologies widely appealing in many domain areas and comes at enhanced costs to those surveilled, monitored, or targeted by their use~\citep{kostka2021between}. Many previous studies which examine and critique these technologies through algorithmic audits do so only up to the point of the algorithm's biases. They stop short of comparing these biases to that of their human alternatives. In this study, we question how the bias of the algorithm compares to human bias in order to fill in one of the largest omissions in the facial recognition bias literature.

We investigate these questions by creating a dataset through extensive hand curation which improves upon previous facial recognition bias auditing datasets, using images from two common facial recognition datasets~\citep{huang2008labeled,liu2015faceattributes} and fixing many of the imbalances and erroneous labels.  Common academic datasets contain many flaws that make them unacceptable for this purpose.  For example, they contain many duplicate image pairs that differ only in their compression scheme or cropping.  As a result, it is quite common for an image to appear in both the gallery and test set when evaluating image models, which distorts accuracy statistics when evaluating on either humans or machines. Standard datasets also contain many incorrect labels and low quality images, the prevalence of which may be unequal across different demographic groups.

We also create a survey instrument that we administer to a sample of non-expert human participants ($n=\nsurvey{}$) and ask machine models (both through academically trained models and commercial APIs) the same survey questions. In comparing the results of these two modalities, we conclude that:
\begin{enumerate}
    \item Humans and academic models both perform better on questions with male subjects,
    \item Humans and academic models both perform better on questions with light-skinned subjects,
    \item Humans perform better on questions where the subject looks like they do, and
    \item Commercial APIs are phenomenally accurate at facial recognition and we could not evaluate any major disparities in their performance across racial or gender lines.
\end{enumerate}
Overall we found that computer systems, while far more accurate than non-expert humans, sometimes have biases that are detectable at a statistically significant level on $t$-tests and logistic regressions.  However, when bias was detected in our studies it was comparable in magnitude to human biases.

\section{Background and Prior Work}

We provide a brief overview of facial recognition and additional related work. We further detail similar comparative studies which contrast the performance of humans and machines. Much of the discussion of bias overlaps with the sub-field of machine learning that focuses on social and societal harms. We refer the reader to~\citet{chouldechova2018frontiers} and~\citet{fairmlbook} for additional background of that broader ecosystem and discussion around bias in machine learning.

\paragraph{Facial Recognition}
In this overview, we focus on a review of the types of facial recognition technology rather than contrasting different implementations thereof. Within facial recognition, there are two large categories of tasks: verification and identification. Verification asks a 1-to-1 question: is the person in the source image the same person as in the target image? Identification asks a 1-to-many question: given the person in the source image, where does the person appear within a gallery composed of many target identities and their associated images, if at all? Modern facial recognition algorithms, such as~\citet{he2016deep,chen2018mobilefacenets,wang2018cosface} and~\citet{deng2019arcface}, use deep neural networks to extract feature representations of faces and then compare those to match individuals. An overview of recent research on these topics can be found in~\citet{wang2018deep}. Other types of facial analysis technology include face detection, gender or age estimation, and facial expression recognition.

\paragraph{Bias in Facial Recognition}
Bias has been studied in facial recognition for the past decade. Early work, like that of~\citet{klare2012face} and~\citet{o2012demographic}, focused on single-demographic effects (specifically, race and gender), whereas the more recent work of~\citet{buolamwini2018gendershades} uncovers unequal performance from an intersectional perspective, specifically between gender and skin tone.  The latter work has been and continues to be hugely impactful both within academia and at the industry level. For example, the 2019 update to NIST FRVT specifically focused on demographic mistreatment from commercial platforms by focusing on performance at the group and subgroup level~\citep{grother2019face}. 

While our work focuses on the identification and comparison of bias, existing work on remedying the ills of socially impactful technology and  unfair systems can be split into three (or, arguably, four~\citep{savani2020posthoc}) focus areas: pre-, in-, and post-processing. Pre-processing work largely focuses on dataset curation and preprocessing~\citep[e.g.,][]{Feldman2015Certifying, ryu2018inclusivefacenet, quadrianto2019discovering, wang2020mitigating}. In-processing often constrains the ML training method or optimization algorithm itself~\citep[e.g.,][]{zafar2017aistats, Zafar2017www, zafar2019jmlr, donini2018empirical, goel2018non,Padala2020achieving, agarwal2018reductions, wang2020mitigating,martinez2020minimax,diana2020convergent,lahoti2020fairness}, or focuses explicitly on so-called fair representation learning~\citep[e.g.,][]{adeli2021representation,dwork2012fairness,zemel13learning,edwards2016censoring,madras2018learning,beutel2017data,wang2019balanced}. Post-processing techniques adjust decisioning at inference time to align with quantitative fairness definitions~\citep[e.g.,][]{hardt2016equality,wang2020fairness}.

\paragraph{Human Performance Comparisons}
No work in the past to our knowledge has specifically focused on the question of comparing bias or disparity between humans and machines. Some prior work has looked at comparing overall performance or accuracy between the two groups. \citet{tang2004face,o2007face,phillips2014comparison} compare human and computer-based face verification performance.  \citet{lu2015surpassing} was the first paper to show machine accuracy outpacing human accuracy. \citet{hu2017person,phillips2018face,robertson2016face} compared face recognition performance of human specific sub-populations 
whereas \citet{white2015error} looked at comparing overall performance of humans who use the \emph{outputs} of face recognition systems. 

\section{\dataname{} Dataset Curation}

We endeavor to answer two research questions: 
{\bf (RQ1) How and to what extent do humans exhibit bias in their accuracy in facial recognition tasks? (RQ2) How does this compare to machine learning-based models?} 
In order to answer these questions, we created a set of challenging identification and verification questions which we posed to humans and machines from a novel dataset called \dataname{} for its application in intersectional facial recognition. The protocol around those experiments are described in Section~\ref{sec:experiments}.

To create our dataset, we first ensured that we had accurately labeled and balanced metadata. This required us to hand-check all the labels in the dataset.  After removing poor quality and redundant images, we found that LFW lacked identities with dark skin tones, which is why further identities were drawn from CelebA. 
Though LFW does have an errata page, CelebA and other facial recognition datasets are known to have many missing or incomplete metadata, and so all CelebA images were examined by an author of this paper before adding them to the dataset. Finally, after randomly generating survey questions, we hand checked that there were no questions for which the answer is apparent or unclear for reasons other than properties of the faces (see Figure~\ref{fig:shortcomings}). In this section we detail our findings about the shortcomings in the metadata labels from LFW and CelebA and outline the steps we took to rectify and supplement these in the creation of the \dataname{} identities.

\subsection{The shortcomings of previous datasets}

\begin{figure}[t]
        \centering
        \begin{subfigure}[t]{.35\textwidth}
            \centering
            \includegraphics[width=.3543\linewidth]{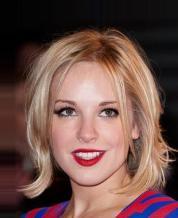}
            \includegraphics[width=.3543\linewidth]{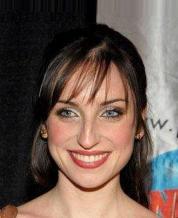}
            \caption{Incorrect Identities: these are labeled as the same; but the left is Zoë Lister and the right is Zoe Lister-Jones. }
            \label{fig:identities}
        \end{subfigure}\hfill
        \begin{subfigure}[t]{.28\textwidth}
            \centering
            \includegraphics[width=.4\linewidth]{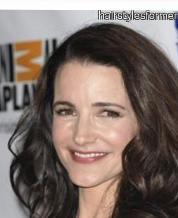}
            \caption{Incorrect Labelling: this individual was labeled as not being pale skinned.}
            \label{fig:labelling}
        \end{subfigure}\hfill
        \begin{subfigure}[t]{.33\textwidth}
            \centering
            \includegraphics[width=.3543\linewidth]{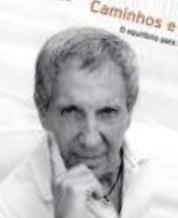}
            \includegraphics[width=.3543\linewidth]{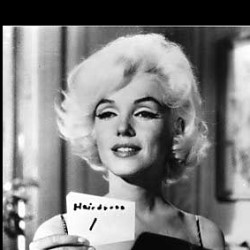}
            \caption{Black and White Images: some identities only have black and white photos.}
            \label{fig:bw}
        \end{subfigure}
        
        \medskip
        
        \begin{subfigure}[t]{.4\textwidth}
            \centering
            \includegraphics[width=.3\linewidth]{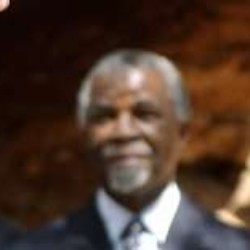}\hfill
            \includegraphics[width=.3\linewidth]{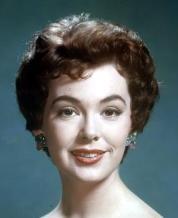}\hfill
            \includegraphics[width=.3\linewidth]{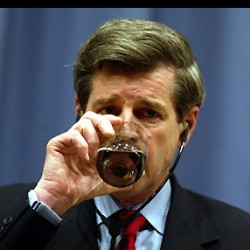}
            \caption{Low-Quality Images}
            \label{fig:qual}
        \end{subfigure}\hfill
        \begin{subfigure}[t]{.35\textwidth}
            \centering
            \includegraphics[width=.4\linewidth]{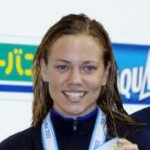}
            \includegraphics[width=.4\linewidth]{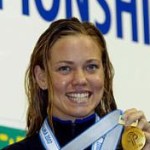}
            \caption{Identical Attire/Background}
            \label{fig:att}
        \end{subfigure}\hfill
        \begin{subfigure}[t]{.25\textwidth}
            \centering
            \includegraphics[width=.5\linewidth]{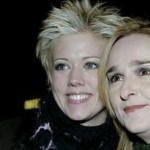}
            \caption{Multiple Distinct Faces}
            \label{fig:mult}
        \end{subfigure}
        
        \caption{Shortcomings present in existing facial identification datasets}
        \label{fig:shortcomings}
    \end{figure}

    In the process of trying to create a reasonable set of identification and verification questions, we identified that the LFW and CelebA datasets generally suffer from a range of problems that distort accuracy and bias metrics. We summarized these problems in~Figure~\ref{fig:shortcomings}.
    
    The first challenge we had to overcome is {\bf incorrect identities}; this includes incorrect names, duplicated identities, as well as clearly incorrect matching between image and name. 
    This problem is particularly harmful for facial recognition models which would be provided with galleries containing incorrect information about identities. 
    In some cases, identities were split across multiple labels due to spellings. We found that this happened almost exclusively with non-canonically western names. E.g., Mesut Ozil (labelled as ``Mesut Zil"), Jithan Ramesh (labelled as ``Githan Ramesh"), Isha Koppikhar (labelled as ``Eesha Koppikhar"), etc. 
    Examples of incorrect identity labels include Neela Rasgotra, a fictional character played by Parminder Singh and ``All That Remains," a band name with the pictured individual being Philip Labonte.
    In other cases, multiple distinct identities were merged into the same label.
    In CelebA, Jennifer Lopez was grouped with Jennifer Driver, and Zoë Lister and Zoe Lister-Jones were both listed under ``Zoe Lister" (pictured in Figure~\ref{fig:identities}). 
        
    Additionally, these datasets exhibit {\bf metadata labelling problems} that manifest in two ways: (1) clearly defined labels being incorrectly or non-uniformly applied, and (2) vague and sometimes harmful metadata. In the first category, CelebA has features such as gender and age which often are incorrect or mislabeled (i.e. a pale-skinned person being labelled as not having pale skin, Figure~\ref{fig:labelling}). Further, many categories in CelebA are subjective and/or harmful. For example, there is a label for ``Attractive," ``Big Nose/Lips," or ``Chubby."

    We found that some identities have {\bf exclusively black and white images} (Figure~\ref{fig:bw}), making it trivial to identity two photos as being of the same label.

    We filtered out {\bf low-quality images} that could not be easily identified for reasons beyond properties of the face, such as poor light exposure, blurriness, facial obstruction, etc. 
    We also removed ``old-timey" photos that were easily associated with a specific time period, as this makes it easy to match them with other similar photos. 
        
    We found that many questions could be answered without considering face features at all, and these were removed. For example if the subject is {\bf wearing identical attire and/or standing in front of an identical background in two images}. Many identities contained multiple images from the same red carpet event or award reception (Figure~\ref{fig:att}).  It {\em very} often happens that the same image appears multiple times in the dataset, but with slightly different crops, compression, or contrast adjustments.  

    Finally, some images {\bf contained multiple faces}. Some of these pictures clearly have one person in the foreground and are therefore not problematic, but in others this is not the case, creating ambiguity as to which person is the target individual. See Figure~\ref{fig:mult}.

    The image types above create inaccuracies when evaluating face recognition systems and distort measurements of bias when these problems occur at rates that differ across groups.  For this reason, many datasets designed for training face analysis systems are not appropriate for evaluating bias. 
    
\subsection{The \dataname{} Identities}
After a thorough review of the LFW and CelebA datasets, random generation of survey questions, and rigorous hand-checking of questions to remove irregularities, we obtained a battery of survey questions for evaluating both humans and machines. We also endeavored to select survey questions that were balanced across gender, age, and skin type. Since LFW is highly skewed towards lighter identities, we included CelebA images and identities as well. We selected identities from LFW with at least two images of an individual, and then we hand labeled each identity for the following: their (1) birth date, (2) country of origin, (3) gender presentation, and (4) Fitzpatrick skin type. Labels 1-3 were assigned by an author of this paper, then that label was checked by at least two other researchers, and modifications were made to achieve agreement among the labelers. Skin type labels (4) were assigned by 8 raters, and the mode was used as the final label.

We note that part of this work does reify categories of gender and skin type that have broader social and political implications. Further, we undertook a task of labeling and categorizing individuals who we do not know and have not received consent from for this task. Every identity for which we created these labels is indeed a celebrity in the public space with Wikipedia entries. Gender labels were rendered from the celebrity's public comments on their own gender identity and/or used pronouns.

The {\bf Fitzpatrick scale}~\citep{fitzpatrick1988validity} was used to help balance the survey to include subjects with diverse skin types. 
This scale is widely used to classify skin complexions into 6 categories. While the Fitzpatrick scale is not perfect, it is the best systematic option currently for ensuring a broadly Representative sample.   

We looked up each celebrity's birth date online, mostly citing Wikipedia, and if we could not find it there, we continued to search on other websites.  However, if we could still not find an individual’s date of birth, we did not list it.  
To find an individual’s {\bf country of origin}, we again cited Wikipedia.  If the individual came from a country that no longer existed (i.e. East and West Germany), we listed the current country.
To label a person’s {\bf gender presentation}, we took note of the person’s preferred pronouns online and in interviews. In the event that their pronouns were not available online, we labeled their gender presentation. A major limitation of the CelebA and LFW datasets is that there were no individuals in our process who identified outside the gender binary or as gender queer.


At the end of our data collection, we collected metadata on 2,545 identities which comprised a total of 7,447 images. The identities themselves are rather imbalanced, though we selected a subgroup from these identities to create a balanced survey, discussed in Section~\ref{sec:experiments}. There are 1,744 lighter-skinned individuals (as defined by Fitzpatrick skin types I-III) and 801 darker-skinned individuals (skin types IV-VI). There are 1,660 males and 885 females. This sample is an improvement over previous datasets as it has been extensively evaluated to remove any errors in labeling and has a robust labeling for a wider array of skin types, unlike previous datasets which chose to label individuals as ``pale.''  
These data have a range of potential future use cases, such as being used for more evaluative facial recognition studies and commercial system audits.

\section{Experiments}\label{sec:experiments}

With the high-quality metadata provided in the \dataname{} identities, we conduct two experiments that aim to answer our main research questions regarding the performance disparities of humans and machines.  In this section, we outline how we selected the survey questions, administered the survey to human participants, and evaluated machine models. We describe the results in Section~\ref{sec:results}.

\begin{figure}
    \centering
    \includegraphics[width=.4\linewidth]{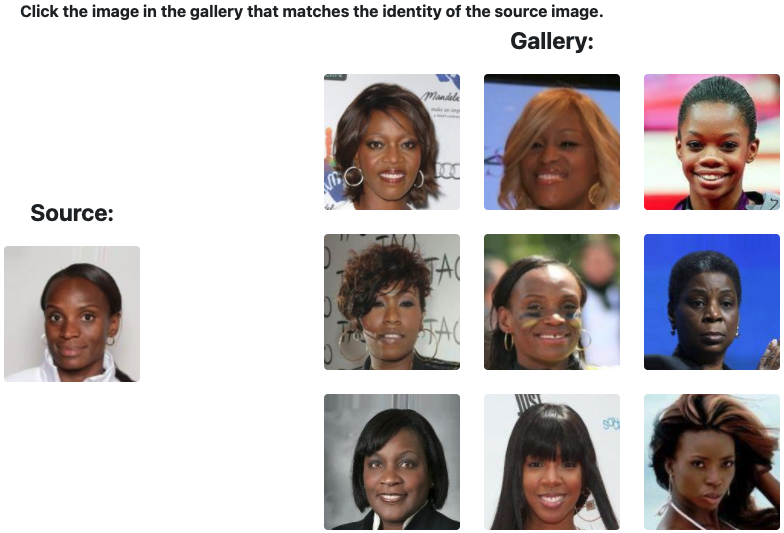}\hfill
    \includegraphics[width=.4\linewidth]{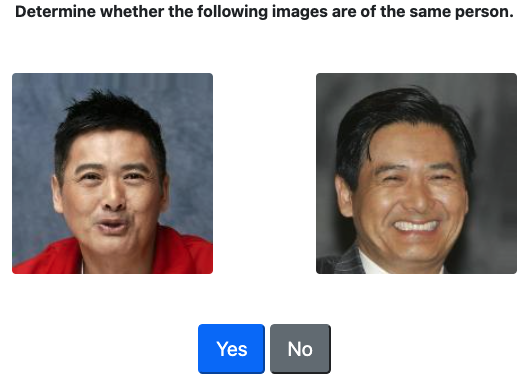}
    \caption{Example questions from the \dataname{} question bank. (Left) An example of an identification question. (Right) An example of a verification question. Notice that the demographics of all identities appearing in a question are matched to ensure the questions are not trivial.}
    \label{fig:queestion_examples}
\end{figure}

For both experiments, we create two types of questions: {\bf identification} and {\bf verification}. Both tasks contain a ``source'' image. In the identification task, 9 other images are presented in a grid, with one being of the same identity as the source and the others being of the same gender and skin type. For the verification task, a second image is selected with equal probability of being the same identity as the source image, or some other of the same gender and skin type as the source.
Examples of these two types of questions can be seen in Figure~\ref{fig:queestion_examples}.

We generated a static question bank with 78 identification questions and 78 verification questions for each of the 12 combinations of gender of skin type.  Of those demographics with more than 78 identities, the source identity for the 78 questions were randomly chosen without replacement. This provided a total of 936 questions for each task.
Finally, a pass was done over all questions to remove any for which context around the face (e.g., background or clothes) could be used to identify a person (e.g., a verification question where both images feature the same sports jersey).
This resulted in a final set of 901 identification questions and 905 verification questions.

\subsection{Human Experiment}\label{sec:human-explain}

\begin{table}
\centering
\captionof{table}{Demographic breakdown of human survey respondents used in final analysis. }
\footnotesize
\resizebox{.6\linewidth}{!}{
\begin{tabular}{ll|ccccc|c}
\toprule
\toprule
                        & Fitzpatrick & Age  & Age   & Age   & Age   & Age & \multirow{2}{*}{Total}  \\
                        &             & 0-19 & 20-39 & 40-59 & 60-79 & 80+ &                         \\ 
\midrule
\multirow{3}{*}{Male}   & I-II        & 0    & 23    & 37    & 33    & 2   & 95                      \\
                        & III-IV      & 1    & 35    & 18    & 24    & 1   & 79                      \\
                        & V-VI        & 4    & 43    & 33    & 17    & 0   & 97                      \\ 
\midrule
\multirow{3}{*}{Female} & I-II        & 0    & 31    & 26    & 36    & 0   & 93                      \\
                        & III-IV      & 4    & 33    & 26    & 27    & 0   & 90                      \\
                        & V-VI        & 1    & 43    & 27    & 20    & 0   & 91                      \\
\bottomrule
\end{tabular}
}
\label{tbl:demo_resp}
\end{table}

We conducted an institutional review board-approved survey. We collected responses through the crowdsource platform Cint. The survey was split into two parts (whose order was randomized), one for each type of question: identification and verification

Each respondent was asked 36 identification questions and 72 verification questions, for a target survey length of around 10 minutes. The questions for each user were randomly sampled from the total question bank such that an even distribution of questions were asked for each demographic group. As such, each respondent was asked 3 identification questions and 6 verification questions for each intersectional demographic identity. When the user first entered the survey they were prompted with a consent form. 
After completing both tasks, respondents filled out a demographic self-identification form which asked the participants their age range, gender, and skin type. When asking respondents to evaluate their own Fitzpatrick skin type scale, we provided a brief description of the scale and respondents were also shown three examples of each skin type from our dataset. 

Within each task, an attention check question was presented after the first five questions and before the last five. For the identification task, the attention check questions used an identical image for the target and in the gallery. For verification, one question consisted of pairing a light skinned female with a dark skin male (obvious negative example), and the other contained two identical images (obvious positive). The images used in these questions do not appear elsewhere in the survey. If a user failed to answer an attention check question correctly, they were screened out and any of their responses were ignored in our analysis. Additionally, any user who passed the attention checks but took fewer than $4$ minutes to complete the survey was dropped from the final analysis. The first $3$ verification and identification questions seen by each user were removed, to account for the possibility that the user may have taken some time to adjust to the format of the questions. 

Our survey sampled English-speaking participants who were 18 years or older and were US residents. Our final sample includes \nsurvey{} participants. There are $146$ self-identified as dark-skinned (Fitzpatrick IV-VI) females, $128$ light-skinned (Fitzpatrick I-III) females, $140$ dark-skinned males, and $131$ light-skinned males. Most respondents ($375$) came from the $20-39$ and $40-59$ age demographics. Participants were compensated between \$2.50 and \$5.00 depending on whether the respondent belongs to a part of the population that is harder or easier to reach. Differential incentive amounts, standard in many survey panels~\citep{pewresearchcenter}, were designed to increase panel survey participation among groups that traditionally have low survey response propensities.

\subsection{Machine Experiments}\label{sec:machine-explain}
We conducted experiments with two types of machine models: academic models which we trained ourselves and commercially-deployed models which we evaluated through APIs. Since we do not have to be concerned about question fatigue with machines, we presented all 901 identification and 905 verification questions to the machines.

\paragraph{Academic Models} To measure algorithmic disparities, we trained 6 face recognition models and evaluated them on \dataname{} questions. We trained ResNet-18, ResNet-50 \citep{he2016deep} and MobileFaceNet \citep{chen2018mobilefacenets} neural networks with CosFace \citep{wang2018cosface} and ArcFace \citep{deng2019arcface} losses, which are designed to improve angular separation of the learned features. For the training data, we used images of 9,277 CelebA identities disjoint from identities selected for the \dataname{} dataset. At inference time, the models solve identification questions by finding the closest gallery image in the angular feature space. To solve verification questions, we threshold the cosine similarity between features extracted from images in the pair. 

\paragraph{Commercial Models} We evaluated three commercial APIs: AWS Rekognition, Microsoft Azure, and Megvii Face++. We were able to evaluate face verification and identification on AWS and Azure, and only face verification on Face++. 
The AWS CompareFace function, which compares a source and target image, was used for both identification and verification; the target image for identification was one image comprised of the nine gallery images stitched together. Azure has native identification and verification built into their Cognitive Services Face API. Face++ has a similar set up to AWS, however they only compare the largest detected faces in the source and target images; thus we were only able to perform face verification.

\subsection{Analysis Strategy}\label{sec:analysis}

We use a two-tailed $t$-test with matched pairs (with a given pair corresponding to a single respondent's or computer model's scores on the two sections) to compare the accuracy rates between tasks. We also use two-tailed, unpaired $t$-tests to compare the overall accuracy of humans on verification questions with the overall accuracy of computer models on verification questions, and the overall accuracy of humans on identification questions with the overall accuracy of computer models on identification questions. The latter $t$-tests and all $t$-tests referred to in the rest of this section are conducted on the question-level: for instance, when comparing the verification accuracy of humans and machines, we use all verification responses from all human test-takers as one sample, and all verification responses from all machines as the other. 

We then analyze the disparity along gender and skin-type categories within our computer algorithms and human survey results. Users and question subjects are binned by skin type. Since the Fitzpatrick is heavily skewed towards Western conceptions of skin tone, we use two categorizations: a binary categorization of ``lighter'' (I-III) and ``darker'' (IV-VI); and categorization by (I-II), (III-IV) and (V-VI). We use two-tailed unpaired $t$-tests to detect the presence of accuracy disparities based on the gender or Fitzpatrick type of the identities that formed the questions. We perform tests of this kind on data from the six individual computer models, and also on the aggregate data sets of all human question responses and all computer algorithm responses. 

We use logistic regression in our analysis to allow us to control for confounding variables. Results are reported as odds ratios, which compare the ratio of odds for a baseline event with the odds for a different event. We consider a main model for human subjects which predicts whether an individual question taken by a respondent was answered correctly, with independent variables as the question target gender and skin-type, and test-taker age, gender, and skin-type. The logistic regressions we run on the computer model responses are similar, but do not include test-taker demographics. We do report separate results for different architectures.

\section{Results}\label{sec:results}

We first provide some overview information about the performance of humans and machines before we move on to answering RQ1 (measuring human bias) and RQ2 (comparing to machine bias). 

\paragraph{Verification is Easier Than Identification; Computers are More Accurate Than Humans}
Humans achieved higher accuracy on verification ($78.9\%$) than identification ($68.3\%$, significant with a two-tailed matched-pair $t$-test with $p < 0.001$). For computer models as a whole, this gap persists but is substantially narrowed -- performance on verification is $94.1\%$, with $92.5\%$ on identification ($p = 0.005$). 

The performance difference between machines and humans is highly significant ($p< 0.001$) on both tasks using unpaired $t$-tests which explore group-level changes between the two tasks. Furthermore, even when controlling for demographic effects in a logistic model, humans have a much lower odds compared to computers of getting a question right (OR = $0.23$ for verification, $p < 0.001$, OR = $0.17$ for identification, $p < 0.001$).  

\begin{table}
\centering
\captionof{table}{Overall gender and skin type disparities exhibited by the human survey respondents, academic models, and commercial APIs. }
\footnotesize
\resizebox{.65\linewidth}{!}{
\begin{tabular}{lll|cccc}
\toprule
                         &       &  & Human & Academic  & Commercial  \\
                         &       &  &  &  Models &  Models \\\midrule
\multirow{4}{*}{Identification} & \multirow{2}{*}{Darker}   & Female & 55.5\% & 89.9\% & 96.7\%       \\
 &    & Male & 73.1\% & 94.1\% & 97.6\%          \\
 & \multirow{2}{*}{Lighter}   & Female & 67.2\% & 91.3\% & 96.7\%                       \\
 &    & Male & 78.3\% & 94.7\% & 98.7\%                       \\\midrule
\multirow{4}{*}{Verification} & \multirow{2}{*}{Darker}   & Female & 73.4\% & 92.0\% & 97.8\%             \\
 &    & Male & 80.1\% & 94.7\% & 99.9\%                       \\
 & \multirow{2}{*}{Lighter}   & Female & 78.7\% & 94.9\% & 97.6\%                       \\
 &    & Male & 83.1\% & 94.9\% & 98.9\%                       \\\bottomrule
\end{tabular}
}
\label{tbl:metric_comp}
\end{table}

\paragraph{Humans and Computers Perform Better on Male Subjects}

For identification questions, we do not observe statistically significant performance gaps for the MobileFaceNet models ($p = 0.3043$ for ArcFace and $p = 0.4752$ for CosFace), but we do observe statistically significant disparities in favor of males for each of the four ResNet models (all $p< 0.04$). 
In logistic regression, we observe an odds ratio for computer models on male identification subjects of $1.76$ ($p < 0.001$). Similarly, humans have  significantly ($p<0.001$) better accuracy on identification questions with male subjects: $75.7\%$ on male subjects versus $61.4\%$ on female subjects. The same holds true for humans on verification questions: they attain an accuracy of $81.6\%$ on male subjects, versus $76.1\%$ on female subjects ($p < 0.001$). Interestingly, all demographics of survey respondents (when grouped by gender and skin-type) perform substantially better on males than on females for each task. The results of the human-only logistic models confirm human biases towards male subjects in both verification (OR = $1.39$, $p < 0.001$) and identification (OR = $1.97$, $p < 0.001$). Academic models are found, through logistic regression, to exhibit a statistically significant difference in performance between verification questions with male or female subjects (OR = $1.28$, $p = 0.03$). 

\paragraph{Humans and Computers Perform Worse on Darker-Skinned Subjects}

Humans collectively are proportionally $5.2\%$ worse on dark-skinned subjects than light-skinned subjects for verification questions ($80.9\%$ versus $76.7\%$, $p < 0.001$) when we aggregate the Fitzpatrick scale as binary. On identification questions, this proportional difference grew to $11.7\%$ in favor of light-skinned subjects ($72.7\%$ versus $64.2\%$, $p < 0.001$).
This holds even when controlling for the demographics of the respondent: the odds ratio of dark-skinned compared to light-skinned question subjects for verification is  $0.78$ ($p < 0.001$) while for identification it is $0.67$ ($p < 0.001$).  
When we aggregate the Fitzpatrick scale as three groups, I-II, III-VI, and V-VI, verification logistic regression finds statistically significant biases in favor of Fitzpatrick types I-II, over both III-VI and V-VI questions compared (OR = 0.93, $p = 0.023$ for III-VI; OR=0.85, $p<0.001$ for V-VI). For the identification task, even when controlling for respondent demographic, question subjects with Fitzpatrick values I-II have higher correct responses than that of values III-VI and V-VI (OR = 0.92, $p = 0.04$ for III-VI; OR=0.70, $p<0.001$ for V-VI).

The results are more nuanced for machines. When we aggregate the Fitzpatrick scale as just ``light'' and ``dark'', we observe a statistically significant proportional disparity of $1.6\%$ in favor of light-skinned question subjects on the verification task ($p = 0.02$); for identification, we do not find evidence of a skin type bias ($p=0.18$).
When we aggregate the Fitzpatrick scale into three categories, I-II, III-IV, and V-VI, we see a disparity for both tasks between the lightest (I-II) and darkest groups (V-VI) ($p=0.004$ and $p=0.04$ for verification and identification respectively).
Academic model performance  is revealed to be significantly different, even when controlling for gender, between the types I-II and V-VI (OR = 0.78, $p = 0.042$ for identification; OR=0.67, $p=0.005$ for verification). However, I-II and III-VI do not show statistically significant differences for academically-trained models (OR = 1.07, $p = 0.591$ for identification; OR=0.93, $p=0.632$ for verification).

\paragraph{Human Test-Takers Perform Better on Subjects of Similar Demographic}

We hypothesized that humans would be more accurate on questions that contained subjects that looked like them. We find evidence to support this hypothesis in our data. On the verification task, humans perform significantly better on questions where the subjects match their gender identity ($1.2\%$, $p = 0.02$), skin type ($1.0\%$, $p = 0.046$), and gender identity and skin type ($1.6\%$, $p = 0.009$). On the identification task, humans perform significantly better on questions where subjects match their skin type ($4.5\%$, $p < 0.001$) and both their gender identity and skin type ($4.7\%$, $p < 0.001$).

\paragraph{Humans and Machines Exhibit Comparable Levels of Disparity}

To test for whether the levels or disparity described above are comparable between humans and machines, we look at the confidence intervals for the odds ratios of comparable models. For both tasks, recall that we observed a disparity on gender and skin type for humans and machines. For verification, we observe that the magnitude of the gender disparities are similar (OR 95\% confidence intervals for humans are [1.33, 1.46] and for academic models are [1.02, 1.61]). For identification, we observe that the magnitude of the gender disparities are also similar (OR 95\% confidence intervals for humans are [1.84,2.10] and for academic models are [1.43,2.17]). As for the skin type disparity, we see similar overlapping confidence intervals between humans and machines for both skin type as binary (light/dark) and ternary (I-II/III-IV/V-VI). This allows us to conclude that when there is a demographic disparity displayed by both humans and machines, the magnitudes and directions of that disparity are statistically similar.

\paragraph{Commercial Facial Recognition Models Are Very Accurate}
The commercial models have very high accuracy, particularly AWS and Face++ which each scored above 97.3\% accuracy on both verification and identification. As a result, these systems do not have enough incorrect responses to have any statistically significant conclusions. On the other hand, Azure achieves verification accuracy of 93.3\% and identification accuracy of 82.9\%. In this case, we see a bias towards question gender in favor of males (OR = 1.76; $p=0.041$) which is comparable to the bias observed with humans and academic models. 
\section{Discussion}

The study described in this work is the first to compare disparities and bias between humans and machines. We see that the gender and skin type biases of humans are also present in academic models. Interestingly the level of the disparities present in humans are comparable to that of the machines. These human disparities are present even when controlling for the demographics of the participant. We also find that humans perform better when the demographics of the question match their own. This is not altogether surprising as humans generally spend more time with people of their similar demographics and are more practiced at discriminating faces that look like them. 

One key limitation of our study is that we analyze a crowdsourced sample. While it is demographically diverse, it does not represent a sample of expert facial recognizers. Our results should not be extrapolated too far outside the sample of non-expert crowd workers located in the US. Additionally, the results we have for the computer models are limited to those which we included and do not represent how all models work or behave. 

Our findings contribute meaningfully to the ongoing work of understanding the benefits and harms presented by facial recognition technology. Specifically, we see that automated methods outperform non-expert humans across the board.  When bias is detected in a machine, that bias is comparable to those exhibited by non-expert humans. In the future, further work should examine more targeted populations, such as the direct users of facial recognition technology (e.g., forensic examiners or police officers), to understand how their native bias compares to the biases of machines or human-machine teams.

While our dataset was used here for one specific purpose, we hope that our dataset and survey can be used for future evaluations of the accuracy and bias of facial analysis systems.  Furthermore, we hope our dataset curation process helps bring attention to the many pitfalls and weaknesses of academic datasets.

\paragraph{Ethics Statement}
Our human subjects research was conducted in accordance with the rules, policies, and oversight of our institutional review board (IRB) which deemed our survey collection process to be Exempt. As is common practice with public figures, the data collected was done without the consent of those depicted in the images. This work contributes meaningfully by helping us better understand the tendencies of both humans and machines in this socially important area of facial recognition. The work could potentially be used to improve facial recognition outcomes, concretize the inevitability of facial recognition technology even in morally questionable scenarios, or argue against the future development of facial recognition technologies on the basis of ongoing biases we describe. 

\paragraph{Acknowledgements}
Dooley and Dickerson were supported in part by NSF CAREER Award IIS-1846237, NSF D-ISN Award \#2039862, NSF Award CCF-1852352, NIH R01 Award NLM-013039-01, NIST MSE Award \#20126334, DARPA GARD \#HR00112020007, DoD WHS Award \#HQ003420F0035, ARPA-E Award \#4334192 and a Google Faculty Research Award.  Downing, Wei, Shankar, Thymes, Thorkelsdottir, Kurtz-Miott, Mattson, and Obiwumi were supported by NSF Award CCF-1852352 through the University of Maryland’s REU-CAAR: Combinatorics and Algorithms Applied to Real Problems. We thank Bill Gasarch for his standing commitment to building and maintaining a strong REU program at the University of Maryland.

\chapter{Field Evidence in COVID-19 App Attractiveness} 

\label{chpt:CovidAds} 
This work was done in collaboration with Dana Turjeman, John P. Dickerson, and Elissa M. Redmiles. See~\cite{dooley2022field}.

\section{Introduction}
To combat
SARS-CoV-2 -- also known as "coronavirus" -- and its associated illness COVID-19, countries and other entities have worked to develop vaccines and a variety of other mitigation tools. One such tool is contact-tracing technology that serves as the foundation for exposure-notification apps (COVID-19 apps) that can alert users when they have been exposed to coronavirus. %
These apps have been developed and deployed in 77 countries and U.S. states.\footnote{See the Linux Public Health Foundation dashboard (\url{https://landscape.lfph.io/}) for a running list of deployed COVID-19 apps.}

Similar to other pro-social COVID-19 behaviors such as vaccination and mask adoption, greater adoption of COVID-19 apps improves their efficacy. Yet, adoption has been low, with the highest adoption rates per jurisdiction hovering around 30\% and typical adoption rates closer to 10\%.\footnote{There has been little official reporting of COVID-19 app adoption rates outside of the popular press; we refer to \url{https://time.com/5905772/covid-19-contact-tracing-apps/} for these adoption statistics.}

Prior work has sought to understand people's considerations for adopting such apps through self-report and lab-based studies. These works suggest that people's adoption is likely driven by concerns regarding the app's privacy and data collection practices, as well as perceptions of whether the benefits of the apps -- to themselves or to society -- outweigh their privacy and data concerns\cite{simko2020covid19, redmiles_user_2020, li2021makes}.

However, none of this prior work observed how these considerations affected actual adoption of these apps in the wild. Self-report studies on privacy have known flaws due to the ``Privacy Paradox'': people have been shown to state that they would choose a more privacy-oriented product or would not be willing to share information, but will quickly forgo of these protection measures when faced with a decision in real life \cite{athey2017digital, obar2020biggest, langheinrich2018privacy}. Thus, to offer real-world insights into users' adoption of a privacy-sensitive health application in the context of COVID-19, we conduct the first, to our knowledge, field study of COVID-19 app adoption.

We collaborated with the state of Louisiana to conduct a randomized, controlled field experiment on the impact of tailored messaging addressing the attributes found most relevant to app adoption in prior work -- the app's benefits, privacy, and data collection -- on adoption of the state's COVID-19 exposure-notification app, CovidDefense. Specifically, we test the impact of advertisements that contain two types of messaging addressing factors identified in prior work and recommended in policy guidance~\cite{national2020encouraging}: (a) app benefits framed as either a collective- or individual-good and (b) transparency regarding privacy and/or data collection.
We conducted our field experiment on the Google Ads Platform using 14 different ads. Ads were randomly displayed to Louisiana residents and generated 7,010,271 impressions.\footnote{As is typical in digital marketing campaigns, ads may be displayed during Google search more than once to the same user/IP address and thus the number of impressions is larger than the population of Louisiana.}
The outcome measured was whether the user clicked the respective ad; those who clicked were redirected to the Louisiana Department of Public Health app download page (\url{http://coviddefensela.com/}).\footnote{A user may have seen more than one ad because Google does not allow us to control this. However, if the user clicked on an ad, the click was associated with the specific ad on which they clicked.}

Using these data we address four research questions:
\begin{enumerate}[label=\textbf{RQ\arabic*:},leftmargin=*, align=left]
    \item Is messaging that presents the benefit of app installation as a \textit{collective-good} appeal (i.e., with societal benefit) more effective than messaging that appeals to \textit{individual-good}?
    \item Is messaging that makes \textit{privacy transparent} more effective than messaging that does not? And, which privacy transparency statements are most/least effective, those that: (a) broadly reassure people about privacy concerns, or those that specifically focus on enhanced control over data collection -- through a statement emphasizing either (b) general, non-technical privacy control or (c) technically concrete privacy control?
    \item Is messaging that makes \textit{data collection transparent} (i.e., stating clearly what data is being collected) by the app more effective than messaging that does not inform potential users what data will be collected?
    \item How do demographics (age, gender, geography) moderate the adoption of CovidDefense and the experimental effects observed in RQs1-3?
    \end{enumerate}

Collective-good appeals (i.e., pro-social messages that speak to community benefit) are suggested as a best practice by existing policy guidance~\cite{national2020encouraging}. However, the efficacy of such appeals is empirically debated in the context of COVID19~\cite{rabb2021no} on the basis of evidence from self-report data, laboratory experiments, and hybrid self-report tracking ~\cite{korn2020vaccination, munzert2021tracking, seberger2021us} and the impact of these appeals in other privacy-sensitive technology settings has not been well studied.

Existing policy guidance also encourages transparency in advertising promoting pro-social health behaviors, and prior work on people's intent to adopt COVID-19 apps emphasizes the importance of privacy and data collection concerns on people's adoption intent~\cite{redmiles_user_2020,li2021makes,simko2020covid19,zhang_americans_2020}. However, there is little field evidence regarding how individuals respond to data transparency and privacy statements in a privacy-sensitive health technology context. While prior research in the privacy domain (e.g., \citealp{gefen2020privacy, tucker2014social, brandimarte2013misplaced}) has found that increased transparency and sense of control regarding existing privacy and data collection may reduce concerns and increase willingness to share data, it is an open question whether such transparency can be effectively provided through tailored messaging~\cite{schaub2017designing}. Findings from some prior work~\cite{kim2019seeing} suggest that it can; but this and other work suggest that increasing the salience of privacy through transparency at the time of the choice to adopt the app may artificially increase people's concerns about privacy~\cite{schaub2017designing, kim2019seeing}.

The results of our field study show that tailored messaging can effectively influence the pro-social behavior of installing a COVID-19 app. We find that significantly more people click on messages that use collective-good appeals than those that use individual-good appeals (RQ1). Furthermore, in a series of moderation analyses, we find that transparency about privacy (RQ2) and data collection (RQ3) moderate this effect. Specifically, collective-goods appeals are even more effective when paired with a privacy transparency statement, but are less effective when additionally paired with a data transparency statement. Individual-goods appeals exhibit the opposite effects. Such differences suggest that digital privacy and data transparency can be effectively provided through tailored messaging, but we must think carefully about how an application's purpose and framing may impact people's privacy considerations and reasoning. Finally, our results shed light on how priming with an individual-good appeal increases gender and age differences in receptiveness to the ads and to the different privacy controls presented (RQ4).

Our findings offer insight into how users make privacy-benefits trade-offs when making decisions to adopt an app in the wild. We confirm in the field prior self-report results on the importance of individualist vs. collectivist mindset~\cite{seberger2021us}, and expand the existing body of literature to provide insight into the real world impact of the tension between our desire to improve community health by sharing personal data and our individual desire for privacy.

\section{Related Work}
Here, we review the prior work most closely related to our study: on the factors that influence COVID-19 app adoption and on privacy and data transparency statements in the context of digital health.

\subsection{Factors Influencing Intent to Adopt COVID-19 Apps}
When considering whether to use an app for COVID-19 contact tracing, previous research has shown three main considerations: the functionality of the app, concerns regarding privacy, and concerns regarding data collection.

The two main functions of COVID-19 contact tracing apps are to indicate to a user if they have been exposed (an individual-good) and to help the broader community reduce the spread of the virus (a collective-good)~\citep{redmiles_user_2020}. \citet{li2021makes} found that of these dual purposes of the app were more influential in determining intention to install than security or privacy concerns. \citet{williams_public_2021} finds that even the possibility of a collective-good outcome can convince otherwise hesitant users to participate in COVID-19 apps, sometimes begrudgingly. However,
some individuals indicate a reluctance to install a COVID-19 app regardless of how well the app works~\citep{kaptchuk_how_2020, li2021makes, simko2020covid19}.

User privacy is well-documented as a main source of hesitancy for individuals to download and use COVID-19 apps, stemming from both a general privacy concern as as well as specific concerns around contract-tracing apps~\citep{williams_public_2021, ladyzhets_we_2021, kaptchuk_how_2020, zimmermann2021early, li2021makes, geber-typology-based_2021, simko2020covid19, zhang_americans_2020}. However, there is still debate about whether people's stated privacy concerns are actually influencing COVID-19 app adoption once controlling for other factors such as incentives~\citep{frimpong_financial_2020}, institutional trust~\citep{velicia-martin_researching_2021, horvath_citizens_2020, julienne2020behavioural}, political ideology~\citep{lockey2021profiling}, or general perceptions of COVID-19~\citep{walrave_adoption_2020, li2021makes, velicia-martin_researching_2021, frimpong_financial_2020, chan2021privacy}. Additionally, there is indication that privacy concerns can be linked to app functionality. For instance, there is evidence that people's privacy considerations about COVID-19 apps can be moderated by the way in which the app works, specifically the centralization of the contact-tracing mechanism. These two things are linked because some COVID-19 apps are structured in a centralized system, with user data going to a central source to make decision about when to notify potentially exposed users, and others are structured in decentralized systems, with exposure risk managed locally by an app. There is not a consensus in prior work regarding whether users prefer one system or the other; \citet{zhang_americans_2020} find from their conjoint analysis that a decentralized system had higher app adoption whereas other studies find the opposite~\citep{li2021makes,horvath_citizens_2020}. Regardless, the majority of deployed contact-tracing apps are decentralized~\cite{LFPublic32:online}.

Relatedly, people may have concerns specifically related to the data used for contact tracing, whether in a centralized or decentralized scheme. Some COVID-19 contact-tracing apps operate using only proximity data, relying on bluetooth to detect proximity between devices, while other apps rely on GPS location data. Prior work~\cite{simko2020covid19,redmiles_user_2020} finds that users do worry about data collection and finds that users are more comfortable with apps that use proximity vs. location data. These concerns and considerations interplay with users' concerns about their privacy, as some of these concerns focus on the privacy of the data collected by the app, even if it is stored only on their device as is the case for decentralized apps.~\footnote{Note that it is possible for even decentralized apps to have privacy leaks~\cite{kaptchuk_how_2020,raskar2020covid,bengio2021inherent,GooglePr12:online}, and thus user's privacy concerns are not unfounded.}

\paragraph{Prior field work on COVID-19 App Adoption} Due to the emerging nature of the pandemic, there has been little field work studying how people's adoption considerations influence their behavior in the real world. Our work seeks to build on findings from prior self-report work while filling the gap of empirical field evidence.

Most closely related to our work, \citet{munzert2021tracking} tested the effect of presenting collective-good appeals in combination with privacy and functionality-related information in a video intervention.  Subjects in their study participated in an opt-in survey panel in Germany on the COVID-19 app adoption behavior so that their digital behavior could be tracked by the survey panel. Their experiment found a null result, though this might be an artifact of some experimental limitations, as identified in ~\citet{toussaert2021upping}, which  include 1. the nature of the intervention, which involved exposure to a training video during a survey-based study rather than as part of real-world installation behavior and which combined multiple experimental messages, preventing isolation of the impact of the collective-good appeal from the other experimental factors, 2. the sample size, and 3. the opt-in nature of the participant pool. In contrast, our work isolates and focuses specifically on the impact of appeals in tailored messaging, presenting the first, to our knowledge\footnote{Banker and Park conducted a field study on the impact of collective-good appeals on clicks to CDC guidelines at the very beginning of the pandemic~\cite{banker2020evaluating}. However, health information consumption and pro-social health behavior are importantly different constructs.}, direct field evaluation in the general population of the efficacy of collective-good appeals in encouraging pro-social COVID-19 behavior. Specifically, we tested this in tailored advertising messaging used to encourage adoption of an exposure notification app at the time it was released to the population. Importantly, our work does not rely on surveys, online studies or an opt-in sample. Instead, we directly measure the outcome of interest: whether a prospective user clicks to download the app.

\subsection{Privacy and Digital Health}
Outside of a COVID-19 setting, privacy concerns in digital and mobile health applications (often termed mHealth apps)
have an extensive research history. Prior contact-tracing mHealth apps have been studied before in other settings such as tuberculosis~\citep{begun2013contact}, influenza~\citep{ginsberg2009detecting}, and H1N1~\citep{signorini2011use}, though much of the framing of these studies has been from the perspective of the individual benefits they provide~\citep{stowell2018designing}, and an emphasis on privacy in these studies is generally lacking given the nascentness of mHealth at the time of their study.

More recently, privacy more broadly has become a central topic that definitively influences an individual's interest in adopting mHealth applications~\citep{prasad2012understanding,preuveneers2016privacy,fritz2014persuasive,baldauf2020trust}. Early on in these studies, Klasajna explored the privacy perspectives of different data types and found that GPS location data was particularly sensitive. \citet{prasad2012understanding} found that an individual may have differing attitudes towards sharing the same data with different individuals, reporting that individuals who wore fitness trackers were less likely to share data with friends and family than with strangers. Among other things,
Demographics may also play a role in privacy perceptions around mHealth applications~\citep{karampela2019connected, serrano2016willingness}, such as age, education, occupation, and digital prevalence.

Prior work typically finds that
the main predictors of mHealth app adoption are trust, utility, and ease of use~\citep{venkatesh2003user, cocosila2010adoption, zhang2014understanding, nunes2019acceptance, balapour2019mobile, deng2018predicts, sun2013understanding} with  moderating effects from age, gender, location, and education. These findings align well with the literature on adoption intention for mHealth apps specifically designed for COVID-19 as described above.

Focusing specifically on privacy and sensitivity around data collection, \citet{jacobs2015comparing} explores the data sharing preferences of different groups of people by role in a data ecosystem around breast cancer. They find that patients, doctors, and navigators have different comfort levels with data sharing, e.g., patients are hesitant to share data about their emotional state. \citet{warner2018privacy} explore the specific privacy concerns within group of HIV-positive men using a geo-social dating app and find that some users disclose their status to reduce their exposure to stigma while others avoid disclosure to avoid being stigmaized.

More broadly, there is limited prior work on the effect of privacy and sense of control control on the sharing of personal health related data, perhaps because individuals do not have much control over their own health data, and sometimes are not even able to access it themselves. While HIPAA and other health-related privacy policies have been developed to let users exercise informed control over sharing health information, such mechanisms are dated and are not suitable for mHealth\cite{munns2017privacy}. Outside of the health domain, prior work on privacy more broadly has found that when people have greater sense of control over their data, they are likely to be willing to share more data (e.g. \cite{brandimarte2013misplaced}), even if this control is merely an artifact of transparency and not of actual usage of the data. Therefore, in this study, we build upon this prior work specifically in the health domain: we explore the role of messaging related to privacy control, and the transparency of the data being collected, on the likelihood to adopt a COVID-19 app, and how these factors intersect with how the appeal of the app is framed, and the socio-demographics of the adopter.

\section{Methods}\label{sec:experimental_design}
To answer our research questions, we conducted a randomized, controlled field experiment using Google Ads. Here, we review our experimental design, data collection, ethical considerations, analysis approach, and the limitations of our work.

\subsection{Experimental Design}

\begin{figure}
    \centering
    \small
    \includegraphics[width=0.8\textwidth]{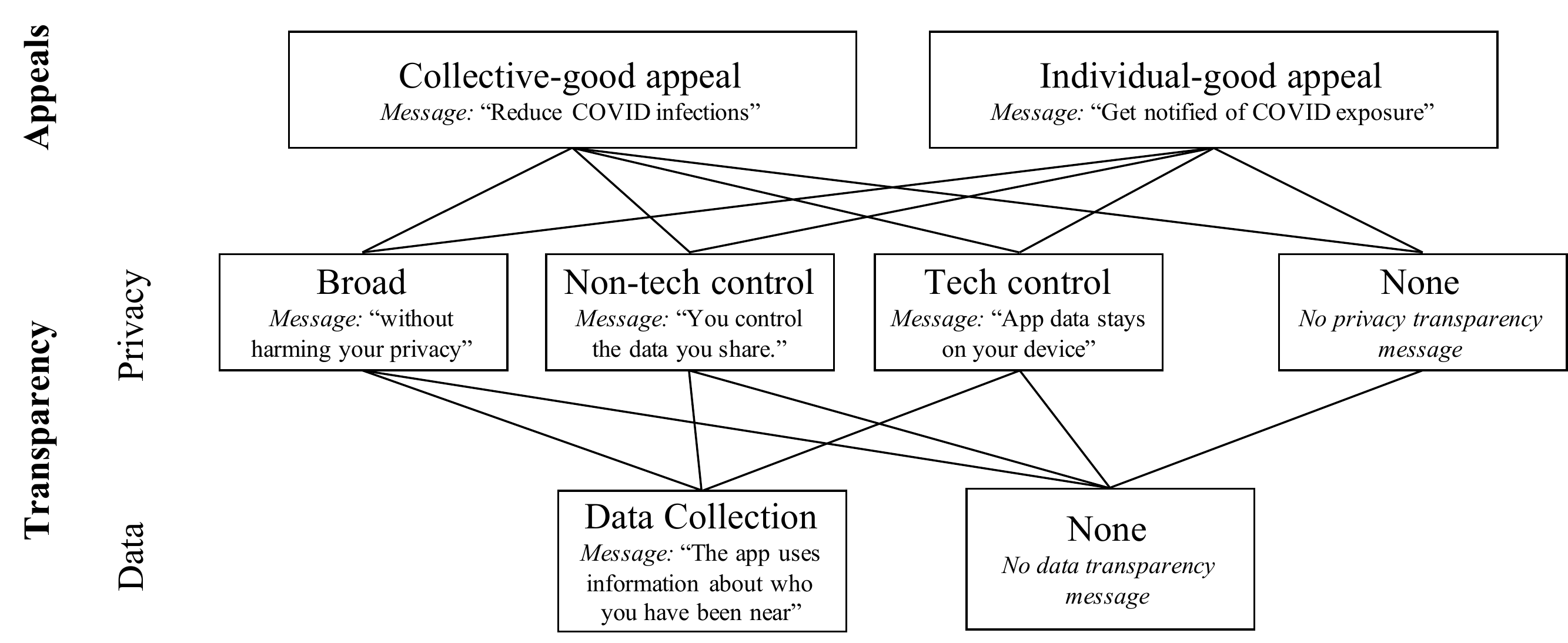}
    \caption{Experimental design for the 14 messages shown in the field study.}
    \label{fig:design}
\end{figure}

\begin{figure}
    \centering
    \small
    \includegraphics[width=0.4\textwidth]{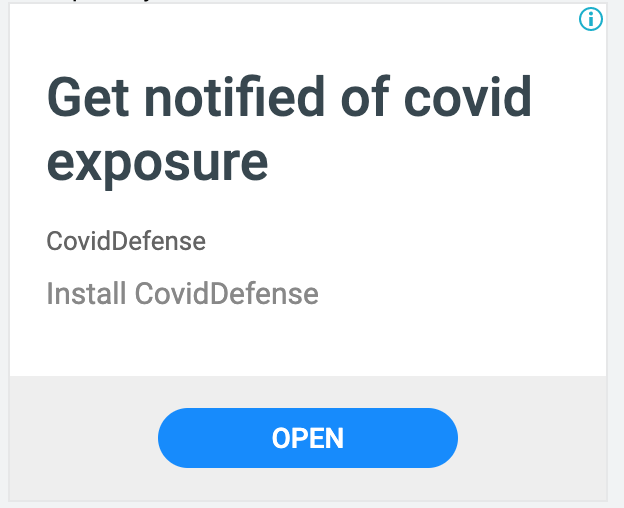}
    \includegraphics[width=0.4\textwidth]{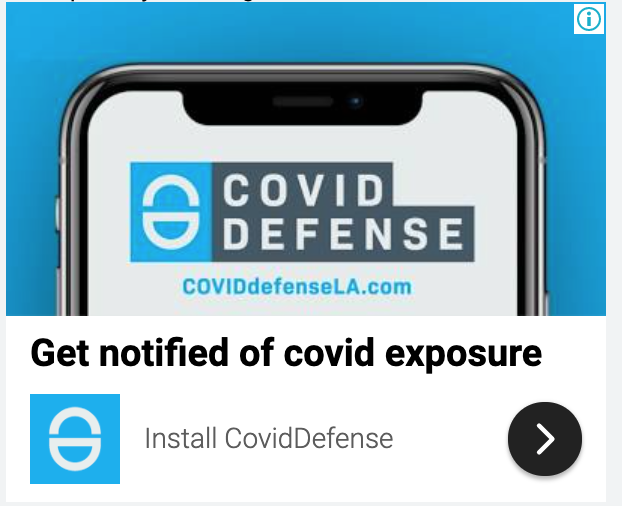}
    \caption{Two examples of how Ad \#1 is displayed on a computer.}
    \label{fig:example_ad}
\end{figure}

Upon the public release of the CovidDefense app, we ran 14 separate Google display ad campaigns from February 1 to 26, 2021. In collaboration with the state of Louisiana, these were the only Google Display ads run for CovidDefense during that time. Each campaign was targeted, via IP address, at people who reside in Louisiana.\footnote{Prior work has validated the accuracy of this state-level targeting~\cite{bandy2021errors}.} All campaigns used the same settings, ad destination, and ad image from the state of Louisiana's CovidDefense marketing materials. The 14 ads varied only in their text data in alignment with the 14 conditions summarized in Figure~\ref{fig:design}. Two examples of how an ad was presented to a user on a computer though Google Ads are depicted in Figure~\ref{fig:example_ad}.

The text of the 14 ads were chosen in the following manner. One of two appeals appeared at the beginning of the ad text: one individual (``Get notified of COVID exposure'') and one collective (``Reduce COVID infections'').\footnote{These phrases were limited to 30 characters by the Google Ads platform. We selected these phrases based on pilot testing in collaboration with the State of Louisiana in which a market research firm conducted an approximately 800 respondent survey to identify the best message phrasings that were most appealing on a variety of criteria. This allowed us to adopt already-successful messages and investigate through this randomized study how the type of appeal, as well as privacy and data transparency messaging, influenced app adoption.} Following the appeal, either one of three privacy transparency statements were made or there was no privacy statement. These three privacy statements either were broadly stated (``...without harming your privacy''), or had a technical (``App data stays on your device.'') or non-technical (``You control the data you share.'') statement. Finally, a privacy statement could also have been paired with a data collection statement (``The app uses information about who you have been near.'').

\subsection{Data Collection}
We observe a total of 7,010,271 impressions. Google Ads does not allow for a user to limit impressions on campaigns, so we manually monitored campaign performance and aimed to stop each campaign at 500,000 $\pm$ 35,000 impressions, with an average of 500,733.6 impressions per campaign.
In total, we observe 28,026 clicks on our 14 campaigns. The average Click Through Rate (CTR -- the outcome of interest -- the proportion between number of clicks and number of impressions) of the 14 campaigns was 0.398\%, with standard deviation of 0.100\%.

Along with the number of clicks and number of impressions, we also observe measures of demographics (age, gender and community density: urban vs. rural). Demographics are provided through Google Ads metadata. Age and gender are inferred by the Google Ads platform through past browsing behavior; the accuracy of these inferences has been validated against gold-standard social scientific probabilistic survey panels and other self-report data sources~\cite{mcdonald2012comparing}. To label participants' community density we map participant counties (called Parishes in Louisiana), which are determined by Google Ads based on IP address, using the Census mapping to community density. $97.8$\% of the impressions (6,858,820) had an associated Parish while $55.9\%$ of the impressions (3,920,232) had both age and gender labels. For privacy reasons, Google Ads separates demographic and location data and thus we cannot analyze age, gender, and community density in one statistical model. As such, we construct separate binomial logistic regression models to analyze the results of our experiment, the first consisting of data for all impressions, the second controlling for age and gender, and the third controlling for community density.

\subsection{Analysis}
Our main analysis examines differences in click through rates (CTRs) for different ads based on their messaging text. For statements about statistical significance, we report $\alpha = 0.05$.

For RQ1, we perform an analysis with a two-sided two proportion $z$-test on the CTRs of collective-good and individual-good ads.

We report statistics as odds ratios for each regression. The odds ratio compares the ratio of odds for a baseline event to the odds for the contrasting event. Additionally, since the geographic and demographic data are a subset of our entire dataset (97.8\% and 55.9\% respectively), it is natural to be concerned that analyzing these subsets may lead to different conclusions. However, we analyzed the results of RQ2 through RQ4 with just the dataset subsets and found that the results are robust to such modeling specifications, defined by overlapping confidence intervals on the odds ratios for the same regressions with the different subsets.

\paragraph{Data Archival}
Data and analysis scripts for this experiment are available in an anonymized form at \dataset.

\subsection{Ethics}
Our study was approved by the our institution's ethics review board and exempted from review by the Louisiana Department of Health IRB. All data collection occurred within the the publicly available Google Ads platform, and we only had limited access to information about users in the manner in which Google provided them. The data were presented to us in an aggregate manner which preserved the users privacy in accordance with Google Ads privacy policy\footnote{\url{https://safety.google/privacy/ads-and-data/}}. This included geographic location of an impression with resolution to the parish level. Google also provided some demographic information which are either user-supplied or inferred by browsing habits\footnote{\url{https://support.google.com/adsense/answer/140378}}.

Further, we were careful to consider the effects that the messages of our ads might have on the user. To that end, we were aware that some combinations of messaging could act as a deterrent to future adoption of these technologies. Since our research takes the view that adoption of contract tracing apps should be encouraged, we chose to not include combinations of messages that might work against that goal -- that is why in our experimental design we only showed a data transparency statement in combination with a privacy transparency statement.

\subsection{Limitations}
Our results rely on a single study, from a single state (Louisiana) and with demographic data that rely on Google's ability to accurately classify gender, region, and age.
The largest limitation of our study design is that we only capture clicks on ads instead of full downloads and app use. Our study was designed this way with a privacy focus, however future studies could examine app adoption and use in conjunction with pro-social messaging questions.
Additionally, while we chose the language used in our ad messages carefully, other forms of appeals or transparency statements could have been used.
We encourage future studies to further investigate the impact of collective- vs. individual-goods appeals, as well as privacy and data transparency, on encouraging pro-social digital health behavior.

\section{Results}\label{sec:results_and_discussion}
Our experimental factors -- appeal, privacy transparency, and data transparency -- all significantly relate to CTR.

\begin{figure*}
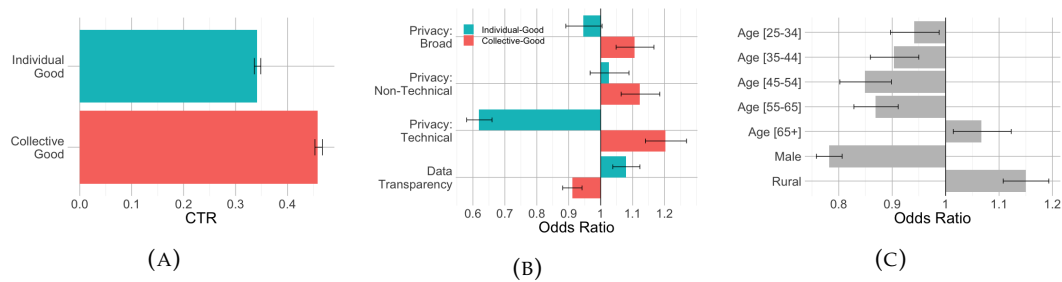

     \centering
     \begin{subfigure}[!t]{0.32\linewidth}
         \centering
        \includegraphics[width=\linewidth]{Chapters/CovidAds/figures/2a.pdf}
        \caption{}
        \label{fig:overall ctr}

     \end{subfigure}
     \hfill
     \begin{subfigure}[!t]{0.32\linewidth}
         \centering
        \includegraphics[width=\linewidth]{Chapters/CovidAds/figures/2b.pdf}
        \caption{}
        \label{fig:interactions}
     \end{subfigure}
     \hfill
     \begin{subfigure}[!t]{0.32\linewidth}
         \centering
        \includegraphics[width=\linewidth]{Chapters/CovidAds/figures/2c.pdf}
        \caption{}
        \label{fig:demo}
     \end{subfigure}
     \label{fig:second}
     \caption{Main results from regression models. Error bars of 95\% confidence are given.
     (a) Overall CTRs for individual- and collective-good appeals.
     (b) Odds ratio for each transparency statement modeled twice for ads presented with each of the two appeals. The individual-good appeal (in red) is above the collective-good appeal (in blue). Regression tables are Appendix Tables~\ref{tbl:model_all_cg} and \ref{tbl:model_all_ig}.
     (c) Overall odds ratio for each demographic and geographic variable. Regression table is Appendix Table~\ref{tbl:model_demo_geo}.}
\end{figure*}

\subsection{RQ1: Collective-good appeals are superior}
Advertisements mentioning individual-good perform significantly worse ($\chi^2=598.54$; df=1, $p<0.001$) than those that use collective-good appeals (CTR of 0.341\% vs. 0.458\%) in ads containing only an appeal, see Figure \ref{fig:overall ctr}. This is still the case even when controlling for other experimental factors (O.R.=0.745), and the interactions between them (O.R.=0.880; $p<$ 0.001), demographics (O.R.=0.747; $p<0.001$), and community density (O.R.=0.744; $p<0.001$).

\subsection{RQ2 and RQ3: Effect of transparency statements depends on appeal}
The type of appeal (collective or individual) moderates the impact of the transparency statements. We conclude this by observing significant interactions between the appeal and the transparency statements in logistic regression models. Subsequently, performing a regression on each appeal individually, we report the odds ratios and errors for the transparency statements for each appeal in Figure \ref{fig:interactions}.

Messages with a collective-good appeal have a higher CTR when they have additional privacy transparency statements -- all three of the statements have O.R.s that range from 1.106 to 1.203, all with $p<0.001$ -- but result in lower CTR responses when paired with a data transparency statement, i.e., when ads explicitly mentions what data is being collected (O.R. = 0.911, $p<0.001$). On the other hand, messages with an individual-good appeal have a lower CTR when paired with a technical privacy transparency statement (O.R. = 0.619, $p<0.001$). However, the broad and non-technical control privacy transparency statements cannot be deemed to affect the CTR of the same individual-good appeal ($p=0.070$ and $p=0.396$, respectively). Moreover, contrary to the negative effect of data transparency on CTR in the collective-good appeal condition, when data collection is made transparent in the individual-good appeal condition, we observe a higher CTR than in messages without such data transparency (O.R.=1.08, $p<0.001$). We observe that, in a single regression model containing interactions between the transparency statements and the appeals, a data transparency statement reduces the difference in CTR between messages with collective- vs. individual-good appeals, while inclusion of a privacy statement increases the difference in CTR between messages with the two different appeals.

\begin{figure*}
    \centering
    \includegraphics[width=.7\linewidth]{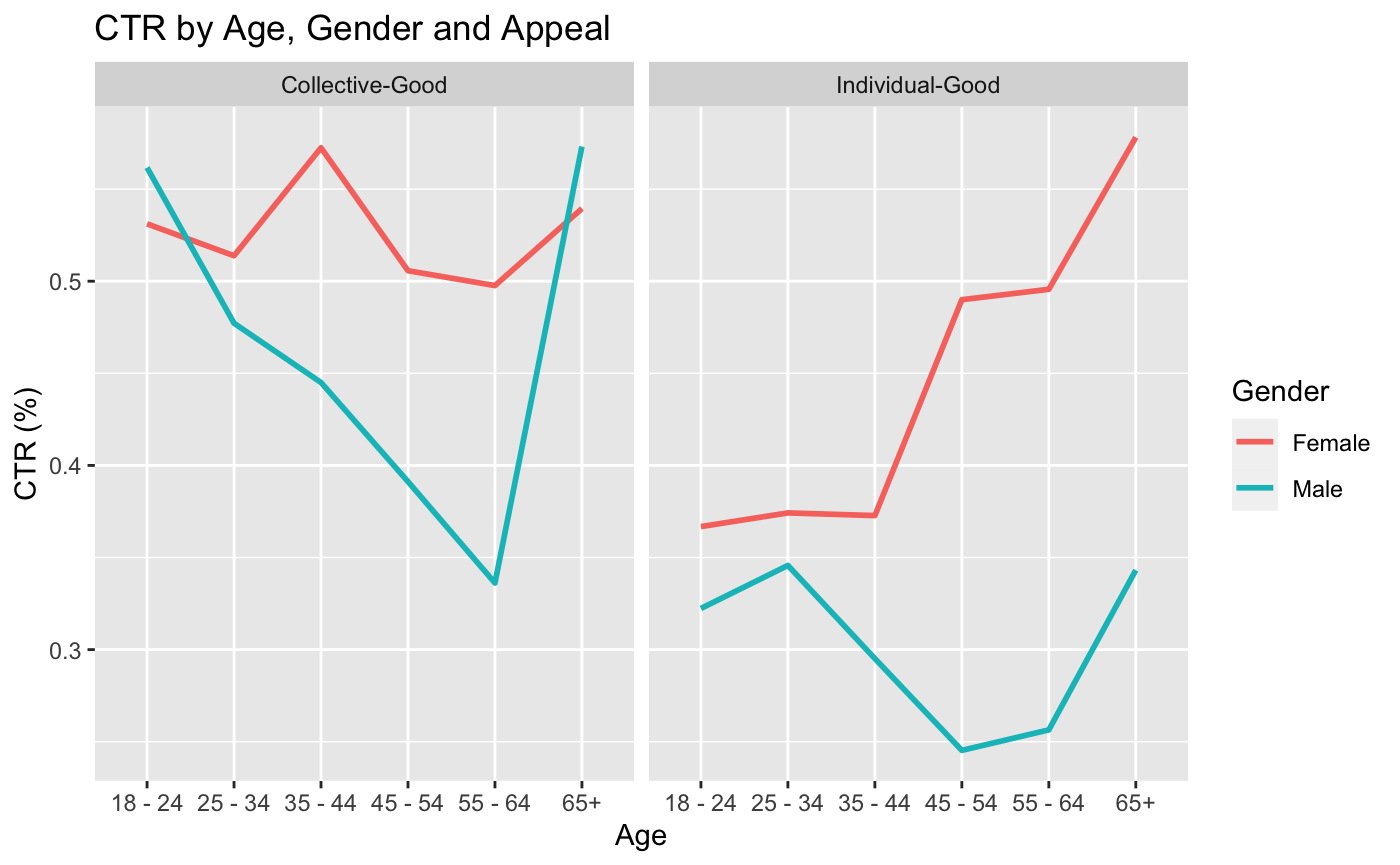}
    \caption{Click through rates (CTR) for each age group are reported with each curve representing an appeal and a gender. Collective-good appeals to females have no difference across the ages, whereas males see a significant drop in CTR in the middle ages. Individual-good appeals for females are non-decreasing across ages and generally flat for males. Error bars are 95\% confidence Clopper-Pearson intervals.}
    \label{fig:appeal}
\end{figure*}

\begin{table}[!htbp] \centering
  \caption{Modeling the statement differences by Appeal and Gender}
  \label{tbl:appeal_gender}
\begin{tabular}{@{\extracolsep{5pt}}lcccc}
\\[-1.8ex]\hline
\hline \\[-1.8ex]
 & \multicolumn{4}{c}{\textit{Dependent variable:}} \\
\cline{2-5}
\\[-1.8ex] & \multicolumn{4}{c}{Clicks} \\
\\[-1.8ex] & \multicolumn{2}{c}{Collective-Good} & \multicolumn{2}{c}{Individual-Good} \\
 & Male & Female & Male & Female\\
\\[-1.8ex] & (1) & (2) & (3) & (4)\\
\hline \\[-1.8ex]
 Privacy.Broad & {\bf 1.160} & 0.993 & {\bf 0.796} & {\bf 1.198} \\
  & (1.049, 1.283) & (0.906, 1.087) & (0.707, 0.896) & (1.076, 1.335) \\
  & p = 0.004$^{**}$ & p = 0.877 & p = 0.0002$^{**}$ & p = 0.002$^{**}$ \\
  & & & & \\
 NonTech.Control & {\bf 1.110} & 1.083 & 1.018 & {\bf 1.333} \\
  & (1.009, 1.222) & (0.986, 1.188) & (0.899, 1.152) & (1.185, 1.498) \\
  & p = 0.032$^{*}$ & p = 0.095 & p = 0.784 & p = 0.00001$^{**}$ \\
  & & & & \\
 Technical.Control & {\bf 1.236} & 0.980 & {\bf 0.460} & {\bf 0.827} \\
  & (1.118, 1.366) & (0.897, 1.071) & (0.402, 0.526) & (0.738, 0.927) \\
  & p = 0.00004$^{**}$ & p = 0.660 & p < 0.001$^{**}$ & p = 0.002$^{**}$ \\
  & & & & \\
 Data.Transparency & {\bf 0.849} & 0.961 & {\bf 1.275} & 1.058 \\
  & (0.795, 0.907) & (0.909, 1.015) & (1.178, 1.379) & (0.987, 1.134) \\
  & p = 0.00001$^{**}$ & p = 0.157 & p < 0.001$^{**}$ & p = 0.113 \\
  & & & & \\
 Constant & {\bf 0.004} & {\bf 0.005} & {\bf 0.003} & {\bf 0.004} \\
  & (0.004, 0.005) & (0.005, 0.006) & (0.003, 0.004) & (0.004, 0.004) \\
  & p < 0.001$^{**}$ & p < 0.001$^{**}$ & p < 0.001$^{**}$ & p < 0.001$^{**}$ \\
  & & & & \\
\hline \\[-1.8ex]
Observations & 916,470 & 1,111,417 & 1,026,261 & 866,084 \\
Log Likelihood & $-$27,378.620 & $-$36,413.650 & $-$20,733.730 & $-$24,634.080 \\
Akaike Inf. Crit. & 54,767.230 & 72,837.310 & 41,477.460 & 49,278.150 \\
\hline
\hline \\[-1.8ex]
\textit{Note:}  & \multicolumn{4}{r}{$^{*}$p$<$0.05; $^{**}$p$<$0.01} \\
\end{tabular}
\end{table}

\subsection{RQ4: Demographic and geographic influences}
Next, we consider demographic differences in responses to CovidDefense advertisements. Thus far it has been debated, on the basis of self-report evidence~\cite{white2020men,barber2021covid,cassino2020masks}, whether men are less likely to adopt pro-social behaviors such as mask wearing. In this work we offer field evidence of a gender difference in receptiveness to COVID-19 pro-social messaging and behavior: men are significantly less likely to click on ads for CovidDefense (O.R. men = $0.794$; $p < 0.001$). We find that this effect varies in size based on the appeal shown in the ad. Both men and women consistently prefer collective-goods ads and men are less likely to click on both collective- and individual-goods ads. However, the gap in CTR between men and women is significantly larger for individual-goods ads: males click 23\% less often than females when shown individual-good ads compared to 10\% less for collective-good ads (with $p<0.001$ for both). 

Overall, we find that users between 34-64 are significantly less likely to click on the advertisements than those between 18 and 24 (O.R.s range from 0.874 - 0.951 with $p<0.05$; see Figure \ref{fig:demo}). On the other hand, those over 65 -- who are also at high risk for developing and dying from COVID-19~\cite{centers2020older}-- are significantly more likely to click than those 18-24 (O.R. = $1.13$; $p < 0.001$). However, these effects are moderated both by the appeal of the ad shown and the gender of the ad viewer.
Specifically, CTRs do not vary among different ages of women shown a collective-good ad (O.R.s range from 0.937-1.08 with $p>0.1$), likely because of the strength of this appeal for women, whom prior research shows are more relational -- focused on the collective good -- than men~\cite{kashima1995culture}. On the other hand, when shown an individual-good appeal, women 45 to 65+ are significantly more likely to click than younger women (each age group above 45 sees an increase in CTR over the previous; O.R.s range from 1.34-1.58 with $p<0.001$). There is no difference between CTRs for ages 18-44 (O.R.s = 1.02 with $p>0.7$). We hypothesize that, aligned with \cite{fan2020heterogeneous}, older women are cognizant of their higher COVID-19 risk and thus are willing to click even on an ad with an appeal that is less preferable and does not align with their broad tendency toward relationally-guided behavior.

Among men who were shown a collective-good appeal, young men (18-24 years old) and older men (65+ years old) are equally likely to click when shown a collective-good ad (O.R.=1.02 with $p=0.728$), and are significantly more likely to click than middle aged men (O.R.s range from 0.597-0.849 with $p<0.001$). Considering individual-good ads, we observe a similar pattern, with no significant differences in the likelihood of clicking among men aged 18-44 and those over 65 (O.R.s range from 0.915-1.07 with $p>0.1$), while men aged 45-64 are significantly less likely to click on the same ads (O.R.s range from 0.760-0.785 with $p<0.001$).  We find that men, especially those who are middle-aged and when presented with individual-good appeal, are far less likely than women to click to install CovidDefense. 
Prior literature finds that men have lower perceptions of their COVID-19 risk~\cite{fan2020heterogeneous}. We hypothesize that the large gap in CTR between men and women, which is especially pronounced when presented with an individual-good appeal, is driven from the gender-based risk tolerance differences documented in the literature. When presented with an individual-good appeal that primes the viewer to especially focus on their own risk, the gender differences are even more pronounced.

Beyond the moderating effects of age and gender on the appeal used to advertise CovidDefense, we also find gender differences in the effect of our privacy and data transparency statements (see Table~\ref{tbl:appeal_gender}). While the effects shown in Figure~\ref{fig:interactions} are consistent across ages, we find that the overall effects of the privacy and data transparency statements are primarily driven by men's response to these statements. When facing collective-good appeal, men are more likely to click if presented with privacy controls of any sort, but are less likely to click if there is an explicit data transparency statement. Women, on the other hand, show different responses to the transparency statements. Specifically, when combined with a collective-good appeal, both privacy and data transparency statements have no impact on women's likelihood to click.

Further, while both men and women are less likely to click on individual-good ads that include a technical privacy control, this reduced likelihood to click is larger among men (CI for O.R. for men: (0.402, 0.526); CI for O.R. for women: (0.738, 0.927)). In addition, while men also respond negatively, or have no significant response, to the non-technical privacy transparency statements, women exhibit higher CTRs when these statements are paired with the individual-good appeal. Additionally, women are  unaffected by the inclusion of a data-privacy transparency statement compared to men who are more likely to click ads with an individual-good appeal when a data transparency statement is included.

Finally, we examine the impact of geography on willingness to click on ads for CovidDefense. In contrast to the existing body of literature around urban-rural differences in COVID-19 behavior, which finds that rural residents are less concerned about COVID-19 and less likely to adopt pro-social COVID-19 behaviors~\cite{callaghan2021rural,chen2020differences,huang2021urban,haischer2020wearing}, we find that Louisiana residents in rural communities are significantly more likely to click on any of the proposed ads (O.R. = $1.15$; $p < 0.001$).
This finding is robust across both appeals and all transparency statements with the appeal preference also being robust between geographies; we observe that rural residents also find that collective-good statements as preferable to individual-good appeals.

\section{Discussion}

\paragraph{The role of transparency in advertising privacy-sensitive public-health technology} The results of our work offer implications for thinking about transparent messaging in the promotion of privacy-sensitive health technologies. Given that the U.S. is a highly individualistic country, it is perhaps surprising that collective-good appeals were more effective than individual-good appeals in encouraging people in Louisiana to click to adopt the CovidDefense app. Prior work finds that Louisiana is the most collectivist U.S. state~\cite{allik2004individualism}. Collectivists tend to engage in pro-social behavior that benefits the in-group, rather than pro-social behavior that they perceive as benefiting themselves individually~\cite{baldassarri2020diversity, kemmelmeier2006individualism}. This effect may explain why residents of Louisiana respond best to societally-oriented benefits. On the other hand, it may simply be the case that people understand that the primary benefit of a COVID-19 exposure notification app is indeed collective and that people respond best to messages that are honest and transparent about the true benefit of the app.

Our findings for transparency regarding privacy and data collection are, however, more nuanced. Transparency about individual data collection improves the efficacy of messages that are already individually focused -- those with individual-good appeals -- while the same transparency statement applied in a collectivist setting appears to conflict with people's sense of collective purpose. Relatedly, transparency regarding privacy and how individual data is protected in a collective setting may reduce concerns about personal privacy risk in a communal context, which prior work finds may be especially elevated, while the same transparency may be ineffective or even detrimental when placed in the context of individualistic privacy-benefit trade-offs~\cite{trepte2017cross}.

\paragraph{Gender effects}
Complicating our transparency findings, we observe gender effects in response to both the appeals and the privacy/data transparency statements.
Overall, we find that men, especially those who are middle-aged and when presented with individual-good appeal, are far less likely than women to click to install CovidDefense.
Prior literature finds that men have lower perceptions of their COVID-19 risk~\cite{fan2020heterogeneous}. We hypothesize that the large gap in CTR between men and women, which is especially pronounced when presented with an individual-good appeal, is driven from the gender-based risk tolerance differences documented in the literature. When presented with an individual-good appeal that primes the viewer to especially focus on their own risk, the gender differences are becoming more pronounced.
Differences between men and women also exist for the privacy and data transparency statements. The overall transparency effects described above are primarily driven by men's response to these statements. Women are unaffected by the inclusion of privacy and data-transparency statements in the more-effective collective-good ads. We hypothesize that this is the case because women are more relational -- focused on the collective good -- than men~\cite{kashima1995culture} and thus, similar to the lack of age effect observed for women when presented with collective-good appeals, we hypothesize that there is such strong alignment between women's tendency toward relational choices and the collective-good appeal that other factors (e.g., age-specific risk perceptions, transparency statements) loose their significant effect

When presented with an individual-good appeal, both men and women are less likely to click when a technical privacy transparency statement is included; the size of this effect is significantly larger for men. Women are \textit{more} likely to click when a non-technical privacy statement is paired with an individual-good appeal. Finally, men are more likely to click -- and women are less likely -- when a data transparency statement is added. Taken together, these findings -- that women are affected positively by less technical statements of privacy while men are positively affected by technical and data related privacy statements -- align with prior work finding that men and women focus on different privacy controls: men have been found to focus more on technical privacy controls, while women are more focused on privacy sentiment and non-technical controls~\cite{habib2018user,kuo2007assessing,mathur2018characterizing,redmiles2018net}.

\paragraph{Implications for Privacy Research}
Our results shed light on the role of data transparency and privacy controls. With our results and similar tailored studies, privacy research and privacy messages can be made more accurate, in order to allow customers to make choices that aligns with their perceived privacy risk and desired controls. We would like to further emphasize that in either option, the COVID-19 app was privacy preserving; we chose messages that were accurate, albeit redundant, given that the privacy was assured by the mere design of the app. With the rise of privacy preserving technologies and mobile health technologies, accurate communication of the privacy measures and controls may allow an increase in the adoption of the service or product. For example, a menstrual cycle tracking app, mostly targeted at women, will benefit from avoiding mentioning technical controls, if they want to appeal to the average woman.

\section{Conclusion}
In a large-scale randomized field study, we find that residents of Louisiana are more likely to click on ads for exposure notification apps if the ad included a collective-good appeal. This effect was moderated (especially for men) by transparency regarding the individual data being collected and privacy protections offered for that data, likely due to the conflict between the sense of collective purpose and the associated cost for individual privacy. Moreover, we find gender and age differences in the likelihood to click the ads, fitting with past literature on gender differences and varying risks of COVID-19 across ages. These differences included lower probability to click on ads with technical privacy controls for women and higher likelihood for older people to click on ads due to the higher risk for COVID-19 associated with older people. We also find that the gender differences were made larger when the ads were individually-focused (with individual-good appeal), suggesting that the priming for individualism is enhancing gender differences. These findings may aid companies and policy makers when promoting digital tools to improve public health, especially those tools that have implications for privacy.

\include{Chapters/CtrlShift}

\bibliographystyle{plainnat}
\bibliography{references}

\begin{thebibliography}{317}
\providecommand{\natexlab}[1]{#1}
\providecommand{\url}[1]{\texttt{#1}}
\expandafter\ifx\csname urlstyle\endcsname\relax
  \providecommand{\doi}[1]{doi: #1}\else
  \providecommand{\doi}{doi: \begingroup \urlstyle{rm}\Url}\fi

\bibitem[AWS()]{AWSGuidelines}
Guidelines on face attributes.
\newblock
  \url{https://docs.aws.amazon.com/rekognition/latest/dg/guidance-face-attributes.html}.
\newblock Accessed: 2021-08-29.

\bibitem[LFP()]{LFPublic32:online}
{Linux Foundation Public Health Landscape}.
\newblock \url{https://landscape.lfph.io/}.
\newblock (Accessed on 08/19/2021).

\bibitem[Acquisti et~al.(2015)Acquisti, Brandimarte, and
  Loewenstein]{acquisti2015privacy}
Alessandro Acquisti, Laura Brandimarte, and George Loewenstein.
\newblock Privacy and human behavior in the age of information.
\newblock \emph{Science}, 347\penalty0 (6221):\penalty0 509--514, 2015.

\bibitem[Adel et~al.(2019)Adel, Valera, Ghahramani, and Weller]{tameem2019one}
Tameem Adel, Isabel Valera, Zoubin Ghahramani, and Adrian Weller.
\newblock One-network adversarial fairness.
\newblock In \emph{AAAI Conference on Artificial Intelligence (AAAI)}, pages
  2412--2420, 2019.

\bibitem[Adeli et~al.(2021)Adeli, Zhao, Pfefferbaum, Sullivan, Fei-Fei,
  Niebles, and Pohl]{adeli2021representation}
Ehsan Adeli, Qingyu Zhao, Adolf Pfefferbaum, Edith~V Sullivan, Li~Fei-Fei,
  Juan~Carlos Niebles, and Kilian~M Pohl.
\newblock Representation learning with statistical independence to mitigate
  bias.
\newblock In \emph{Proceedings of the IEEE/CVF Winter Conference on
  Applications of Computer Vision}, pages 2513--2523, 2021.

\bibitem[Agarwal et~al.(2014)Agarwal, Hsu, Kale, Langford, Li, and
  Schapire]{Agarwal14:Taming}
Alekh Agarwal, Daniel Hsu, Satyen Kale, John Langford, Lihong Li, and Robert
  Schapire.
\newblock Taming the monster: A fast and simple algorithm for contextual
  bandits.
\newblock In \emph{International Conference on Machine Learning (ICML)}, pages
  1638--1646, 2014.

\bibitem[Agarwal et~al.(2018)Agarwal, Beygelzimer, Dudik, Langford, and
  Wallach]{agarwal2018reductions}
Alekh Agarwal, Alina Beygelzimer, Miroslav Dudik, John Langford, and Hanna
  Wallach.
\newblock A reductions approach to fair classification.
\newblock In \emph{International Conference on Machine Learning}, pages 60--69,
  2018.

\bibitem[Ahmed(2019)]{ahmed2019bridging}
Alex~A Ahmed.
\newblock Bridging social critique and design: Building a health informatics
  tool for transgender voice.
\newblock In \emph{Extended Abstracts of the 2019 CHI Conference on Human
  Factors in Computing Systems}, pages 1--4, 2019.

\bibitem[Allik and Realo(2004)]{allik2004individualism}
J{\"u}ri Allik and Anu Realo.
\newblock Individualism-collectivism and social capital.
\newblock \emph{Journal of cross-cultural psychology}, 35\penalty0
  (1):\penalty0 29--49, 2004.

\bibitem[Alonso et~al.(2018)Alonso, de~la Torre-D{\'{i}}ez, Hamrioui,
  L{\'{o}}pez-Coronado, Barreno, Nozaleda, and Franco]{Alonso2018}
Susel~G{\'{o}}ngora Alonso, Isabel de~la Torre-D{\'{i}}ez, Sofiane Hamrioui,
  Miguel L{\'{o}}pez-Coronado, Diego~Calvo Barreno, Lola~Mor{\'{o}}n Nozaleda,
  and Manuel Franco.
\newblock {Data Mining Algorithms and Techniques in Mental Health: A Systematic
  Review}, sep 2018.
\newblock ISSN 1573689X.

\bibitem[{American Psychological
  Association}(2020)]{AmericanPsychologicalAssociation}
{American Psychological Association}.
\newblock {Ethical Principles of Psychologists and Code of Conduct}, 2020.
\newblock URL \url{apa.org/ethics/code/}.

\bibitem[Andalibi and Buss(2020)]{andalibi2020human}
Nazanin Andalibi and Justin Buss.
\newblock The human in emotion recognition on social media: Attitudes,
  outcomes, risks.
\newblock In \emph{Proceedings of the 2020 CHI Conference on Human Factors in
  Computing Systems}, pages 1--16, 2020.

\bibitem[Ashkan et~al.(2015)Ashkan, Kveton, Berkovsky, and
  Wen]{ashkan2015optimal}
Azin Ashkan, Branislav Kveton, Shlomo Berkovsky, and Zheng Wen.
\newblock Optimal greedy diversity for recommendation.
\newblock In \emph{Twenty-Fourth International Joint Conference on Artificial
  Intelligence}, 2015.

\bibitem[Athey et~al.(2017)Athey, Catalini, and Tucker]{athey2017digital}
Susan Athey, Christian Catalini, and Catherine Tucker.
\newblock The digital privacy paradox: Small money, small costs, small talk.
\newblock Technical report, National Bureau of Economic Research, 2017.

\bibitem[Audibert et~al.(2010)Audibert, Bubeck, and Munos]{audibert2010best}
Jean-Yves Audibert, S{\'e}bastien Bubeck, and R{\'e}mi Munos.
\newblock Best arm identification in multi-armed bandits.
\newblock In \emph{Conference on Learning Theory (COLT)}, 2010.

\bibitem[Auer et~al.(2002)Auer, Cesa-Bianchi, and Fischer]{Auer02:Finite-time}
Peter Auer, Nicolo Cesa-Bianchi, and Paul Fischer.
\newblock Finite-time analysis of the multiarmed bandit problem.
\newblock \emph{Machine Learning}, 47\penalty0 (2-3):\penalty0 235--256, 2002.

\bibitem[Bahdanau et~al.(2014)Bahdanau, Cho, and Bengio]{bahdanau2014neural}
Dzmitry Bahdanau, Kyunghyun Cho, and Yoshua Bengio.
\newblock Neural machine translation by jointly learning to align and
  translate.
\newblock \emph{arXiv preprint arXiv:1409.0473}, 2014.

\bibitem[Balakrishnan et~al.(2019)Balakrishnan, Bouneffouf, Mattei, and
  Rossi]{Balakrishnan19:Incorporating}
Avinash Balakrishnan, Djallel Bouneffouf, Nicholas Mattei, and Francesca Rossi.
\newblock Incorporating behavioral constraints in online ai systems.
\newblock In \emph{Conference on Artificial Intelligence (AAAI)}, volume~33,
  pages 3--11, 2019.

\bibitem[Balapour et~al.(2019)Balapour, Reychav, Sabherwal, and
  Azuri]{balapour2019mobile}
Ali Balapour, Iris Reychav, Rajiv Sabherwal, and Joseph Azuri.
\newblock Mobile technology identity and self-efficacy: Implications for the
  adoption of clinically supported mobile health apps.
\newblock \emph{International Journal of Information Management}, 49:\penalty0
  58--68, 2019.

\bibitem[Baldassarri and Abascal(2020)]{baldassarri2020diversity}
Delia Baldassarri and Maria Abascal.
\newblock Diversity and prosocial behavior.
\newblock \emph{Science}, 369\penalty0 (6508):\penalty0 1183--1187, 2020.

\bibitem[Baldauf et~al.(2020)Baldauf, Fr{\"o}ehlich, and
  Endl]{baldauf2020trust}
Matthias Baldauf, Peter Fr{\"o}ehlich, and Rainer Endl.
\newblock Trust me, i?ma doctor--user perceptions of ai-driven apps for mobile
  health diagnosis.
\newblock In \emph{19th International Conference on Mobile and Ubiquitous
  Multimedia}, pages 167--178, 2020.

\bibitem[Bandy and Hecht(2021)]{bandy2021errors}
Jack Bandy and Brent Hecht.
\newblock Errors in geotargeted display advertising: Good news for local
  journalism?
\newblock \emph{Proceedings of the ACM on Human-Computer Interaction},
  5\penalty0 (CSCW), 2021.

\bibitem[Banker and Park(2020)]{banker2020evaluating}
Sachin Banker and Joowon Park.
\newblock Evaluating prosocial covid-19 messaging frames: Evidence from a field
  study on facebook.
\newblock \emph{Judgment and Decision Making}, 15\penalty0 (6):\penalty0
  1037--1043, 2020.

\bibitem[Barber and Kim(2021)]{barber2021covid}
Sarah~J Barber and Hyunji Kim.
\newblock Covid-19 worries and behavior changes in older and younger men and
  women.
\newblock \emph{The Journals of Gerontology: Series B}, 76\penalty0
  (2):\penalty0 e17--e23, 2021.

\bibitem[Barocas and Selbst(2016)]{barocas2016big}
Solon Barocas and Andrew~D Selbst.
\newblock Big data's disparate impact.
\newblock \emph{California Law Review}, 104:\penalty0 671, 2016.

\bibitem[Barocas et~al.(2019)Barocas, Hardt, and Narayanan]{fairmlbook}
Solon Barocas, Moritz Hardt, and Arvind Narayanan.
\newblock \emph{Fairness and Machine Learning}.
\newblock fairmlbook.org, 2019.
\newblock \url{http://www.fairmlbook.org}.

\bibitem[Begun et~al.(2013)Begun, Newall, Marks, and Wood]{begun2013contact}
Matt Begun, Anthony~T Newall, Guy~B Marks, and James~G Wood.
\newblock Contact tracing of tuberculosis: a systematic review of transmission
  modelling studies.
\newblock \emph{PLoS One}, 8\penalty0 (9):\penalty0 e72470, 2013.

\bibitem[Bengio et~al.(2021)Bengio, Ippolito, Janda, Jarvie, Prud'homme,
  Rousseau, Sharma, and Yu]{bengio2021inherent}
Yoshua Bengio, Daphne Ippolito, Richard Janda, Max Jarvie, Benjamin Prud'homme,
  Jean-Fran{\c{c}}ois Rousseau, Abhinav Sharma, and Yun~William Yu.
\newblock Inherent privacy limitations of decentralized contact tracing apps.
\newblock \emph{Journal of the American Medical Informatics Association},
  28\penalty0 (1):\penalty0 193--195, 2021.

\bibitem[Benthall and Haynes(2019)]{benthall2019racial}
Sebastian Benthall and Bruce~D Haynes.
\newblock Racial categories in machine learning.
\newblock In \emph{Proceedings of the conference on fairness, accountability,
  and transparency}, pages 289--298, 2019.

\bibitem[Benton et~al.(2017)Benton, Coppersmith, and Dredze]{Benton2017ethical}
Adrian Benton, Glen Coppersmith, and Mark Dredze.
\newblock {Ethical Research Protocols for Social Media Health Research}.
\newblock In \emph{Proceedings of the First ACL Workshop on Ethics in Natural
  Language Processing}, 2017.
\newblock \doi{10.18653/v1/w17-1612}.

\bibitem[Best-Rowden and Jain(2017)]{best2017longitudinal}
Lacey Best-Rowden and Anil~K Jain.
\newblock Longitudinal study of automatic face recognition.
\newblock \emph{IEEE transactions on pattern analysis and machine
  intelligence}, 40\penalty0 (1):\penalty0 148--162, 2017.

\bibitem[Beutel et~al.(2017)Beutel, Chen, Zhao, and Chi]{beutel2017data}
Alex Beutel, Jilin Chen, Zhe Zhao, and Ed~H Chi.
\newblock Data decisions and theoretical implications when adversarially
  learning fair representations.
\newblock \emph{arXiv preprint arXiv:1707.00075}, 2017.

\bibitem[Beygelzimer et~al.(2011)Beygelzimer, Langford, Li, Reyzin, and
  Schapire]{Beygelzimer11:Contextual}
Alina Beygelzimer, John Langford, Lihong Li, Lev Reyzin, and Robert Schapire.
\newblock Contextual bandit algorithms with supervised learning guarantees.
\newblock In \emph{International Conference on Artificial Intelligence and
  Statistics (AISTATS)}, pages 19--26, 2011.

\bibitem[Biega et~al.(2018)Biega, Gummadi, and Weikum]{beiga2018equity}
Asia~J. Biega, Krishna~P. Gummadi, and Gerhard Weikum.
\newblock Equity of attention: Amortizing individual fairness in rankings.
\newblock In \emph{ACM Conference on Research and Development in Information
  Retrieval (SIGIR)}, page 405?414, 2018.

\bibitem[Binns(2017)]{Binns17:Fairness}
Reuben Binns.
\newblock Fairness in machine learning: Lessons from political philosophy.
\newblock \emph{Proceedings of Machine Learning Research}, 81:\penalty0 1--11,
  2017.

\bibitem[Bloom et~al.(2011)Bloom, Cafiero, Jan{\'{e}}-Llopis, Abrahams-Gessel,
  Bloom, Fathima, Feigl, Gaziano, Hamandi, Mowafi, Pandya, Prettner, Rosenberg,
  Seligman, Stein, and Weinstein]{Bloom}
DE~Bloom, ET~Cafiero, E~Jan{\'{e}}-Llopis, S~Abrahams-Gessel, LR~Bloom,
  S~Fathima, AB~Feigl, T~Gaziano, A~Hamandi, M~Mowafi, A~Pandya, K~Prettner,
  L~Rosenberg, B~Seligman, AZ~Stein, and C.~Weinstein.
\newblock {The global economic burden of noncommunicable diseases.}
\newblock \emph{Geneva: World Economic Forum}, 2011.

\bibitem[Bogen and Rieke(2018)]{BoRi18a}
M.~Bogen and A.~Rieke.
\newblock Help wanted: {A}n examination of hiring algorithms, equity, and bias.
\newblock Technical report, Upturn, 2018.

\bibitem[Boser et~al.(1992)Boser, Guyon, and Vapnik]{boser1992svm}
Bernhard~E. Boser, Isabelle~M. Guyon, and Vladimir~N. Vapnik.
\newblock A training algorithm for optimal margin classifiers.
\newblock In \emph{Conference on Learning Theory (COLT)}, page 144?152, 1992.

\bibitem[Brandimarte et~al.(2013)Brandimarte, Acquisti, and
  Loewenstein]{brandimarte2013misplaced}
Laura Brandimarte, Alessandro Acquisti, and George Loewenstein.
\newblock Misplaced confidences: Privacy and the control paradox.
\newblock \emph{Social psychological and personality science}, 4\penalty0
  (3):\penalty0 340--347, 2013.

\bibitem[Bresler et~al.(2014)Bresler, Chen, and Shah]{Bresler14:Latent}
Guy Bresler, George~H Chen, and Devavrat Shah.
\newblock A latent source model for online collaborative filtering.
\newblock In \emph{Advances in Neural Information Processing Systems
  (NeurIPS)}, pages 3347--3355, 2014.

\bibitem[Bubeck et~al.(2011)Bubeck, Munos, and Stoltz]{bubeck2011pure}
S{\'e}bastien Bubeck, R{\'e}mi Munos, and Gilles Stoltz.
\newblock Pure exploration in finitely-armed and continuous-armed bandits.
\newblock \emph{Theoretical Computer Science}, 412\penalty0 (19):\penalty0
  1832--1852, 2011.

\bibitem[Bubeck et~al.(2012)Bubeck, Cesa-Bianchi, et~al.]{bubeck2012regret}
S{\'e}bastien Bubeck, Nicolo Cesa-Bianchi, et~al.
\newblock Regret analysis of stochastic and nonstochastic multi-armed bandit
  problems.
\newblock \emph{Foundations and Trends{\textregistered} in Machine Learning},
  5\penalty0 (1):\penalty0 1--122, 2012.

\bibitem[Bubeck et~al.(2013)Bubeck, Wang, and Viswanathan]{bubeck2013multiple}
S{\'e}ebastian Bubeck, Tengyao Wang, and Nitin Viswanathan.
\newblock Multiple identifications in multi-armed bandits.
\newblock In \emph{International Conference on Machine Learning}, pages
  258--265, 2013.

\bibitem[Buolamwini and Gebru(2018{\natexlab{a}})]{buolamwini2018gender}
Joy Buolamwini and Timnit Gebru.
\newblock Gender shades: Intersectional accuracy disparities in commercial
  gender classification.
\newblock In \emph{ACM Conference on Fairness, Accountability, and Transparency
  (FAccT)}, pages 77--91, 2018{\natexlab{a}}.

\bibitem[Buolamwini and Gebru(2018{\natexlab{b}})]{buolamwini2018gendershades}
Joy Buolamwini and Timnit Gebru.
\newblock Gender shades: Intersectional accuracy disparities in commercial
  gender classification.
\newblock In \emph{Proceedings of the 1st Conference on Fairness,
  Accountability and Transparency}, volume~81, pages 77--91,
  2018{\natexlab{b}}.
\newblock URL \url{http://proceedings.mlr.press/v81/buolamwini18a.html}.

\bibitem[Callaghan et~al.(2021)Callaghan, Lueck, Trujillo, and
  Ferdinand]{callaghan2021rural}
Timothy Callaghan, Jennifer~A Lueck, Kristin~Lunz Trujillo, and Alva~O
  Ferdinand.
\newblock Rural and urban differences in covid-19 prevention behaviors.
\newblock \emph{The Journal of Rural Health}, 2021.

\bibitem[Calmon et~al.(2017)Calmon, Wei, Vinzamuri, Natesan~Ramamurthy, and
  Varshney]{calmon2017optimized}
Flavio Calmon, Dennis Wei, Bhanukiran Vinzamuri, Karthikeyan
  Natesan~Ramamurthy, and Kush~R Varshney.
\newblock Optimized pre-processing for discrimination prevention.
\newblock In \emph{Advances in Neural Information Processing Systems 30},
  NIPS'17, pages 3992--4001. 2017.
\newblock URL
  \url{http://papers.nips.cc/paper/6988-optimized-pre-processing-for-discrimination-prevention.pdf}.

\bibitem[Carlini and Wagner(2017)]{carlini2017towards}
Nicholas Carlini and David Wagner.
\newblock Towards evaluating the robustness of neural networks.
\newblock In \emph{2017 IEEE Symposium on Security and Privacy (S\&P)}, pages
  39--57, 2017.

\bibitem[Cassino and Besen-Cassino(2020)]{cassino2020masks}
Dan Cassino and Yasemin Besen-Cassino.
\newblock Of masks and men? gender, sex, and protective measures during
  covid-19.
\newblock \emph{Politics \& Gender}, 16\penalty0 (4):\penalty0 1052--1062,
  2020.

\bibitem[{Centers for Disease Control and Prevention}
  et~al.(2020)]{centers2020older}
{Centers for Disease Control and Prevention} et~al.
\newblock Older adults at greater risk of requiring hospitalization or dying if
  diagnosed with {COVID-19}, 2020.

\bibitem[Chan and Saqib(2021)]{chan2021privacy}
Eugene~Y Chan and Najam~U Saqib.
\newblock Privacy concerns can explain unwillingness to download and use
  contact tracing apps when covid-19 concerns are high.
\newblock \emph{Computers in Human Behavior}, 119:\penalty0 106718, 2021.

\bibitem[Chen(2014)]{chen2014laborers}
Adrian Chen.
\newblock The laborers who keep dick pics and beheadings out of your facebook
  feed.
\newblock \emph{Wired}, 23:\penalty0 14, 2014.

\bibitem[Chen et~al.(2018)Chen, Liu, Gao, and Han]{chen2018mobilefacenets}
Sheng Chen, Yang Liu, Xiang Gao, and Zhen Han.
\newblock Mobilefacenets: Efficient cnns for accurate real-time face
  verification on mobile devices.
\newblock In \emph{Chinese Conference on Biometric Recognition}, pages
  428--438. Springer, 2018.

\bibitem[Chen et~al.(2014)Chen, Lin, King, Lyu, and Chen]{Chen14:Combinatorial}
Shouyuan Chen, Tian Lin, Irwin King, Michael~R Lyu, and Wei Chen.
\newblock Combinatorial pure exploration of multi-armed bandits.
\newblock In \emph{Advances in Neural Information Processing Systems}, pages
  379--387, 2014.

\bibitem[Chen and Chen(2020)]{chen2020differences}
Xuewei Chen and Hongliang Chen.
\newblock Differences in preventive behaviors of covid-19 between urban and
  rural residents: lessons learned from a cross-sectional study in china.
\newblock \emph{International journal of environmental research and public
  health}, 17\penalty0 (12):\penalty0 4437, 2020.

\bibitem[Cherepanova et~al.(2021)Cherepanova, Goldblum, Foley, Duan, Dickerson,
  Taylor, and Goldstein]{cherepanova2021lowkey}
Valeriia Cherepanova, Micah Goldblum, Harrison Foley, Shiyuan Duan, John~P.
  Dickerson, Gavin Taylor, and Tom Goldstein.
\newblock Lowkey: leveraging adversarial attacks to protect social media users
  from facial recognition.
\newblock In \emph{International Conference on Learning Representations
  (ICLR)}, 2021.

\bibitem[Cherepanova et~al.(2022)Cherepanova, Reich, Dooley, Souri, Goldblum,
  and Goldstein]{cherepanova2022deep}
Valeriia Cherepanova, Steven Reich, Samuel Dooley, Hossein Souri, Micah
  Goldblum, and Tom Goldstein.
\newblock A deep dive into dataset imbalance and bias in face identification.
\newblock \emph{arXiv preprint arXiv:2203.08235}, 2022.

\bibitem[Chong et~al.(2021)Chong, Maudet, Harima, and
  Igarashi]{chong2021exploring}
Toby Chong, Nolwenn Maudet, Katsuki Harima, and Takeo Igarashi.
\newblock Exploring a makeup support system for transgender passing based on
  automatic gender recognition.
\newblock In \emph{Proceedings of the 2021 CHI Conference on Human Factors in
  Computing Systems}, pages 1--13, 2021.

\bibitem[Chouldechova(2017)]{Chouldechova17:Fair}
Alexandra Chouldechova.
\newblock Fair prediction with disparate impact: A study of bias in recidivism
  prediction instruments.
\newblock \emph{Big Data}, 5\penalty0 (2):\penalty0 153--163, 2017.

\bibitem[Chouldechova and Roth(2018)]{chouldechova2018frontiers}
Alexandra Chouldechova and Aaron Roth.
\newblock The frontiers of fairness in machine learning.
\newblock \emph{arXiv preprint arXiv:1810.08810}, 2018.

\bibitem[Chow and Chang(2008)]{Chow08:Adaptive}
Shein-Chung Chow and Mark Chang.
\newblock Adaptive design methods in clinical trials--a review.
\newblock \emph{Orphanet Journal of Rare Diseases}, 3\penalty0 (1):\penalty0
  11, 2008.

\bibitem[Cocosila and Archer(2010)]{cocosila2010adoption}
Mihail Cocosila and Norm Archer.
\newblock Adoption of mobile ict for health promotion: an empirical
  investigation.
\newblock \emph{Electronic Markets}, 20\penalty0 (3):\penalty0 241--250, 2010.

\bibitem[Cohen et~al.(2019)Cohen, Rosenfeld, and Kolter]{cohen2019smoothing}
Jeremy~M. Cohen, Elan Rosenfeld, and J.~Zico Kolter.
\newblock Certified adversarial robustness via randomized smoothing.
\newblock In \emph{International Conference on Machine Learning (ICML)}, 2019.

\bibitem[Conway and O'Connor(2016)]{Conway2016}
Mike Conway and Daniel O'Connor.
\newblock {Social media, big data, and mental health: Current advances and
  ethical implications}, jun 2016.
\newblock ISSN 2352250X.

\bibitem[Cook et~al.(2019)Cook, Howard, Sirotin, Tipton, and
  Vemury]{cook2019demographic}
Cynthia~M Cook, John~J Howard, Yevgeniy~B Sirotin, Jerry~L Tipton, and Arun~R
  Vemury.
\newblock Demographic effects in facial recognition and their dependence on
  image acquisition: An evaluation of eleven commercial systems.
\newblock \emph{IEEE Transactions on Biometrics, Behavior, and Identity
  Science}, 1\penalty0 (1):\penalty0 32--41, 2019.

\bibitem[Coppersmith et~al.(2014{\natexlab{a}})Coppersmith, Dredze, and
  Harman]{Coppersmith2014}
Glen Coppersmith, Mark Dredze, and Craig Harman.
\newblock {Quantifying Mental Health Signals in Twitter}.
\newblock In \emph{Proceedings of the Workshop on Computational Linguistics and
  Clinical Psychology: From Linguistic Signal to Clinical Reality}, pages
  51--60, Stroudsburg, PA, USA, 2014{\natexlab{a}}. Association for
  Computational Linguistics.

\bibitem[Coppersmith et~al.(2014{\natexlab{b}})Coppersmith, Dredze, and
  Harman]{Coppersmith2015g}
Glen Coppersmith, Mark Dredze, and Craig Harman.
\newblock Quantifying mental health signals in twitter.
\newblock In \emph{Proceedings of the Workshop on Computational Linguistics and
  Clinical Psychology: From Linguistic Signal to Clinical Reality}, pages
  51--60, Baltimore, Maryland, USA, June 2014{\natexlab{b}}. Association for
  Computational Linguistics.
\newblock \doi{10.3115/v1/W14-3207}.
\newblock URL \url{https://www.aclweb.org/anthology/W14-3207}.

\bibitem[Coppersmith et~al.(2018)Coppersmith, Leary, Crutchley, and
  Fine]{Coppersmith2018}
Glen Coppersmith, Ryan Leary, Patrick Crutchley, and Alex Fine.
\newblock {Natural Language Processing of Social Media as Screening for Suicide
  Risk}.
\newblock \emph{Biomedical Informatics Insights}, 10:\penalty0 117822261879286,
  jan 2018.
\newblock ISSN 1178-2226.

\bibitem[Corbitt-Hall et~al.(2019)Corbitt-Hall, Gauthier, and
  Troop-Gordon]{Corbitt-Hall2019}
Darcy~J. Corbitt-Hall, Jami~M. Gauthier, and Wendy Troop-Gordon.
\newblock {Suicidality Disclosed Online: Using a Simulated Facebook Task to
  Identify Predictors of Support Giving to Friends at Risk of Self-harm}.
\newblock \emph{Suicide and Life-Threatening Behavior}, 2019.
\newblock ISSN 1943278X.

\bibitem[Corcoran et~al.(2019)Corcoran, Benavides, and Cecchi]{Corcoran2019}
Cheryl~M. Corcoran, Caridad Benavides, and Guillermo Cecchi.
\newblock {Natural language processing: Opportunities and challenges for
  patients, providers, and hospital systems}.
\newblock \emph{Psychiatric Annals}, 49\penalty0 (5):\penalty0 202--208, may
  2019.
\newblock ISSN 00485713.

\bibitem[Cramer et~al.(2019)Cramer, Vaughan, Holstein, Wallach,
  Garcia-Gathright, III, Dudok, and Reddy]{cramer2019challenges}
Henriette Cramer, Jenn~Wortman Vaughan, Ken Holstein, Hanna Wallach, Jean
  Garcia-Gathright, Hal~Daume III, Miroslav Dudok, and Sravana Reddy.
\newblock Challenges of incorporating algorithmic fairness into industry
  practice.
\newblock \emph{FAT* Tutorial}, 2019.
\newblock URL
  \url{https://drive.google.com/file/d/1rUQkVS0NzSH3IEqZDsczSxBbhYHbjamN/view}.

\bibitem[Crawford and Paglen(2019)]{crawford2019excavating}
Kate Crawford and Trevor Paglen.
\newblock Excavating ai: The politics of images in machine learning training
  sets.
\newblock 2019.
\newblock URL \url{https://www.excavating.ai/}.

\bibitem[Daniels(2016)]{Daniels16:Resource}
Norman Daniels.
\newblock Resource allocation and priority setting.
\newblock In \emph{Public Health Ethics: Cases Spanning the Globe}, pages
  61--94. Springer, 2016.

\bibitem[De~Choudhury(2014)]{de2014opportunities}
Munmun De~Choudhury.
\newblock Opportunities of social media in health and well-being.
\newblock \emph{XRDS: Crossroads, The ACM Magazine for Students}, 21\penalty0
  (2):\penalty0 23--27, 2014.

\bibitem[{De Choudhury}(2015)]{DeChoudhury2015}
Munmun {De Choudhury}.
\newblock {Opportunities of social media in health and well-being}.
\newblock \emph{XRDS: Crossroads, The ACM Magazine for Students}, 21\penalty0
  (2):\penalty0 23--27, dec 2015.
\newblock ISSN 15284972.

\bibitem[{De Choudhury} et~al.(2016){De Choudhury}, Kiciman, Dredze,
  Coppersmith, and Kumar]{DeChoudhury2016}
Munmun {De Choudhury}, Emre Kiciman, Mark Dredze, Glen Coppersmith, and Mrinal
  Kumar.
\newblock {Discovering Shifts to Suicidal Ideation from Mental Health Content
  in Social Media}.
\newblock In \emph{Proceedings of the 2016 CHI Conference on Human Factors in
  Computing Systems - CHI '16}, 2016.
\newblock ISBN 9781450333627.
\newblock \doi{10.1145/2858036.2858207}.

\bibitem[Deng et~al.(2019)Deng, Guo, Xue, and Zafeiriou]{deng2019arcface}
Jiankang Deng, Jia Guo, Niannan Xue, and Stefanos Zafeiriou.
\newblock Arcface: Additive angular margin loss for deep face recognition.
\newblock In \emph{Proceedings of the IEEE/CVF Conference on Computer Vision
  and Pattern Recognition}, pages 4690--4699, 2019.

\bibitem[Deng et~al.(2018)Deng, Hong, Ren, Zhang, and Xiang]{deng2018predicts}
Zhaohua Deng, Ziying Hong, Cong Ren, Wei Zhang, and Fei Xiang.
\newblock What predicts patients? adoption intention toward mhealth services in
  china: empirical study.
\newblock \emph{JMIR mHealth and uHealth}, 6\penalty0 (8):\penalty0 e172, 2018.

\bibitem[Derringer(2019)]{derringer2019surveillance}
William Derringer.
\newblock A surveillance net blankets china?s cities, giving police vast
  powers.
\newblock \emph{The New York Times}, Dec. 17 2019.
\newblock URL
  \url{https://www.nytimes.com/2019/12/17/technology/china-surveillance.html}.

\bibitem[Diana et~al.(2020)Diana, Gill, Kearns, Kenthapadi, and
  Roth]{diana2020convergent}
Emily Diana, Wesley Gill, Michael Kearns, Krishnaram Kenthapadi, and Aaron
  Roth.
\newblock Convergent algorithms for (relaxed) minimax fairness.
\newblock \emph{arXiv preprint arXiv:2011.03108}, 2020.

\bibitem[Ding et~al.(2013)Ding, Qin, Zhang, and Liu]{ding2013multi}
Wenkui Ding, Tao Qin, Xu-Dong Zhang, and Tie-Yan Liu.
\newblock Multi-armed bandit with budget constraint and variable costs.
\newblock In \emph{Twenty-Seventh AAAI Conference on Artificial Intelligence},
  2013.

\bibitem[Donini et~al.(2018)Donini, Oneto, Ben-David, Shawe-Taylor, and
  Pontil]{donini2018empirical}
Michele Donini, Luca Oneto, Shai Ben-David, John Shawe-Taylor, and Massimiliano
  Pontil.
\newblock Empirical risk minimization under fairness constraints.
\newblock In \emph{Proceedings of the 32nd International Conference on Neural
  Information Processing Systems}, NIPS'18, page 2796?2806, 2018.

\bibitem[Dooley and Dickerson(2020)]{dooley2020affiliate}
Samuel Dooley and John~P Dickerson.
\newblock The affiliate matching problem: On labor markets where firms are also
  interested in the placement of previous workers.
\newblock \emph{arXiv preprint arXiv:2009.11867}, 2020.

\bibitem[Dooley et~al.()Dooley, Wei, Goldstein, and
  Dickerson]{dooleyrobustness}
Samuel Dooley, George~Zhihong Wei, Tom Goldstein, and John~P Dickerson.
\newblock Robustness disparities in face detection.
\newblock In \emph{Thirty-sixth Conference on Neural Information Processing
  Systems Datasets and Benchmarks Track}.

\bibitem[Dooley et~al.(2021{\natexlab{a}})Dooley, Downing, Wei, Shankar,
  Thymes, Thorkelsdottir, Kurtz-Miott, Mattson, Obiwumi, Cherepanova,
  et~al.]{dooley2021comparing}
Samuel Dooley, Ryan Downing, George Wei, Nathan Shankar, Bradon Thymes, Gudrun
  Thorkelsdottir, Tiye Kurtz-Miott, Rachel Mattson, Olufemi Obiwumi, Valeriia
  Cherepanova, et~al.
\newblock Comparing human and machine bias in face recognition.
\newblock \emph{arXiv preprint arXiv:2110.08396}, 2021{\natexlab{a}}.

\bibitem[Dooley et~al.(2021{\natexlab{b}})Dooley, Goldstein, and
  Dickerson]{dooley2021robustness}
Samuel Dooley, Tom Goldstein, and John~P Dickerson.
\newblock Robustness disparities in commercial face detection.
\newblock \emph{arXiv preprint arXiv:2108.12508}, 2021{\natexlab{b}}.

\bibitem[Dooley et~al.(2022{\natexlab{a}})Dooley, Turjeman, Dickerson, and
  Redmiles]{dooley2022field}
Samuel Dooley, Dana Turjeman, John~P Dickerson, and Elissa~M Redmiles.
\newblock Field evidence of the effects of privacy, data transparency, and
  pro-social appeals on covid-19 app attractiveness.
\newblock In \emph{CHI Conference on Human Factors in Computing Systems}, pages
  1--21, 2022{\natexlab{a}}.

\bibitem[Dooley et~al.(2022{\natexlab{b}})Dooley, Wei, Goldstein, and
  Dickerson]{dooley2022commercial}
Samuel Dooley, George~Z Wei, Tom Goldstein, and John~P Dickerson.
\newblock Are commercial face detection models as biased as academic models?
\newblock \emph{arXiv preprint arXiv:2201.10047}, 2022{\natexlab{b}}.

\bibitem[Dwork and Ilvento(2018)]{dwork2018fairness}
Cynthia Dwork and Christina Ilvento.
\newblock Fairness under composition.
\newblock In \emph{Innovations in Theoretical Computer Science Conference
  (ITCS)}, 2018.

\bibitem[Dwork et~al.(2012)Dwork, Hardt, Pitassi, Reingold, and
  Zemel]{dwork2012fairness}
Cynthia Dwork, Moritz Hardt, Toniann Pitassi, Omer Reingold, and Richard Zemel.
\newblock Fairness through awareness.
\newblock In \emph{Innovations in Theoretical Computer Science Conference
  (ITCS)}, 2012.

\bibitem[Edwards and Storkey(2016)]{edwards2016censoring}
Harrison Edwards and Amos~J. Storkey.
\newblock Censoring representations with an adversary.
\newblock In \emph{4th International Conference on Learning Representations,
  {ICLR} 2016, San Juan, Puerto Rico, May 2-4, 2016, Conference Track
  Proceedings}, 2016.
\newblock URL \url{http://arxiv.org/abs/1511.05897}.

\bibitem[Eidinger et~al.(2014{\natexlab{a}})Eidinger, Enbar, and
  Hassner]{adience}
Eran Eidinger, Roee Enbar, and Tal Hassner.
\newblock Age and gender estimation of unfiltered faces.
\newblock \emph{IEEE Transactions on Information Forensics and Security},
  9\penalty0 (12):\penalty0 2170--2179, 2014{\natexlab{a}}.

\bibitem[Eidinger et~al.(2014{\natexlab{b}})Eidinger, Enbar, and
  Hassner]{eidinger2014age}
Eran Eidinger, Roee Enbar, and Tal Hassner.
\newblock Age and gender estimation of unfiltered faces.
\newblock \emph{IEEE Transactions on Information Forensics and Security},
  9\penalty0 (12):\penalty0 2170--2179, 2014{\natexlab{b}}.

\bibitem[El~Khiyari and Wechsler(2016)]{el2016face}
Hachim El~Khiyari and Harry Wechsler.
\newblock Face verification subject to varying (age, ethnicity, and gender)
  demographics using deep learning.
\newblock \emph{Journal of Biometrics and Biostatistics}, 7\penalty0
  (323):\penalty0 11, 2016.

\bibitem[Ernala et~al.(2019)Ernala, Birnbaum, Candan, Rizvi, Sterling, Kane,
  and {De Choudhury}]{Ernala2019b}
Sindhu~Kiranmai Ernala, Michael~L. Birnbaum, Kristin~A. Candan, Asra~F. Rizvi,
  William~A. Sterling, John~M. Kane, and Munmun {De Choudhury}.
\newblock {Methodological gaps in predicting mental health states from social
  media: Triangulating diagnostic signals}.
\newblock In \emph{Conference on Human Factors in Computing Systems -
  Proceedings}. Association for Computing Machinery, may 2019.
\newblock ISBN 9781450359702.

\bibitem[Eubanks(2018)]{eubanks2018automating}
Virginia Eubanks.
\newblock \emph{Automating inequality: How high-tech tools profile, police, and
  punish the poor}.
\newblock St. Martin's Press, 2018.

\bibitem[Fan et~al.(2020)Fan, Orhun, and Turjeman]{fan2020heterogeneous}
Ying Fan, A~Ye{\c{s}}im Orhun, and Dana Turjeman.
\newblock Heterogeneous actions, beliefs, constraints and risk tolerance during
  the covid-19 pandemic.
\newblock Technical report, National Bureau of Economic Research, 2020.

\bibitem[Feldman et~al.(2015{\natexlab{a}})Feldman, Friedler, Moeller,
  Scheidegger, and Venkatasubramanian]{Feldman15:Certifying}
Michael Feldman, Sorelle~A Friedler, John Moeller, Carlos Scheidegger, and
  Suresh Venkatasubramanian.
\newblock Certifying and removing disparate impact.
\newblock In \emph{International Conference on Knowledge Discovery and Data
  Mining (KDD)}, pages 259--268, 2015{\natexlab{a}}.

\bibitem[Feldman et~al.(2015{\natexlab{b}})Feldman, Friedler, Moeller,
  Scheidegger, and Venkatasubramanian]{Feldman2015Certifying}
Michael Feldman, Sorelle~A Friedler, John Moeller, Carlos Scheidegger, and
  Suresh Venkatasubramanian.
\newblock Certifying and removing disparate impact.
\newblock In \emph{Knowledge Discovery and Data Mining}, pages 259--268,
  2015{\natexlab{b}}.

\bibitem[Ferreira et~al.(2018)Ferreira, Simchi-Levi, and
  Wang]{Ferreira18:Online}
Kris~Johnson Ferreira, David Simchi-Levi, and He~Wang.
\newblock Online network revenue management using thompson sampling.
\newblock \emph{Operations Research}, 66\penalty0 (6):\penalty0 1586--1602,
  2018.

\bibitem[Fitzpatrick(1988)]{fitzpatrick1988validity}
Thomas~B Fitzpatrick.
\newblock The validity and practicality of sun-reactive skin types i through
  vi.
\newblock \emph{Archives of dermatology}, 124\penalty0 (6):\penalty0 869--871,
  1988.

\bibitem[Ford et~al.(2019)Ford, Curlewis, Wongkoblap, and
  Curcin]{ford2019public}
Elizabeth Ford, Keegan Curlewis, Akkapon Wongkoblap, and Vasa Curcin.
\newblock Public opinions on using social media content to identify users with
  depression and target mental health care advertising: mixed methods survey.
\newblock \emph{JMIR Mental Health}, 6\penalty0 (11):\penalty0 e12942, 2019.

\bibitem[Franklin et~al.(2017)Franklin, Ribeiro, Fox, Bentley, Kleiman, Huang,
  Musacchio, Jaroszewski, Chang, and Nock]{Franklin2017}
Joseph~C. Franklin, Jessica~D. Ribeiro, Kathryn~R. Fox, Kate~H. Bentley,
  Evan~M. Kleiman, Xieyining Huang, Katherine~M. Musacchio, Adam~C.
  Jaroszewski, Bernard~P. Chang, and Matthew~K. Nock.
\newblock {Risk factors for suicidal thoughts and behaviors: A meta-analysis of
  50 years of research.}
\newblock \emph{Psychological Bulletin}, 143\penalty0 (2):\penalty0 187--232,
  2017.
\newblock ISSN 1939-1455.

\bibitem[Frimpong and Helleringer(2020)]{frimpong_financial_2020}
Jemima~A. Frimpong and Stephane Helleringer.
\newblock Financial {Incentives} for {Downloading} {COVID}?19 {Digital}
  {Contact} {Tracing} {Apps}.
\newblock preprint, SocArXiv, June 2020.
\newblock URL \url{https://osf.io/9vp7x}.

\bibitem[Fritz et~al.(2014)Fritz, Huang, Murphy, and
  Zimmermann]{fritz2014persuasive}
Thomas Fritz, Elaine~M Huang, Gail~C Murphy, and Thomas Zimmermann.
\newblock Persuasive technology in the real world: a study of long-term use of
  activity sensing devices for fitness.
\newblock In \emph{Proceedings of the SIGCHI conference on human factors in
  computing systems}, pages 487--496, 2014.

\bibitem[Galhotra et~al.(2017)Galhotra, Brun, and Meliou]{sainyam2017fairness}
Sainyam Galhotra, Yuriy Brun, and Alexandra Meliou.
\newblock Fairness testing: Testing software for discrimination.
\newblock In \emph{Proceedings of the 2017 11th Joint Meeting on Foundations of
  Software Engineering}, ESEC/FSE 2017, page 498?510, New York, NY, USA, 2017.
\newblock \doi{10.1145/3106237.3106277}.
\newblock URL \url{https://doi.org/10.1145/3106237.3106277}.

\bibitem[Garvie(2016)]{garvie2016perpetual}
Clare Garvie.
\newblock \emph{The perpetual line-up: Unregulated police face recognition in
  America}.
\newblock Georgetown Law, Center on Privacy \& Technology, 2016.

\bibitem[Geber and Friemel(2021)]{geber-typology-based_2021}
Sarah Geber and Thomas Friemel.
\newblock A {Typology}-{Based} {Approach} to {Tracing}-{App} {Adoption}
  {During} the {COVID}-19 {Pandemic}: {The} {Case} of the {SwissCovid} {App}.
\newblock \emph{Journal of Quantitative Description: Digital Media}, 1, April
  2021.
\newblock ISSN 2673-8813.
\newblock \doi{10.51685/jqd.2021.007}.
\newblock URL \url{https://journalqd.org/article/view/2556}.

\bibitem[Gefen et~al.(2020)Gefen, Ben-Porat, Tennenholtz, and
  Yom-Tov]{gefen2020privacy}
Gilie Gefen, Omer Ben-Porat, Moshe Tennenholtz, and Elad Yom-Tov.
\newblock Privacy, altruism, and experience: Estimating the perceived value of
  internet data for medical uses.
\newblock In \emph{Companion Proceedings of the Web Conference 2020}, pages
  552--556, 2020.

\bibitem[Ginsberg et~al.(2009)Ginsberg, Mohebbi, Patel, Brammer, Smolinski, and
  Brilliant]{ginsberg2009detecting}
Jeremy Ginsberg, Matthew~H Mohebbi, Rajan~S Patel, Lynnette Brammer, Mark~S
  Smolinski, and Larry Brilliant.
\newblock Detecting influenza epidemics using search engine query data.
\newblock \emph{Nature}, 457\penalty0 (7232):\penalty0 1012--1014, 2009.

\bibitem[Goel et~al.(2018)Goel, Yaghini, and Faltings]{goel2018non}
Naman Goel, Mohammad Yaghini, and Boi Faltings.
\newblock Non-discriminatory machine learning through convex fairness criteria.
\newblock \emph{Proceedings of the AAAI Conference on Artificial Intelligence},
  32\penalty0 (1), 2018.
\newblock URL \url{https://ojs.aaai.org/index.php/AAAI/article/view/11662}.

\bibitem[{Gomes de Andrade} et~al.(2018){Gomes de Andrade}, Pawson, Muriello,
  Donahue, and Guadagno]{GomesdeAndrade2018}
Norberto~Nuno {Gomes de Andrade}, Dave Pawson, Dan Muriello, Lizzy Donahue, and
  Jennifer Guadagno.
\newblock {Ethics and Artificial Intelligence: Suicide Prevention on Facebook},
  dec 2018.
\newblock ISSN 22105441.

\bibitem[Goodfellow et~al.(2015)Goodfellow, Shlens, and
  Szegedy]{goodfellow2014explaining}
Ian~J. Goodfellow, Jonathon Shlens, and Christian Szegedy.
\newblock Explaining and harnessing adversarial examples.
\newblock In \emph{International Conference on Learning Representations
  (ICLR)}, 2015.

\bibitem[Google(2021)]{GooglePhotosFRT}
Google.
\newblock How google uses pattern recognition to make sense of images.
\newblock
  \url{https://policies.google.com/technologies/pattern-recognition?hl=en-US},
  2021.
\newblock Accessed: 2021-06-07.

\bibitem[Grgi?-Hla?a et~al.(2018)Grgi?-Hla?a, Zafar, Gummadi, and
  Weller]{nina2018beyond}
Nina Grgi?-Hla?a, Muhammad~Bilal Zafar, Krishna~P. Gummadi, and Adrian Weller.
\newblock Beyond distributive fairness in algorithmic decision making: Feature
  selection for procedurally fair learning.
\newblock In \emph{AAAI Conference on Artificial Intelligence (AAAI)}, 2018.

\bibitem[Grother et~al.(2019)Grother, Ngan, and Hanaoka]{grother2019face}
Patrick Grother, Mei Ngan, and Kayee Hanaoka.
\newblock \emph{Face Recognition Vendor Test (FVRT): Part 3, Demographic
  Effects}.
\newblock National Institute of Standards and Technology, 2019.

\bibitem[Gutman(2021)]{Gutman21:King}
David Gutman.
\newblock {K}ing {C}ounty {C}ouncil bans use of facial recognition technology
  by {S}heriff's {O}ffice, other agencies.
\newblock \emph{The Seattle Times}, June 2021.
\newblock URL
  \url{https://www.seattletimes.com/seattle-news/politics/king-county-council-bans-use-of-facial-recognition-technology-by-sheriffs-office-other-agencies/}.

\bibitem[Habib et~al.(2018)Habib, Naeini, Devlin, Oates, Swoopes, Bauer,
  Christin, and Cranor]{habib2018user}
Hana Habib, Pardis~Emami Naeini, Summer Devlin, Maggie Oates, Chelse Swoopes,
  Lujo Bauer, Nicolas Christin, and Lorrie~Faith Cranor.
\newblock User behaviors and attitudes under password expiration policies.
\newblock In \emph{Fourteenth Symposium on Usable Privacy and Security
  ($\{$SOUPS$\}$ 2018)}, pages 13--30, 2018.

\bibitem[Haischer et~al.(2020)Haischer, Beilfuss, Hart, Opielinski, Wrucke,
  Zirgaitis, Uhrich, and Hunter]{haischer2020wearing}
Michael~H Haischer, Rachel Beilfuss, Meggie~Rose Hart, Lauren Opielinski, David
  Wrucke, Gretchen Zirgaitis, Toni~D Uhrich, and Sandra~K Hunter.
\newblock Who is wearing a mask? gender-, age-, and location-related
  differences during the covid-19 pandemic.
\newblock \emph{PloS one}, 15\penalty0 (10):\penalty0 e0240785, 2020.

\bibitem[Hamidi et~al.(2018)Hamidi, Scheuerman, and Branham]{hamidi2018gender}
Foad Hamidi, Morgan~Klaus Scheuerman, and Stacy~M Branham.
\newblock Gender recognition or gender reductionism? the social implications of
  embedded gender recognition systems.
\newblock In \emph{Proceedings of the 2018 chi conference on human factors in
  computing systems}, pages 1--13, 2018.

\bibitem[Han et~al.(2017)Han, Kim, and Kim]{han2017pyramidnet}
Dongyoon Han, Jiwhan Kim, and Junmo Kim.
\newblock Deep pyramidal residual networks.
\newblock \emph{IEEE CVPR}, 2017.

\bibitem[Hanna et~al.(2020)Hanna, Denton, Smart, and
  Smith-Loud]{hanna2020towards}
Alex Hanna, Emily Denton, Andrew Smart, and Jamila Smith-Loud.
\newblock Towards a critical race methodology in algorithmic fairness.
\newblock In \emph{ACM Conference on Fairness, Accountability, and Transparency
  (FAccT)}, pages 501--512, 2020.

\bibitem[Hardt et~al.(2016)Hardt, Price, and Srebro]{hardt2016equality}
Moritz Hardt, Eric Price, and Nathan Srebro.
\newblock Equality of opportunity in supervised learning.
\newblock In \emph{Proceedings of the Annual Conference on Neural Information
  Processing Systems (NeurIPS)}, 2016.

\bibitem[Hartzog(2020)]{hartzog2020secretive}
Woodrow Hartzog.
\newblock The secretive company that might end privacy as we know it.
\newblock \emph{The New York Times}, Jan. 18 2020.
\newblock URL
  \url{https://www.nytimes.com/2020/01/18/technology/clearview-privacy-facial-recognition.html}.

\bibitem[Hashimoto et~al.(2018)Hashimoto, Srivastava, Namkoong, and
  Liang]{hashimoto2018fairness}
Tatsunori~B. Hashimoto, Megha Srivastava, Hongseok Namkoong, and Percy Liang.
\newblock Fairness without demographics in repeated loss minimization.
\newblock In \emph{International Conference on Machine Learning (ICML)}, 2018.

\bibitem[Hazirbas et~al.(2021)Hazirbas, Bitton, Dolhansky, Pan, Gordo, and
  Ferrer]{ccd}
Caner Hazirbas, Joanna Bitton, Brian Dolhansky, Jacqueline Pan, Albert Gordo,
  and Cristian~Canton Ferrer.
\newblock Towards measuring fairness in ai: the casual conversations dataset.
\newblock \emph{arXiv preprint arXiv:2104.02821}, 2021.

\bibitem[He et~al.(2016)He, Zhang, Ren, and Sun]{he2016deep}
Kaiming He, Xiangyu Zhang, Shaoqing Ren, and Jian Sun.
\newblock Deep residual learning for image recognition.
\newblock In \emph{Computer Vision and Pattern Recognition (CVPR)}, pages
  770--778, 2016.

\bibitem[Hedegaard et~al.(2018)Hedegaard, Curtin, and
  Warner]{hedegaard2018suicide}
Holly Hedegaard, Sally~C Curtin, and Margaret Warner.
\newblock Suicide rates in the united states continue to increase.
\newblock \emph{NCHS Data Brief No. 309}, June 2018.

\bibitem[Heidari and Krause(2018)]{heidari2018preventing}
Hoda Heidari and Andreas Krause.
\newblock Preventing disparate treatment in sequential decision making.
\newblock In \emph{Proceedings of the International Joint Conference on
  Artificial Intelligence (IJCAI)}, 2018.

\bibitem[Heidari et~al.(2019)Heidari, Nanda, and Gummadi]{heidari2019longterm}
Hoda Heidari, Vedant Nanda, and Krishna~P. Gummadi.
\newblock On the long-term impact of algorithmic decision policies: Effort
  unfairness and feature segregation through social learning.
\newblock In \emph{International Conference on Machine Learning (ICML)}, 2019.

\bibitem[Hendrycks and Dietterich(2019)]{hendrycks2019benchmarking}
Dan Hendrycks and Thomas Dietterich.
\newblock Benchmarking neural network robustness to common corruptions and
  perturbations.
\newblock 2019.

\bibitem[Holstein et~al.(2019)Holstein, Wortman~Vaughan, Daum\'{e}, Dudik, and
  Wallach]{holstein2019improving}
Kenneth Holstein, Jennifer Wortman~Vaughan, Hal Daum\'{e}, Miro Dudik, and
  Hanna Wallach.
\newblock Improving fairness in machine learning systems: What do industry
  practitioners need?
\newblock In \emph{Proceedings of the 2019 CHI Conference on Human Factors in
  Computing Systems}, CHI '19, page 1?16, 2019.
\newblock ISBN 9781450359702.
\newblock \doi{10.1145/3290605.3300830}.
\newblock URL \url{https://doi.org/10.1145/3290605.3300830}.

\bibitem[Horvath et~al.(2020)Horvath, Banducci, and
  James]{horvath_citizens_2020}
Laszlo Horvath, Susan Banducci, and Oliver James.
\newblock Citizens? {Attitudes} to {Contact} {Tracing} {Apps}.
\newblock \emph{Journal of Experimental Political Science}, pages 1--13,
  September 2020.
\newblock ISSN 2052-2630, 2052-2649.
\newblock \doi{10.1017/XPS.2020.30}.
\newblock URL
  \url{https://www.cambridge.org/core/journals/journal-of-experimental-political-science/article/citizens-attitudes-to-contact-tracing-apps/F9B8B8CFE051E6D89C3C9ADD6DF76019}.

\bibitem[Horvitz and Mulligan(2015)]{Horvitz2015}
Eric Horvitz and Deirdre Mulligan.
\newblock {Data, privacy, and the greater good}.
\newblock \emph{Science}, 2015.
\newblock ISSN 10959203.
\newblock \doi{10.1126/science.aac4520}.

\bibitem[Hosseini et~al.(2017)Hosseini, Xiao, and
  Poovendran]{hosseini2017google}
Hossein Hosseini, Baicen Xiao, and Radha Poovendran.
\newblock {G}oogle's cloud vision {API} is not robust to noise.
\newblock In \emph{2017 16th IEEE international conference on machine learning
  and applications (ICMLA)}, pages 101--105. IEEE, 2017.

\bibitem[Hu et~al.(2017)Hu, Jackson, Yates, White, Phillips, and
  O?Toole]{hu2017person}
Ying Hu, Kelsey Jackson, Amy Yates, David White, P~Jonathon Phillips, and
  Alice~J O?Toole.
\newblock Person recognition: Qualitative differences in how forensic face
  examiners and untrained people rely on the face versus the body for
  identification.
\newblock \emph{Visual Cognition}, 25\penalty0 (4-6):\penalty0 492--506, 2017.

\bibitem[Huang et~al.(2017)Huang, Liu, Van Der~Maaten, and
  Weinberger]{huang2017densely}
Gao Huang, Zhuang Liu, Laurens Van Der~Maaten, and Kilian~Q Weinberger.
\newblock Densely connected convolutional networks.
\newblock In \emph{Computer Vision and Pattern Recognition (CVPR)}, 2017.

\bibitem[Huang et~al.(2008)Huang, Mattar, Berg, and
  Learned-Miller]{huang2008labeled}
Gary~B Huang, Marwan Mattar, Tamara Berg, and Eric Learned-Miller.
\newblock Labeled faces in the wild: A database forstudying face recognition in
  unconstrained environments.
\newblock In \emph{Workshop on faces in'Real-Life'Images: detection, alignment,
  and recognition}, 2008.

\bibitem[Huang et~al.(2021)Huang, Jackson, Derakhshan, Lee, Pham, Jackson, and
  Cutter]{huang2021urban}
Qian Huang, Sarah Jackson, Sahar Derakhshan, Logan Lee, Erika Pham, Amber
  Jackson, and Susan~L Cutter.
\newblock Urban-rural differences in covid-19 exposures and outcomes in the
  south: A preliminary analysis of south carolina.
\newblock \emph{PloS one}, 16\penalty0 (2):\penalty0 e0246548, 2021.

\bibitem[Iandola et~al.(2016)Iandola, Han, Moskewicz, Ashraf, Dally, and
  Keutzer]{iandola2016squeezenet}
Forrest~N Iandola, Song Han, Matthew~W Moskewicz, Khalid Ashraf, William~J
  Dally, and Kurt Keutzer.
\newblock {S}queeze{N}et: {A}lex{N}et-level accuracy with 50x fewer parameters
  and {\textless}0.5mb model size.
\newblock \emph{CoRR}, abs/1602.07360, 2016.

\bibitem[Insel(2008)]{Insel2008}
Thomas~R. Insel.
\newblock {Assessing the economic costs of serious mental illness}.
\newblock \emph{American Journal of Psychiatry}, 165\penalty0 (6):\penalty0
  663--665, jun 2008.
\newblock ISSN 0002953X.

\bibitem[Irani(2016)]{Irani2016}
Lilly Irani.
\newblock {The hidden faces of automation}.
\newblock \emph{XRDS: Crossroads, The ACM Magazine for Students}, 23\penalty0
  (2):\penalty0 34--37, dec 2016.
\newblock ISSN 15284972.
\newblock \doi{10.1145/3014390}.
\newblock URL \url{http://dl.acm.org/citation.cfm?doid=3026779.3014390}.

\bibitem[Iter et~al.(2018)Iter, Yoon, and Jurafsky]{Iter2018}
Dan Iter, Jong Yoon, and Dan Jurafsky.
\newblock {Automatic Detection of Incoherent Speech for Diagnosing
  Schizophrenia}.
\newblock In \emph{Proceedings of the Fifth Workshop on Computational
  Linguistics and Clinical Psychology: From Keyboard to Clinic}, pages
  136--146, Stroudsburg, PA, USA, 2018. Association for Computational
  Linguistics.

\bibitem[Jacobs et~al.(2015)Jacobs, Clawson, and Mynatt]{jacobs2015comparing}
Maia~L Jacobs, James Clawson, and Elizabeth~D Mynatt.
\newblock Comparing health information sharing preferences of cancer patients,
  doctors, and navigators.
\newblock In \emph{Proceedings of the 18th ACM Conference on Computer Supported
  Cooperative Work \& Social Computing}, pages 808--818, 2015.

\bibitem[Jain and Parsheera(2021)]{jain2021Million}
Gaurav Jain and Smriti Parsheera.
\newblock 1.4 billion missing pieces? auditing the accuracy of facial
  processing tools on indian faces.
\newblock \emph{First Workshop on Ethical Considerations in Creative
  applications of Computer Vision}, 2021.

\bibitem[Jaroszewski et~al.(2019)Jaroszewski, Morris, and
  Nock]{Jaroszewski2019}
Adam~C. Jaroszewski, Robert~R. Morris, and Matthew~K. Nock.
\newblock {Randomized controlled trial of an online machine learning-driven
  risk assessment and intervention platform for increasing the use of crisis
  services}.
\newblock \emph{Journal of Consulting and Clinical Psychology}, 87\penalty0
  (4):\penalty0 370--379, apr 2019.
\newblock ISSN 19392117.

\bibitem[Julienne et~al.(2020)Julienne, Lavin, Belton, Barjakov{\'a}, Timmons,
  and Lunn]{julienne2020behavioural}
Hannah Julienne, Ciar{\'a}n Lavin, Cameron Belton, Martina Barjakov{\'a}, Shane
  Timmons, and Peter~D Lunn.
\newblock Behavioural pre-testing of covid tracker, ireland?s contact-tracing
  app.
\newblock 2020.

\bibitem[{Kaiser Family Foundation}(2019)]{kff_2019}
{Kaiser Family Foundation}.
\newblock Mental health care health professional shortage areas (hpsas).
\newblock
  \url{https://www.kff.org/other/state-indicator/mental-health-care-health-professional-shortage-areas-hpsas},
  Nov 2019.

\bibitem[Kantayya(2020)]{CodedBias}
Shalini Kantayya.
\newblock Coded bias, 2020.
\newblock Feature-length documentary.

\bibitem[Kaptchuk et~al.(2020)Kaptchuk, Goldstein, Hargittai, Hofman, and
  Redmiles]{kaptchuk_how_2020}
Gabriel Kaptchuk, Daniel~G. Goldstein, Eszter Hargittai, Jake Hofman, and
  Elissa~M. Redmiles.
\newblock How good is good enough for {COVID19} apps? {The} influence of
  benefits, accuracy, and privacy on willingness to adopt.
\newblock \emph{arXiv:2005.04343 [cs]}, May 2020.
\newblock URL \url{http://arxiv.org/abs/2005.04343}.
\newblock arXiv: 2005.04343.

\bibitem[Karampela et~al.(2019)Karampela, Ouhbi, and
  Isomursu]{karampela2019connected}
Maria Karampela, Sofia Ouhbi, and Minna Isomursu.
\newblock Connected health user willingness to share personal health data:
  questionnaire study.
\newblock \emph{Journal of medical Internet research}, 21\penalty0
  (11):\penalty0 e14537, 2019.

\bibitem[Kashima et~al.(1995)Kashima, Yamaguchi, Kim, Choi, Gelfand, and
  Yuki]{kashima1995culture}
Yoshihisa Kashima, Susumu Yamaguchi, Uichol Kim, Sang-Chin Choi, Michele~J
  Gelfand, and Masaki Yuki.
\newblock Culture, gender, and self: a perspective from
  individualism-collectivism research.
\newblock \emph{Journal of personality and social psychology}, 69\penalty0
  (5):\penalty0 925, 1995.

\bibitem[Kelly et~al.(2019)Kelly, Spaderna, Hodzic, Nair, Kitchen, Werkheiser,
  Powell, Feldman, Liu, Espy-Wilson, Coppersmith, and Resnik]{Kelly}
Deanna~L. Kelly, Max Spaderna, Vedrana Hodzic, Suraj Nair, Christopher Kitchen,
  Anne Werkheiser, Megan Powell, Stephanie Feldman, Fang Liu, Carol
  Espy-Wilson, Glen Coppersmith, and Philip Resnik.
\newblock {Blinded Clinical Ratings of Social Media Data are Correlated with
  In-Person Clinical Ratings in Participants Diagnosed with Either Depression,
  Schizophrenia, or Healthy Controls}.
\newblock 2019.

\bibitem[Kemmelmeier et~al.(2006)Kemmelmeier, Jambor, and
  Letner]{kemmelmeier2006individualism}
Markus Kemmelmeier, Edina~E Jambor, and Joyce Letner.
\newblock Individualism and good works: Cultural variation in giving and
  volunteering across the united states.
\newblock \emph{Journal of Cross-Cultural Psychology}, 37\penalty0
  (3):\penalty0 327--344, 2006.

\bibitem[Keyes(2018)]{keyes2018misgendering}
Os~Keyes.
\newblock The misgendering machines: Trans/hci implications of automatic gender
  recognition.
\newblock \emph{Proceedings of the ACM on human-computer interaction},
  2\penalty0 (CSCW):\penalty0 1--22, 2018.

\bibitem[Khandani et~al.(2010)Khandani, Kim, and Lo]{khandani2010consumer}
Amir~E. Khandani, Adlar~J. Kim, and Andrew~W. Lo.
\newblock Consumer credit-risk models via machine-learning algorithms.
\newblock \emph{Journal of Banking \& Finance}, 34\penalty0 (11):\penalty0
  2767--2787, 2010.

\bibitem[Khani and Liang(2019)]{khani2019noise}
Fereshte Khani and Percy Liang.
\newblock Noise induces loss discrepancy across groups for linear regression,
  2019.

\bibitem[Kilbertus et~al.(2017)Kilbertus, Carulla, Parascandolo, Hardt,
  Janzing, and Sch{\"o}lkopf]{Kilbertus17:Avoiding}
Niki Kilbertus, Mateo~Rojas Carulla, Giambattista Parascandolo, Moritz Hardt,
  Dominik Janzing, and Bernhard Sch{\"o}lkopf.
\newblock Avoiding discrimination through causal reasoning.
\newblock In \emph{Proceedings of the Annual Conference on Neural Information
  Processing Systems (NeurIPS)}, pages 656--666, 2017.

\bibitem[Kim et~al.(2019)Kim, Barasz, and John]{kim2019seeing}
Tami Kim, Kate Barasz, and Leslie~K John.
\newblock Why am i seeing this ad? the effect of ad transparency on ad
  effectiveness.
\newblock \emph{Journal of Consumer Research}, 45\penalty0 (5):\penalty0
  906--932, 2019.

\bibitem[Klare et~al.(2012)Klare, Burge, Klontz, Bruegge, and
  Jain]{klare2012face}
Brendan~F Klare, Mark~J Burge, Joshua~C Klontz, Richard W~Vorder Bruegge, and
  Anil~K Jain.
\newblock Face recognition performance: Role of demographic information.
\newblock \emph{IEEE Transactions on Information Forensics and Security},
  7\penalty0 (6):\penalty0 1789--1801, 2012.

\bibitem[Knittel et~al.(2022)Knittel, Dooley, and
  Dickerson]{knittel2022dichotomous}
Marina Knittel, Samuel Dooley, and John~P Dickerson.
\newblock The dichotomous affiliate stable matching problem: Approval-based
  matching with applicant-employer relations.
\newblock \emph{arXiv preprint arXiv:2202.11095}, 2022.

\bibitem[Korn et~al.(2020)Korn, B{\"o}hm, Meier, and
  Betsch]{korn2020vaccination}
Lars Korn, Robert B{\"o}hm, Nicolas~W Meier, and Cornelia Betsch.
\newblock Vaccination as a social contract.
\newblock \emph{Proceedings of the National Academy of Sciences}, 117\penalty0
  (26):\penalty0 14890--14899, 2020.

\bibitem[Kostka et~al.(2021)Kostka, Steinacker, and Meckel]{kostka2021between}
Genia Kostka, L{\'e}a Steinacker, and Miriam Meckel.
\newblock Between security and convenience: Facial recognition technology in
  the eyes of citizens in china, germany, the united kingdom, and the united
  states.
\newblock \emph{Public Understanding of Science}, page 09636625211001555, 2021.

\bibitem[Krause and Golovin(2014)]{krause2014submodular}
Andreas Krause and Daniel Golovin.
\newblock Submodular function maximization., 2014.

\bibitem[Krizhevsky(2009)]{krizhevsky2009learning}
Alex Krizhevsky.
\newblock Learning multiple layers of features from tiny images.
\newblock \emph{Master's thesis, University of Toronto}, 2009.

\bibitem[Krizhevsky(2014)]{krizhevsky2014one}
Alex Krizhevsky.
\newblock One weird trick for parallelizing convolutional neural networks.
\newblock \emph{CoRR}, abs/1404.5997, 2014.

\bibitem[Kuo et~al.(2007)Kuo, Lin, and Hsu]{kuo2007assessing}
Feng-Yang Kuo, Cathy~S Lin, and Meng-Hsiang Hsu.
\newblock Assessing gender differences in computer professionals?
  self-regulatory efficacy concerning information privacy practices.
\newblock \emph{Journal of business ethics}, 73\penalty0 (2):\penalty0
  145--160, 2007.

\bibitem[Kuo et~al.(2020)Kuo, Ostuni, Horishny, Curry, Dooley, Chiang,
  Goldstein, and Dickerson]{kuo2020proportionnet}
Kevin Kuo, Anthony Ostuni, Elizabeth Horishny, Michael~J Curry, Samuel Dooley,
  Ping-yeh Chiang, Tom Goldstein, and John~P Dickerson.
\newblock Proportionnet: Balancing fairness and revenue for auction design with
  deep learning.
\newblock \emph{arXiv preprint arXiv:2010.06398}, 2020.

\bibitem[Kusner et~al.(2017)Kusner, Loftus, Russell, and
  Silva]{kusner2017counterfactual}
Matt~J Kusner, Joshua Loftus, Chris Russell, and Ricardo Silva.
\newblock Counterfactual fairness.
\newblock In \emph{Advances in Neural Information Processing Systems 30}, pages
  4066--4076. 2017.
\newblock URL
  \url{http://papers.nips.cc/paper/6995-counterfactual-fairness.pdf}.

\bibitem[Kveton et~al.(2014)Kveton, Wen, Ashkan, Eydgahi, and
  Eriksson]{Kveton14:Matroid}
Branislav Kveton, Zheng Wen, Azin Ashkan, Hoda Eydgahi, and Brian Eriksson.
\newblock Matroid bandits: Fast combinatorial optimization with learning.
\newblock In \emph{Conference on Uncertainty in Artificial Intelligence (UAI)},
  2014.

\bibitem[Ladyzhets(2021)]{ladyzhets_we_2021}
Betty Ladyzhets.
\newblock We investigated whether digital contact tracing actually worked in
  the {US}, June 2021.
\newblock URL
  \url{https://www.technologyreview.com/2021/06/16/1026255/us-digital-contact-tracing-exposure-notification-analysis/}.

\bibitem[Lahoti et~al.(2020)Lahoti, Beutel, Chen, Lee, Prost, Thain, Wang, and
  Chi]{lahoti2020fairness}
Preethi Lahoti, Alex Beutel, Jilin Chen, Kang Lee, Flavien Prost, Nithum Thain,
  Xuezhi Wang, and Ed~H. Chi.
\newblock Fairness without demographics through adversarially reweighted
  learning.
\newblock \emph{arXiv preprint arXiv:2006.13114}, 2020.

\bibitem[Lai and Robbins(1985)]{Lai85:Asymptotically}
Tze~Leung Lai and Herbert Robbins.
\newblock Asymptotically efficient adaptive allocation rules.
\newblock \emph{Advances in Applied Mathematics}, 6\penalty0 (1):\penalty0
  4--22, 1985.

\bibitem[Lam et~al.(2018)Lam, Kuzma, McGee, Dooley, Laielli, Klaric, Bulatov,
  and McCord]{lam2018xview}
Darius Lam, Richard Kuzma, Kevin McGee, Samuel Dooley, Michael Laielli, Matthew
  Klaric, Yaroslav Bulatov, and Brendan McCord.
\newblock xview: Objects in context in overhead imagery.
\newblock \emph{arXiv preprint arXiv:1802.07856}, 2018.

\bibitem[Langford and Zhang(2008)]{Langford08:Epoch-greedy}
John Langford and Tong Zhang.
\newblock The epoch-greedy algorithm for multi-armed bandits with side
  information.
\newblock In \emph{Conference on Neural Information Processing Systems
  (NeurIPS)}, pages 817--824, 2008.

\bibitem[Langheinrich and Schaub(2018)]{langheinrich2018privacy}
Marc Langheinrich and Florian Schaub.
\newblock Privacy in mobile and pervasive computing.
\newblock \emph{Synthesis Lectures on Mobile and Pervasive Computing},
  10\penalty0 (1):\penalty0 1--139, 2018.

\bibitem[Leben(2020)]{Leben20:Normative}
Derek Leben.
\newblock Normative principles for evaluating fairness in machine learning.
\newblock In \emph{Conference on Artificial Intelligence, Ethics, and Society
  (AIES)}, pages 86--92, 2020.

\bibitem[Lee(2014)]{lee2014:samaritan}
Naomi Lee.
\newblock Trouble on the radar.
\newblock \emph{The Lancet Technology}, 384\penalty0 (9958):\penalty0 1917,
  November 2014.
\newblock https://doi.org/10.1016/S0140-6736(14)62267-4.

\bibitem[Li et~al.(2010)Li, Chu, Langford, and Schapire]{Li10:Contextual}
Lihong Li, Wei Chu, John Langford, and Robert~E Schapire.
\newblock A contextual-bandit approach to personalized news article
  recommendation.
\newblock In \emph{International Conference on World Wide Web (WWW)}, pages
  661--670, 2010.

\bibitem[Li et~al.(2021)Li, Cobb, Yang, Baviskar, Agarwal, Li, Bauer, and
  Hong]{li2021makes}
Tianshi Li, Camille Cobb, Jackie Yang, Sagar Baviskar, Yuvraj Agarwal, Beibei
  Li, Lujo Bauer, and Jason~I Hong.
\newblock What makes people install a covid-19 contact-tracing app?
  understanding the influence of app design and individual difference on
  contact-tracing app adoption intention.
\newblock \emph{Pervasive and Mobile Computing}, page 101439, 2021.

\bibitem[Lim and Li(2018)]{Lim18:Optimal}
Jooseop Lim and Tieshan Li.
\newblock The optimal advertising-allocation rules for sequentially released
  products: The case of the motion picture industry.
\newblock \emph{Journal of Advertising Research}, 58\penalty0 (2):\penalty0
  228--239, 2018.

\bibitem[Lin and Bilmes(2011)]{lin2011class}
Hui Lin and Jeff Bilmes.
\newblock A class of submodular functions for document summarization.
\newblock In \emph{Proceedings of the 49th Annual Meeting of the Association
  for Computational Linguistics: Human Language Technologies-Volume 1}, pages
  510--520. Association for Computational Linguistics, 2011.

\bibitem[Linthicum et~al.(2019)Linthicum, Schafer, and Ribeiro]{Linthicum2019}
Kathryn~P. Linthicum, Katherine~Musacchio Schafer, and Jessica~D. Ribeiro.
\newblock {Machine learning in suicide science: Applications and ethics}.
\newblock \emph{Behavioral Sciences {\&} the Law}, 37\penalty0 (3):\penalty0
  214--222, may 2019.
\newblock ISSN 0735-3936.

\bibitem[Littman(1996)]{littman1996algorithms}
Michael~Lederman Littman.
\newblock \emph{Algorithms for sequential decision making}.
\newblock Brown University Providence, RI, 1996.

\bibitem[Liu et~al.(2018)Liu, Dean, Rolf, Simchowitz, and
  Hardt]{liu2018delayed}
Lydia~T Liu, Sarah Dean, Esther Rolf, Max Simchowitz, and Moritz Hardt.
\newblock Delayed impact of fair machine learning.
\newblock In \emph{International Conference on Machine Learning (ICML)}, 2018.

\bibitem[Liu et~al.(2015)Liu, Luo, Wang, and Tang]{liu2015faceattributes}
Ziwei Liu, Ping Luo, Xiaogang Wang, and Xiaoou Tang.
\newblock Deep learning face attributes in the wild.
\newblock In \emph{Proceedings of International Conference on Computer Vision
  (ICCV)}, December 2015.

\bibitem[Lockey et~al.(2021)Lockey, Edwards, Hornsey, Gillespie, Akhlaghpour,
  and Colville]{lockey2021profiling}
Steven Lockey, Martin~R Edwards, Matthew~J Hornsey, Nicole Gillespie, Saeed
  Akhlaghpour, and Shannon Colville.
\newblock Profiling adopters (and non-adopters) of a contact tracing mobile
  application: insights from australia.
\newblock \emph{International Journal of Medical Informatics}, 149:\penalty0
  104414, 2021.

\bibitem[Lohr(2018)]{lohr2018facial}
Steve Lohr.
\newblock Facial recognition is accurate, if you?re a white guy.
\newblock \emph{New York Times}, 9, 2018.

\bibitem[Losada et~al.(2018)Losada, Crestani, and Parapar]{Losada2018}
David~E. Losada, Fabio Crestani, and Javier Parapar.
\newblock {Overview of eRisk: Early risk prediction on the internet}.
\newblock In \emph{Lecture Notes in Computer Science (including subseries
  Lecture Notes in Artificial Intelligence and Lecture Notes in
  Bioinformatics)}, volume 11018 LNCS, pages 343--361. Springer Verlag, 2018.
\newblock ISBN 9783319989310.

\bibitem[Losada et~al.(2019)Losada, Crestani, and Parapar]{Losada2019}
David~E. Losada, Fabio Crestani, and Javier Parapar.
\newblock Overview of erisk 2019 early risk prediction on the internet.
\newblock In Fabio Crestani, Martin Braschler, Jacques Savoy, Andreas Rauber,
  Henning M{\"u}ller, David~E. Losada, Gundula Heinatz~B{\"u}rki, Linda
  Cappellato, and Nicola Ferro, editors, \emph{Experimental IR Meets
  Multilinguality, Multimodality, and Interaction}. Springer International
  Publishing, 2019.
\newblock ISBN 978-3-030-28577-7.

\bibitem[Lu and Tang(2015)]{lu2015surpassing}
Chaochao Lu and Xiaoou Tang.
\newblock Surpassing human-level face verification performance on lfw with
  gaussianface.
\newblock In \emph{Twenty-ninth AAAI conference on artificial intelligence},
  2015.

\bibitem[Ma et~al.(2018)Ma, Gao, Suo, You, Zhou, and Zhang]{ma2018risk}
Fenglong Ma, Jing Gao, Qiuling Suo, Quanzeng You, Jing Zhou, and Aidong Zhang.
\newblock Risk prediction on electronic health records with prior medical
  knowledge.
\newblock In \emph{Proceedings of the 24th ACM SIGKDD International Conference
  on Knowledge Discovery \& Data Mining}, pages 1910--1919, 2018.

\bibitem[MacAvaney et~al.(2018)MacAvaney, Desmet, Cohan, Soldaini, Yates,
  Zirikly, and Goharian]{MacAvaney2018}
Sean MacAvaney, Bart Desmet, Arman Cohan, Luca Soldaini, Andrew Yates, Ayah
  Zirikly, and Nazli Goharian.
\newblock {RSDD-Time: Temporal Annotation of Self-Reported Mental Health
  Diagnoses}.
\newblock In \emph{Proceedings of the Fifth Workshop on Computational
  Linguistics and Clinical Psychology: From Keyboard to Clinic}, pages
  168--173, Stroudsburg, PA, USA, 2018. Association for Computational
  Linguistics.

\bibitem[Mace et~al.(2018)Mace, Manville, Barbu-McInnis, Laielli, Klaric, and
  Dooley]{mace2018overhead}
Eliza Mace, Keith Manville, Monica Barbu-McInnis, Michael Laielli, Matthew
  Klaric, and Samuel Dooley.
\newblock Overhead detection: Beyond 8-bits and rgb.
\newblock \emph{arXiv preprint arXiv:1808.02443}, 2018.

\bibitem[Madras et~al.(2018)Madras, Creager, Pitassi, and
  Zemel]{madras2018learning}
David Madras, Elliot Creager, Toniann Pitassi, and Richard~S. Zemel.
\newblock Learning adversarially fair and transferable representations.
\newblock In \emph{Proceedings of the 35th International Conference on Machine
  Learning, {ICML} 2018, Stockholmsm{\"{a}}ssan, Stockholm, Sweden, July 10-15,
  2018}, volume~80 of \emph{Proceedings of Machine Learning Research}, pages
  3381--3390. {PMLR}, 2018.
\newblock URL \url{http://proceedings.mlr.press/v80/madras18a.html}.

\bibitem[Maitra(2020)]{Maitra2020}
Suvradip Maitra.
\newblock {Artificial Intelligence and Indigenous Perspectives: Protecting and
  Empowering Intelligent Human Beings}.
\newblock In \emph{Proceedings of the AAAI/ACM Conference on AI, Ethics, and
  Society}, pages 320--326, New York, NY, USA, feb 2020.

\bibitem[Marson and Forrest(2021{\natexlab{a}})]{marson2021}
James Marson and Brett Forrest.
\newblock Armed low-cost drones, made by turkey, reshape battlefields and
  geopolitics.
\newblock
  {https://www.wsj.com/articles/armed-low-cost-drones-made-by-turkey-reshape-battlefields-and-geopolitics-11622727370},
  Jun 2021{\natexlab{a}}.
\newblock The Wall Street Journal.

\bibitem[Marson and Forrest(2021{\natexlab{b}})]{marson_forrest_2021}
James Marson and Brett Forrest.
\newblock Armed low-cost drones, made by turkey, reshape battlefields and
  geopolitics.
\newblock \emph{The Wall Street Journal}, Jun 2021{\natexlab{b}}.
\newblock URL
  \url{https://www.wsj.com/articles/armed-low-cost-drones-made-by-turkey-reshape-battlefields-and-geopolitics-11622727370}.

\bibitem[Martinez et~al.(2020)Martinez, Bertran, and
  Sapiro]{martinez2020minimax}
Natalia Martinez, Martin Bertran, and Guillermo Sapiro.
\newblock Minimax pareto fairness: A multi objective perspective.
\newblock In \emph{Proceedings of the 37th International Conference on Machine
  Learning}, volume 119, pages 6755--6764, 2020.
\newblock URL \url{http://proceedings.mlr.press/v119/martinez20a.html}.

\bibitem[Mathur et~al.(2018)Mathur, Vitak, Narayanan, and
  Chetty]{mathur2018characterizing}
Arunesh Mathur, Jessica Vitak, Arvind Narayanan, and Marshini Chetty.
\newblock Characterizing the use of browser-based blocking extensions to
  prevent online tracking.
\newblock In \emph{Fourteenth Symposium on Usable Privacy and Security
  ($\{$SOUPS$\}$ 2018)}, pages 103--116, 2018.

\bibitem[McDonald et~al.(2012)McDonald, Mohebbi, and
  Slatkin]{mcdonald2012comparing}
Paul McDonald, Matt Mohebbi, and Brett Slatkin.
\newblock Comparing google consumer surveys to existing probability and
  non-probability based internet surveys.
\newblock \emph{Google White Paper}, 2012.

\bibitem[Mehrabi et~al.(2020)Mehrabi, Naveed, Morstatter, and
  Galstyan]{mehrabi2020exacerbating}
Ninareh Mehrabi, Muhammad Naveed, Fred Morstatter, and Aram Galstyan.
\newblock Exacerbating algorithmic bias through fairness attacks, 2020.

\bibitem[Mikal et~al.(2016{\natexlab{a}})Mikal, Hurst, and Conway]{Mikal2016}
Jude Mikal, Samantha Hurst, and Mike Conway.
\newblock {Ethical issues in using Twitter for population-level depression
  monitoring: A qualitative study}.
\newblock \emph{BMC Medical Ethics}, 17\penalty0 (1):\penalty0 22, dec
  2016{\natexlab{a}}.
\newblock ISSN 14726939.

\bibitem[Mikal et~al.(2016{\natexlab{b}})Mikal, Hurst, and
  Conway]{mikal2016ethical}
Jude Mikal, Samantha Hurst, and Mike Conway.
\newblock Ethical issues in using {Twitter} for population-level depression
  monitoring: A qualitative study.
\newblock \emph{BMC Medical Ethics}, 17\penalty0 (1):\penalty0 1--11,
  2016{\natexlab{b}}.

\bibitem[Milne et~al.(2016{\natexlab{a}})Milne, Pink, Hachey, and
  Calvo]{Milne2016}
David~N Milne, Glen Pink, Ben Hachey, and Rafael~A Calvo.
\newblock {Triaging content in online peer-support forums}.
\newblock pages 118--127, 2016{\natexlab{a}}.
\newblock URL \url{https://www.aclweb.org/anthology/W16-0312}.

\bibitem[Milne et~al.(2016{\natexlab{b}})Milne, Pink, Hachey, and
  Calvo]{Milne2016a}
David~N. Milne, Glen Pink, Ben Hachey, and Rafael~A. Calvo.
\newblock {CLP}sych 2016 shared task: Triaging content in online peer-support
  forums.
\newblock In \emph{Proceedings of the Third Workshop on Computational
  Linguistics and Clinical Psychology}, pages 118--127, San Diego, CA, USA,
  June 2016{\natexlab{b}}. Association for Computational Linguistics.
\newblock \doi{10.18653/v1/W16-0312}.
\newblock URL \url{https://www.aclweb.org/anthology/W16-0312}.

\bibitem[Milne et~al.(2019)Milne, McCabe, and Calvo]{Milne2019}
David~N. Milne, Kathryn~L. McCabe, and Rafael~A. Calvo.
\newblock {Improving moderator responsiveness in online peer support through
  automated triage}.
\newblock \emph{Journal of Medical Internet Research}, 21\penalty0 (4), apr
  2019.
\newblock ISSN 14388871.

\bibitem[Monahan(2008)]{Monahan2008}
Torin Monahan.
\newblock {Editorial: surveillance and inequality}.
\newblock Technical Report~3, 2008.
\newblock URL \url{http://www.surveillance-and-society.org}.

\bibitem[Moosavi-Dezfooli et~al.(2016)Moosavi-Dezfooli, Fawzi, and
  Frossard]{moosavi2016deepfool}
Seyed-Mohsen Moosavi-Dezfooli, Alhussein Fawzi, and Pascal Frossard.
\newblock Deepfool: a simple and accurate method to fool deep neural networks.
\newblock In \emph{Computer Vision and Pattern Recognition (CVPR)}, pages
  2574--2582, 2016.

\bibitem[Munns and Basu(2017)]{munns2017privacy}
Christina Munns and Subhajit Basu.
\newblock \emph{Privacy and healthcare data:?Choice of Control?to ?Choice?and
  ?Control?}
\newblock Routledge, 2017.

\bibitem[Munzert et~al.(2021)Munzert, Selb, Gohdes, Stoetzer, and
  Lowe]{munzert2021tracking}
Simon Munzert, Peter Selb, Anita Gohdes, Lukas~F Stoetzer, and Will Lowe.
\newblock Tracking and promoting the usage of a covid-19 contact tracing app.
\newblock \emph{Nature Human Behaviour}, 5\penalty0 (2):\penalty0 247--255,
  2021.

\bibitem[Nanda et~al.(2021)Nanda, Dooley, Singla, Feizi, and
  Dickerson]{nanda2021fairness}
Vedant Nanda, Samuel Dooley, Sahil Singla, Soheil Feizi, and John~P Dickerson.
\newblock Fairness through robustness: Investigating robustness disparity in
  deep learning.
\newblock In \emph{Proceedings of the 2021 ACM Conference on Fairness,
  Accountability, and Transparency}, pages 466--477, 2021.

\bibitem[{National Academies of Sciences, Engineering, and Medicine and
  others}(2020)]{national2020encouraging}
{National Academies of Sciences, Engineering, and Medicine and others}.
\newblock Encouraging adoption of protective behaviors to mitigate the spread
  of covid-19: Strategies for behavior change, 2020.

\bibitem[Ng(2021)]{GooglePr12:online}
Alfred Ng.
\newblock Google promised its contact tracing app was completely private - but
  it wasn?t.
\newblock
  \url{https://themarkup.org/privacy/2021/04/27/google-promised-its-contact-tracing-app-was-completely-private-but-it-wasnt},
  April 2021.
\newblock (Accessed on 08/19/2021).

\bibitem[Nicholas et~al.(2020)Nicholas, Onie, and Larsen]{nicholas2020ethics}
Jennifer Nicholas, Sandersan Onie, and Mark~E Larsen.
\newblock Ethics and privacy in social media research for mental health.
\newblock \emph{Current Psychiatry Reports}, 22\penalty0 (12):\penalty0 1--7,
  2020.

\bibitem[Nunes et~al.(2019)Nunes, Limpo, and Castro]{nunes2019acceptance}
Andreia Nunes, Teresa Limpo, and S{\~a}o~Lu{\'\i}s Castro.
\newblock Acceptance of mobile health applications: examining key determinants
  and moderators.
\newblock \emph{Frontiers in psychology}, 10:\penalty0 2791, 2019.

\bibitem[Obar and Oeldorf-Hirsch(2020)]{obar2020biggest}
Jonathan~A Obar and Anne Oeldorf-Hirsch.
\newblock The biggest lie on the internet: Ignoring the privacy policies and
  terms of service policies of social networking services.
\newblock \emph{Information, Communication \& Society}, 23\penalty0
  (1):\penalty0 128--147, 2020.

\bibitem[O'Toole et~al.(2007)O'Toole, Phillips, Jiang, Ayyad, Penard, and
  Abdi]{o2007face}
Alice~J O'Toole, P~Jonathon Phillips, Fang Jiang, Janet Ayyad, Nils Penard, and
  Herve Abdi.
\newblock Face recognition algorithms surpass humans matching faces over
  changes in illumination.
\newblock \emph{IEEE transactions on pattern analysis and machine
  intelligence}, 29\penalty0 (9):\penalty0 1642--1646, 2007.

\bibitem[O'Toole et~al.(2012)O'Toole, Phillips, An, and
  Dunlop]{o2012demographic}
Alice~J O'Toole, P~Jonathon Phillips, Xiaobo An, and Joseph Dunlop.
\newblock Demographic effects on estimates of automatic face recognition
  performance.
\newblock \emph{Image and Vision Computing}, 30\penalty0 (3):\penalty0
  169--176, 2012.

\bibitem[Padala and Gujar(2020)]{Padala2020achieving}
Manisha Padala and Sujit Gujar.
\newblock Fnnc: Achieving fairness through neural networks.
\newblock In \emph{Proceedings of the Twenty-Ninth International Joint
  Conference on Artificial Intelligence, {IJCAI-20}}, pages 2277--2283.
  International Joint Conferences on Artificial Intelligence Organization, 7
  2020.
\newblock \doi{10.24963/ijcai.2020/315}.
\newblock URL \url{https://doi.org/10.24963/ijcai.2020/315}.

\bibitem[Padrez et~al.(2016)Padrez, Ungar, Schwartz, Smith, Hill, Antanavicius,
  Brown, Crutchley, Asch, and Merchant]{Padrez2016}
Kevin~A. Padrez, Lyle Ungar, Hansen~Andrew Schwartz, Robert~J. Smith, Shawndra
  Hill, Tadas Antanavicius, Dana~M. Brown, Patrick Crutchley, David~A. Asch,
  and Raina~M. Merchant.
\newblock {Linking social media and medical record data: A study of adults
  presenting to an academic, urban emergency department}.
\newblock \emph{BMJ Quality and Safety}, 25\penalty0 (6):\penalty0 414--423,
  jun 2016.
\newblock ISSN 20445415.

\bibitem[Papernot et~al.(2016)Papernot, McDaniel, Jha, Fredrikson, Celik, and
  Swami]{papernot2015limitations}
Nicolas Papernot, Patrick McDaniel, Somesh Jha, Matt Fredrikson, Z~Berkay
  Celik, and Ananthram Swami.
\newblock The limitations of deep learning in adversarial settings.
\newblock In \emph{IEEE European Symposium on Security and Privacy (EuroS\&P)},
  pages 372--387. IEEE, 2016.

\bibitem[Paszke et~al.(2019)Paszke, Gross, Massa, Lerer, Bradbury, Chanan,
  Killeen, Lin, Gimelshein, Antiga, Desmaison, Kopf, Yang, DeVito, Raison,
  Tejani, Chilamkurthy, Steiner, Fang, Bai, and Chintala]{pytorch}
Adam Paszke, Sam Gross, Francisco Massa, Adam Lerer, James Bradbury, Gregory
  Chanan, Trevor Killeen, Zeming Lin, Natalia Gimelshein, Luca Antiga, Alban
  Desmaison, Andreas Kopf, Edward Yang, Zachary DeVito, Martin Raison, Alykhan
  Tejani, Sasank Chilamkurthy, Benoit Steiner, Lu~Fang, Junjie Bai, and Soumith
  Chintala.
\newblock Pytorch: An imperative style, high-performance deep learning library.
\newblock In \emph{NeurIPS}, pages 8026--8037. 2019.

\bibitem[Peri et~al.(2021)Peri, Curry, Dooley, and
  Dickerson]{peri2021preferencenet}
Neehar Peri, Michael Curry, Samuel Dooley, and John Dickerson.
\newblock Preferencenet: Encoding human preferences in auction design with deep
  learning.
\newblock \emph{Advances in Neural Information Processing Systems},
  34:\penalty0 17532--17542, 2021.

\bibitem[{Pew Research Center}(2021)]{pewresearchcenter}
{Pew Research Center}.
\newblock In response to climate change, citizens in advanced economies are
  willing to alter how they live and work.
\newblock Technical report, {Pew Research Center}, {Washington, D.C.},
  September 2021.
\newblock URL
  \url{https://www.pewresearch.org/global/wp-content/uploads/sites/2/2021/09/PG_2021.09.14_Climate_FINAL.pdf}.

\bibitem[Phillips and O'toole(2014)]{phillips2014comparison}
P~Jonathon Phillips and Alice~J O'toole.
\newblock Comparison of human and computer performance across face recognition
  experiments.
\newblock \emph{Image and Vision Computing}, 32\penalty0 (1):\penalty0 74--85,
  2014.

\bibitem[Phillips et~al.(2007)Phillips, Scruggs, O?Toole, Flynn, Bowyer,
  Schott, and Sharpe]{phillips2007frvt}
P~Jonathon Phillips, W~Todd Scruggs, Alice~J O?Toole, Patrick~J Flynn, Kevin~W
  Bowyer, Cathy~L Schott, and Matthew Sharpe.
\newblock Frvt 2006 and ice 2006 large-scale results.
\newblock \emph{National Institute of Standards and Technology, NISTIR},
  7408\penalty0 (1):\penalty0 1, 2007.

\bibitem[Phillips et~al.(2011)Phillips, Beveridge, Draper, Givens, O'Toole,
  Bolme, Dunlop, Lui, Sahibzada, and Weimer]{phillips2011introduction}
P~Jonathon Phillips, J~Ross Beveridge, Bruce~A Draper, Geof Givens, Alice~J
  O'Toole, David~S Bolme, Joseph Dunlop, Yui~Man Lui, Hassan Sahibzada, and
  Samuel Weimer.
\newblock An introduction to the good, the bad, \& the ugly face recognition
  challenge problem.
\newblock In \emph{2011 IEEE International Conference on Automatic Face \&
  Gesture Recognition (FG)}, pages 346--353. IEEE, 2011.

\bibitem[Phillips et~al.(2018)Phillips, Yates, Hu, Hahn, Noyes, Jackson,
  Cavazos, Jeckeln, Ranjan, Sankaranarayanan, et~al.]{phillips2018face}
P~Jonathon Phillips, Amy~N Yates, Ying Hu, Carina~A Hahn, Eilidh Noyes, Kelsey
  Jackson, Jacqueline~G Cavazos, G{\'e}raldine Jeckeln, Rajeev Ranjan, Swami
  Sankaranarayanan, et~al.
\newblock Face recognition accuracy of forensic examiners, superrecognizers,
  and face recognition algorithms.
\newblock \emph{Proceedings of the National Academy of Sciences}, 115\penalty0
  (24):\penalty0 6171--6176, 2018.

\bibitem[Pleiss et~al.(2017)Pleiss, Raghavan, Wu, Kleinberg, and
  Weinberger]{Pleiss17:Fairness}
Geoff Pleiss, Manish Raghavan, Felix Wu, Jon Kleinberg, and Kilian~Q
  Weinberger.
\newblock On fairness and calibration.
\newblock In I.~Guyon, U.~V. Luxburg, S.~Bengio, H.~Wallach, R.~Fergus,
  S.~Vishwanathan, and R.~Garnett, editors, \emph{Advances in Neural
  Information Processing Systems 30}, pages 5680--5689. Curran Associates,
  Inc., 2017.
\newblock URL
  \url{http://papers.NeurIPS.cc/paper/7151-on-fairness-and-calibration.pdf}.

\bibitem[Prasad et~al.(2012)Prasad, Sorber, Stablein, Anthony, and
  Kotz]{prasad2012understanding}
Aarathi Prasad, Jacob Sorber, Timothy Stablein, Denise Anthony, and David Kotz.
\newblock Understanding sharing preferences and behavior for mhealth devices.
\newblock In \emph{Proceedings of the 2012 ACM workshop on Privacy in the
  electronic society}, pages 117--128, 2012.

\bibitem[Preuveneers and Joosen(2016)]{preuveneers2016privacy}
Davy Preuveneers and Wouter Joosen.
\newblock Privacy-enabled remote health monitoring applications for resource
  constrained wearable devices.
\newblock In \emph{Proceedings of the 31st Annual ACM Symposium on Applied
  Computing}, pages 119--124, 2016.

\bibitem[Quadrianto et~al.(2019)Quadrianto, Sharmanska, and
  Thomas]{quadrianto2019discovering}
Novi Quadrianto, Viktoriia Sharmanska, and Oliver Thomas.
\newblock Discovering fair representations in the data domain.
\newblock In \emph{{IEEE} Conference on Computer Vision and Pattern
  Recognition, {CVPR} 2019, Long Beach, CA, USA, June 16-20, 2019}, pages
  8227--8236. Computer Vision Foundation / {IEEE}, 2019.
\newblock \doi{10.1109/CVPR.2019.00842}.
\newblock URL
  \url{http://openaccess.thecvf.com/content\_CVPR\_2019/html/Quadrianto\_Discovering\_Fair\_Representations\_in\_the\_Data\_Domain\_CVPR\_2019\_paper.html}.

\bibitem[Rabb et~al.(2021)Rabb, Glick, Houston, Bowers, and Yokum]{rabb2021no}
Nathaniel Rabb, David Glick, Attiyya Houston, Jake Bowers, and David Yokum.
\newblock No evidence that collective-good appeals best promote covid-related
  health behaviors.
\newblock \emph{Proceedings of the National Academy of Sciences}, 118\penalty0
  (14), 2021.

\bibitem[Raji and Buolamwini(2019)]{raji2019actionable}
Inioluwa~Deborah Raji and Joy Buolamwini.
\newblock Actionable auditing: Investigating the impact of publicly naming
  biased performance results of commercial ai products.
\newblock In \emph{Proceedings of the 2019 AAAI/ACM Conference on AI, Ethics,
  and Society}, pages 429--435, 2019.

\bibitem[Raskar et~al.(2020)Raskar, Nadeau, Werner, Barbar, Mehra, Harp,
  Leopoldseder, Wilson, Flakoll, Vepakomma, et~al.]{raskar2020covid}
Ramesh Raskar, Greg Nadeau, John Werner, Rachel Barbar, Ashley Mehra, Gabriel
  Harp, Markus Leopoldseder, Bryan Wilson, Derrick Flakoll, Praneeth Vepakomma,
  et~al.
\newblock Covid-19 contact-tracing mobile apps: evaluation and assessment for
  decision makers.
\newblock \emph{arXiv preprint arXiv:2006.05812}, 2020.

\bibitem[Redmiles(2018)]{redmiles2018net}
Elissa Redmiles.
\newblock Net benefits: Digital inequities in social capital, privacy
  preservation, and digital parenting practices of us social media users.
\newblock In \emph{Proceedings of the International AAAI Conference on Web and
  Social Media}, volume~12, 2018.

\bibitem[Redmiles(2020)]{redmiles_user_2020}
Elissa~M. Redmiles.
\newblock User {Concerns} \& {Tradeoffs} in {Technology}-facilitated {COVID}-19
  {Response}.
\newblock \emph{Digital Government: Research and Practice}, 2\penalty0
  (1):\penalty0 6:1--6:12, November 2020.
\newblock ISSN 2691-199X.
\newblock \doi{10.1145/3428093}.
\newblock URL \url{https://doi.org/10.1145/3428093}.

\bibitem[Ribeiro et~al.(2019)Ribeiro, Saha, Babaei, Henrique, Messias,
  Benevenuto, Goga, Gummadi, and Redmiles]{ribeiro2019microtagging}
Filipe~N. Ribeiro, Koustuv Saha, Mahmoudreza Babaei, Lucas Henrique, Johnnatan
  Messias, Fabricio Benevenuto, Oana Goga, Krishna~P. Gummadi, and Elissa~M.
  Redmiles.
\newblock On microtargeting socially divisive ads: A case study of
  russia-linked ad campaigns on facebook.
\newblock In \emph{Proceedings of the Conference on Fairness, Accountability,
  and Transparency}, FAT* '19, page 140?149, New York, NY, USA, 2019.
\newblock ISBN 9781450361255.
\newblock \doi{10.1145/3287560.3287580}.
\newblock URL \url{https://doi.org/10.1145/3287560.3287580}.

\bibitem[Robertson et~al.(2016)Robertson, Noyes, Dowsett, Jenkins, and
  Burton]{robertson2016face}
David~J Robertson, Eilidh Noyes, Andrew~J Dowsett, Rob Jenkins, and A~Mike
  Burton.
\newblock Face recognition by metropolitan police super-recognisers.
\newblock \emph{PloS one}, 11\penalty0 (2):\penalty0 e0150036, 2016.

\bibitem[Roijers et~al.(2013)Roijers, Vamplew, Whiteson, and
  Dazeley]{roijers2013survey}
Diederik~M Roijers, Peter Vamplew, Shimon Whiteson, and Richard Dazeley.
\newblock A survey of multi-objective sequential decision-making.
\newblock \emph{Journal of Artificial Intelligence Research}, 48:\penalty0
  67--113, 2013.

\bibitem[Rothstein and Siegal(2012)]{Rothstein2012}
Mark~A. Rothstein and Gil Siegal.
\newblock {Health Information Technology and Physicians' Duty to Notify
  Patients of New Medical Developments}.
\newblock \emph{Houston Journal of Health Law {\&} Policy}, pages 93--136,
  2012.
\newblock ISSN 1534-7907.

\bibitem[Ryu et~al.(2018)Ryu, Adam, and Mitchell]{ryu2018inclusivefacenet}
Hee~Jung Ryu, Hartwig Adam, and Margaret Mitchell.
\newblock Inclusivefacenet: Improving face attribute detection with race and
  gender diversity.
\newblock \emph{arXiv preprint arXiv:1712.00193}, 2018.

\bibitem[Saha et~al.(2020)Saha, Schumann, McElfresh, Dickerson, Mazurek, and
  Tschantz]{Saha20:Measuring}
Debjani Saha, Candice Schumann, Duncan~C. McElfresh, John~P. Dickerson,
  Michelle~L Mazurek, and Michael~Carl Tschantz.
\newblock Measuring non-expert comprehension of machine learning fairness
  metrics.
\newblock In \emph{International Conference on Machine Learning (ICML)}, 2020.

\bibitem[Salman et~al.(2019)Salman, Li, Razenshteyn, Zhang, Zhang, Bubeck, and
  Yang]{salman2019provable}
Hadi Salman, Jerry Li, Ilya Razenshteyn, Pengchuan Zhang, Huan Zhang, Sebastien
  Bubeck, and Greg Yang.
\newblock Provably robust deep learning via adversarially trained smoothed
  classifiers.
\newblock In \emph{Proceedings of the Annual Conference on Neural Information
  Processing Systems (NeurIPS)}, pages 11292--11303. 2019.

\bibitem[Savani et~al.(2020)Savani, White, and
  Govindarajulu]{savani2020posthoc}
Yash Savani, Colin White, and Naveen~Sundar Govindarajulu.
\newblock Intra-processing methods for debiasing neural networks.
\newblock In \emph{Proceedings of Advances in Neural Information Processing
  Systems}, 2020.

\bibitem[Schaub et~al.(2017)Schaub, Balebako, and Cranor]{schaub2017designing}
Florian Schaub, Rebecca Balebako, and Lorrie~Faith Cranor.
\newblock Designing effective privacy notices and controls.
\newblock \emph{IEEE Internet Computing}, 2017.

\bibitem[Schumann et~al.(2019{\natexlab{a}})Schumann, Counts, Foster, and
  Dickerson]{Schumann19:Diverse}
Candice Schumann, Samsara~N Counts, Jeffrey~S Foster, and John~P Dickerson.
\newblock The diverse cohort selection problem.
\newblock In \emph{Proceedings of the 18th International Conference on
  Autonomous Agents and MultiAgent Systems}, pages 601--609. International
  Foundation for Autonomous Agents and Multiagent Systems, 2019{\natexlab{a}}.

\bibitem[Schumann et~al.(2019{\natexlab{b}})Schumann, Counts, Foster, and
  Dickerson]{schumann2019diverse}
Candice Schumann, Samsara~N. Counts, Jeffrey~S. Foster, and John~P. Dickerson.
\newblock The diverse cohort selection problem.
\newblock In \emph{International Conference on Autonomous Agents and
  Multi-Agent Systems (AAMAS)}, page 601?609, 2019{\natexlab{b}}.

\bibitem[Schumann et~al.(2019{\natexlab{c}})Schumann, Lang, Foster, and
  Dickerson]{Schumann19:Making}
Candice Schumann, Zhi Lang, Jeffrey Foster, and John~P. Dickerson.
\newblock Making the cut: A bandit-based approach to tiered interviewing.
\newblock In \emph{Conference on Neural Information Processing Systems
  (NeurIPS)}, 2019{\natexlab{c}}.

\bibitem[Schumann et~al.(2019{\natexlab{d}})Schumann, Lang, Foster, and
  Dickerson]{Schumann2019}
Candice Schumann, Zhi Lang, Jeffrey~S Foster, and John~P Dickerson.
\newblock {Making the Cut: A Bandit-based Approach to Tiered Interviewing}.
\newblock In \emph{Neural Information Processing Systems}, 2019{\natexlab{d}}.

\bibitem[Schumann et~al.(2019{\natexlab{e}})Schumann, Lang, Mattei, and
  Dickerson]{Schumann2019a}
Candice Schumann, Zhi Lang, Nicholas Mattei, and John~P. Dickerson.
\newblock {Group Fairness in Bandit Arm Selection}.
\newblock dec 2019{\natexlab{e}}.
\newblock URL \url{http://arxiv.org/abs/1912.03802}.

\bibitem[Schumann et~al.(2020{\natexlab{a}})Schumann, Foster, Mattei, and
  Dickerson]{schumann2020hiring}
Candice Schumann, Jeffrey~S. Foster, Nicholas Mattei, and John~P. Dickerson.
\newblock We need fairness and explainability in algorithmic hiring.
\newblock In \emph{International Conference on Autonomous Agents and
  Multi-Agent Systems (AAMAS)}, page 1716?1720, 2020{\natexlab{a}}.

\bibitem[Schumann et~al.(2020{\natexlab{b}})Schumann, Foster, Mattei, and
  Dickerson]{schumann2020we}
Candice Schumann, Jeffrey~S Foster, Nicholas Mattei, and John~P Dickerson.
\newblock We need fairness and explainability in algorithmic hiring.
\newblock In \emph{Proceedings of the 19th International Conference on
  Autonomous Agents and MultiAgent Systems}, pages 1716--1720,
  2020{\natexlab{b}}.

\bibitem[Schumann et~al.(2021)Schumann, Pantofaru, Ricco, Prabhu, and
  Ferrari]{miap}
Candice Schumann, Caroline~Rebecca Pantofaru, Susanna Ricco, Utsav Prabhu, and
  Vittorio Ferrari.
\newblock A step toward more inclusive people annotations for fairness.
\newblock In \emph{Proceedings of the AAAI/ACM Conference on AI, Ethics, and
  Society}, 2021.

\bibitem[Seberger and Patil(2021)]{seberger2021us}
John~S Seberger and Sameer Patil.
\newblock Us and them (and it): Social orientation, privacy concerns, and
  expected use of pandemic-tracking apps in the united states.
\newblock In \emph{Proceedings of the 2021 CHI Conference on Human Factors in
  Computing Systems}, pages 1--19, 2021.

\bibitem[Serrano et~al.(2016)Serrano, Yu, Riley, Patel, Hughes, Marchesini, and
  Atienza]{serrano2016willingness}
Katrina~J Serrano, Mandi Yu, William~T Riley, Vaishali Patel, Penelope Hughes,
  Kathryn Marchesini, and Audie~A Atienza.
\newblock Willingness to exchange health information via mobile devices:
  findings from a population-based survey.
\newblock \emph{The Annals of Family Medicine}, 14\penalty0 (1):\penalty0
  34--40, 2016.

\bibitem[Shan et~al.(2020)Shan, Wenger, Zhang, Li, Zheng, and
  Zhao]{shan2020fawkes}
Shawn Shan, Emily Wenger, Jiayun Zhang, Huiying Li, Haitao Zheng, and Ben~Y
  Zhao.
\newblock Fawkes: Protecting privacy against unauthorized deep learning models.
\newblock In \emph{29th $\{$USENIX$\}$ Security Symposium ($\{$USENIX$\}$
  Security 20)}, pages 1589--1604, 2020.

\bibitem[Shing et~al.(2018)Shing, Nair, Zirikly, Friedenberg, {Daum{\'{e}}
  III}, and Resnik]{shing2018}
Han-Chin Shing, Suraj Nair, Ayah Zirikly, Meir Friedenberg, Hal {Daum{\'{e}}
  III}, and Philip Resnik.
\newblock {Expert, Crowdsourced, and Machine Assessment of Suicide Risk via
  Online Postings}.
\newblock In \emph{Proceedings of the Fifth Workshop on Computational
  Linguistics and Clinical Psychology: From Keyboard to Clinic}, pages 25--36,
  Stroudsburg, PA, USA, 2018. Association for Computational Linguistics.

\bibitem[Shing et~al.(2019)Shing, Wang, and Resnik]{eldan2019}
Han-Chin Shing, Guoli Wang, and Philip Resnik.
\newblock Assigning medical codes at the encounter level by paying attention to
  documents.
\newblock In \emph{Machine Learning for Health ({ML4H}) at {NeurIPS} 2019},
  2019.
\newblock Extended Abstract.

\bibitem[Shing et~al.(2020)Shing, Resnik, and Oard]{shing2020}
Han-Chin Shing, Philip Resnik, and Douglas Oard.
\newblock A prioritization model for suicidality risk assessment.
\newblock In \emph{Conference of the Association for Computational Linguistics
  (ACL 2020)}, July 2020.

\bibitem[Signorini et~al.(2011)Signorini, Segre, and
  Polgreen]{signorini2011use}
Alessio Signorini, Alberto~Maria Segre, and Philip~M Polgreen.
\newblock The use of twitter to track levels of disease activity and public
  concern in the us during the influenza a h1n1 pandemic.
\newblock \emph{PloS one}, 6\penalty0 (5):\penalty0 e19467, 2011.

\bibitem[Simko et~al.(2020)Simko, Chang, Jiang, Calo, Roesner, and
  Kohno]{simko2020covid19}
Lucy Simko, Jack~Lucas Chang, Maggie Jiang, Ryan Calo, Franziska Roesner, and
  Tadayoshi Kohno.
\newblock Covid-19 contact tracing and privacy: A longitudinal study of public
  opinion, 2020.

\bibitem[Simonyan and Zisserman(2015)]{simonyan2014very}
Karen Simonyan and Andrew Zisserman.
\newblock Very deep convolutional networks for large-scale image recognition.
\newblock In \emph{International Conference on Learning Representations
  (ICLR)}, 2015.

\bibitem[Singer(2018)]{singer2018microsoft}
Natasha Singer.
\newblock Microsoft urges congress to regulate use of facial recognition.
\newblock \emph{The New York Times}, 2018.

\bibitem[Singh and Joachims(2018)]{singh2018fairness}
Ashudeep Singh and Thorsten Joachims.
\newblock Fairness of exposure in rankings.
\newblock In \emph{International Conference on Knowledge Discovery and Data
  Mining (KDD)}, 2018.

\bibitem[Singh et~al.(2020)Singh, Agarwal, Singh, Nagpal, and
  Vatsa]{singh2020robustness}
Richa Singh, Akshay Agarwal, Maneet Singh, Shruti Nagpal, and Mayank Vatsa.
\newblock On the robustness of face recognition algorithms against attacks and
  bias.
\newblock In \emph{Proceedings of the AAAI Conference on Artificial
  Intelligence}, volume~34, pages 13583--13589, 2020.

\bibitem[Singla and Feizi(2020)]{singla2020curvature}
Sahil Singla and Soheil Feizi.
\newblock Second-order provable defenses against adversarial attacks.
\newblock In \emph{International Conference on Machine Learning (ICML)}, 2020.

\bibitem[Siroker and Koomen(2013)]{Siroker13:AB}
Dan Siroker and Pete Koomen.
\newblock \emph{A/B testing: The most powerful way to turn clicks into
  customers}.
\newblock John Wiley \& Sons, 2013.

\bibitem[Slivkins(2019)]{Slivkins19:Introduction}
Aleksandrs Slivkins.
\newblock Introduction to multi-armed bandits.
\newblock \emph{CoRR}, abs/1904.07272, 2019.

\bibitem[Solans et~al.(2020)Solans, Biggio, and Castillo]{solans2020poisoning}
David Solans, Battista Biggio, and Carlos Castillo.
\newblock Poisoning attacks on algorithmic fairness, 2020.

\bibitem[Speicher et~al.(2018)Speicher, Ali, Venkatadri, Ribeiro, Arvanitakis,
  Benevenuto, Gummadi, Loiseau, and Mislove]{speicher2018potential}
Till Speicher, Muhammad Ali, Giridhari Venkatadri, Filipe~Nunes Ribeiro, George
  Arvanitakis, Fabricio Benevenuto, Krishna~P. Gummadi, Patrick Loiseau, and
  Alan Mislove.
\newblock Potential for discrimination in online targeted advertising.
\newblock In \emph{ACM Conference on Fairness, Accountability, and Transparency
  (FAccT)}, 2018.

\bibitem[Stowell et~al.(2018)Stowell, Lyson, Saksono, Wurth, Jimison, Pavel,
  and Parker]{stowell2018designing}
Elizabeth Stowell, Mercedes~C Lyson, Herman Saksono, Rene{\'e}~C Wurth, Holly
  Jimison, Misha Pavel, and Andrea~G Parker.
\newblock Designing and evaluating mhealth interventions for vulnerable
  populations: A systematic review.
\newblock In \emph{Proceedings of the 2018 CHI Conference on Human Factors in
  Computing Systems}, pages 1--17, 2018.

\bibitem[Sun et~al.(2013)Sun, Wang, Guo, and Peng]{sun2013understanding}
Yongqiang Sun, Nan Wang, Xitong Guo, and Zeyu Peng.
\newblock Understanding the acceptance of mobile health services: a comparison
  and integration of alternative models.
\newblock \emph{Journal of electronic commerce research}, 14\penalty0
  (2):\penalty0 183, 2013.

\bibitem[Suykens and Vandewalle(1999)]{suykens1999least}
J.~A.~K. Suykens and J.~Vandewalle.
\newblock Least squares support vector machine classifiers.
\newblock \emph{Neural Processing Letters}, 9\penalty0 (3):\penalty0 293?300,
  June 1999.

\bibitem[Szegedy et~al.(2014)Szegedy, Zaremba, Sutskever, Bruna, Erhan,
  Goodfellow, and Fergus]{szegedy2013intriguing}
Christian Szegedy, Wojciech Zaremba, Ilya Sutskever, Joan Bruna, Dumitru Erhan,
  Ian Goodfellow, and Rob Fergus.
\newblock Intriguing properties of neural networks.
\newblock In \emph{International Conference on Learning Representations
  (ICLR)}, 2014.

\bibitem[Tang and Wang(2004)]{tang2004face}
Xiaoou Tang and Xiaogang Wang.
\newblock Face sketch recognition.
\newblock \emph{IEEE Transactions on Circuits and Systems for video
  Technology}, 14\penalty0 (1):\penalty0 50--57, 2004.

\bibitem[Thorbj{\o}rnsen et~al.(2020)Thorbj{\o}rnsen, Dahl{\'e}n, and
  Lange]{Thorbjornsen20:Tomorrow}
Helge Thorbj{\o}rnsen, Micael Dahl{\'e}n, and Fredrik Lange.
\newblock Tomorrow never dies: preadvertised sequels boost movie satisfaction
  and {WOM}.
\newblock \emph{International Journal of Advertising}, 39\penalty0
  (3):\penalty0 433--444, 2020.

\bibitem[Toussaert(2021)]{toussaert2021upping}
S{\'e}verine Toussaert.
\newblock Upping uptake of covid contact tracing apps.
\newblock \emph{Nature Human Behaviour}, 5\penalty0 (2):\penalty0 183--184,
  2021.

\bibitem[Tramer et~al.(2020)Tramer, Carlini, Brendel, and
  Madry]{tramer2020adaptive}
Florian Tramer, Nicholas Carlini, Wieland Brendel, and Aleksander Madry.
\newblock On adaptive attacks to adversarial example defenses, 2020.

\bibitem[Trepte et~al.(2017)Trepte, Reinecke, Ellison, Quiring, Yao, and
  Ziegele]{trepte2017cross}
Sabine Trepte, Leonard Reinecke, Nicole~B Ellison, Oliver Quiring, Mike~Z Yao,
  and Marc Ziegele.
\newblock A cross-cultural perspective on the privacy calculus.
\newblock \emph{Social Media+ Society}, 3\penalty0 (1):\penalty0
  2056305116688035, 2017.

\bibitem[Tucker(2014)]{tucker2014social}
Catherine~E Tucker.
\newblock Social networks, personalized advertising, and privacy controls.
\newblock \emph{Journal of marketing research}, 51\penalty0 (5):\penalty0
  546--562, 2014.

\bibitem[Velicia-Martin et~al.(2021)Velicia-Martin, Cabrera-Sanchez,
  Gil-Cordero, and Palos-Sanchez]{velicia-martin_researching_2021}
Felix Velicia-Martin, Juan-Pedro Cabrera-Sanchez, Eloy Gil-Cordero, and
  Pedro~R. Palos-Sanchez.
\newblock Researching {COVID}-19 tracing app acceptance: incorporating theory
  from the technological acceptance model.
\newblock \emph{PeerJ Computer Science}, 7:\penalty0 e316, January 2021.
\newblock ISSN 2376-5992.
\newblock \doi{10.7717/peerj-cs.316}.
\newblock URL \url{https://peerj.com/articles/cs-316}.

\bibitem[Venkatesh et~al.(2003)Venkatesh, Morris, Davis, and
  Davis]{venkatesh2003user}
Viswanath Venkatesh, Michael~G Morris, Gordon~B Davis, and Fred~D Davis.
\newblock User acceptance of information technology: Toward a unified view.
\newblock \emph{MIS quarterly}, pages 425--478, 2003.

\bibitem[Wadsworth et~al.(2018)Wadsworth, Vera, and
  Piech]{wadsworth2018achieving}
Christina Wadsworth, Francesca Vera, and Chris Piech.
\newblock Achieving fairness through adversarial learning: an application to
  recidivism prediction.
\newblock \emph{CoRR}, abs/1807.00199, 2018.

\bibitem[Walrave et~al.(2020)Walrave, Waeterloos, and
  Ponnet]{walrave_adoption_2020}
Michel Walrave, Cato Waeterloos, and Koen Ponnet.
\newblock Adoption of a {Contact} {Tracing} {App} for {Containing} {COVID}-19:
  {A} {Health} {Belief} {Model} {Approach}.
\newblock \emph{JMIR Public Health and Surveillance}, 6\penalty0 (3):\penalty0
  e20572, September 2020.
\newblock ISSN 2369-2960.
\newblock \doi{10.2196/20572}.
\newblock URL \url{http://publichealth.jmir.org/2020/3/e20572/}.

\bibitem[Wang et~al.(2018)Wang, Wang, Zhou, Ji, Gong, Zhou, Li, and
  Liu]{wang2018cosface}
Hao Wang, Yitong Wang, Zheng Zhou, Xing Ji, Dihong Gong, Jingchao Zhou, Zhifeng
  Li, and Wei Liu.
\newblock Cosface: Large margin cosine loss for deep face recognition.
\newblock In \emph{Proceedings of the IEEE conference on computer vision and
  pattern recognition}, pages 5265--5274, 2018.

\bibitem[Wang and Deng(2018)]{wang2018deep}
Mei Wang and Weihong Deng.
\newblock Deep face recognition: A survey.
\newblock \emph{arXiv preprint arXiv:1804.06655}, 2018.

\bibitem[Wang and Deng(2020)]{wang2020mitigating}
Mei Wang and Weihong Deng.
\newblock Mitigating bias in face recognition using skewness-aware
  reinforcement learning.
\newblock In \emph{Proceedings of the IEEE/CVF Conference on Computer Vision
  and Pattern Recognition}, pages 9322--9331, 2020.

\bibitem[Wang et~al.(2019)Wang, Zhao, Yatskar, Chang, and
  Ordonez]{wang2019balanced}
Tianlu Wang, Jieyu Zhao, Mark Yatskar, Kai-Wei Chang, and Vicente Ordonez.
\newblock Balanced datasets are not enough: Estimating and mitigating gender
  bias in deep image representations.
\newblock In \emph{Proceedings of the IEEE International Conference on Computer
  Vision}, pages 5310--5319, 2019.

\bibitem[Wang et~al.(2020{\natexlab{a}})Wang, Qinami, Karakozis, Genova, Nair,
  Hata, and Russakovsky]{wang2020fairness}
Zeyu Wang, Klint Qinami, Ioannis~Christos Karakozis, Kyle Genova, Prem Nair,
  Kenji Hata, and Olga Russakovsky.
\newblock Towards fairness in visual recognition: Effective strategies for bias
  mitigation, 2020{\natexlab{a}}.

\bibitem[Wang et~al.(2020{\natexlab{b}})Wang, Qinami, Karakozis, Genova, Nair,
  Hata, and Russakovsky]{Wang2020Towards}
Zeyu Wang, Klint Qinami, Yannis Karakozis, Kyle Genova, P.~Nair, Kenji Hata,
  and Olga Russakovsky.
\newblock Towards fairness in visual recognition: Effective strategies for bias
  mitigation.
\newblock \emph{2020 IEEE/CVF Conference on Computer Vision and Pattern
  Recognition (CVPR)}, pages 8916--8925, 2020{\natexlab{b}}.

\bibitem[Warner et~al.(2018)Warner, Gutmann, Sasse, and
  Blandford]{warner2018privacy}
Mark Warner, Andreas Gutmann, M~Angela Sasse, and Ann Blandford.
\newblock Privacy unraveling around explicit hiv status disclosure fields in
  the online geosocial hookup app grindr.
\newblock \emph{Proceedings of the ACM on human-computer interaction},
  2\penalty0 (CSCW):\penalty0 1--22, 2018.

\bibitem[Weise and Singer(2020{\natexlab{a}})]{weise2020}
Karen Weise and Natasha Singer.
\newblock Amazon pauses police use of its facial recognition software.
\newblock \emph{The New York Times}, Jul 2020{\natexlab{a}}.
\newblock URL
  \url{{https://www.nytimes.com/2020/06/10/technology/amazon-facial-recognition-backlash.html}}.

\bibitem[Weise and Singer(2020{\natexlab{b}})]{weise2020amazon}
Karen Weise and Natasha Singer.
\newblock Amazon pauses police use of its facial recognition software.
\newblock \emph{The New York Times}, Jul. 10 2020{\natexlab{b}}.
\newblock URL
  \url{https://www.nytimes.com/2020/06/10/technology/amazon-facial-recognition-backlash.html}.

\bibitem[White(2020)]{white2020men}
Alan White.
\newblock Men and covid-19: the aftermath.
\newblock \emph{Postgraduate Medicine}, 132\penalty0 (sup4):\penalty0 18--27,
  2020.

\bibitem[White et~al.(2015)White, Dunn, Schmid, and Kemp]{white2015error}
David White, James~D Dunn, Alexandra~C Schmid, and Richard~I Kemp.
\newblock Error rates in users of automatic face recognition software.
\newblock \emph{PloS one}, 10\penalty0 (10):\penalty0 e0139827, 2015.

\bibitem[Whittle(1988)]{Whittle88:Restless}
Peter Whittle.
\newblock Restless bandits: Activity allocation in a changing world.
\newblock \emph{Journal of Applied Probability}, 25\penalty0 (A):\penalty0
  287--298, 1988.

\bibitem[Wilber et~al.(2016)Wilber, Shmatikov, and Belongie]{wilber2016can}
Michael~J Wilber, Vitaly Shmatikov, and Serge Belongie.
\newblock Can we still avoid automatic face detection?
\newblock In \emph{2016 IEEE Winter Conference on Applications of Computer
  Vision (WACV)}, pages 1--9. IEEE, 2016.

\bibitem[Williams et~al.(2021)Williams, Armitage, Tampe, and
  Dienes]{williams_public_2021}
Simon~N. Williams, Christopher~J. Armitage, Tova Tampe, and Kimberly Dienes.
\newblock Public attitudes towards {COVID}?19 contact tracing apps: {A}
  {UK}?based focus group study.
\newblock \emph{Health Expectations}, 24\penalty0 (2):\penalty0 377--385, April
  2021.
\newblock ISSN 1369-6513, 1369-7625.
\newblock \doi{10.1111/hex.13179}.
\newblock URL \url{https://onlinelibrary.wiley.com/doi/10.1111/hex.13179}.

\bibitem[Woodroofe(1979)]{Woodroofe79:One-armed}
Michael Woodroofe.
\newblock A one-armed bandit problem with a concomitant variable.
\newblock \emph{Journal of the American Statistical Association}, 74\penalty0
  (368):\penalty0 799--806, 1979.

\bibitem[Xia et~al.(2016)Xia, Qin, Ma, Yu, and Liu]{xia2016budgeted}
Yingce Xia, Tao Qin, Weidong Ma, Nenghai Yu, and Tie-Yan Liu.
\newblock Budgeted multi-armed bandits with multiple plays.
\newblock In \emph{IJCAI}, pages 2210--2216, 2016.

\bibitem[Yang et~al.(2016)Yang, Yang, Dyer, He, Smola, and
  Hovy]{yang2016hierarchical}
Zichao Yang, Diyi Yang, Chris Dyer, Xiaodong He, Alex Smola, and Eduard Hovy.
\newblock Hierarchical attention networks for document classification.
\newblock In \emph{Proceedings of the 2016 Conference of the North American
  Chapter of the Association for Computational Linguistics: Human Language
  Technologies}, pages 1480--1489, 2016.

\bibitem[Zafar et~al.(2017{\natexlab{a}})Zafar, Valera, Gomez~Rodriguez, and
  Gummadi]{Zafar2017www}
Muhammad~Bilal Zafar, Isabel Valera, Manuel Gomez~Rodriguez, and Krishna~P.
  Gummadi.
\newblock Fairness beyond disparate treatment \& disparate impact.
\newblock \emph{Proceedings of the 26th International Conference on World Wide
  Web}, Apr 2017{\natexlab{a}}.
\newblock \doi{10.1145/3038912.3052660}.
\newblock URL \url{http://dx.doi.org/10.1145/3038912.3052660}.

\bibitem[Zafar et~al.(2017{\natexlab{b}})Zafar, Valera, Gomez{-}Rodriguez, and
  Gummadi]{zafar2017aistats}
Muhammad~Bilal Zafar, Isabel Valera, Manuel Gomez{-}Rodriguez, and Krishna~P.
  Gummadi.
\newblock Fairness constraints: Mechanisms for fair classification.
\newblock In \emph{Proceedings of the 20th International Conference on
  Artificial Intelligence and Statistics, {AISTATS} 2017, 20-22 April 2017,
  Fort Lauderdale, FL, {USA}}, volume~54 of \emph{Proceedings of Machine
  Learning Research}, pages 962--970. {PMLR}, 2017{\natexlab{b}}.
\newblock URL \url{http://proceedings.mlr.press/v54/zafar17a.html}.

\bibitem[Zafar et~al.(2017{\natexlab{c}})Zafar, Valera, Rodriguez, Gummadi, and
  Weller]{zafar2017parity}
Muhammad~Bilal Zafar, Isabel Valera, Manuel~Gomez Rodriguez, Krishna~P.
  Gummadi, and Adrian Weller.
\newblock From parity to preference-based notions of fairness in
  classification.
\newblock In \emph{Proceedings of the Annual Conference on Neural Information
  Processing Systems (NeurIPS)}, 2017{\natexlab{c}}.

\bibitem[Zafar et~al.(2019{\natexlab{a}})Zafar, Valera, Gomez-Rodriguez, and
  Gummadi]{zafar2019constraints}
Muhammad~Bilal Zafar, Isabel Valera, Manuel Gomez-Rodriguez, and Krishna~P.
  Gummadi.
\newblock Fairness constraints: A flexible approach for fair classification.
\newblock \emph{Journal of Machine Learning Research}, 20\penalty0
  (75):\penalty0 1--42, 2019{\natexlab{a}}.

\bibitem[Zafar et~al.(2019{\natexlab{b}})Zafar, Valera, Gomez-Rodriguez, and
  Gummadi]{zafar2019jmlr}
Muhammad~Bilal Zafar, Isabel Valera, Manuel Gomez-Rodriguez, and Krishna~P.
  Gummadi.
\newblock Fairness constraints: A flexible approach for fair classification.
\newblock \emph{Journal of Machine Learning Research}, 20\penalty0
  (75):\penalty0 1--42, 2019{\natexlab{b}}.
\newblock URL \url{http://jmlr.org/papers/v20/18-262.html}.

\bibitem[Zemel et~al.(2013{\natexlab{a}})Zemel, Wu, Swersky, Pitassi, and
  Dwork]{zemel13learning}
Rich Zemel, Yu~Wu, Kevin Swersky, Toni Pitassi, and Cynthia Dwork.
\newblock Learning fair representations.
\newblock volume~28 of \emph{Proceedings of Machine Learning Research}, pages
  325--333, Atlanta, Georgia, USA, 17--19 Jun 2013{\natexlab{a}}. PMLR.
\newblock URL \url{http://proceedings.mlr.press/v28/zemel13.html}.

\bibitem[Zemel et~al.(2013{\natexlab{b}})Zemel, Wu, Swersky, Pitassi, and
  Dwork]{zemel2013learning}
Rich Zemel, Yu~Wu, Kevin Swersky, Toni Pitassi, and Cynthia Dwork.
\newblock Learning fair representations.
\newblock In \emph{International Conference on Machine Learning (ICML)}, pages
  325--333, 2013{\natexlab{b}}.

\bibitem[Zhang et~al.(2020)Zhang, Kreps, McMurry, and
  McCain]{zhang_americans_2020}
Baobao Zhang, Sarah Kreps, Nina McMurry, and R.~Miles McCain.
\newblock Americans? perceptions of privacy and surveillance in the {COVID}-19
  pandemic.
\newblock \emph{PLOS ONE}, 15\penalty0 (12):\penalty0 e0242652, December 2020.
\newblock ISSN 1932-6203.
\newblock \doi{10.1371/journal.pone.0242652}.
\newblock URL \url{https://dx.plos.org/10.1371/journal.pone.0242652}.

\bibitem[Zhang et~al.(2019)Zhang, Tang, Dodge, Zhou, and
  Wang]{zhang2019metapred}
Xi~Sheryl Zhang, Fengyi Tang, Hiroko~H Dodge, Jiayu Zhou, and Fei Wang.
\newblock Metapred: Meta-learning for clinical risk prediction with limited
  patient electronic health records.
\newblock In \emph{Proceedings of the 25th ACM SIGKDD International Conference
  on Knowledge Discovery \& Data Mining}, pages 2487--2495, 2019.

\bibitem[Zhang et~al.(2014)Zhang, Guo, Lai, Guo, and
  Li]{zhang2014understanding}
Xiaofei Zhang, Xitong Guo, Kee-hung Lai, Feng Guo, and Chenlei Li.
\newblock Understanding gender differences in m-health adoption: a modified
  theory of reasoned action model.
\newblock \emph{Telemedicine and e-Health}, 20\penalty0 (1):\penalty0 39--46,
  2014.

\bibitem[Zhang et~al.(2017{\natexlab{a}})Zhang, Song, and Qi]{utk}
Zhifei Zhang, Yang Song, and Hairong Qi.
\newblock Age progression/regression by conditional adversarial autoencoder.
\newblock In \emph{Proceedings of the IEEE conference on computer vision and
  pattern recognition}, pages 5810--5818, 2017{\natexlab{a}}.

\bibitem[Zhang et~al.(2017{\natexlab{b}})Zhang, Song, and Qi]{zhifei2017cvpr}
Zhifei Zhang, Yang Song, and Hairong Qi.
\newblock Age progression/regression by conditional adversarial autoencoder.
\newblock In \emph{IEEE Conference on Computer Vision and Pattern Recognition
  (CVPR)}. IEEE, 2017{\natexlab{b}}.

\bibitem[Zimmermann et~al.(2021)Zimmermann, Fiske, Prainsack, Hangel, McLennan,
  and Buyx]{zimmermann2021early}
Bettina~Maria Zimmermann, Amelia Fiske, Barbara Prainsack, Nora Hangel, Stuart
  McLennan, and Alena Buyx.
\newblock Early perceptions of covid-19 contact tracing apps in german-speaking
  countries: comparative mixed methods study.
\newblock \emph{Journal of medical Internet research}, 23\penalty0
  (2):\penalty0 e25525, 2021.

\bibitem[Zirikly et~al.(2019)Zirikly, Resnik, Uzuner, and
  Hollingshead]{Zirikly2019}
Ayah Zirikly, Philip Resnik, {\"{O}}zlem Uzuner, and Kristy Hollingshead.
\newblock {CLPsych 2019 Shared Task: Predicting the Degree of Suicide Risk in
  Reddit Posts}.
\newblock In \emph{Proceedings of the Sixth Workshop on Computational
  Linguistics and Clinical Psychology}, pages 24--33, Stroudsburg, PA, USA,
  2019. Association for Computational Linguistics.

\end{thebibliography}


\end{document}